\documentclass[pdflatex,sn-mathphys-ay,iicol]{sn-jnl}

\usepackage{graphicx}%
\usepackage{multirow}%
\usepackage{amsmath,amssymb,amsfonts}%
\usepackage{amsthm}%
\usepackage[title]{appendix}%
\usepackage{textcomp}%
\usepackage{manyfoot}%
\usepackage{booktabs}%
\usepackage{algorithm}%
\usepackage{algorithmicx}%
\usepackage{algpseudocode}%
\usepackage{listings}%
\usepackage[mathscr]{eucal}
\usepackage{pifont} 
\usepackage{array} 
\usepackage{imakeidx}
\usepackage[table]{xcolor}%
\usepackage{hyperref}   
\usepackage{cleveref}   
\usepackage[normalem]{ulem}

\crefname{figure}{fig.}{figs.}
\Crefname{figure}{Fig.}{Figs.}
\Crefname{section}{Sec.}{Secs.}

\usepackage{caption}
\usepackage{subcaption}

\definecolor{catgray}{RGB}{230,230,230}
\usepackage{tcolorbox}
\usepackage{threeparttable}
\usepackage{tabularx}

\definecolor{rankfirst}{HTML}{A9C8EF}
\definecolor{ranksecond}{HTML}{C9DCF5}
\definecolor{rankthird}{HTML}{E5EFFB}
\definecolor{rankfail}{HTML}{F9DADA}

\definecolor{myred}{RGB}{230, 25, 75}
\definecolor{myblue}{RGB}{67, 99, 216}
\definecolor{mylime}{RGB}{191, 239, 69}
\definecolor{myyellow}{RGB}{255, 225, 25}
\definecolor{mymagenta}{RGB}{240, 50, 230}
\definecolor{mycyan}{RGB}{66, 212, 244}
\definecolor{mypurple}{RGB}{145, 30, 180}

\newcommand{\blfootnotetext}[1]{%
  \begingroup
  \renewcommand{\thefootnote}{}%
  \footnotetext{#1}%
  \endgroup
}

\usepackage[nolist,nohyperlinks]{acronym}
\makeatletter
\let\AC@placelabel\@gobble
\let\AC@placelabel@part\@gobble
\makeatother
\acrodef{E2E}[E2E]{end-to-end}
\acrodef{MVS}[MVS]{multi-view stereo}
\acrodef{SfM}[SfM]{structure from motion}
\acrodef{SLAM}[SLAM]{Simultaneous Localization and Mapping}
\acrodef{OOM}[OOM]{out-of-memory}
\acrodef{SOTA}[SOTA]{state-of-the-art}
\acrodef{DA3}[DA3]{Depth Anything 3}
\acrodef{DA-Next}[DA-Next]{Depth-Anything-Next}
\acrodef{sMSE}[sMSE]{Scale-invariant Mean Square Error}
\acrodef{RRA}[RRA]{relative rotation accuracy}
\acrodef{RTA}[RTA]{relative translation accuracy}
\acrodef{AUC}[AUC]{area under curve}
\acrodef{GSD}[GSD]{ground sampling distance}

\def\eg{\emph{e.g.}} 
\def\ie{\emph{i.e}}

\makeatother

\theoremstyle{thmstyleone}%
\theoremstyle{thmstyletwo}%

\theoremstyle{thmstylethree}%

\gdef\orcidlogo{%
\includegraphics{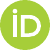}%
}%

\renewcommand{\orcidlogo}{%
  \includegraphics[height=1em]{Orcidlogo.pdf}%
}

\newcommand{\plotdir}{figs/sup_materials/plots/}
\newcommand{\plotw}{0.32\linewidth}
\newcommand{\plotcell}[1]{\includegraphics[width=\plotw,keepaspectratio]{\plotdir #1}}

\begin{document}

\title[A 3D Reconstruction Benchmark for Asset Inspection]{A 3D Reconstruction Benchmark for Asset Inspection}

\author*[]{\fnm{James L.} \sur{Gray} \orcid{https://orcid.org/0009-0008-3943-6588}}\email{jim.lyndon.gray@gmail.com}

\author[]{\fnm{Nikolai} \sur{Goncharov} \orcid{https://orcid.org/0009-0004-3477-9638}}\email{nikolai.goncharov@sydney.edu.au}

\author[]{\fnm{Alexandre} \sur{Cardaillac} \orcid{https://orcid.org/0000-0003-1918-4893}}\email{alexandre.cardaillac@sydney.edu.au}

\author[]{\fnm{Joshua} \sur{Knights} \orcid{https://orcid.org/0000-0002-1215-6956}}\email{joshua.knights@sydney.edu.au}

\author[]{\fnm{Jack} \sur{Naylor} \orcid{https://orcid.org/0000-0003-2185-7102}}\email{jack.naylor@sydney.edu.au}

\author[]{\fnm{Ryan} \sur{Griffiths} \orcid{https://orcid.org/0000-0002-2126-2523}}\email{r.griffiths@sydney.edu.au}

\author[]{\fnm{Donald G.} \sur{Dansereau} \orcid{https://orcid.org/0000-0003-2540-1639}}\email{donald.dansereau@sydney.edu.au}

\affil[]{\orgdiv{Australian Centre for Robotics}, \orgname{University of Sydney}, \orgaddress{\street{Street}, \city{Sydney}, \postcode{2006}, \state{NSW}, \country{Australia}}}

\abstract{
 Asset management requires accurate 3D models to inform the maintenance, repair, and assessment of buildings, maritime vessels, and other key structures as they age. These downstream applications rely on high-fidelity models produced from aerial surveys in close proximity to the asset, enabling operators to locate and characterise deterioration or damage and plan repairs. Captured images typically have high overlap between adjacent camera poses, sufficient detail at millimetre scale, and challenging visual appearances such as reflections and transparency. However, existing 3D reconstruction datasets lack examples of these conditions, making it difficult to benchmark methods for this task. We present a new dataset of 364k images with ground truth depth maps, camera poses, and mesh models of seven synthetic scenes with simulated inspection trajectories and varying surface conditions and reflectivities on non-Lambertian scene content. 
 We perform an extensive evaluation of state-of-the-art reconstruction methods on this dataset. 
 Our results show that current approaches struggle with thin structures such as lattices and cables. 
 Additionally, the evaluation reveals a trade-off between detail and 3D reconstruction accuracy, as camera trajectories closer to the asset yield worse 3D reconstruction performance on average. Project page:
 \url{https://roboticimaging.org/Projects/asset-inspection-dataset/}
}

\keywords{3D Reconstruction, Asset Inspection, Structure from Motion, Multi-View Stereo, Feed-Forward Model }

\maketitle
\section{Introduction}
\label{sec:intro}

3D reconstruction recovers the geometry and appearance of a scene at high resolution by establishing correspondences between camera views, and can do so from inexpensive RGB imagery alone.
In asset management, these reconstructions serve as digital twins of structures such as buildings, bridges and ships. The 3D model spatially organises thousands of captured photos so that inspectors can navigate to any region of the asset, while its geometry allows defects to be localised and assessed to plan maintenance~\citep{koch_review_2015, sharma_digital_2022}.

\begin{figure*}
    \centering
    \includegraphics[width=\linewidth]{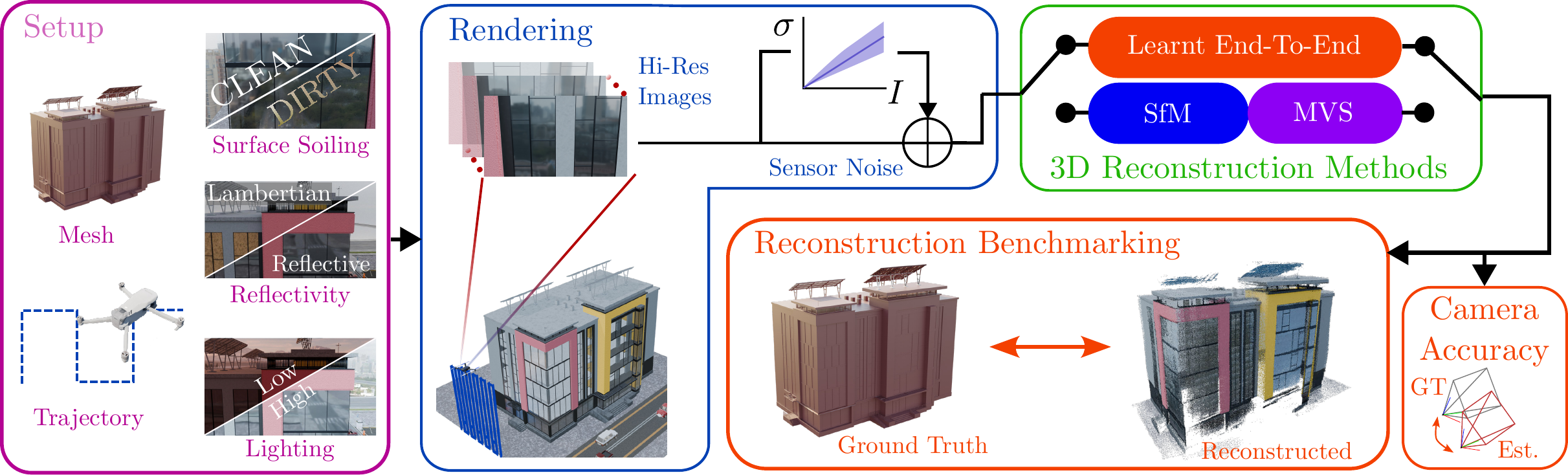}
    \caption{\small
    We propose a new benchmark for 3D reconstruction from imagery captured in asset inspection operations. By considering surface soiling, reflectivity, drone trajectories, lighting and asset meshes, we generate photorealistic synthetic data that matches the close proximity and high overlap of data captured in industrial asset inspection operations. After adding emulated signal-dependent sensor noise, we benchmark the camera pose and 3D reconstruction accuracy of both end-to-end transformer architectures and traditional structure from motion and multi-view stereo pipelines with ground truth data from our simulator. 
    }
    \label{fig:teaster_photo}
\end{figure*}

3D reconstruction has advanced rapidly in recent years.
\Ac{E2E} learning-based models now achieve \ac{SOTA} accuracy on standard benchmarks~\citep{wang_vggt_2025, wang2025pi, depthanything3, peng2026spatialbenchspatialfoundationmodel}, while advances in \ac{SfM} and \ac{MVS} have improved both efficiency~\citep{pan2024glomap} and accuracy~\citep{zeng_apde_2025, chen2025dual}.
Public benchmarks~\citep{schoeps2017cvpr, Knapitsch2017_Tanks, jensen2014large, yao2020blendedmvs, peng2026spatialbenchspatialfoundationmodel} have driven this progress by providing a common basis for quantitative comparison and large-scale training data for learning-based methods.
However, these benchmarks predominantly contain object-centric or distant views in which a whole object or a large portion of the scene is visible in each frame.
On the other hand, asset-inspection photos are typically taken with an emphasis on detail, with an aim of 1.5--10mm/px of resolution~\citep{Trendspek_capture_guide, t2_capture_guide}.
As a result, for large structures, each view may only cover a small part of it and a single survey can comprise thousands of images. 
To our knowledge, no publicly available dataset reproduces these capture conditions, leaving a significant domain gap between the existing benchmarks and asset inspection data and leaving an open question as to whether learning-based models show diminished performance on such data.
Our benchmark targets this gap directly.
Each scene was rendered along an inspection trajectory that follows industry capture guidelines~\citep{Trendspek_capture_guide}, and along trajectories at three distances from the asset corresponding to different spatial resolutions.
Each trajectory is also rendered under four lighting conditions. 
By rendering these scenes under different lighting conditions and from different trajectories, we can measure each method's sensitivity to changes in trajectory and illumination.

Asset inspection also depends on local geometric accuracy, since defects must be resolved where they occur, yet global point cloud and pose metrics can remain favourable despite substantial local errors.
Real benchmarks rarely provide the dense per-view depth needed to measure this.
Because our dataset is synthetic, it provides exact, dense depth maps for every view alongside meshes, camera poses and point clouds, enabling both local and global evaluation.

Glass facades are now common on modern buildings, making them a high-value target for remote asset inspection.
However, reflective surfaces remain a persistent failure case for 3D reconstruction~\citep{ihrke_transparent_2010, guo_nerfren_2022}: reflections violate photometric consistency, and planar glass in particular leads methods to reconstruct a virtual surface behind the pane~\citep{whelan2018reconstructing, tan2021mirror3d}.
No existing dataset varies reflectivity and soiling independently of geometry in an inspection setting.
To address this gap, our dataset contains two scenes with glass facades at five reflectivity and four soiling levels with fixed geometry and camera poses.

We conduct an evaluation of several \ac{E2E}, \ac{E2E} chunking, SLAM-based and \ac{SfM} methods on our new dataset.
Our evaluation reveals several key insights:
(1) Thin structures such as cables, lattices and trusses remain an open challenge for 3D reconstruction methods.
(2) For most 3D reconstruction approaches, there is a substantial trade-off between image detail and reconstruction accuracy. Camera trajectories further from the asset yield better reconstruction results for most 3D reconstruction methods, and the \textit{manual} trajectory that followed industry capture guidelines proved to be an additional challenge for learning-based methods.
(3) Pure \ac{E2E} methods experience \ac{OOM} errors on long sequences, making them impractical for asset-inspection applications.
(4) Low-light conditions pose an additional challenge to 3D reconstruction methods.
(5) Higher reflectivities reduce depth map accuracy for classical and learning-based methods to a lesser extent and produce unpredictable effects on point cloud quality.
(6) The best \ac{E2E} and their chunking and SLAM equivalents produce higher-quality clouds and more accurate depth maps than \ac{SfM} methods, while the best \ac{SfM} methods produce more accurate camera poses with a lower pose registration rate.
(7) Global metrics that measure scene-level performance do not properly capture local performance, which is necessary for asset inspection applications.

To address the limitations of existing datasets and benchmarks in an asset inspection context, this paper presents three key contributions:
\begin{enumerate}
    \item A synthetic benchmark containing 364k images for 3D reconstruction in asset inspection that isolates the effects of capture trajectory, ground sampling distance, lighting, reflectivity and surface soiling, with dense depth maps, meshes, camera poses and point clouds as ground truth for evaluating both local and global accuracy. 
    \item An extensive evaluation of widely used and \ac{SOTA} 3D reconstruction methods, spanning \ac{SfM}, \ac{E2E} feedforward, and \ac{SLAM}-based \ac{E2E} approaches. 
    \item An open-source inference, evaluation and visualisation pipeline that allows for the straightforward evaluation of 3D reconstruction approaches on this dataset.
\end{enumerate}
 Our dataset is available at \url{https://huggingface.co/datasets/RoboticImaging/3D-Reconstruction-Asset-Inspection-Benchmark}, and our evaluation pipeline is available at \url{https://github.com/RoboticImaging/inspection_dataset_tools}.
 
\section{Related Work}
\label{sec:RelatedWork}
\subsection{3D reconstruction benchmarks and datasets}

\begin{table*}[ht]
    \centering
    \caption{\small
    A comparison between our dataset and existing 3D reconstruction and drone imagery datasets.
    We compare whether each dataset has asset inspection-style imagery (Insp.), whether it can vary reflectivity (Ref.), surface condition or materials (Surf.), lighting and environment maps (Env.), and camera trajectories (Cam.).
    We also detail whether each dataset has ground truth depth data (Depth), whether it is captured from real scenes (Real), whether it has point clouds and camera poses (3D GT) and whether it has dynamic scenes (Dyn.).
    Image numbers with an * symbol have multiple copies of the same scenes captured or rendered under multiple different conditions, which yields an increased frame count.
    ETH3D has sparse ground truth depth maps, so it is given {\color{orange}$\boldsymbol{\sim}$}
    in the Depth column.
    }
    \label{tab:related_datasets}
    \footnotesize
    \begin{tabular}{c c c c c c c c c c c}
    \hline
    \textbf{Dataset} &
    {\textbf{Insp.}} & 
    {\textbf{Ref.}} & 
    {\textbf{Surf.}} & 
    {\textbf{Env.}} & 
    {\textbf{Cam.}} & 
    {\textbf{Depth}} & 
    {\textbf{Real}} & 
    {\textbf{3D GT}} & 
    {\textbf{Dyn.}} & 
    {\textbf{Images}} \\
    \hline
    BlendedMVS\footnotemark[1] &
    \color{red}{\ding{55}} & 
    \color{red}{\ding{55}} & 
    \color{red}{\ding{55}} & 
    \color{red}{\ding{55}} & 
    \color{green}{\ding{51}} & 
    \color{red}{\ding{55}} & 
    \color{red}{\ding{55}} & 
    \color{red}{\ding{55}} & 
    \color{red}{\ding{55}} & 
    110k* 
    \\

    Smith et al.\footnotemark[2] &
    \color{red}{\ding{55}} & 
    \color{red}{\ding{55}} & 
    \color{red}{\ding{55}} & 
    \color{red}{\ding{55}} & 
    \color{green}{\ding{51}} & 
    \color{red}{\ding{55}} & 
    \color{red}{\ding{55}} & 
    \color{green}{\ding{51}} & 
    \color{red}{\ding{55}} & 
    3274*
    \\ 

    DTU\footnotemark[3] & 
    \color{red}{\ding{55}} & 
    \color{red}{\ding{55}} & 
    \color{red}{\ding{55}} & 
    \color{green}{\ding{51}} & 
    \color{red}{\ding{55}} & 
    \color{red}{\ding{55}} & 
    \color{green}{\ding{51}} & 
    \color{green}{\ding{51}} & 
    \color{red}{\ding{55}} & 
    30k* \\

    Urban SaR\footnotemark[4] &
    \color{red}{\ding{55}} & 
    \color{red}{\ding{55}} & 
    \color{red}{\ding{55}} & 
    \color{red}{\ding{55}} & 
    \color{red}{\ding{55}} & 
    \color{red}{\ding{55}} & 
    \color{green}{\ding{51}} & 
    \color{green}{\ding{51}} & 
    \color{green}{\ding{51}} & 
    3065 
    \\

    Tanks \& Temples\footnotemark[5] & 
    \color{red}{\ding{55}} & 
    \color{red}{\ding{55}} & 
    \color{red}{\ding{55}} & 
    \color{red}{\ding{55}} & 
    \color{red}{\ding{55}} & 
    \color{red}{\ding{55}} & 
    \color{green}{\ding{51}} & 
    \color{green}{\ding{51}} & 
    \color{green}{\ding{51}} & 
    4100 
    \\

    ETH3D\footnotemark[6] &
    \color{red}{\ding{55}} & 
    \color{red}{\ding{55}} & 
    \color{red}{\ding{55}} & 
    \color{red}{\ding{55}} & 
    \color{red}{\ding{55}} & 
    \color{orange}{$\boldsymbol{\sim}$} & 
    \color{green}{\ding{51}} & 
    \color{green}{\ding{51}} & 
    \color{green}{\ding{51}} & 
    898 
    \\

    MatrixCity\footnotemark[7] & 
    \color{red}{\ding{55}} & 
    \color{red}{\ding{55}} & 
    \color{red}{\ding{55}} & 
    \color{green}{\ding{51}} & 
    \color{green}{\ding{51}} & 
    \color{green}{\ding{51}} & 
    \color{red}{\ding{55}} & 
    \color{green}{\ding{51}} & 
    \color{green}{\ding{51}} & 
    519k*
    \\

    LightCity\footnotemark[8] & 
    \color{red}{\ding{55}} & 
    \color{red}{\ding{55}} & 
    \color{red}{\ding{55}} & 
    \color{green}{\ding{51}} & 
    \color{green}{\ding{51}} & 
    \color{green}{\ding{51}} & 
    \color{red}{\ding{55}} & 
    \color{green}{\ding{51}} & 
    \color{red}{\ding{55}} & 
    50k* 
    \\

    DA-Next-5M\footnotemark[9] &
    \color{red}{\ding{55}} & 
    \color{red}{\ding{55}} & 
    \color{red}{\ding{55}} & 
    \color{red}{\ding{55}} & 
    \color{red}{\ding{55}} & 
    \color{green}{\ding{51}} & 
    \color{green}{\ding{51}} & 
    \color{green}{\ding{51}} & 
    \color{green}{\ding{51}} & 
    5M \\

    UrbanScene3D\footnotemark[10] &
    \color{red}{\ding{55}} & 
    \color{red}{\ding{55}} & 
    \color{red}{\ding{55}} & 
    \color{red}{\ding{55}} & 
    \color{green}{\ding{51}} & 
    \color{green}{\ding{51}} & 
    \color{green}{\ding{51}} & 
    \color{green}{\ding{51}} & 
    \color{green}{\ding{51}} & 
    128k* 
    \\

    PASMVS\footnotemark[11] &
    \color{red}{\ding{55}} & 
    \color{green}{\ding{51}} & 
    \color{green}{\ding{51}} & 
    \color{green}{\ding{51}} & 
    \color{green}{\ding{51}} & 
    \color{green}{\ding{51}} & 
    \color{red}{\ding{55}} & 
    \color{red}{\ding{55}} & 
    \color{red}{\ding{55}} & 
    18k* \\

    ClaraVid\footnotemark[12] &
    \color{red}{\ding{55}} & 
    \color{red}{\ding{55}} & 
    \color{red}{\ding{55}} & 
    \color{red}{\ding{55}} & 
    \color{green}{\ding{51}} & 
    \color{green}{\ding{51}} & 
    \color{red}{\ding{55}} & 
    \color{green}{\ding{51}} & 
    \color{green}{\ding{51}} & 
    17k*
    \\

    GigaMVS\footnotemark[13] &
    \color{red}{\ding{55}} & 
    \color{red}{\ding{55}} & 
    \color{red}{\ding{55}} & 
    \color{red}{\ding{55}} & 
    \color{red}{\ding{55}} & 
    \color{red}{\ding{55}} & 
    \color{green}{\ding{51}} & 
    \color{green}{\ding{51}} & 
    \color{green}{\ding{51}} & 
    3599 \\
    \hline
    \textbf{Ours} &
    \color{green}{\ding{51}} & 
    \color{green}{\ding{51}} & 
    \color{green}{\ding{51}} & 
    \color{green}{\ding{51}} & 
    \color{green}{\ding{51}} & 
    \color{green}{\ding{51}} & 
    \color{red}{\ding{55}} & 
    \color{green}{\ding{51}} & 
    \color{green}{\ding{51}} & 
    364k* \\
    \hline
    \end{tabular}
\end{table*}

\Cref{tab:related_datasets} compares existing datasets along the factors relevant to asset inspection.
Most real benchmarks image objects or scenes from the ground: DTU captures tabletop objects with a robot-mounted camera~\citep{jensen2014large}, Tanks and Temples provides handheld video of large indoor and outdoor scenes~\citep{Knapitsch2017_Tanks}, ETH3D combines high-resolution DSLR images with video~\citep{Schops_ETH3D_benchmark}, and GigaMVS covers large scenes with gigapixel imagery~\citep{GigaMVS2022}. More recent benchmarks add wrist-mounted and egocentric views~\citep{peng2026spatialbenchspatialfoundationmodel}.
While aerial datasets simulate drone capture, they do not follow asset-inspection trajectories.
BlendedMVS~\citep{yao2020blendedmvs}, Urban SaR~\citep{martell_benchmarking_2018}, UrbanScene3D~\citep{lin2022capturing}, MatrixCity~\citep{li2023matrixcity} and ClaraVid~\citep{beche2025claravid} target block- or city-scale reconstruction from altitude, and \citet{smith2018aerial} introduce a dataset that plans trajectories for completeness at building scale.
In all of these, each frame sees a whole object or a large part of the scene. 
None contain the sustained close-range, high-overlap capture of a single asset that inspection guidelines prescribe~\citep{Trendspek_capture_guide, t2_capture_guide}.

Additionally, real benchmarks are limited by their ability to collect ground truth data.
Typically, the ground truth data is derived from laser or structured-light scans~\citep{jensen2014large, Knapitsch2017_Tanks, Schops_ETH3D_benchmark, GigaMVS2022}, which are incomplete in occluded regions, unreliable on glass and other reflective surfaces~\citep{whelan2018reconstructing, tan2021mirror3d}, cannot capture moving objects, and must be registered to the images.
Tanks and Temples therefore evaluates against a scanned point cloud, and ETH3D's depth maps, rendered from its scans, are sparse.
Evaluation is consequently restricted to scene-level point cloud and pose metrics, which are insensitive to the localised failures that matter most for inspection: thin structures such as railings, cabling and scaffolding, large textureless regions, and non-Lambertian surfaces such as glass and metal.

Furthermore, with real datasets, it is difficult to isolate individual factors of variation, since scene geometry and appearance often change between captures and repeated field campaigns are costly.
DTU, which images each object under multiple controlled lighting conditions, is the main exception.
Synthetic datasets can isolate individual factors while providing dense ground truth: ClaraVid and UrbanScene3D vary camera trajectories, MatrixCity and LightCity~\citep{wang2025lightcity} vary lighting and environmental conditions, and PASMVS~\citep{broekman_pasmvs_2020} renders objects at several levels of specularity.
PASMVS is the only dataset that varies reflectivity with fixed geometry.
However, unlike the glass facades of real buildings, its objects have highly complex curved geometry and they are rendered against a black background with little to reflect.
As \Cref{tab:related_datasets} shows, no existing dataset combines inspection-style capture with dense ground truth and independent control of trajectory, lighting, reflectivity and surface condition.

\blfootnotetext{1. \citep{yao2020blendedmvs}; 2. \citep{smith2018aerial}; 3. \citep{jensen2014large}; 4. \citep{martell_benchmarking_2018}; 5. \citep{Knapitsch2017_Tanks}; 6. \citep{Schops_ETH3D_benchmark}; 7. \citep{li2023matrixcity}; 8. \citep{wang2025lightcity}; 9. \citep{peng2026spatialbenchspatialfoundationmodel}; 10. \citep{lin2022capturing}; 11. \citep{broekman_pasmvs_2020}; 12. \citep{beche2025claravid}; 13. \citep{GigaMVS2022}}

\subsection{3D reconstruction approaches}
\label{sec:learning_vs_traditional}

Traditionally, 3D reconstruction from a series of unposed 2D images is performed in two stages: an \ac{SfM} stage and an \ac{MVS} stage.
The \ac{SfM} stage recovers sparse points, camera poses, and intrinsics by finding correspondences between images using image features~\citep{sarlin2020superglue,lowe2004distinctive} and triangulating 3D points using the epipolar geometry of calibrated camera models~\citep{hartley2003multiple, schoenberger2016sfm}.
After the \ac{SfM} stage, an \ac{MVS} stage matches patches between images, producing depth maps and a dense point cloud~\citep{furukawa2009accurate, galliani2015massively, zheng2014patchmatch, schoenberger2016mvs}.
Both stages typically struggle in low-texture regions since feature and patch matches are ambiguous~\citep{martell_benchmarking_2018}.
Furthermore, in non-Lambertian regions, classic photometric constancy assumptions used in both \ac{SfM} and \ac{MVS} stages are violated.
Scenes that feature these challenges often yield incorrect or incomplete reconstructions.
To address these issues in the SfM stage, 
detector-free matching has been proposed to deal with low-texture scenes~\citep{he2024dfsfm} and, more recently, MP-SfM integrated monocular depth and normal cues into the \ac{SfM} pipeline, improving performance in scenes with low overlap between views~\citep{pataki2025mpsfm}.
In addition, recent advances in monocular depth estimation have been integrated into the \ac{MVS} stage, yielding improved results~\citep{guo_multi-view_2025}.
Fully learned \ac{MVS} stages can also be used, such as the work of~\cite{fan_revisiting_2026} that yields highly consistent and accurate results.

A compelling alternative to \ac{SfM} pipelines is \ac{E2E} learning-based models.
Beginning with DUSt3R~\citep{wang2024dust3r}, these \ac{E2E} approaches use transformer-based architectures trained extensively on 3D reconstruction datasets.
By leveraging learned priors, in addition to epipolar geometry~\citep{bratulic_geometric_2026}, these models can produce accurate point clouds from unconstrained image collections, even in challenging scenes where classical methods struggle, such as those with little overlap between frames and/or complex appearances.
These models are appealing: they achieve \ac{SOTA} accuracy within their domains and use a feedforward structure that parallelises 3D reconstruction, and therefore avoids lengthy iterative optimisation stages~\citep{wang_vggt_2025, wang2025pi, depthanything3, wang2026vggtomega, peng2026spatialbenchspatialfoundationmodel}. 
Unfortunately, these models perform significantly worse out-of-domain~\citep{peng2026spatialbenchspatialfoundationmodel, li_when_2026} and require prohibitive amounts of VRAM on long sequences with hundreds or thousands of images~\citep{peng2026spatialbenchspatialfoundationmodel, deng2025vggtlongchunkitloop}.
To address this issue, chunking approaches, \eg{} VGGT-Long, divide long sequences into overlapping chunks, which are then aligned to reconstruct the scene~\citep{deng2025vggtlongchunkitloop}.
An alternative is to incorporate \ac{E2E} models into a \ac{SLAM} system for real-time 3D reconstruction~\citep{murai2024_mast3rslam, maggio2025vggt-slam, maggio2025vggt-slam2}.
Note that asset-inspection-style data typically features close-proximity images to large objects and long sequences of hundreds and thousands of images. 
To our knowledge, there has been no exploration as to whether this style of data is in- or out-of-domain for \ac{E2E} models and whether these chunking and \ac{SLAM} approaches are appropriate for this application.
To investigate this and identify open challenges, we benchmark \ac{SfM} methods, \ac{E2E} models, and their chunked variants on our novel asset-inspection 3D reconstruction dataset.

\subsection{Computer vision in asset inspection}

The goal of 3D reconstruction for asset inspection is to build a digital twin of the asset, a highly detailed representation that captures its appearance and geometry~\citep{Dong2025DigitalTwin}.
Computer vision supports the whole inspection workflow, from collecting imagery with drones~\citep{ham2016visual, xu2023vision, mirzazade2023semi} to detecting and segmenting defects such as cracks and corrosion~\citep{koch_review_2015, cha2017crack, spencer2019advances}.
Increasingly, these image-level detections are projected onto a 3D reconstruction of the asset so that damage can be localised, measured and tracked over time~\citep{hoskere2018condition, kalfarisi2020crack}.
The value of such condition-aware models depends directly on the local accuracy of the underlying reconstruction: a defect mapped onto misplaced geometry is mislocated and mismeasured.

Trajectory planning for 3D reconstruction has received attention,  largely on large-scale completeness~\citep{feng2024fc, feng2023predrecon} and visual fidelity~\citep{you_openflyscan_2026} rather than the fine surface detail and geometric accuracy required in asset management tasks~\citep{wang2025path}.
\citet{smith2018aerial} introduced a framework benchmarking 3D reconstruction from different trajectory plans and demonstrated higher completeness and improved reconstruction from optimised per-scene trajectories, motivating structured evaluation of reconstruction quality under controlled scene capture conditions.

Radiance field representations, \eg{} NeRF~\citep{mildenhall_nerf_2022} and Gaussian Splatting~\citep{kerbl_3d_2023}, have found increasing use in asset reconstruction~\citep{tang2025dronesplat, jiang2025horizon, choi2025uavtwinneuraldigitaltwins, Dong2025DigitalTwin}.
Extensions targeting non-Lambertian surfaces~\citep{verbin2022refnerf, naylor2025surf, guo_nerfren_2022, ye_3d_2024, gu_irgs_2025, jiang2024gaussianshader} show particular promise for photorealistic reconstruction of challenging appearances.
However, these methods depend on upstream pose and geometry estimates from 3D reconstruction approaches, so improvements in reconstruction quality propagate directly into these representations~\citep{pan2024glomap}.

Across this literature, the reconstruction is either taken as given by downstream inspection tasks or evaluated only at scene level, for completeness or visual fidelity.
No benchmark measures how inspection-specific capture choices and surface appearance affect the local accuracy on which defect localisation depends.
Our benchmark addresses this by evaluating per-view depth alongside point clouds and camera poses under controlled variation of trajectory, imaging resolution, lighting, reflectivity and soiling.

\section{Asset inspection dataset}
\label{sec:AssetInspectDataset}
\begin{figure}
    \centering
    \includegraphics[width=\linewidth]{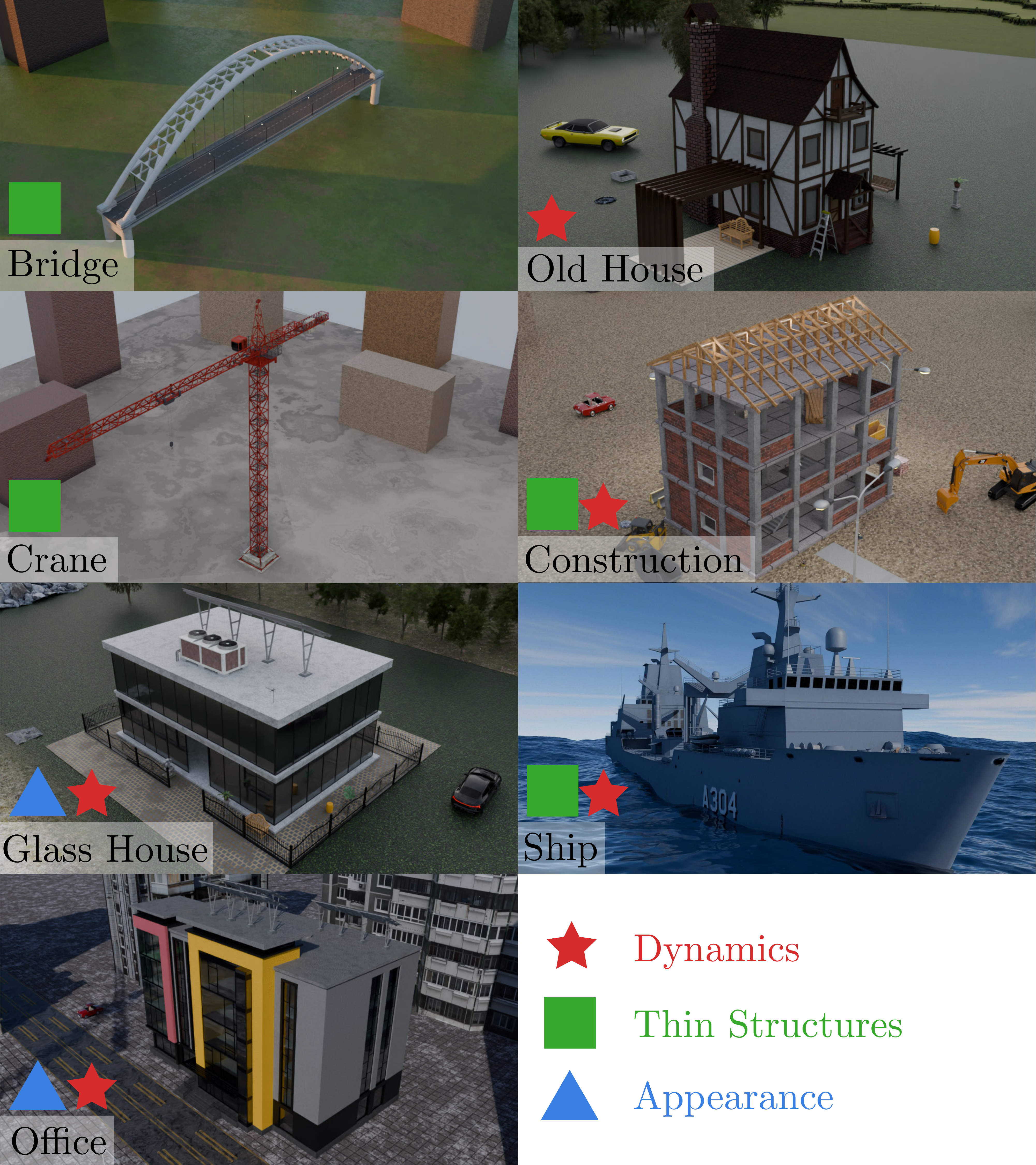}
    \caption{ \small
    An overview of our dataset scenes. Each scene simulates an asset under inspection under four different lighting conditions and four inspection trajectories. Five scenes contain dynamic objects simulating capture in unstructured conditions, and four contain thin structures which are common in urban environments. We vary the appearance of the windows in Glass House and Office by changing their reflectivity or level of surface soiling. }
    \label{fig:overview_figure}
\end{figure}

\begin{figure} 
    \centering
    \includegraphics[width=\linewidth]{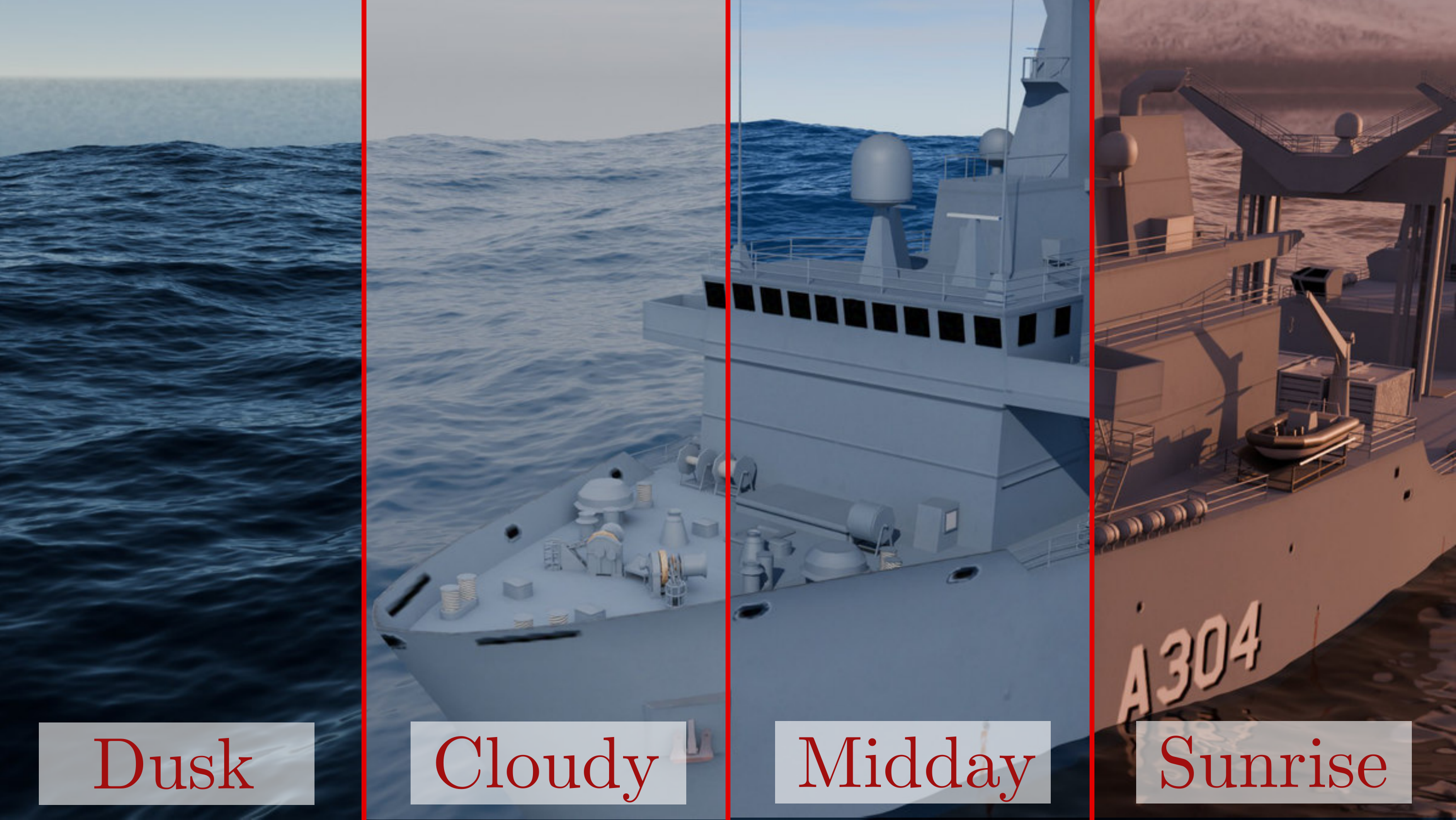}
    \caption{\small
    Lighting conditions on \textit{Ship}. Asset inspection operations encounter a range of lighting conditions, so we consider four lighting conditions across all scenes. By comparing reconstructions across the same poses, we can isolate the effect of lighting on downstream 3D reconstruction quality.}
    \label{fig:lighting}
\end{figure}

\begin{figure}
    \centering
    \includegraphics[width=\linewidth]{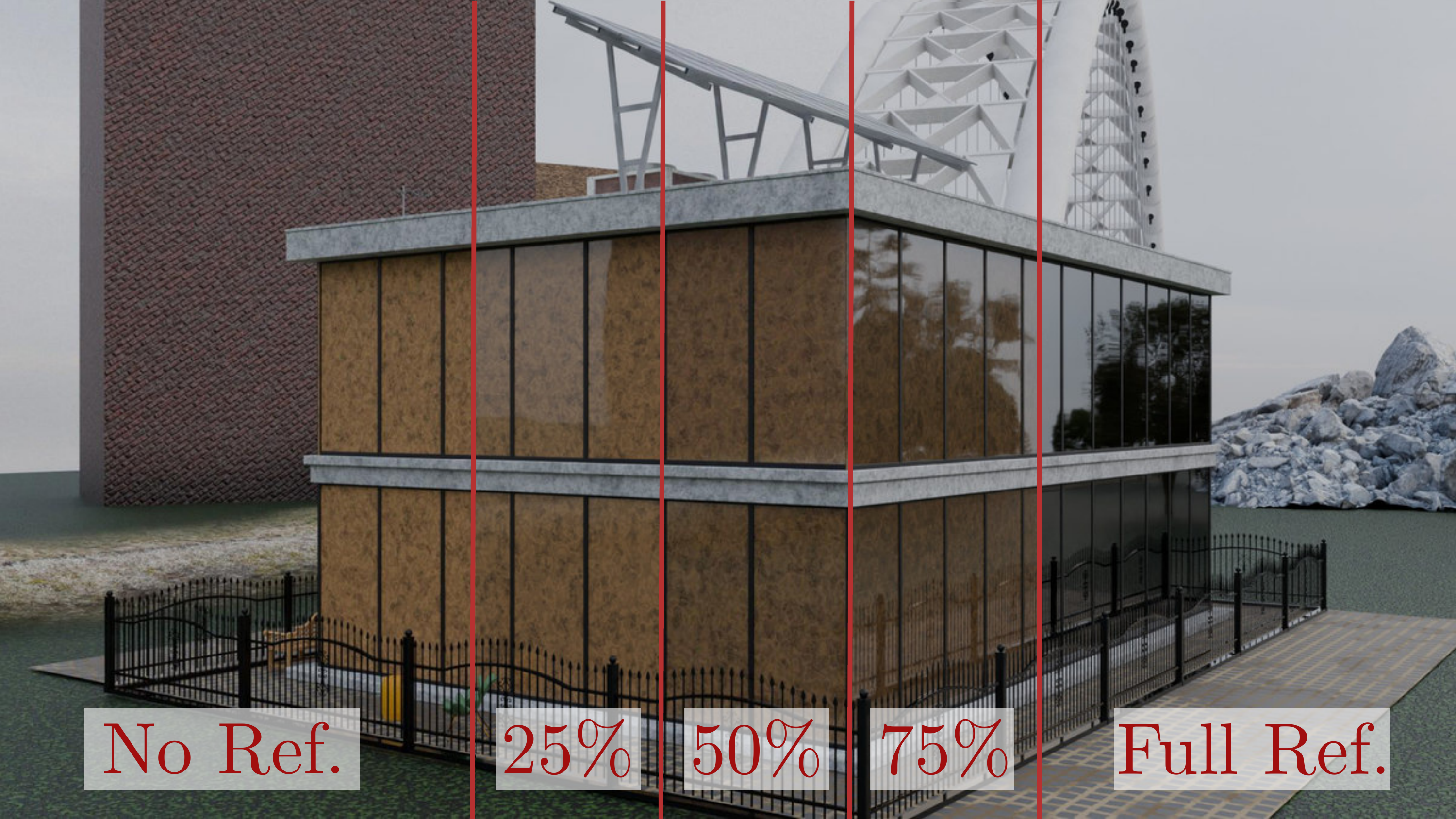}
    \caption{\small
    Reflectivity on \textit{Glass House}. We vary surface reflectivity across two scenes with windows. This allows one to compare reconstructions with different levels of reflectivity while holding all other variables constant.}
    \label{fig:reflectivity}
\end{figure}

\begin{figure} 
    \centering
    \includegraphics[width=\linewidth]{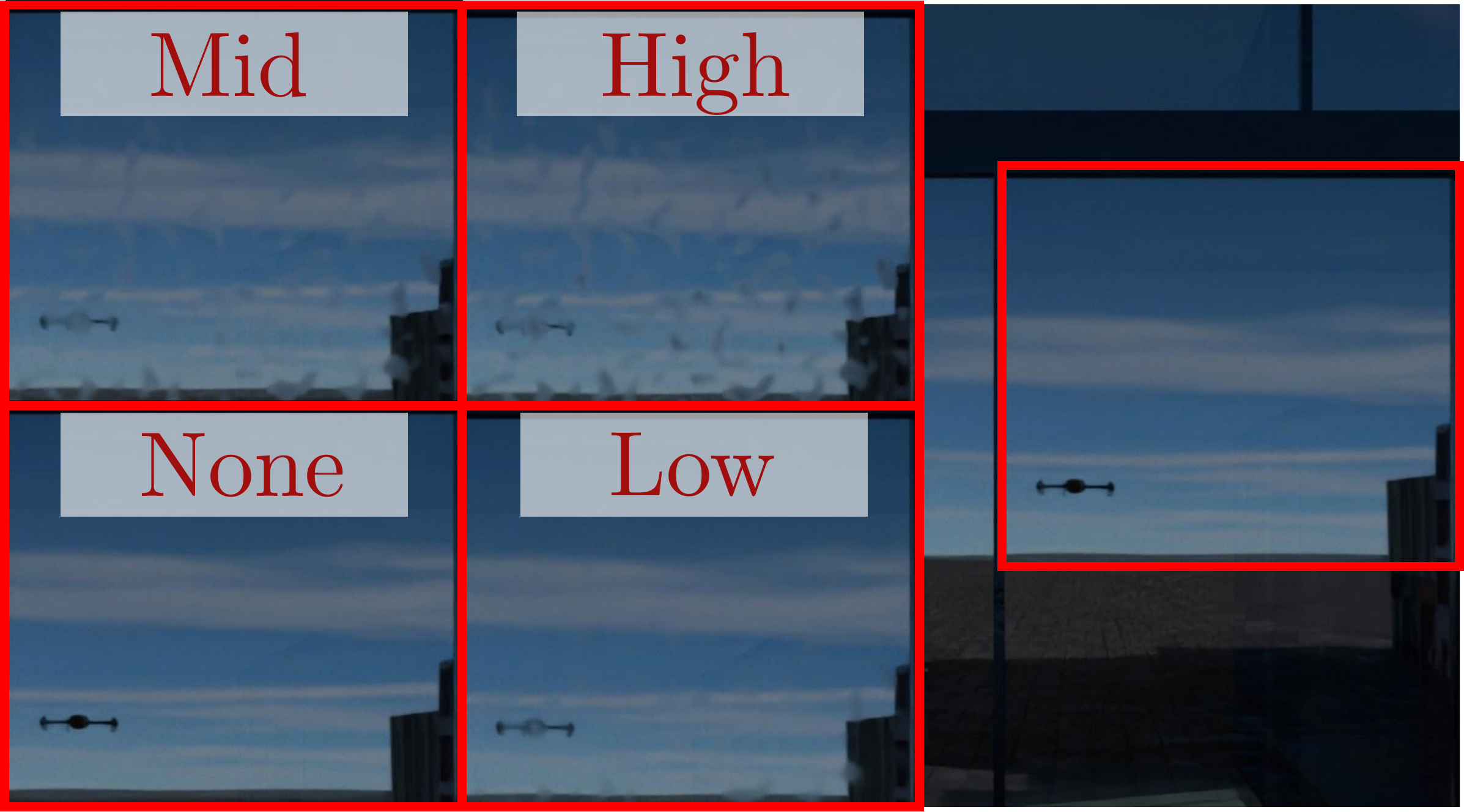}
    \caption{\small
    Surface soiling on \textit{Office}. We vary surface soiling across two scenes with windows. Comparing reconstructions across the same poses and lighting conditions isolates the effect of soiling on downstream 3D reconstruction quality.}
    \label{fig:soiling}
\end{figure}

We present a synthetic dataset designed to simulate drone imagery taken during asset inspection operations. 
Our dataset consists of 7 scenes, 
each containing an asset under inspection as shown in \Cref{fig:overview_figure}.
We rendered each scene under 4 lighting conditions (see \Cref{fig:lighting}) and from 4 different camera trajectories, yielding 16 lighting and trajectory combinations.
One capture pattern, \textit{manual}, mimics an asset-inspection style capture pattern, where camera poses are kept close to each object with around 75-80\% overlap between views~\citep{t2_capture_guide}.
For larger structures, such as \textit{Bridge}, \textit{Crane}, \textit{Office}, \textit{Ship}, only a small portion of each object is visible in each view.
Around corners, the camera smoothly transitions from one side to another as suggested in~\citep{Trendspek_capture_guide}.
For the other capture patterns, \textit{close}, \textit{medium} and \textit{far}, the cameras are pseudorandomly distributed at offsets from a proxy surface surrounding the asset, corresponding to different spatial resolutions on the asset's surface.
These poses are joined via a travelling salesperson algorithm to form a trajectory which visits each pose once. 
All images are rendered at $1920 \text{px} \times 1080 \text{px}$ resolution with an ideal pinhole camera (36mm sensor width, no distortion), using a 50mm lens (horizontal field of view $39.6^\circ$) for the \textit{manual} capture pattern.
For the proxy-surface capture patterns, a 35mm lens ($54.4^\circ$) was used on all scenes except \textit{Construction}, where a 50mm lens was used.
The wider-angle lens was used on most scenes to maintain coverage of the asset and avoid the camera trajectory intersecting with background objects.
\textit{Construction} had no such background objects for the camera to intersect with so a 50mm lens was used. 
\Cref{tab:gsd} reports, for each scene and capture pattern, the closest camera-to-asset distance and the corresponding finest \ac{GSD} in each frame, averaged across the trajectory.
At the rendered resolution, the \textit{manual} and \textit{close} patterns span 1.29--16.5mm/px.
As is, most trajectories yield a \ac{GSD} within the 1.5--10\,mm/px range recommended by industry guidelines~\citep{Trendspek_capture_guide, t2_capture_guide}.
However, the \textit{Bridge} and \textit{Ship} scenes have trajectories yielding a \ac{GSD} outside of this range.
We note that cameras used for asset inspection often have resolutions of 20\,MP or more, at $6000\text{px} \times 4000\text{px} $, with the same field of view yielding a \ac{GSD} about $3\times$ finer than what was rendered.
Once this is taken into account, all sequences meet the 1.5--10\,mm/px guideline.
For a given asset, the manual and close trajectories always have a smaller \ac{GSD} than the \textit{medium} and \textit{far} trajectories for the same asset.
\textit{Glass House} and \textit{Office} were rendered with 5 reflectivity settings (as shown in \Cref{fig:reflectivity}) and 4 soiling settings (as shown in \Cref{fig:soiling}). 
A breakdown of each sequence is shown in \Cref{tab:Scene_Details}.
By rendering scenes under multiple lighting conditions, asset inspection-style capture patterns, reflectivities, and surface conditions, we can perform a controlled evaluation of each 3D reconstruction method against these variables, which is not possible with previous datasets.
Furthermore, we can interrogate the priors of learning-based methods and determine where traditional \ac{SfM} methods are best applied.

Further details on the scenes, including scene descriptions and sources for the models, environment maps and textures are available in the supplementary information. 

\begin{figure}
    \centering
    \includegraphics[width=\linewidth]{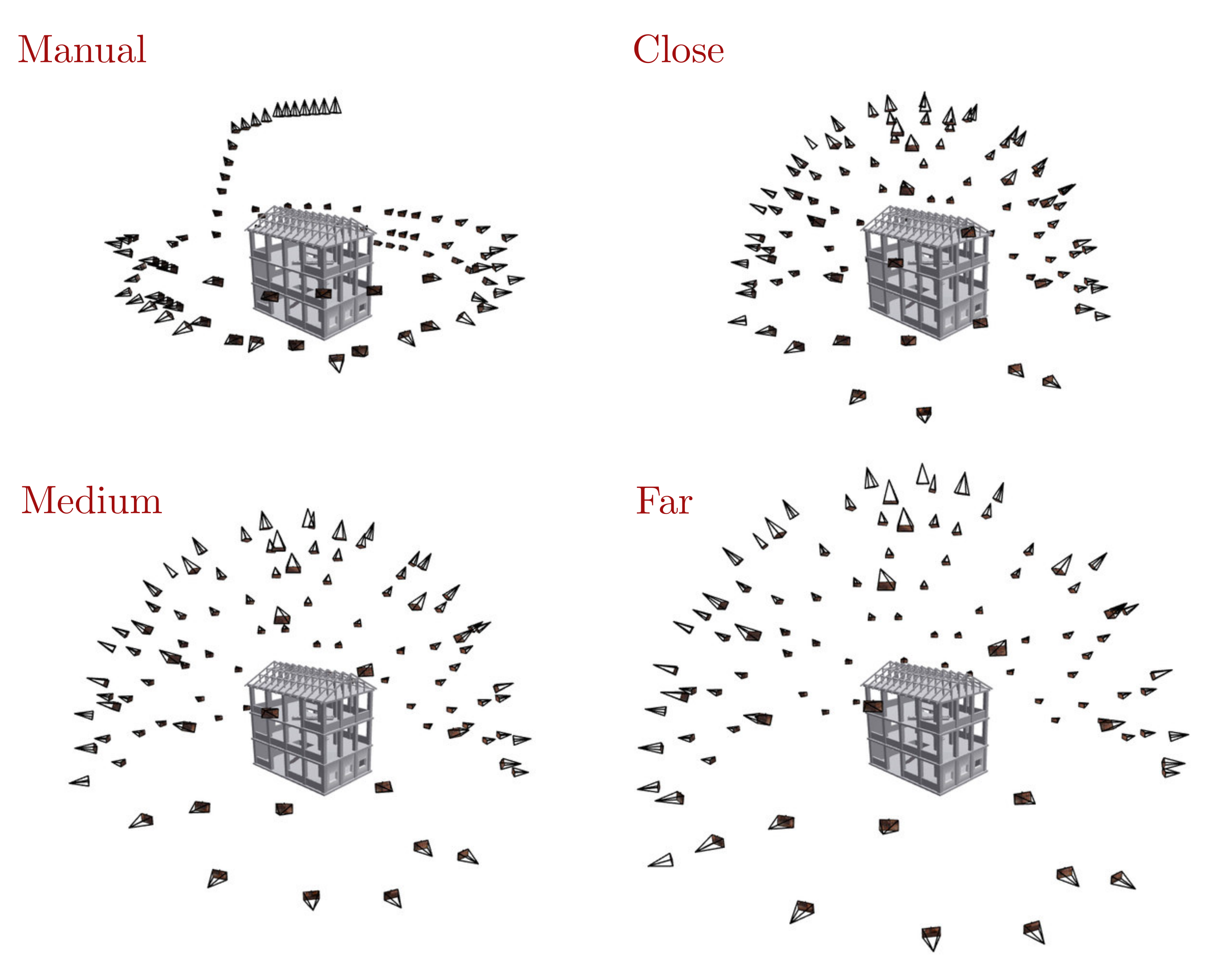}
    \caption{\small
    Asset inspection trajectories for the \textit{Construction} scene. In this work, we consider a manually derived trajectory for each scene that mimics the overlap and transition between frames of aerial platforms in inspection tasks, and three pseudorandom trajectories that mimic three imaging resolutions at close, medium and far range.}
    \label{fig:trajectories}
\end{figure}

\begin{table*}[t]
\centering
\caption{\small
For each scene, there are 4 capture patterns (\textit{manual}, \textit{close}, \textit{medium} and \textit{far}) and 4 lighting conditions (\textit{Dusk}, \textit{Cloudy}, \textit{Midday} and \textit{Sunrise}).
For the two scenes where the asset is reflective, there are 5 reflectivity and 4 soiling settings, yielding a total of 9 surface conditions.
In total, we have 368 sequences and 364k frames.
}
\footnotesize
\setlength{\tabcolsep}{4pt}
\begin{tabular}{l cccc cc c}
\toprule
& \multicolumn{4}{c}{\textbf{Capture Pattern Frames}}
& \multicolumn{2}{c}{\textbf{Surface Conditions}}
& \\
\cmidrule(lr){2-5}\cmidrule(lr){6-7}
\textbf{Scene}
& \textbf{Manual}
& \textbf{Close}
& \textbf{Medium}
& \textbf{Far}
& \textbf{Reflectivity}
& \textbf{Soiling}
& \textbf{Sequences} \\
\midrule
Bridge       & 1073 & 566  & 566  & 566  & 1 & 0 & 16  \\
Construction & 100  & 100  & 100  & 100  & 1 & 0 & 16  \\
Crane        & 642  & 518  & 518  & 518  & 1 & 0 & 16  \\
Glass House  & 100  & 100  & 100  & 96   & 5 & 4 & 144 \\
Office       & 4763 & 1311 & 1311 & 1311 & 5 & 4 & 144 \\
Old House    & 100  & 97   & 99   & 100  & 1 & 0 & 16  \\
Ship         & 1029 & 803  & 801  & 803  & 1 & 0 & 16  \\
\bottomrule
\end{tabular}
\label{tab:Scene_Details}
\end{table*}

\begin{table*}[t]
\centering
\caption{\small
Camera focal lengths, stand-off distances and imaging resolutions for each scene and capture pattern.
All images are rendered at $1920\text{px} \times 1080\text{px}$ with an ideal pinhole camera of 36mm sensor width; \textit{Proxy} denotes the lens shared by the \textit{close}, \textit{medium} and \textit{far} patterns.
For each frame we take the minimum depth, $d$ over asset pixels and the corresponding finest \ac{GSD} (mm/px), and report $d$ and \ac{GSD} as the means of these per-frame values across the trajectory.
}
\footnotesize
\setlength{\tabcolsep}{4pt}
\begin{tabular}{l cc cc cc cc cc}
\toprule
& \multicolumn{2}{c}{\textbf{Focal length (mm)}}
& \multicolumn{2}{c}{\textbf{Manual}}
& \multicolumn{2}{c}{\textbf{Close}}
& \multicolumn{2}{c}{\textbf{Medium}}
& \multicolumn{2}{c}{\textbf{Far}} \\
\cmidrule(lr){2-3}\cmidrule(lr){4-5}\cmidrule(lr){6-7}\cmidrule(lr){8-9}\cmidrule(lr){10-11}
\textbf{Scene} & \textbf{Manual} & \textbf{Proxy}
& $d$ (m) & GSD & $d$ (m) & GSD & $d$ (m) & GSD & $d$ (m) & GSD \\
\midrule
Bridge       & 50 & 36 & 3.32  & 1.29 & 12.08 & 6.71 & 16.87 & 8.52 & 21.85 & 10.9 \\
Construction & 50 & 50 & 14.01 & 5.17 & 13.45 & 4.91 & 16.98 & 6.26 & 20.80 & 7.66 \\
Crane        & 50 & 36 & 4.16  & 1.53 & 3.28  & 1.66 & 6.74  & 3.44 & 10.23 & 5.23 \\
Glass House  & 50 & 36 & 11.92 & 6.03 & 10.25 & 5.26 & 14.13 & 7.28 & 17.98 & 9.37 \\
Office       & 50 & 36 & 8.39  & 3.10 & 5.05  & 2.60 & 8.82  & 4.52 & 12.71 & 6.48 \\
Old House    & 50 & 36 & 13.33 & 4.90 & 8.86  & 4.58 & 11.17 & 5.83 & 13.45 & 7.07 \\
Ship         & 50 & 36 & 45.03 & 16.5 & 32.90 & 16.5 & 36.37 & 18.2 & 39.88 & 20.0 \\
\bottomrule
\end{tabular}
\label{tab:gsd}
\end{table*}

\subsection{Dataset production}
To create this dataset, we used Blender~\citep{blender} and its photorealistic physically-based path tracing renderer Cycles.
In addition, we employed several strategies to further reduce the sim2real gap: 
(1) We used procedurally generated textures where appropriate to avoid unrealistic repetition and self-similarity of tiled image textures, while maintaining the challenges of visually scene-content;
(2) To properly simulate the slightly warped reflections in real windows, we added low-frequency noise to the window normal maps;
(3) In scenes where the asset was reflective, we animated a drone to follow the camera trajectory so that reflections of the drone are visible from the correct poses;
(4) We simulated scene dynamics such as moving vehicles or waves in 5 of the scenes;
(5) We simulated realistic signal-dependent sensor noise to each image during post-processing, following~\citep{foi_clipped_2009}.

As part of our open source dataset and evaluation pipeline, we also provide our Blender scenes allowing others to modify and extend the dataset to custom trajectories, lighting conditions and more.

\subsection{Ground truth}
\label{sec:GroundTruth}
All ground truth is exported directly from Blender in metres, in a common world frame.
For each frame we export the camera-to-world pose, converted from Blender's camera convention to the OpenCV convention, together with the pinhole intrinsics.
Depth maps are rendered from Cycles at full resolution and stored in separate files from the RGB images, as \texttt{.exr} files containing planar depth along the optical axis. Pixels that see the sky have infinite depth and are excluded from all depth metrics.
Depth corresponds to the first surface intersected by each camera ray: on windows it is the depth of the glass pane itself, not of the reflected scene or of anything visible through the glass.
This is what an inspection requires, since defects lie on the pane, and it means that reconstructing the virtual scene behind a reflection is penalised. It also means that ground-truth depth is identical across all reflectivity and soiling settings, so differences in error across these settings are due to appearance alone.
Because depth is rendered per frame, it includes dynamic objects, such as vehicles and waves, at their positions in that frame.
For point cloud evaluation we export two static meshes per scene, excluding dynamic objects: an accuracy mesh containing all static geometry and a completeness mesh containing only the asset and its immediate surroundings, each sampled evenly along mesh surfaces to produce 1 million points.
Predicted points are scored against the accuracy point cloud, so that floaters anywhere in the scene count against accuracy, while completeness is measured only on the asset.
Per-frame asset and window masks are rendered using object and material index passes, and are used for the masked depth evaluation and the window-region analyses.

\section{Evaluation}
\label{sec:Evaluation}

\subsection{Task description and metrics}
\label{sec:Metrics}

This benchmark is designed to evaluate the performance of a 3D reconstruction method for asset inspection. 
For a digital twin to be useful for remotely inspecting an asset, the reconstruction must be accurate and detailed.
Accordingly, we assess point cloud metrics to evaluate the global accuracy of the reconstruction, as well as per-view depth estimation metrics to evaluate the local accuracy.
To ensure that the local accuracy in the depth maps can be connected to the overall global accuracy, we also evaluate the estimated camera poses. 

To evaluate point clouds, we use 
{Chamfer distance} as it is computed against the accuracy and completeness point clouds described in \Cref{sec:GroundTruth}.
To evaluate the depth maps, we use {\ac{sMSE}}, as per~\citep{arampatzakis_monocular_2024}, since inspecting an otherwise accurate model at an incorrect scale is still valuable.
However, since several approaches perform metric depth estimation, such as \ac{DA3}, we also report {AbsRel}.
Because accurate asset reconstruction matters more than the background, we present these metrics under two masking conditions. 
The first is unmasked, evaluated over the entirety of each frame, and the second is masked, where an asset mask is applied to each frame.
We use the standard measure of AUC@30 to evaluate the estimated camera poses, as it combines both the \ac{RTA} and \ac{RRA} scores.
Definitions of these metrics are included in \Cref{supp:metrics} of the supplementary information.

\subsection{Evaluated methods}
\label{sec: Evaluated Methods}

We focus on methods that perform a reconstruction of an asset from unposed images.
Often, drone imagery is taken in the form of videos, or as a series of stills in temporal order, making SLAM- and chunking-style approaches feasible. 
Therefore, in this work, we evaluate a representative sample of benchmark and \ac{SOTA} methods that perform this task.
These methods are divided into the following categories:
\begin{enumerate}
    \item {Structure from Motion}, including COLMAP~\citep{schoenberger2016sfm, schoenberger2016mvs} 
    VidMap~\citep{pataki2026vidmap}, 
    and GLOMAP~\citep{pan2024glomap}.
    \item {End to End Feed Forward}, including \ac{DA3}~\citep{depthanything3}, VGGT~\citep{wang_vggt_2025}, $\pi^3$ and $\pi^3$-\text{X}~\citep{wang2025pi}, 
    and VGGT-$\Omega$~\citep{wang2026vggtomega}.
    \item {Chunk-based} and {SLAM-based} adaptations of \ac{E2E} methods, including DA3-Streaming~\citep{da3streamingcode2025}, Pi-Long~\citep{pilongcode2025}, VGGT-Long~\citep{deng2025vggtlongchunkitloop} and VGGT-SLAM 2.0~\citep{maggio2025vggt-slam2}.
\end{enumerate}

Each method was run with its default parameters unless otherwise indicated in \Cref{supp:hyperparameters} of the supplementary information.

\subsection{Results}
\label{sec:Results}

\begin{table*}[t]
    \centering
    \caption{\small
    Chamfer Distance (m) $\downarrow$ averaged across each scene.  
        The best, second-best, and third-best results in each column are highlighted in
        \colorbox{rankfirst}{deep blue},
        \colorbox{ranksecond}{medium blue}, and
        \colorbox{rankthird}{light blue}, respectively.
            \ac{OOM} errors are highlighted in \colorbox{rankfail}{light red}.
            Within each method subcategory, the best-performing result is indicated in
            \textbf{bold}.
            Cons. refers to \textit{Construction} and H. refers to \textit{House}.
            The columns corresponding to the scene are sorted according to sequence length in descending order.
            The Avg. column reports the unweighted mean across all seven scenes.
    }
    \footnotesize
    \begin{tabular}{c c c c c c c c c}
    \hline
        \multirow{2}{*}{\textbf{Method}} & \multicolumn{7}{c}{\textbf{Scene}} & \multirow{2}{*}{\textbf{Avg.}} \\
         & \textbf{Office} & \textbf{Bridge} & \textbf{Ship} & \textbf{Crane} & \textbf{Old H.} & \textbf{Glass H.} & \textbf{Cons.} & \\
    \hline
        \multicolumn{9}{c}{\cellcolor{catgray}\textbf{Structure from Motion}} \\
        \hline
 COLMAP & 2.457 & 2.697 & 5.821 & 9.026 & 0.499 & 0.911 & 1.216 & 3.233 \\
 GLOMAP & 6.480 & 5.902 & 29.621 & 10.350 & 0.726 & 1.569 & 1.471 & 8.017 \\
 VidMap & \cellcolor{rankthird}\textbf{1.446} & \cellcolor{rankfirst}\textbf{1.413} & \textbf{4.025} & \cellcolor{rankthird}\textbf{5.079} & \textbf{0.325} & \textbf{0.461} & \cellcolor{rankthird}\textbf{0.281} & \cellcolor{rankthird}\textbf{1.861} \\
        \hline
        \multicolumn{9}{c}{\cellcolor{catgray}\textbf{End to End Feed-Forward}} \\
        \hline
 VGGT & \cellcolor{rankfail}OOM & \cellcolor{rankfail}OOM & \cellcolor{rankfail}OOM & 8.083 & 0.277 & 0.590 & 1.000 & -- \\
 DA3 & \cellcolor{rankfail}OOM & \cellcolor{rankfail}OOM & \cellcolor{rankfail}OOM & \cellcolor{rankfail}OOM & \cellcolor{rankfirst}\textbf{0.252} & \cellcolor{ranksecond}0.407 & \cellcolor{rankfirst}\textbf{0.242} & -- \\
 VGGT-$\Omega$ & \cellcolor{rankfail}OOM & \cellcolor{rankfail}OOM & \cellcolor{rankfail}OOM & \cellcolor{rankfail}OOM & \cellcolor{rankthird}0.263 & \cellcolor{rankfirst}\textbf{0.406} & 0.328 & -- \\
 $\pi^3$ & \cellcolor{rankfail}OOM & \cellcolor{rankfail}OOM & \cellcolor{rankfail}OOM & \cellcolor{rankfail}OOM & \cellcolor{ranksecond}0.259 & \cellcolor{rankthird}0.414 & 0.324 & -- \\
 $\pi^3$-\text{X} & \cellcolor{rankfail}OOM & \textbf{3.268} & \cellcolor{rankthird}\textbf{3.384} & \textbf{6.229} & 0.511 & 0.595 & 0.434 & -- \\
        \hline
        \multicolumn{9}{c}{\cellcolor{catgray}\textbf{Chunk and SLAM Based E2E}} \\
        \hline
 Pi-Long & \cellcolor{ranksecond}1.254 & \cellcolor{rankthird}1.884 & \cellcolor{ranksecond}3.085 & \cellcolor{ranksecond}5.075 & 0.266 & 0.421 & 0.291 & \cellcolor{ranksecond}1.754 \\
 DA3-Streaming & \cellcolor{rankfirst}\textbf{1.232} & \cellcolor{ranksecond}\textbf{1.490} & \cellcolor{rankfirst}\textbf{2.462} & \cellcolor{rankfirst}\textbf{3.774} & \cellcolor{ranksecond}\textbf{0.259} & \textbf{0.419} & \cellcolor{ranksecond}\textbf{0.256} & \cellcolor{rankfirst}\textbf{1.413} \\
 VGGT-Long & 1.646 & 2.333 & 8.357 & 5.763 & 0.310 & 1.052 & 0.752 & 2.887 \\
 VGGT-SLAM 2.0 & 2.953 & 3.490 & 6.527 & 5.773 & 0.300 & 1.289 & 1.200 & 3.076 \\
        \hline
    \end{tabular}
    \label{tab:Scene_Level_Results_Chamfer}
\end{table*}

\begin{table*}[t]
\centering
\caption{\small
Point cloud accuracy (Acc., m) $\downarrow$ and completeness (Comp., m) $\downarrow$ averaged across each scene.
}
\footnotesize
\begin{tabular}{c c c c c c c c c c}
\hline
\multirow{2}{*}{\textbf{Method}} &
\multirow{2}{*}{\textbf{Metric}} &
\multicolumn{7}{c}{\textbf{Scene}} & \multirow{2}{*}{\textbf{Avg.}} \\
& & \textbf{Office} & \textbf{Bridge} & \textbf{Ship} & \textbf{Crane} & \textbf{Old H.} & \textbf{Glass H.} & \textbf{Cons.} & \\
\hline
\multicolumn{10}{c}{\cellcolor{catgray}\textbf{Structure from Motion}} \\
\hline
\multirow{2}{*}{COLMAP}
  & Acc. & 1.792 & \cellcolor{rankthird}1.792 & 5.817 & 9.337 & 0.372 & 0.808 & 0.943 & 2.980 \\
  & Comp. & 3.123 & 3.602 & 5.825 & 8.716 & 0.626 & 1.014 & 1.489 & 3.485 \\
\hline
\multirow{2}{*}{GLOMAP}
  & Acc. & \cellcolor{rankfirst}\textbf{0.790} & 2.172 & 18.175 & \cellcolor{rankfirst}\textbf{4.634} & 0.669 & 1.262 & 1.063 & 4.109 \\
  & Comp. & 12.171 & 9.632 & 41.066 & 16.067 & 0.783 & 1.877 & 1.880 & 11.925 \\
\hline
\multirow{2}{*}{VidMap}
  & Acc. & \cellcolor{rankthird}1.291 & \cellcolor{ranksecond}\textbf{1.468} & \cellcolor{ranksecond}\textbf{3.722} & 6.140 & \textbf{0.332} & \textbf{0.385} & \cellcolor{rankthird}\textbf{0.331} & \cellcolor{ranksecond}\textbf{1.953} \\
  & Comp. & \textbf{1.601} & \cellcolor{rankfirst}\textbf{1.357} & \textbf{4.328} & \textbf{4.018} & \textbf{0.318} & \textbf{0.536} & \textbf{0.231} & \cellcolor{rankthird}\textbf{1.770} \\
\hline
\multicolumn{10}{c}{\cellcolor{catgray}\textbf{End to End Feed-Forward}} \\
\hline
\multirow{2}{*}{VGGT}
  & Acc. & \cellcolor{rankfail} & \cellcolor{rankfail} & \cellcolor{rankfail} & 9.529 & 0.302 & 0.727 & 1.417 & -- \\
  & Comp. & \cellcolor{rankfail}\multirow{-2}{*}{OOM} & \cellcolor{rankfail}\multirow{-2}{*}{OOM} & \cellcolor{rankfail}\multirow{-2}{*}{OOM} & 6.637 & 0.251 & \cellcolor{rankfirst}\textbf{0.454} & 0.583 & -- \\
\hline
\multirow{2}{*}{DA3}
  & Acc. & \cellcolor{rankfail} & \cellcolor{rankfail} & \cellcolor{rankfail} & \cellcolor{rankfail} & \cellcolor{rankfirst}\textbf{0.265} & \cellcolor{rankfirst}\textbf{0.304} & \cellcolor{rankfirst}\textbf{0.302} & -- \\
  & Comp. & \cellcolor{rankfail}\multirow{-2}{*}{OOM} & \cellcolor{rankfail}\multirow{-2}{*}{OOM} & \cellcolor{rankfail}\multirow{-2}{*}{OOM} & \cellcolor{rankfail}\multirow{-2}{*}{OOM} & \cellcolor{ranksecond}\textbf{0.238} & 0.509 & \cellcolor{ranksecond}\textbf{0.181} & -- \\
\hline
\multirow{2}{*}{VGGT-$\Omega$}
  & Acc. & \cellcolor{rankfail} & \cellcolor{rankfail} & \cellcolor{rankfail} & \cellcolor{rankfail} & \cellcolor{ranksecond}0.266 & \cellcolor{rankthird}0.325 & 0.373 & -- \\
  & Comp. & \cellcolor{rankfail}\multirow{-2}{*}{OOM} & \cellcolor{rankfail}\multirow{-2}{*}{OOM} & \cellcolor{rankfail}\multirow{-2}{*}{OOM} & \cellcolor{rankfail}\multirow{-2}{*}{OOM} & 0.260 & \cellcolor{rankthird}0.486 & 0.284 & -- \\
\hline
\multirow{2}{*}{$\pi^3$}
  & Acc. & \cellcolor{rankfail} & \cellcolor{rankfail} & \cellcolor{rankfail} & \cellcolor{rankfail} & \cellcolor{rankthird}0.271 & \cellcolor{ranksecond}0.324 & 0.455 & -- \\
  & Comp. & \cellcolor{rankfail}\multirow{-2}{*}{OOM} & \cellcolor{rankfail}\multirow{-2}{*}{OOM} & \cellcolor{rankfail}\multirow{-2}{*}{OOM} & \cellcolor{rankfail}\multirow{-2}{*}{OOM} & 0.246 & 0.504 & 0.194 & -- \\
\hline
\multirow{2}{*}{$\pi^3$-X}
  & Acc. &\cellcolor{rankfail}  & \textbf{4.330} & \textbf{4.879} & \textbf{8.555} & 0.472 & 0.624 & 0.545 & -- \\
  & Comp. & \cellcolor{rankfail} \multirow{-2}{*}{OOM} & \textbf{2.206} & \cellcolor{ranksecond}\textbf{1.890} & \textbf{3.903} & 0.549 & 0.566 & 0.324 & -- \\
\hline
\multicolumn{10}{c}{\cellcolor{catgray}\textbf{Chunk and SLAM Based E2E}} \\
\hline
\multirow{2}{*}{Pi-Long}
  & Acc. & 1.372 & 2.118 & \cellcolor{rankthird}4.014 & 8.613 & 0.293 & 0.372 & 0.426 & \cellcolor{rankthird}2.458 \\
  & Comp. & \cellcolor{rankfirst}\textbf{1.136} & \cellcolor{rankthird}1.651 & \cellcolor{rankthird}2.157 & \cellcolor{rankfirst}\textbf{1.538} & \cellcolor{ranksecond}0.238 & \cellcolor{ranksecond}\textbf{0.470} & \cellcolor{rankfirst}\textbf{0.156} & \cellcolor{rankfirst}\textbf{1.049} \\
\hline
\multirow{2}{*}{DA3-Streaming}
  & Acc. & \cellcolor{ranksecond}\textbf{0.991} & \cellcolor{rankfirst}\textbf{1.021} & \cellcolor{rankfirst}\textbf{3.676} & \cellcolor{ranksecond}\textbf{5.094} & \textbf{0.273} & \textbf{0.332} & \cellcolor{ranksecond}\textbf{0.330} & \cellcolor{rankfirst}\textbf{1.674} \\
  & Comp. & \cellcolor{rankthird}1.474 & 1.960 & \cellcolor{rankfirst}\textbf{1.247} & \cellcolor{ranksecond}2.454 & \cellcolor{rankthird}0.245 & 0.507 & \cellcolor{rankthird}0.183 & \cellcolor{ranksecond}1.153 \\
\hline
\multirow{2}{*}{VGGT-Long}
  & Acc. & 1.980 & 3.140 & 9.026 & 8.216 & 0.347 & 1.584 & 0.951 & 3.606 \\
  & Comp. & \cellcolor{ranksecond}1.311 & \cellcolor{ranksecond}\textbf{1.526} & 7.688 & \cellcolor{rankthird}3.309 & 0.273 & 0.521 & 0.553 & 2.169 \\
\hline
\multirow{2}{*}{VGGT-SLAM 2.0}
  & Acc. & 1.708 & 3.727 & 5.872 & \cellcolor{rankthird}5.373 & 0.388 & 1.641 & 1.254 & 2.852 \\
  & Comp. & 4.198 & 3.252 & 7.182 & 6.173 & \cellcolor{rankfirst}\textbf{0.213} & 0.938 & 1.146 & 3.300 \\
\hline
\end{tabular}
\label{tab:Scene_Level_Results_Accuracy_Completeness}
\end{table*}

\begin{table*}[t]
    \centering
    \caption{\small
        AUC@30 $\uparrow$ averaged across each scene.
            }
    \footnotesize
    \begin{tabular}{c c c c c c c c c}
    \hline
        \multirow{2}{*}{\textbf{Method}} & \multicolumn{7}{c}{\textbf{Scene}} & \multirow{2}{*}{\textbf{Avg.}} \\
         & \textbf{Office} & \textbf{Bridge} & \textbf{Ship} & \textbf{Crane} & \textbf{Old H.} & \textbf{Glass H.} & \textbf{Cons.} & \\
    \hline
        \multicolumn{9}{c}{\cellcolor{catgray}\textbf{Structure from Motion}} \\
        \hline
 COLMAP & \cellcolor{rankfirst}\textbf{0.778} & \cellcolor{ranksecond}0.726 & \cellcolor{rankthird}0.685 & \cellcolor{rankfirst}\textbf{0.312} & 0.971 & 0.928 & 0.763 & \cellcolor{ranksecond}0.737 \\
 GLOMAP & 0.153 & 0.486 & 0.198 & 0.018 & 0.853 & 0.774 & 0.684 & 0.452 \\
 VidMap & \cellcolor{rankthird}0.563 & \cellcolor{rankfirst}\textbf{0.752} & \cellcolor{rankfirst}\textbf{0.872} & 0.177 & \cellcolor{rankthird}\textbf{0.983} & \cellcolor{rankfirst}\textbf{0.993} & \cellcolor{rankfirst}\textbf{0.993} & \cellcolor{rankfirst}\textbf{0.762} \\
        \hline
        \multicolumn{9}{c}{\cellcolor{catgray}\textbf{End to End Feed-Forward}} \\
        \hline
 VGGT & \cellcolor{rankfail}OOM & \cellcolor{rankfail}OOM & \cellcolor{rankfail}OOM & 0.031 & 0.839 & 0.682 & 0.547 & -- \\
 DA3 & \cellcolor{rankfail}OOM & \cellcolor{rankfail}OOM & \cellcolor{rankfail}OOM & \cellcolor{rankfail}OOM & \cellcolor{rankfirst}\textbf{0.987} & \cellcolor{ranksecond}\textbf{0.969} & \cellcolor{ranksecond}\textbf{0.961} & -- \\
 VGGT-$\Omega$ & \cellcolor{rankfail}OOM & \cellcolor{rankfail}OOM & \cellcolor{rankfail}OOM & \cellcolor{rankfail}OOM & 0.942 & 0.920 & 0.850 & -- \\
 $\pi^3$ & \cellcolor{rankfail}OOM & \cellcolor{rankfail}OOM & \cellcolor{rankfail}OOM & \cellcolor{rankfail}OOM & 0.926 & 0.908 & 0.819 & -- \\
 $\pi^3$-\text{X} & \cellcolor{rankfail}OOM & \textbf{0.105} & \textbf{0.225} & \textbf{0.035} & 0.714 & 0.772 & 0.588 & -- \\
        \hline
        \multicolumn{9}{c}{\cellcolor{catgray}\textbf{Chunk and SLAM Based E2E}} \\
        \hline
 Pi-Long & 0.507 & 0.480 & 0.676 & \cellcolor{rankthird}0.197 & 0.933 & 0.920 & 0.864 & \cellcolor{rankthird}0.654 \\
 DA3-Streaming & \cellcolor{ranksecond}\textbf{0.625} & \cellcolor{rankthird}\textbf{0.677} & \cellcolor{ranksecond}\textbf{0.715} & \cellcolor{ranksecond}\textbf{0.237} & \cellcolor{ranksecond}\textbf{0.986} & \cellcolor{rankthird}\textbf{0.966} & \cellcolor{rankthird}\textbf{0.952} & \cellcolor{ranksecond}\textbf{0.737} \\
 VGGT-Long & 0.109 & 0.339 & 0.503 & 0.051 & 0.859 & 0.669 & 0.633 & 0.452 \\
 VGGT-SLAM 2.0 & 0.160 & 0.303 & 0.540 & 0.060 & 0.878 & 0.600 & 0.493 & 0.433 \\
        \hline
    \end{tabular}
    \label{tab:Scene_Level_Results_AUC}
\end{table*}

\begin{table*}[t]
    \centering
    \caption{\small
    Camera pose registration rate (\%) $\uparrow$ averaged across each scene.
    }
    \footnotesize
    \begin{tabular}{c c c c c c c c c}
    \hline
        \multirow{2}{*}{\textbf{Method}} & \multicolumn{7}{c}{\textbf{Scene}} & \multirow{2}{*}{\textbf{Avg.}} \\
         & \textbf{Office} & \textbf{Bridge} & \textbf{Ship} & \textbf{Crane} & \textbf{Old H.} & \textbf{Glass H.} & \textbf{Cons.} & \\
        \hline
        \multicolumn{9}{c}{\cellcolor{catgray}\textbf{Structure from Motion}} \\
        \hline
 COLMAP & 62.7\% & \cellcolor{rankthird}77.3\% & 80.0\% & 59.9\% & 75.1\% & 75.6\% & 88.5\% & 74.2\% \\
 GLOMAP & \cellcolor{ranksecond}\textbf{75.7\%} & \cellcolor{ranksecond}\textbf{96.8\%} & \cellcolor{ranksecond}\textbf{97.1\%} & \cellcolor{ranksecond}\textbf{88.1\%} & \cellcolor{ranksecond}\textbf{92.6\%} & \cellcolor{ranksecond}\textbf{98.9\%} & \cellcolor{ranksecond}\textbf{99.9\%} & \cellcolor{ranksecond}\textbf{92.7\%} \\
 VidMap & \cellcolor{rankthird}67.2\% & 65.0\% & 75.7\% & 77.9\% & 65.5\% & 88.3\% & 76.5\% & 73.7\% \\
        \hline
        \multicolumn{9}{c}{\cellcolor{catgray}\textbf{End to End Feed-Forward}} \\
        \hline
 VGGT & \cellcolor{rankfail}OOM & \cellcolor{rankfail}OOM & \cellcolor{rankfail}OOM & \cellcolor{rankfirst}\textbf{100.0\%} & \cellcolor{rankfirst}\textbf{100.0\%} & \cellcolor{rankfirst}\textbf{100.0\%} & \cellcolor{rankfirst}\textbf{100.0\%} & -- \\
 DA3 & \cellcolor{rankfail}OOM & \cellcolor{rankfail}OOM & \cellcolor{rankfail}OOM & \cellcolor{rankfail}OOM & \cellcolor{rankfirst}\textbf{100.0\%} & \cellcolor{rankfirst}\textbf{100.0\%} & \cellcolor{rankfirst}\textbf{100.0\%} & -- \\
 VGGT-$\Omega$ & \cellcolor{rankfail}OOM & \cellcolor{rankfail}OOM & \cellcolor{rankfail}OOM & \cellcolor{rankfail}OOM & \cellcolor{rankfirst}\textbf{100.0\%} & \cellcolor{rankfirst}\textbf{100.0\%} & \cellcolor{rankfirst}\textbf{100.0\%} & -- \\
 $\pi^3$ & \cellcolor{rankfail}OOM & \cellcolor{rankfail}OOM & \cellcolor{rankfail}OOM & \cellcolor{rankfail}OOM & \cellcolor{rankfirst}\textbf{100.0\%} & \cellcolor{rankfirst}\textbf{100.0\%} & \cellcolor{rankfirst}\textbf{100.0\%} & -- \\
 $\pi^3$-\text{X} & \cellcolor{rankfail}OOM & \cellcolor{rankfirst}\textbf{100.0\%} & \cellcolor{rankfirst}\textbf{100.0\%} & \cellcolor{rankfirst}\textbf{100.0\%} & \cellcolor{rankfirst}\textbf{100.0\%} & \cellcolor{rankfirst}\textbf{100.0\%} & \cellcolor{rankfirst}\textbf{100.0\%} & -- \\
        \hline
        \multicolumn{9}{c}{\cellcolor{catgray}\textbf{Chunk and SLAM Based E2E}} \\
        \hline
 Pi-Long & \cellcolor{rankfirst}\textbf{100.0\%} & \cellcolor{rankfirst}\textbf{100.0\%} & \cellcolor{rankfirst}\textbf{100.0\%} & \cellcolor{rankfirst}\textbf{100.0\%} & \cellcolor{rankfirst}\textbf{100.0\%} & \cellcolor{rankfirst}\textbf{100.0\%} & \cellcolor{rankfirst}\textbf{100.0\%} & \cellcolor{rankfirst}\textbf{100.0\%} \\
 DA3-Streaming & \cellcolor{rankfirst}\textbf{100.0\%} & \cellcolor{rankfirst}\textbf{100.0\%} & \cellcolor{rankfirst}\textbf{100.0\%} & \cellcolor{rankfirst}\textbf{100.0\%} & \cellcolor{rankfirst}\textbf{100.0\%} & \cellcolor{rankfirst}\textbf{100.0\%} & \cellcolor{rankfirst}\textbf{100.0\%} & \cellcolor{rankfirst}\textbf{100.0\%} \\
 VGGT-Long & \cellcolor{rankfirst}\textbf{100.0\%} & \cellcolor{rankfirst}\textbf{100.0\%} & \cellcolor{rankfirst}\textbf{100.0\%} & \cellcolor{rankfirst}\textbf{100.0\%} & \cellcolor{rankfirst}\textbf{100.0\%} & \cellcolor{rankfirst}\textbf{100.0\%} & \cellcolor{rankfirst}\textbf{100.0\%} & \cellcolor{rankfirst}\textbf{100.0\%} \\
 VGGT-SLAM 2.0 & 60.9\% & 73.5\% & \cellcolor{rankthird}95.7\% & \cellcolor{rankthird}87.3\% & \cellcolor{rankthird}90.5\% & \cellcolor{rankthird}90.5\% & \cellcolor{rankthird}94.5\% & \cellcolor{rankthird}84.7\% \\
        \hline
    \end{tabular}
    \label{tab:Scene_Level_Results_Registration}
\end{table*}

\begin{table*}[t]
    \centering
    \caption{\small
    sMSE $\downarrow$ averaged across each scene.
    }
    \footnotesize
    \begin{tabular}{c c c c c c c c c c}
    \hline
    \multirow{2}{*}{\textbf{Method}} & \multirow{2}{*}{\textbf{Asset Mask}} & \multicolumn{7}{c}{\textbf{Scene}} & \multirow{2}{*}{\textbf{Avg.}} \\
     & & \textbf{Office} & \textbf{Bridge} & \textbf{Ship} & \textbf{Crane} & \textbf{Old H.} & \textbf{Glass H.} & \textbf{Cons.} & \\
    \hline
    \multicolumn{10}{c}{\cellcolor{catgray}\textbf{Structure from Motion}} \\
    \hline
 \multirow{2}{*}{COLMAP} & Unmasked & 0.231 & 0.452 & 0.176 & 1.159 & 0.136 & 0.206 & 0.251 & 0.373 \\
  & Masked & 0.209 & 0.401 & 0.089 & 1.217 & 0.072 & 0.167 & 0.200 & 0.336 \\
    \hline
 \multirow{2}{*}{GLOMAP} & Unmasked & 0.173 & 0.367 & 0.218 & 0.844 & 0.105 & 0.148 & 0.103 & 0.280 \\
  & Masked & 0.155 & 0.239 & 0.065 & 0.849 & 0.038 & 0.097 & 0.058 & 0.214 \\
    \hline
 \multirow{2}{*}{VidMap} & Unmasked & \textbf{0.039} & \textbf{0.097} & \cellcolor{rankthird}\textbf{0.021} & \textbf{0.615} & \textbf{0.030} & \textbf{0.031} & \textbf{0.017} & \textbf{0.121} \\
  & Masked & \cellcolor{ranksecond}\textbf{0.015} & \textbf{0.080} & \cellcolor{rankthird}\textbf{0.017} & \textbf{0.345} & \textbf{0.011} & \textbf{0.014} & \textbf{0.005} & \textbf{0.070} \\
    \hline
    \multicolumn{10}{c}{\cellcolor{catgray}\textbf{End to End Feed-Forward}} \\
    \hline
 \multirow{2}{*}{VGGT} & Unmasked & \cellcolor{rankfail} & \cellcolor{rankfail} & \cellcolor{rankfail} & 0.488 & 0.010 & 0.017 & 0.011 & -- \\
  & Masked & \cellcolor{rankfail}\multirow{-2}{*}{OOM} & \cellcolor{rankfail}\multirow{-2}{*}{OOM} & \cellcolor{rankfail}\multirow{-2}{*}{OOM} & 0.291 & \cellcolor{ranksecond}0.003 & \cellcolor{rankthird}0.005 & \cellcolor{ranksecond}0.003 & -- \\
    \hline
 \multirow{2}{*}{DA3} & Unmasked & \cellcolor{rankfail} & \cellcolor{rankfail} & \cellcolor{rankfail} & \cellcolor{rankfail} & \cellcolor{rankthird}0.007 & 0.014 & \cellcolor{rankfirst}\textbf{0.005} & -- \\
  & Masked & \cellcolor{rankfail}\multirow{-2}{*}{OOM} & \cellcolor{rankfail}\multirow{-2}{*}{OOM} & \cellcolor{rankfail}\multirow{-2}{*}{OOM} & \cellcolor{rankfail}\multirow{-2}{*}{OOM} & 0.005 & 0.007 & \cellcolor{rankthird}0.004 & -- \\
    \hline
 \multirow{2}{*}{VGGT-$\Omega$} & Unmasked & \cellcolor{rankfail} & \cellcolor{rankfail} & \cellcolor{rankfail} & \cellcolor{rankfail} & \cellcolor{rankfirst}\textbf{0.004} & \cellcolor{ranksecond}0.008 & \cellcolor{rankfirst}\textbf{0.005} & -- \\
  & Masked & \cellcolor{rankfail}\multirow{-2}{*}{OOM} & \cellcolor{rankfail}\multirow{-2}{*}{OOM} & \cellcolor{rankfail}\multirow{-2}{*}{OOM} & \cellcolor{rankfail}\multirow{-2}{*}{OOM} & \cellcolor{rankfirst}\textbf{0.002} & \cellcolor{ranksecond}0.003 & \cellcolor{rankfirst}\textbf{0.002} & -- \\
    \hline
 \multirow{2}{*}{$\pi^3$} & Unmasked & \cellcolor{rankfail} & \cellcolor{rankfail} & \cellcolor{rankfail} & \cellcolor{rankfail} & \cellcolor{rankfirst}\textbf{0.004} & \cellcolor{ranksecond}0.008 & \cellcolor{rankfirst}\textbf{0.005} & -- \\
  & Masked & \cellcolor{rankfail}\multirow{-2}{*}{OOM} & \cellcolor{rankfail}\multirow{-2}{*}{OOM} & \cellcolor{rankfail}\multirow{-2}{*}{OOM} & \cellcolor{rankfail}\multirow{-2}{*}{OOM} & \cellcolor{rankfirst}\textbf{0.002} & \cellcolor{rankfirst}\textbf{0.002} & \cellcolor{rankfirst}\textbf{0.002} & -- \\
    \hline
 \multirow{2}{*}{$\pi^3$-X} & Unmasked & \cellcolor{rankfail} & \cellcolor{rankfirst}\textbf{0.036} & \cellcolor{rankfirst}\textbf{0.008} & \cellcolor{rankfirst}\textbf{0.240} & \cellcolor{rankthird}0.007 & \cellcolor{rankfirst}\textbf{0.007} & \cellcolor{rankfirst}\textbf{0.005} & -- \\
  & Masked & \cellcolor{rankfail}\multirow{-2}{*}{OOM} & \cellcolor{rankfirst}\textbf{0.026} & \cellcolor{rankfirst}\textbf{0.004} & \cellcolor{rankfirst}\textbf{0.193} & \cellcolor{rankfirst}\textbf{0.002} & \cellcolor{ranksecond}0.003 & \cellcolor{rankfirst}\textbf{0.002} & -- \\
    \hline
    \multicolumn{10}{c}{\cellcolor{catgray}\textbf{Chunk and SLAM Based E2E}} \\
    \hline
 \multirow{2}{*}{Pi-Long} & Unmasked & \cellcolor{ranksecond}0.018 & \cellcolor{rankthird}0.040 & \cellcolor{ranksecond}\textbf{0.010} & \cellcolor{ranksecond}\textbf{0.267} & \cellcolor{ranksecond}\textbf{0.006} & \cellcolor{rankthird}\textbf{0.009} & \cellcolor{ranksecond}0.006 & \cellcolor{rankfirst}\textbf{0.051} \\
  & Masked & \cellcolor{ranksecond}0.015 & \cellcolor{rankfirst}\textbf{0.026} & \cellcolor{ranksecond}\textbf{0.005} & \cellcolor{ranksecond}\textbf{0.202} & \cellcolor{ranksecond}\textbf{0.003} & \cellcolor{ranksecond}\textbf{0.003} & \cellcolor{ranksecond}\textbf{0.003} & \cellcolor{rankfirst}\textbf{0.036} \\
    \hline
 \multirow{2}{*}{DA3-Streaming} & Unmasked & \cellcolor{rankfirst}\textbf{0.010} & \cellcolor{ranksecond}\textbf{0.038} & 0.035 & \cellcolor{rankthird}0.339 & 0.009 & 0.015 & \cellcolor{rankfirst}\textbf{0.005} & \cellcolor{ranksecond}0.064 \\
  & Masked & \cellcolor{rankfirst}\textbf{0.006} & 0.050 & 0.019 & 0.313 & 0.005 & 0.007 & \cellcolor{rankthird}0.004 & 0.058 \\
    \hline
 \multirow{2}{*}{VGGT-Long} & Unmasked & \cellcolor{rankthird}0.038 & 0.058 & 0.085 & 0.475 & 0.012 & 0.017 & 0.012 & 0.100 \\
  & Masked & \cellcolor{rankthird}0.027 & \cellcolor{ranksecond}0.046 & 0.022 & 0.284 & \cellcolor{rankthird}0.004 & \cellcolor{rankthird}0.005 & \cellcolor{ranksecond}\textbf{0.003} & \cellcolor{rankthird}0.056 \\
    \hline
 \multirow{2}{*}{VGGT-SLAM 2.0} & Unmasked & 0.042 & 0.059 & 0.083 & 0.459 & 0.016 & 0.020 & \cellcolor{rankthird}0.010 & \cellcolor{rankthird}0.098 \\
  & Masked & 0.031 & \cellcolor{rankthird}0.047 & 0.021 & \cellcolor{rankthird}0.275 & \cellcolor{rankthird}0.004 & \cellcolor{rankthird}0.005 & \cellcolor{ranksecond}\textbf{0.003} & \cellcolor{ranksecond}0.055 \\
    \hline
    \end{tabular}
    \label{tab:DepthResults_sMSE}
\end{table*}

\begin{table*}[t]
    \centering
    \caption{\small
    AbsRel $\downarrow$ averaged across each scene.
    }
    \footnotesize
    \begin{tabular}{c c c c c c c c c c}
    \hline
    \multirow{2}{*}{\textbf{Method}} & \multirow{2}{*}{\textbf{Asset Mask}} & \multicolumn{7}{c}{\textbf{Scene}} & \multirow{2}{*}{\textbf{Avg.}} \\
     & & \textbf{Office} & \textbf{Bridge} & \textbf{Ship} & \textbf{Crane} & \textbf{Old H.} & \textbf{Glass H.} & \textbf{Cons.} & \\
    \hline
    \multicolumn{10}{c}{\cellcolor{catgray}\textbf{Structure from Motion}} \\
    \hline
 \multirow{2}{*}{COLMAP} & Unmasked & 0.722 & 1.195 & 0.353 & 1.799 & 0.219 & 0.374 & 0.525 & 0.741 \\
  & Masked & 0.853 & 1.085 & 0.271 & 3.895 & 0.143 & 0.372 & 0.458 & 1.011 \\
    \hline
 \multirow{2}{*}{GLOMAP} & Unmasked & 0.930 & 0.884 & 0.837 & 1.023 & 0.221 & 0.333 & 0.325 & 0.650 \\
  & Masked & 0.929 & 0.728 & 0.820 & 1.082 & 0.166 & 0.328 & 0.294 & 0.621 \\
    \hline
 \multirow{2}{*}{VidMap} & Unmasked & \cellcolor{rankthird}\textbf{0.312} & \cellcolor{ranksecond}\textbf{0.363} & \cellcolor{rankthird}\textbf{0.249} & \textbf{0.869} & \textbf{0.059} & \textbf{0.076} & \textbf{0.046} & \cellcolor{rankthird}\textbf{0.282} \\
  & Masked & \cellcolor{rankthird}\textbf{0.301} & \cellcolor{rankthird}\textbf{0.450} & \textbf{0.251} & \textbf{0.878} & \textbf{0.045} & \textbf{0.050} & \cellcolor{rankthird}\textbf{0.028} & \cellcolor{rankthird}\textbf{0.286} \\
    \hline
    \multicolumn{10}{c}{\cellcolor{catgray}\textbf{End to End Feed-Forward}} \\
    \hline
 \multirow{2}{*}{VGGT} & Unmasked & \cellcolor{rankfail} & \cellcolor{rankfail} & \cellcolor{rankfail} & \cellcolor{ranksecond}\textbf{0.662} & 0.042 & 0.079 & 0.104 & -- \\
  & Masked & \cellcolor{rankfail}\multirow{-2}{*}{OOM} & \cellcolor{rankfail}\multirow{-2}{*}{OOM} & \cellcolor{rankfail}\multirow{-2}{*}{OOM} & \cellcolor{rankthird}\textbf{0.705} & 0.035 & 0.067 & 0.105 & -- \\
    \hline
 \multirow{2}{*}{DA3} & Unmasked & \cellcolor{rankfail} & \cellcolor{rankfail} & \cellcolor{rankfail} & \cellcolor{rankfail} & \cellcolor{ranksecond}0.022 & \cellcolor{ranksecond}0.030 & \cellcolor{rankfirst}\textbf{0.018} & -- \\
  & Masked & \cellcolor{rankfail}\multirow{-2}{*}{OOM} & \cellcolor{rankfail}\multirow{-2}{*}{OOM} & \cellcolor{rankfail}\multirow{-2}{*}{OOM} & \cellcolor{rankfail}\multirow{-2}{*}{OOM} & \cellcolor{ranksecond}0.018 & \cellcolor{rankfirst}\textbf{0.021} & \cellcolor{rankfirst}\textbf{0.017} & -- \\
    \hline
 \multirow{2}{*}{VGGT-$\Omega$} & Unmasked & \cellcolor{rankfail} & \cellcolor{rankfail} & \cellcolor{rankfail} & \cellcolor{rankfail} & \cellcolor{rankfirst}\textbf{0.019} & \cellcolor{rankfirst}\textbf{0.028} & \cellcolor{rankthird}0.031 & -- \\
  & Masked & \cellcolor{rankfail}\multirow{-2}{*}{OOM} & \cellcolor{rankfail}\multirow{-2}{*}{OOM} & \cellcolor{rankfail}\multirow{-2}{*}{OOM} & \cellcolor{rankfail}\multirow{-2}{*}{OOM} & \cellcolor{rankfirst}\textbf{0.015} & \cellcolor{rankthird}0.023 & 0.030 & -- \\
    \hline
 \multirow{2}{*}{$\pi^3$} & Unmasked & \cellcolor{rankfail} & \cellcolor{rankfail} & \cellcolor{rankfail} & \cellcolor{rankfail} & \cellcolor{rankthird}0.024 & \cellcolor{rankfirst}\textbf{0.028} & 0.039 & -- \\
  & Masked & \cellcolor{rankfail}\multirow{-2}{*}{OOM} & \cellcolor{rankfail}\multirow{-2}{*}{OOM} & \cellcolor{rankfail}\multirow{-2}{*}{OOM} & \cellcolor{rankfail}\multirow{-2}{*}{OOM} & \cellcolor{rankthird}0.019 & \cellcolor{ranksecond}0.022 & 0.037 & -- \\
    \hline
 \multirow{2}{*}{$\pi^3$-X} & Unmasked & \cellcolor{rankfail} & \textbf{1.099} & \textbf{0.473} & 1.036 & 0.256 & 0.155 & 0.103 & -- \\
  & Masked & \cellcolor{rankfail}\multirow{-2}{*}{OOM} & \textbf{1.238} & \textbf{0.471} & 1.454 & 0.256 & 0.156 & 0.102 & -- \\
    \hline
    \multicolumn{10}{c}{\cellcolor{catgray}\textbf{Chunk and SLAM Based E2E}} \\
    \hline
 \multirow{2}{*}{Pi-Long} & Unmasked & \cellcolor{ranksecond}0.299 & \cellcolor{rankthird}0.411 & \cellcolor{rankfirst}\textbf{0.214} & \cellcolor{rankfirst}\textbf{0.487} & 0.056 & 0.086 & 0.074 & \cellcolor{ranksecond}0.233 \\
  & Masked & \cellcolor{ranksecond}0.298 & \cellcolor{ranksecond}0.412 & \cellcolor{ranksecond}0.205 & \cellcolor{ranksecond}0.545 & 0.050 & 0.082 & 0.072 & \cellcolor{rankfirst}\textbf{0.237} \\
    \hline
 \multirow{2}{*}{DA3-Streaming} & Unmasked & \cellcolor{rankfirst}\textbf{0.173} & \cellcolor{rankfirst}\textbf{0.237} & \cellcolor{ranksecond}0.221 & 0.798 & \cellcolor{ranksecond}\textbf{0.022} & \cellcolor{rankthird}\textbf{0.032} & \cellcolor{ranksecond}\textbf{0.019} & \cellcolor{rankfirst}\textbf{0.215} \\
  & Masked & \cellcolor{rankfirst}\textbf{0.172} & \cellcolor{rankfirst}\textbf{0.268} & \cellcolor{rankfirst}\textbf{0.164} & 1.163 & \cellcolor{ranksecond}\textbf{0.018} & \cellcolor{ranksecond}\textbf{0.022} & \cellcolor{ranksecond}\textbf{0.018} & \cellcolor{ranksecond}0.261 \\
    \hline
 \multirow{2}{*}{VGGT-Long} & Unmasked & 1.165 & 0.459 & 0.306 & 0.868 & 0.072 & 0.149 & 0.141 & 0.451 \\
  & Masked & 1.184 & 0.472 & \cellcolor{rankthird}0.230 & 0.995 & 0.062 & 0.133 & 0.136 & 0.459 \\
    \hline
 \multirow{2}{*}{VGGT-SLAM 2.0} & Unmasked & 0.728 & 0.488 & 0.378 & \cellcolor{rankthird}0.681 & 0.121 & 0.182 & 0.235 & 0.402 \\
  & Masked & 0.733 & 0.479 & 0.307 & \cellcolor{rankfirst}\textbf{0.542} & 0.120 & 0.160 & 0.229 & 0.367 \\
    \hline
    \end{tabular}
    \label{tab:DepthResults_AbsRel}
\end{table*}

The \ac{SfM} and the \ac{E2E} chunking and \ac{SLAM} approaches were able to produce reconstructions of every sequence in this dataset.
However, the pure \ac{E2E} models would produce \ac{OOM} errors on long sequences.
Given the same hardware, \ac{DA3}, $\pi^3$ and VGGT-$\Omega$ were unable to reconstruct sequences of 518 frames or longer.
VGGT was able to reconstruct sequences up to 642 in length, but was unable to reconstruct sequences of 801 images or longer.
$\pi^3$-X was able to reconstruct sequences of up to sequences of up to 1073 frames in length, but was unable to reconstruct sequences of 1311 frames or longer.

Overall, the best learning-based methods produce higher-quality point clouds than the \ac{SfM} methods, as shown in \Cref{tab:Scene_Level_Results_Chamfer}. 
DA3-Streaming achieves the lowest chamfer distance, and
Pi-Long produces the second lowest chamfer distance.
However, VidMap produces the third smallest chamfer distance and is competitive with Pi-Long.
As expected, in \Cref{tab:Scene_Level_Results_Accuracy_Completeness} we see that this difference in performance between these different approaches is more apparent in the completeness metric. 
\ac{SfM} methods have relatively competitive accuracies; however, the learning-based methods significantly outperform them in the completeness metrics.
It is also worth noting that VidMap particularly is primarily aimed at estimating camera trajectories accurately, rather than producing complete and accurate point clouds, which may partially explain its relatively sparse point clouds.

On the other hand, COLMAP and VidMap produce more accurate camera poses (see \Cref{tab:Scene_Level_Results_AUC}), albeit with a lower camera registration rate, since learning-based methods estimate the pose of each view (see \Cref{tab:Scene_Level_Results_Registration}). 
However, since these methods can discard uncertain frames, the lower registration rate may in fact bias the camera pose accuracy metrics towards these approaches.
This suggests that video-based priors used by COLMAP's video mode and VidMap are highly beneficial in an asset inspection context.
This is supported by the fact that GLOMAP, which was not set to use video mode, does not perform well at pose estimation. 
We note that DA3-Streaming produces the joint second-most accurate set of camera poses,
 suggesting that the best learning-based methods can be competitive with COLMAP and VidMap in terms of camera pose estimation.

Learning-based methods substantially outperform \ac{SfM} methods in terms of depth map accuracy.
This holds for both the scale-invariant metric, sMSE (shown in \Cref{tab:DepthResults_sMSE}) and the absolute metric, AbsRel (shown in \Cref{tab:DepthResults_AbsRel}) and for both the unmasked and masked evaluations.
This shows that scene-level performance is distinct from local performance, and therefore requires specific evaluation in an asset inspection context.

Interestingly, the relative performance of methods changes significantly when comparing the scale-invariant sMSE results and the metric AbsRel results.
VidMap's performance improves to third, and DA3-Streaming's results improve from second and fourth to first and second for unmasked and masked evaluations, respectively. 
Note that we set VidMap to use \ac{DA3} as a monocular depth estimation stage.
This discrepancy between the scale-invariant results and the metric results is likely because
\ac{DA3} was directly trained for metric depth estimation, whereas $\pi^3$ and VGGT were not.

\bmhead{Effect of camera trajectory}

\begin{figure} 
    \centering
        \centering
        \includegraphics[width=\linewidth]{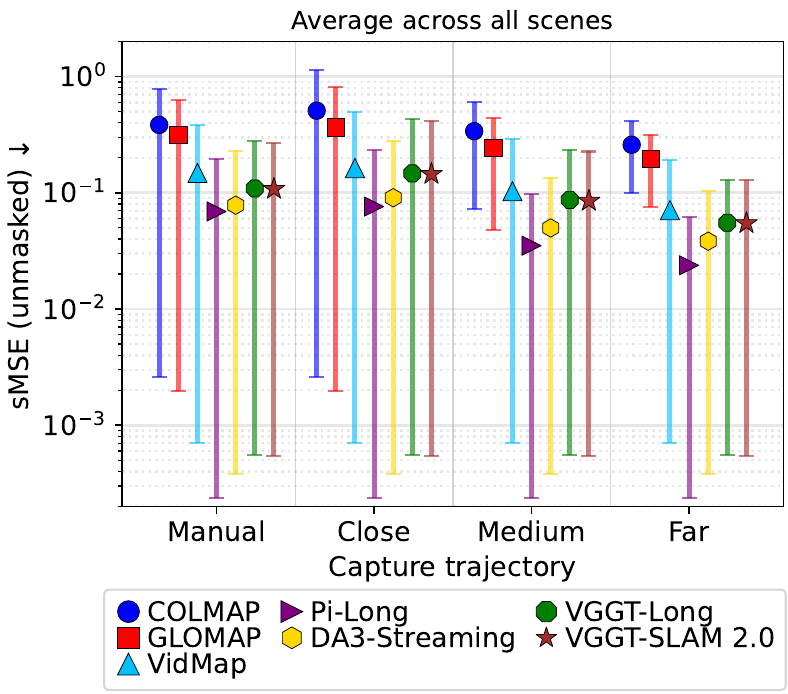}
        \caption{\small
        Average sMSE vs. trajectory across all sequences in this dataset.
        The error bars represent one standard deviation from the mean.
        }
        \label{fig:trajectory_smse_ave}
\end{figure}

\begin{figure} 
    \centering
        \centering
        \includegraphics[width=\linewidth]{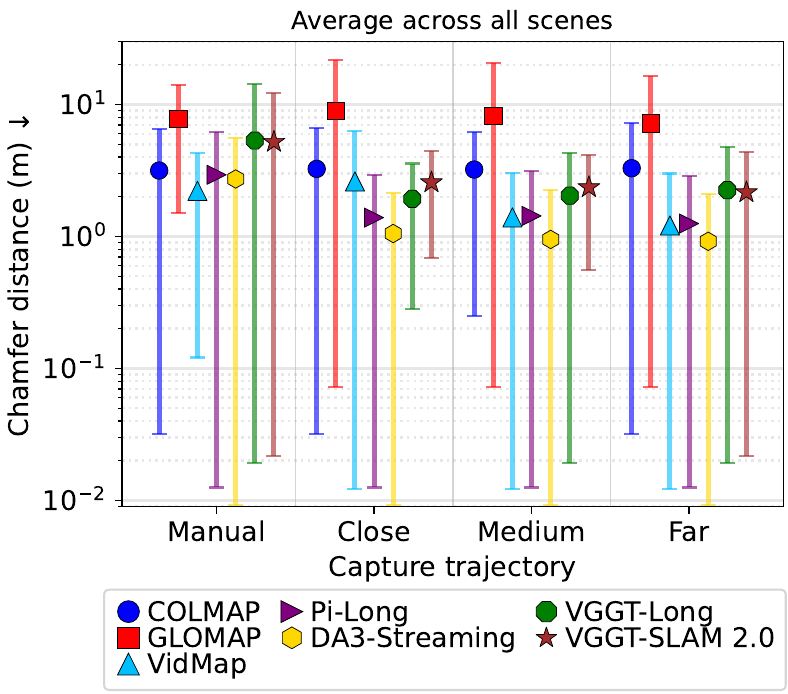}
        \caption{\small
        Average Chamfer distance vs. trajectory across all sequences in this dataset.
        }
        \label{fig:trajectory_cd_ave}
\end{figure}

\begin{figure} 
    \centering
        \centering
        \includegraphics[width=\linewidth]{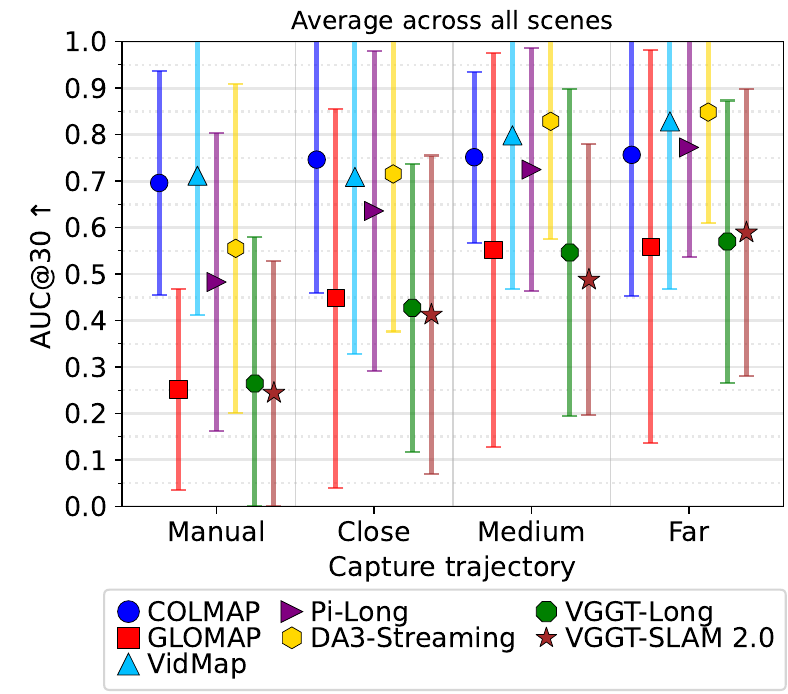}
        \caption{\small
        Average AUC@30 vs. trajectory across all sequences in this dataset.}
        \label{fig:trajectory_poses_ave}
\end{figure}

Camera trajectory significantly affects depth map accuracy, as shown in \Cref{fig:trajectory_smse_ave}. 
Across all methods, on average, depth maps improve from \textit{close} to \textit{far}.
For all methods, \textit{manual} produces more accurate depth maps than \textit{close}, despite being a similar distance from the target in most scenes.
This is likely because the \textit{manual} capture pattern is smoother and contains more overlap between frames. 

Interestingly, the point cloud and pose results somewhat differ from the depth map results.
For all learning-based methods, the Chamfer distance is substantially higher, and the AUC@30 is lower for the \textit{manual} trajectory as shown in \Cref{fig:trajectory_cd_ave,fig:trajectory_poses_ave}.
However, the \ac{SfM}-style approaches behave somewhat differently with higher-quality point clouds for the \textit{manual} capture pattern and a much smaller decrease in pose accuracy.
This may be because \textit{manual} capture resembles video data much more than the pseudorandom trajectories, and VidMap and COLMAP (when placed in video mode, as we did in this evaluation) explicitly utilise priors about video data.
This allows VidMap to achieve the highest quality point clouds and most accurate poses for the \textit{manual} trajectories.
This trend is illustrated in \Cref{fig:construction_point_clouds_pics}, where the VGGT-based methods experience a significant decrease in point cloud fidelity on the \textit{manual} sequence, and VidMap and COLMAP do not.

For the pseudorandom trajectories, on average, most methods show improved point cloud quality and pose accuracy with distance from the asset.
However, this does vary scene-to-scene significantly, as shown in \Cref{supp:trajectory} of the supplementary information, which includes a full breakdown of these results by scene.
This overall trend is reflected in \Cref{fig:construction_point_clouds_pics}, where many methods demonstrate improved point cloud fidelity as distance increases, such as $\pi^3$-X, \ac{DA3} and VGGT-SLAM 2.0.

\begin{figure*} 
    \centering
        \centering
        \includegraphics[width=\linewidth]{construction_point_clouds_pics.pdf}
        \caption{\small
        Predicted point clouds for the \textit{Construction} scene at varying camera trajectories under the \textit{sunrise} lighting condition. 
        The ground truth completeness mesh is provided for comparison.
        For most methods, point cloud quality improves as distance away from the asset increases.
        Note that the manual trajectory presents an additional challenge to the VGGT-based methods. 
        Many methods struggled to reconstruct the roof frame correctly, which consists of thin beams.
        }
        \label{fig:construction_point_clouds_pics}
\end{figure*}

Therefore, we see a significant trade-off in terms of the capture pattern for asset-inspection applications.
Closer distances to the asset yield higher resolutions, greater detail and therefore a greater ability to spot small changes in the asset.
However, for most methods, this results in lower-quality 3D reconstructions, less accurate depth maps and reduced camera pose accuracy.
We note that the classical COLMAP is less affected by this trade-off; however, it produces lower-quality clouds than the \ac{SOTA} learning-based methods for all capture patterns.
One can mitigate this trade-off by increasing the camera resolution and using camera poses further away from the asset.
However, with higher camera resolutions, one could still achieve a higher level of detail with closer captures at the same resolution.

\begin{figure} 
    \centering
        \centering
        \includegraphics[width=\linewidth]{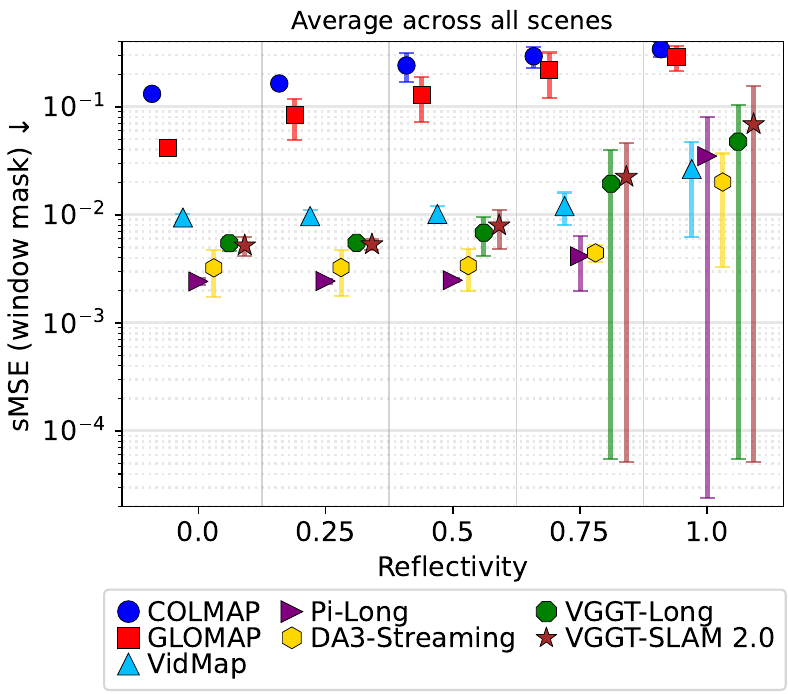}
        \caption{\small
        Average sMSE vs reflectivity across all sequences of \textit{Glass House} and \textit{Office}.
        Reflectivity was rendered at 5 discrete levels, 0, 0.25, 0.5, 0.75 and 1.0. The horizontal separation of each method is used for readability only. 
        In this plot, we only consider reflective areas, as everything else in the scene remains exactly the same.
        Error bars represent one standard deviation from the mean.
        }
        \label{fig:ref_smse_ave}
\end{figure}

\bmhead{Effect of reflectivity}
Classical methods, such as
COLMAP and GLOMAP, show a significant decrease in depth estimation performance in window regions as reflectivity increases, as shown in \Cref{fig:ref_smse_ave}.
This decrease is likely because reflective surfaces violate brightness constancy assumptions which underpin the \ac{MVS} stages of COLMAP and GLOMAP.
On the other hand, methods that use monocular depth estimation, such as VGGT-Long or VidMap, are more robust to reflective scene content.
Only at 50-75\% reflectivity does depth estimation performance decrease for these methods, and even then, their accuracy remains superior to the purely classical approaches.
This clearly shows that monocular depth-enabled methods produce more accurate depth maps in reflective scenes than classical approaches. 
This is likely due to the inclusion of reflective content in the training data for these methods.
However, reflective content still presents a greater challenge for these methods than non-reflective content.
In \Cref{fig:ref_smse_ave}, we see a decrease in depth estimation performance as reflectivity increases.
Additionally, the standard deviation in these metrics increases as reflectivity increases, indicating that performance becomes less predictable.
Because reflectivity 0 corresponds to a textured wooden panel, decreasing reflectivity also adds surface texture, so improvements at low reflectivity cannot be solely attributed to reduced reflection alone.
Plots of reflectivity vs. sMSE for both \textit{Glass House} and \textit{Office} are shown in \Cref{supp:reflectivity} of the supplementary information.

\begin{figure}[t]
  \centering
  \setlength{\tabcolsep}{0pt}
  \renewcommand{\arraystretch}{0}
  \newcommand{\imgw}{0.225\linewidth}   
  \newcommand{\cell}[1]{\includegraphics[width=\imgw]{#1}}
  \newcommand{\hdr}[1]{\makebox[\imgw][c]{\small #1\rule{0pt}{2.5ex}}}

  \newlength{\imgh}
  \settoheight{\imgh}{\cell{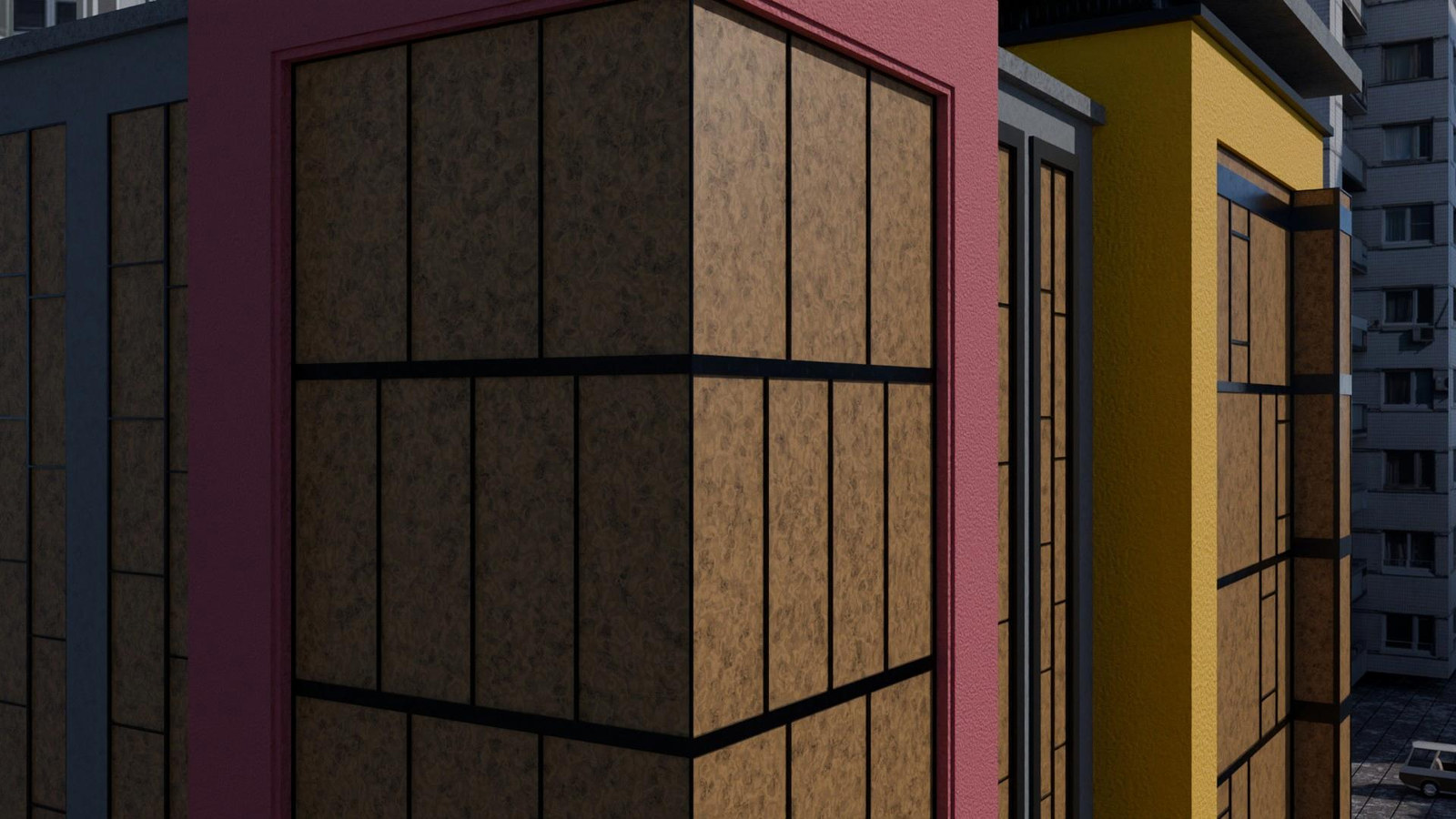}}

  \newcommand{\rlabel}[1]{%
    \raisebox{0.35\imgh}[0pt][0pt]{\rotatebox[origin=c]{90}{\small #1}}}
  \newcommand{\mainlabel}[1]{%
    \raisebox{-1.5\imgh}[0pt][0pt]{\rotatebox[origin=c]{90}{\small #1}}}

  \begin{tabular}{@{}c@{\hspace{4pt}}c@{\hspace{2pt}}c c c c@{}}
    & & \hdr{RGB Image} & \hdr{COLMAP} & \hdr{GLOMAP} & \hdr{VidMap} \\[2pt]

    \mainlabel{Reflectivity}
      & \rlabel{0.0}
      & \cell{office_image_ref0.0}
      & \cell{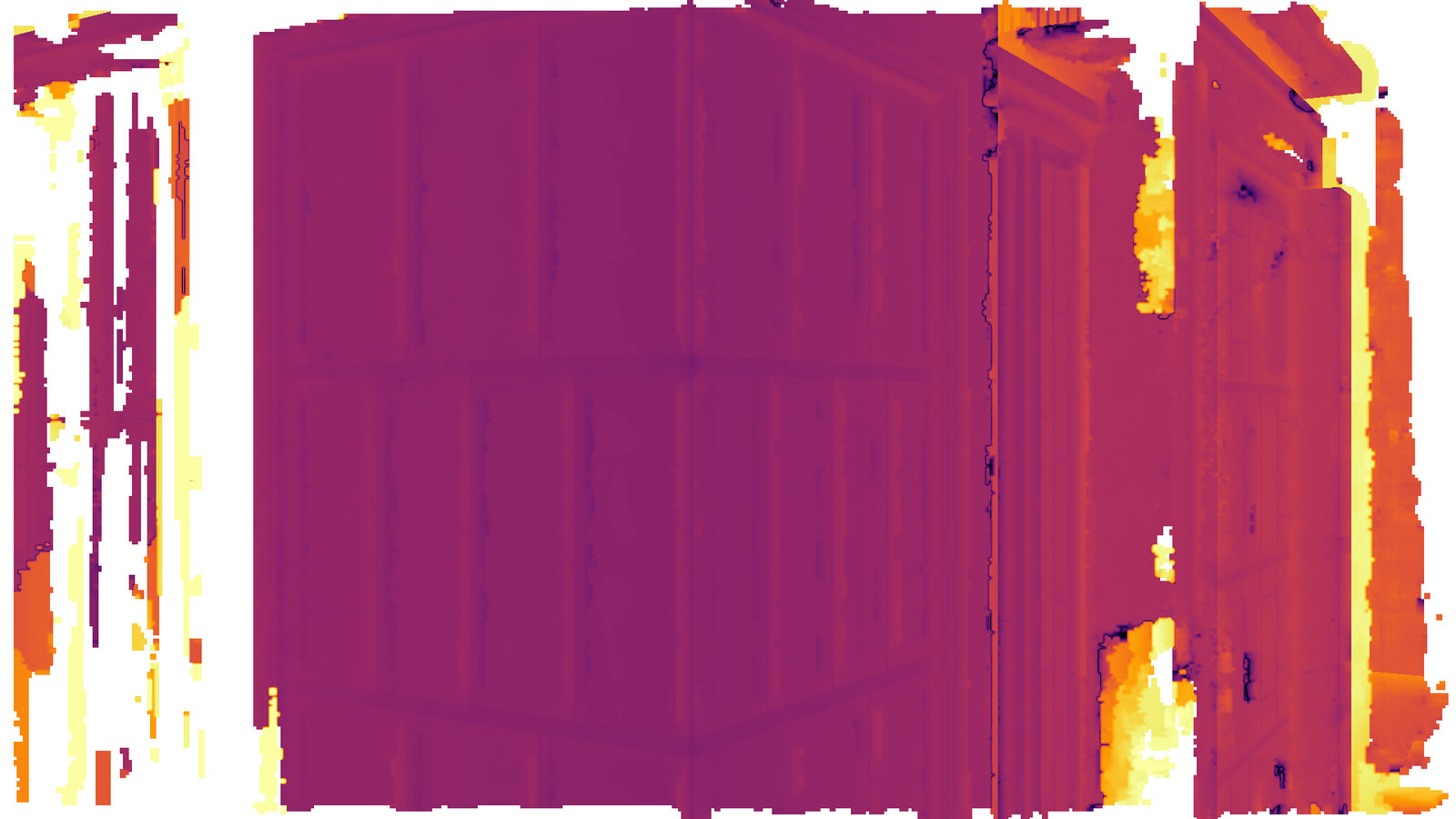}
      & \cell{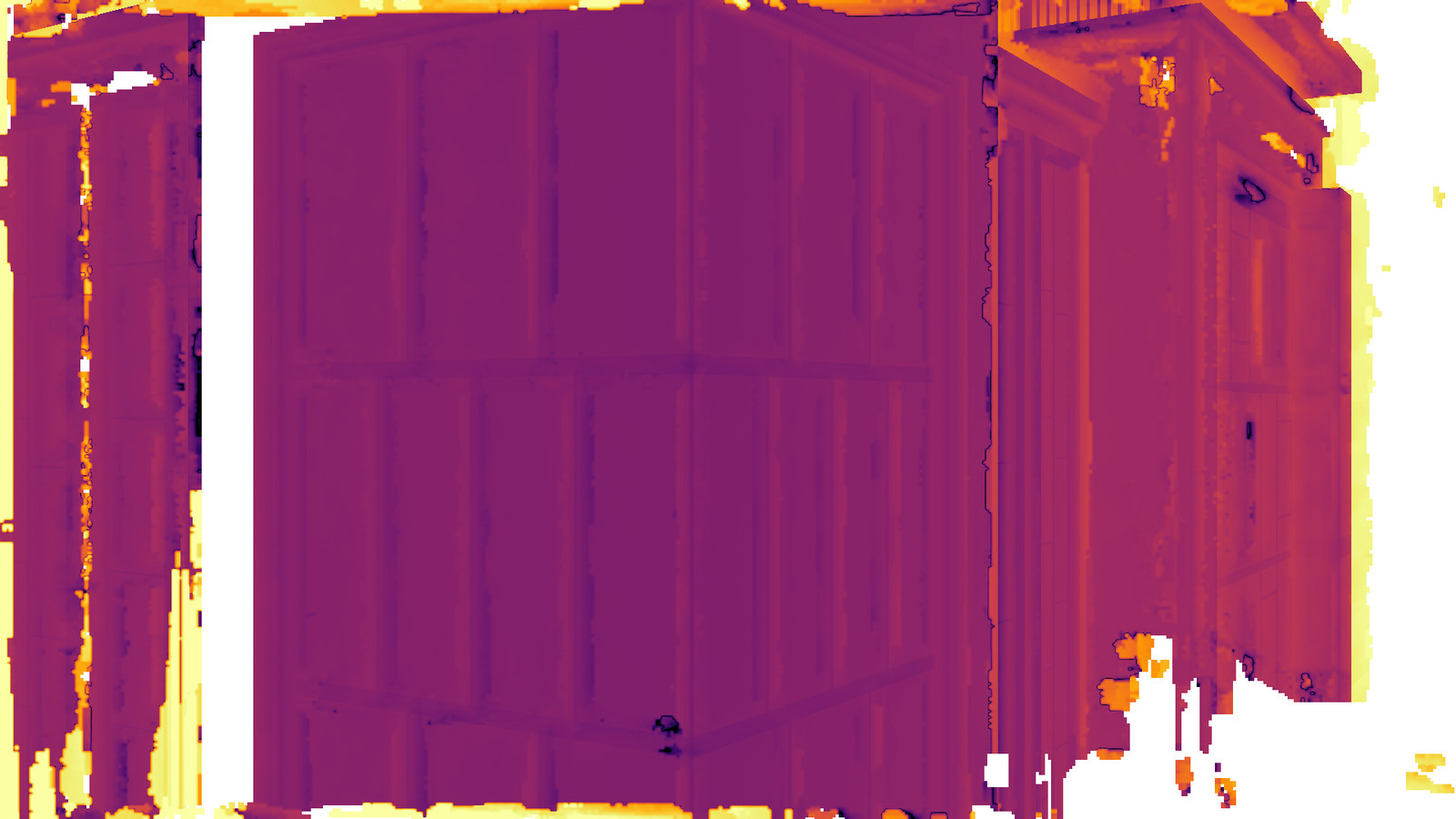}
      & \cell{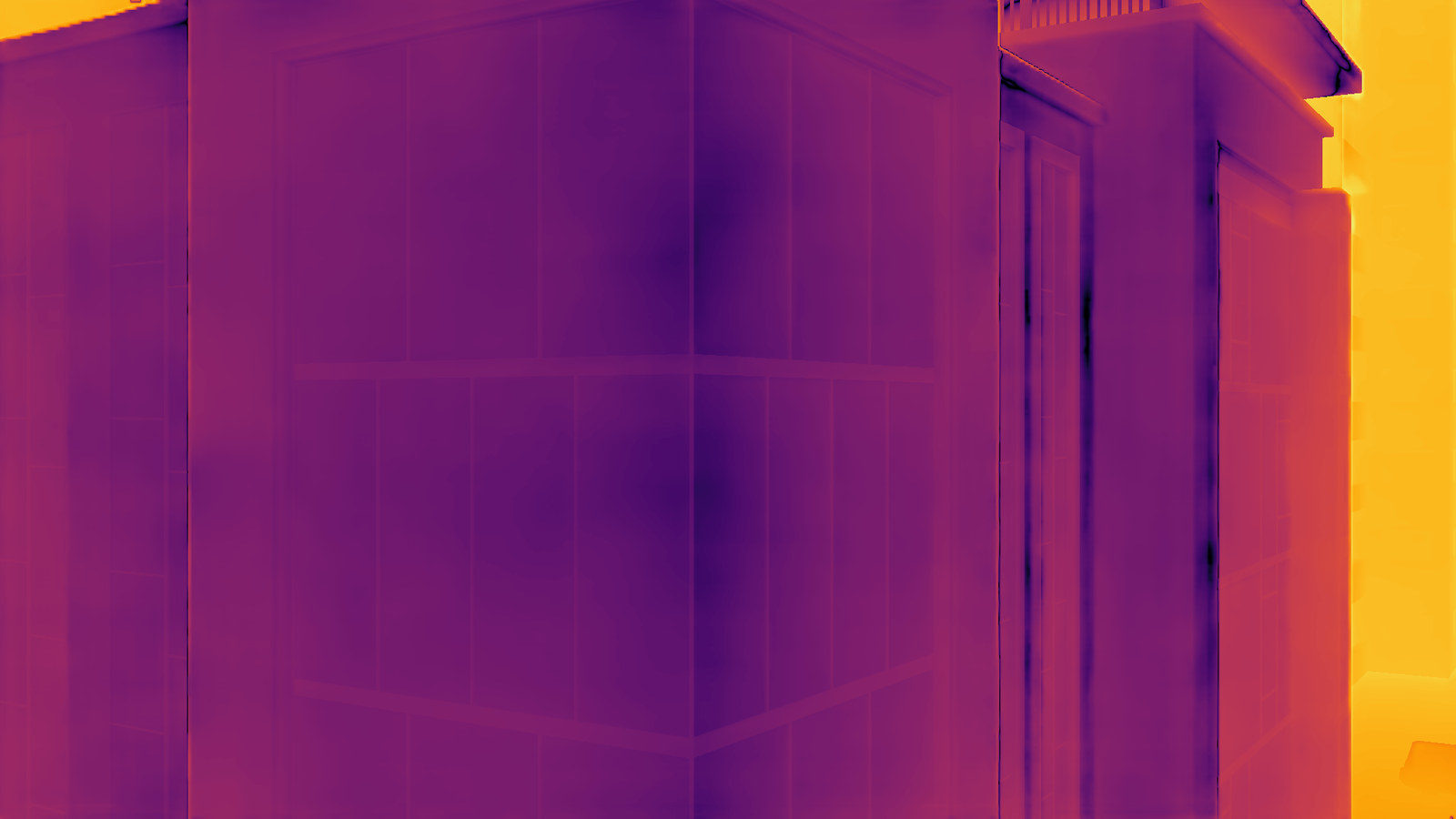}  \\
    & \rlabel{0.25}
      & \cell{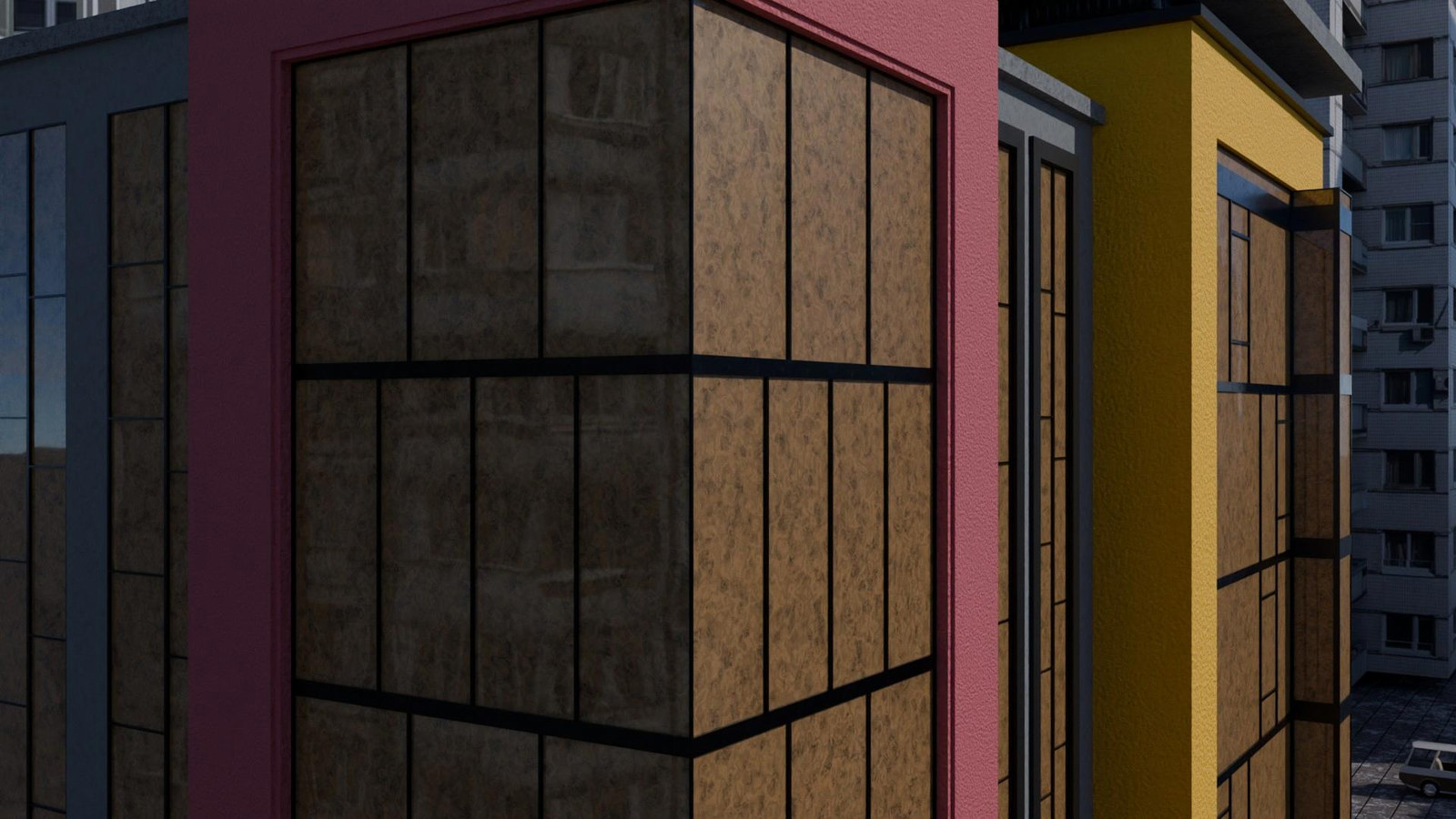}
      & \cell{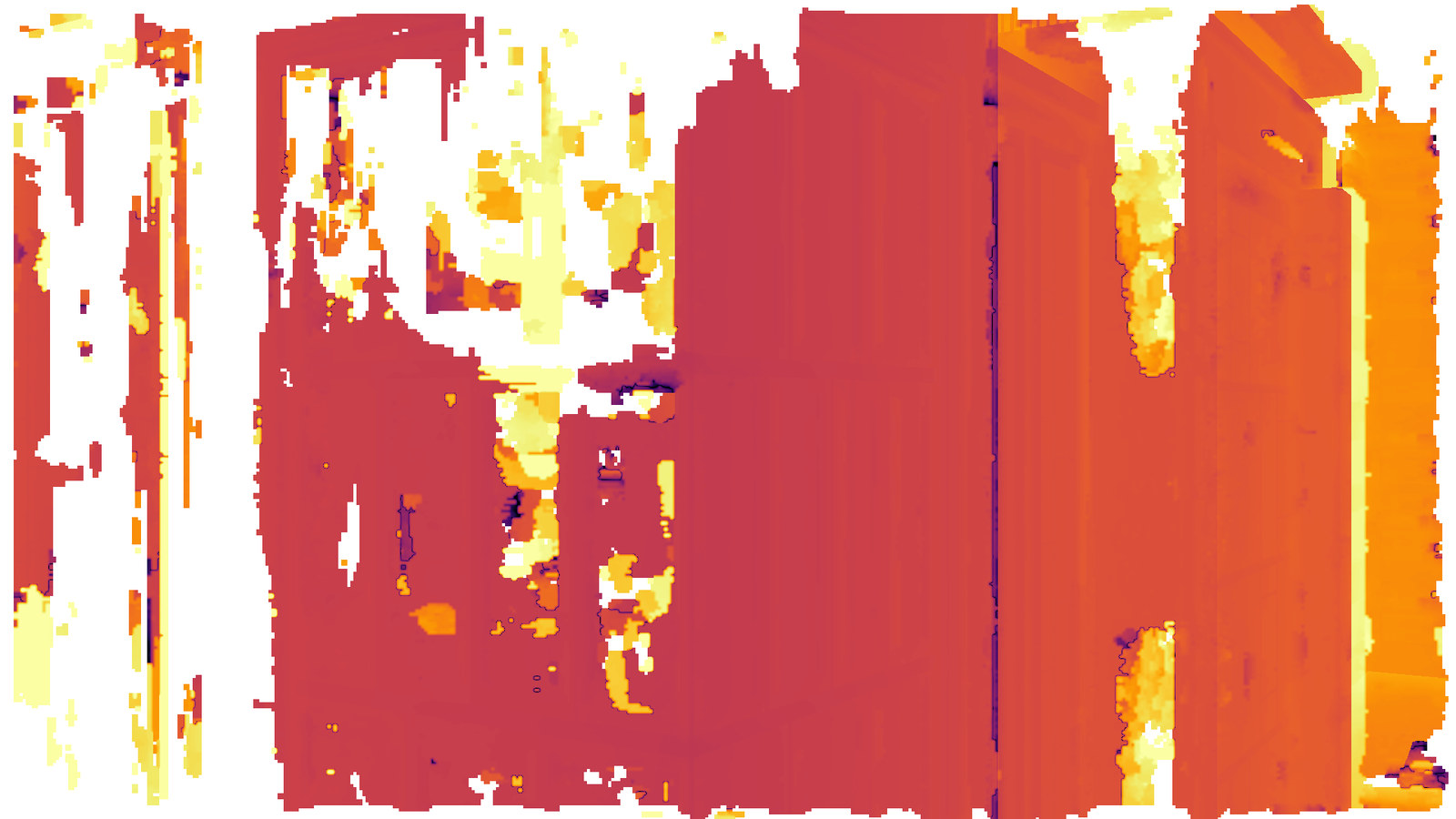}
      & \cell{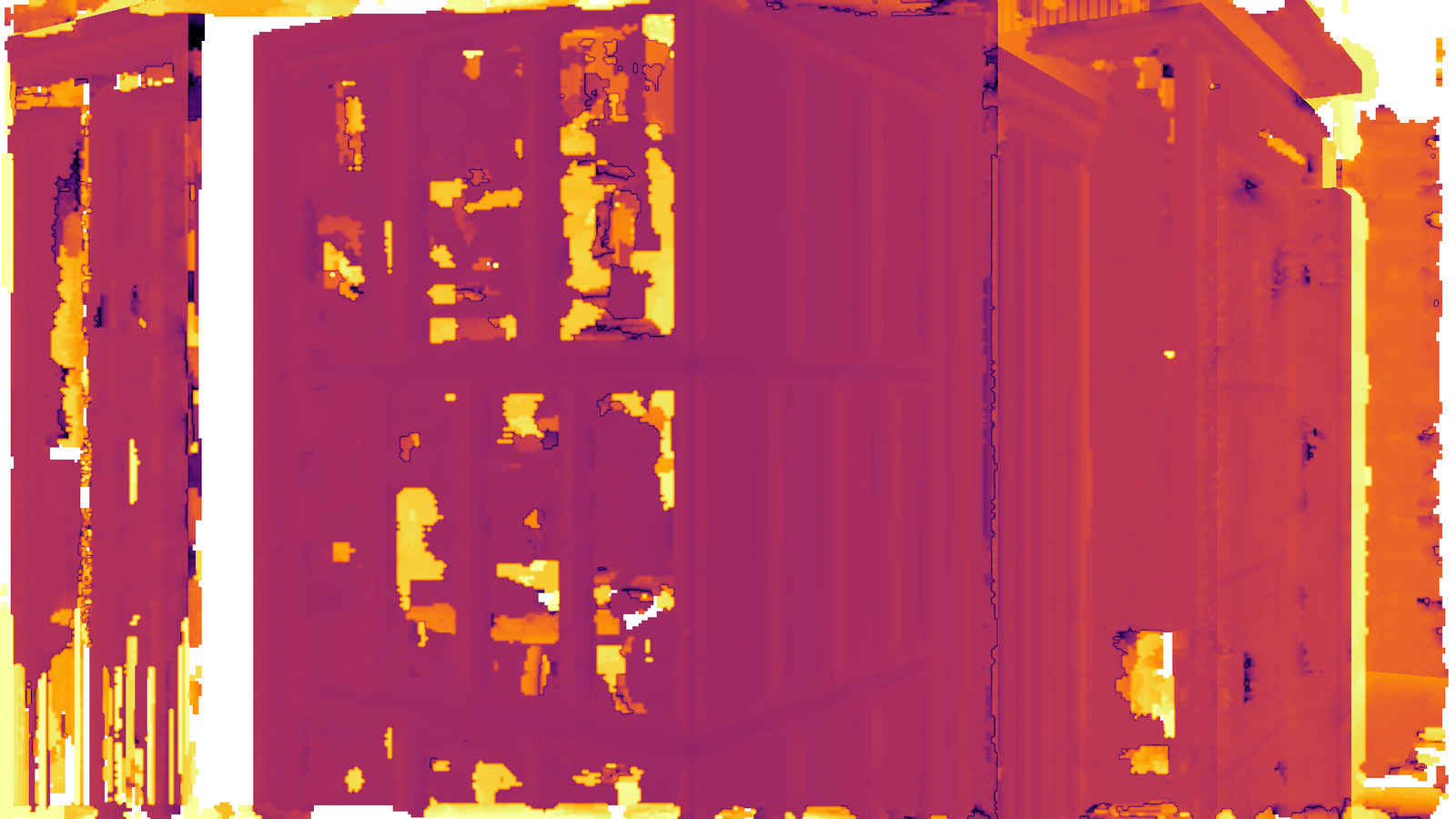}
      & \cell{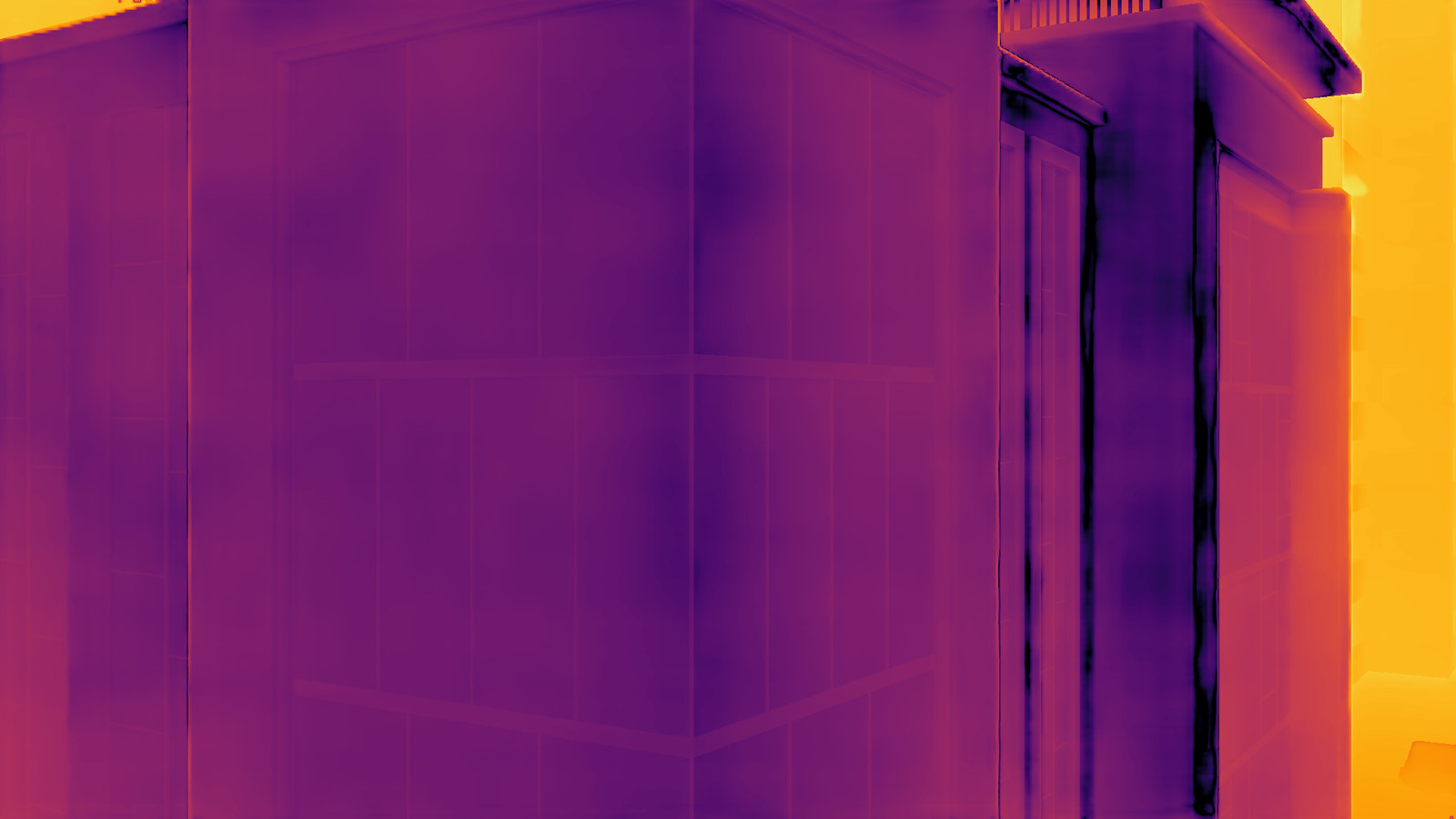}  \\
    & \rlabel{0.5}
      & \cell{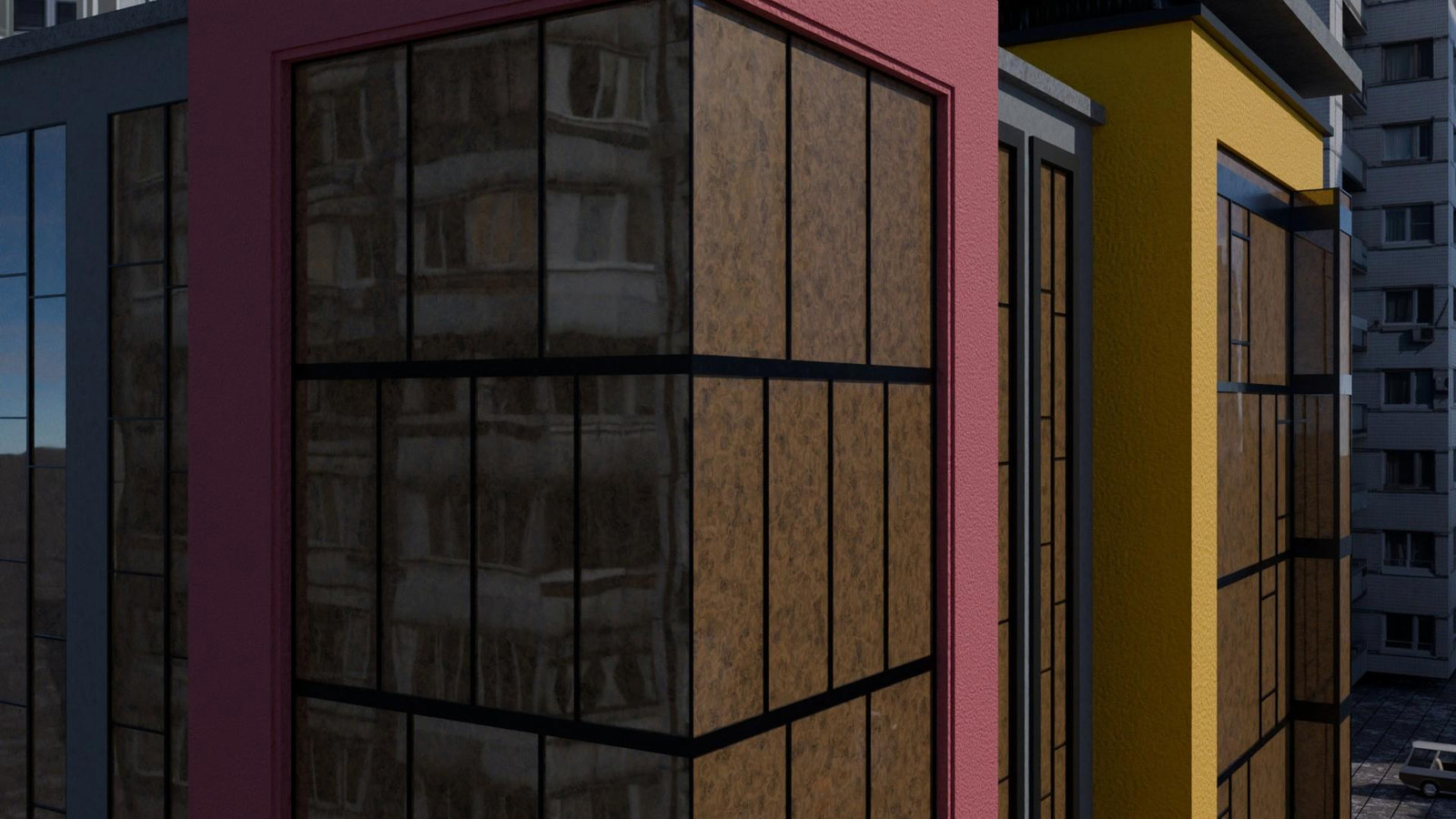}
      & \cell{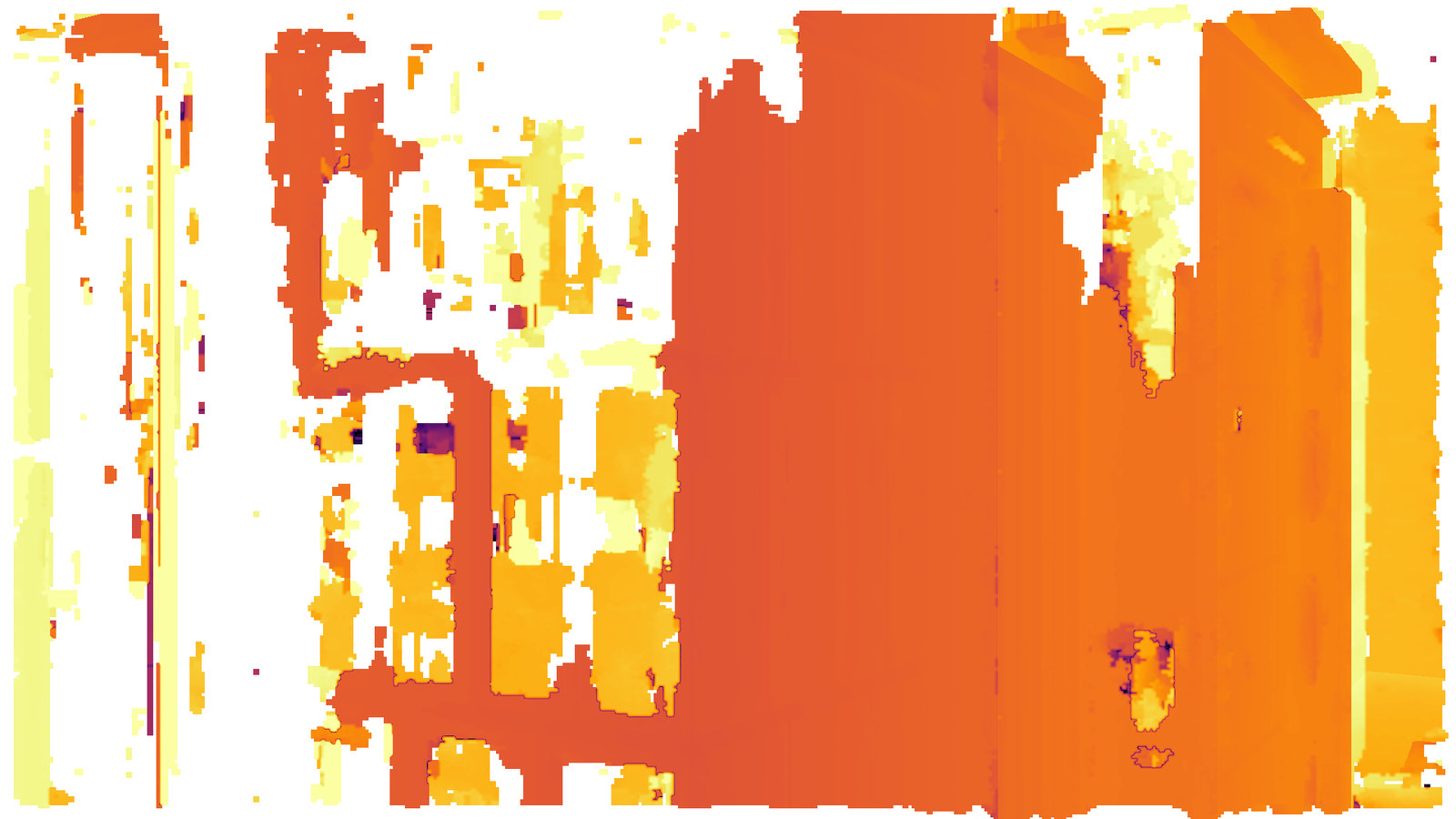}
      & \cell{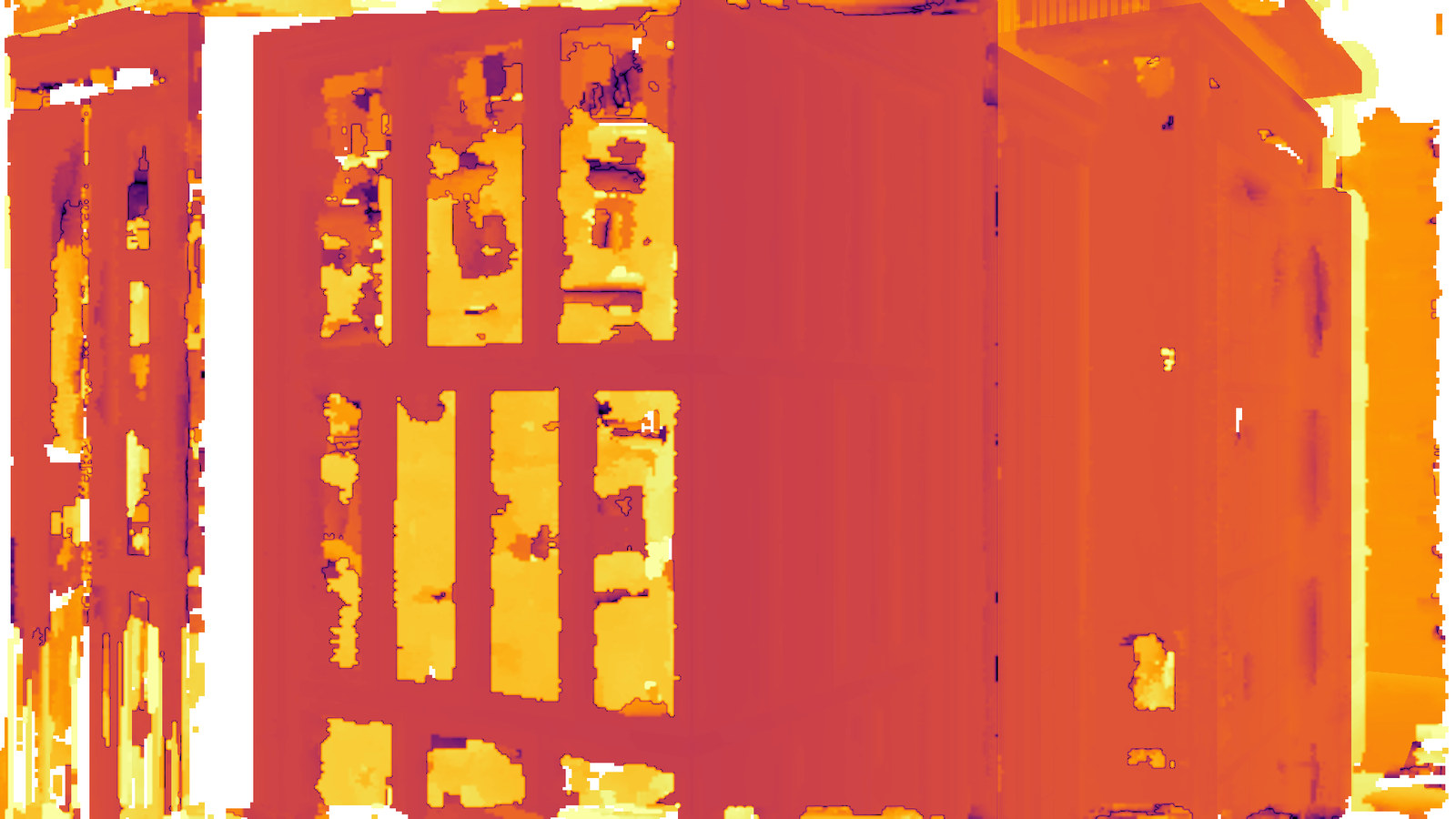}
      & \cell{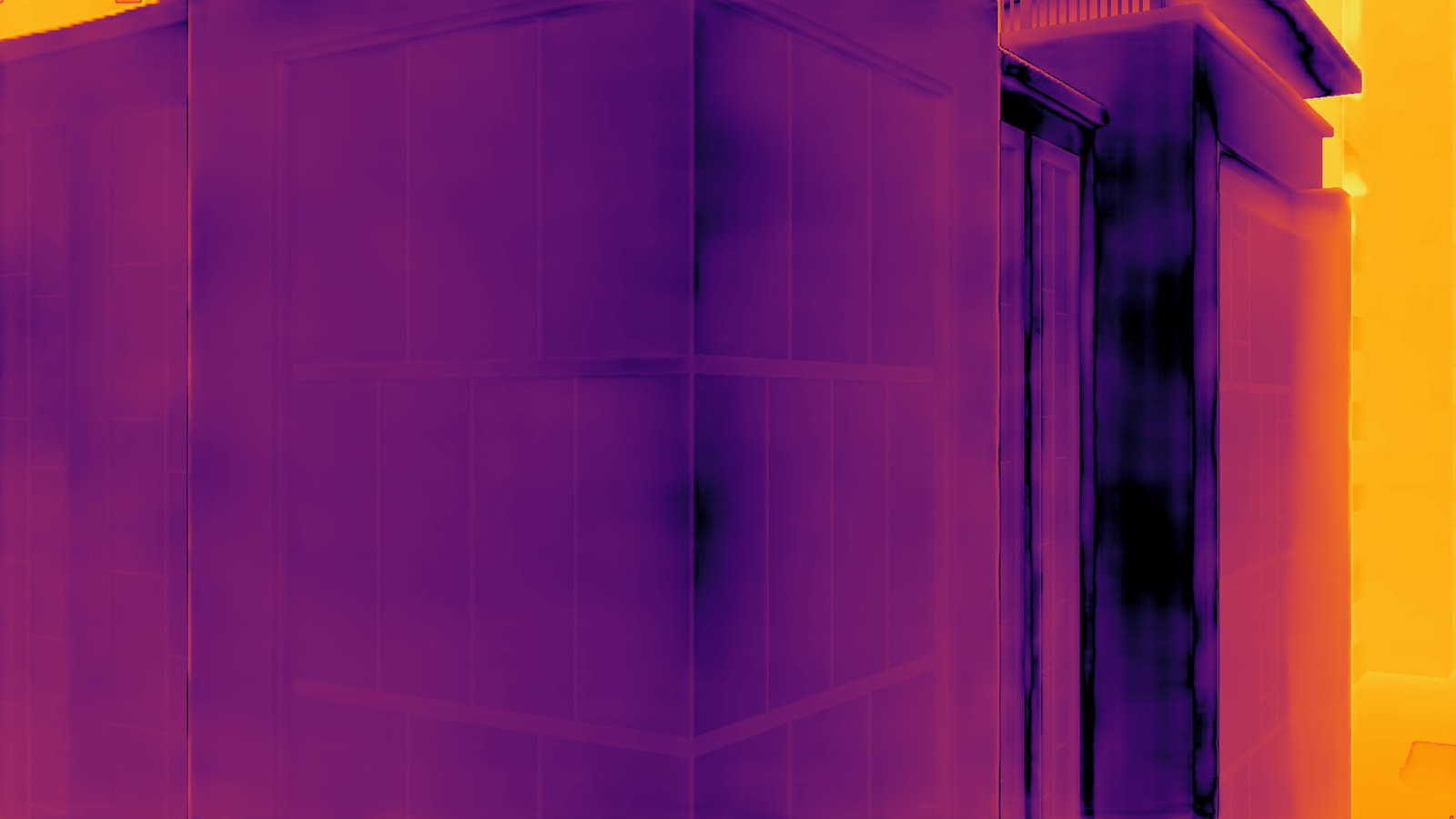}  \\
    & \rlabel{0.75}
      & \cell{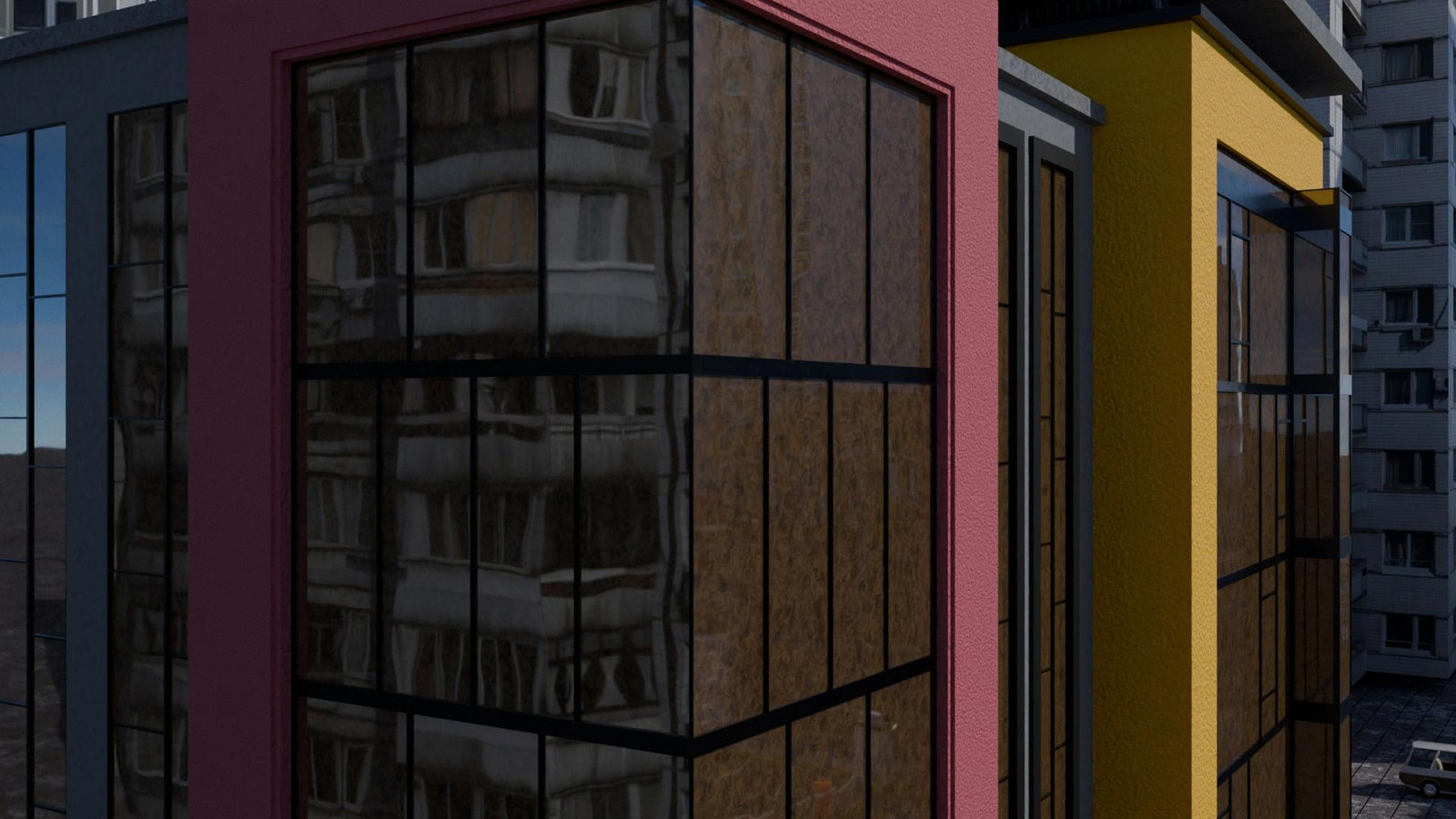}
      & \cell{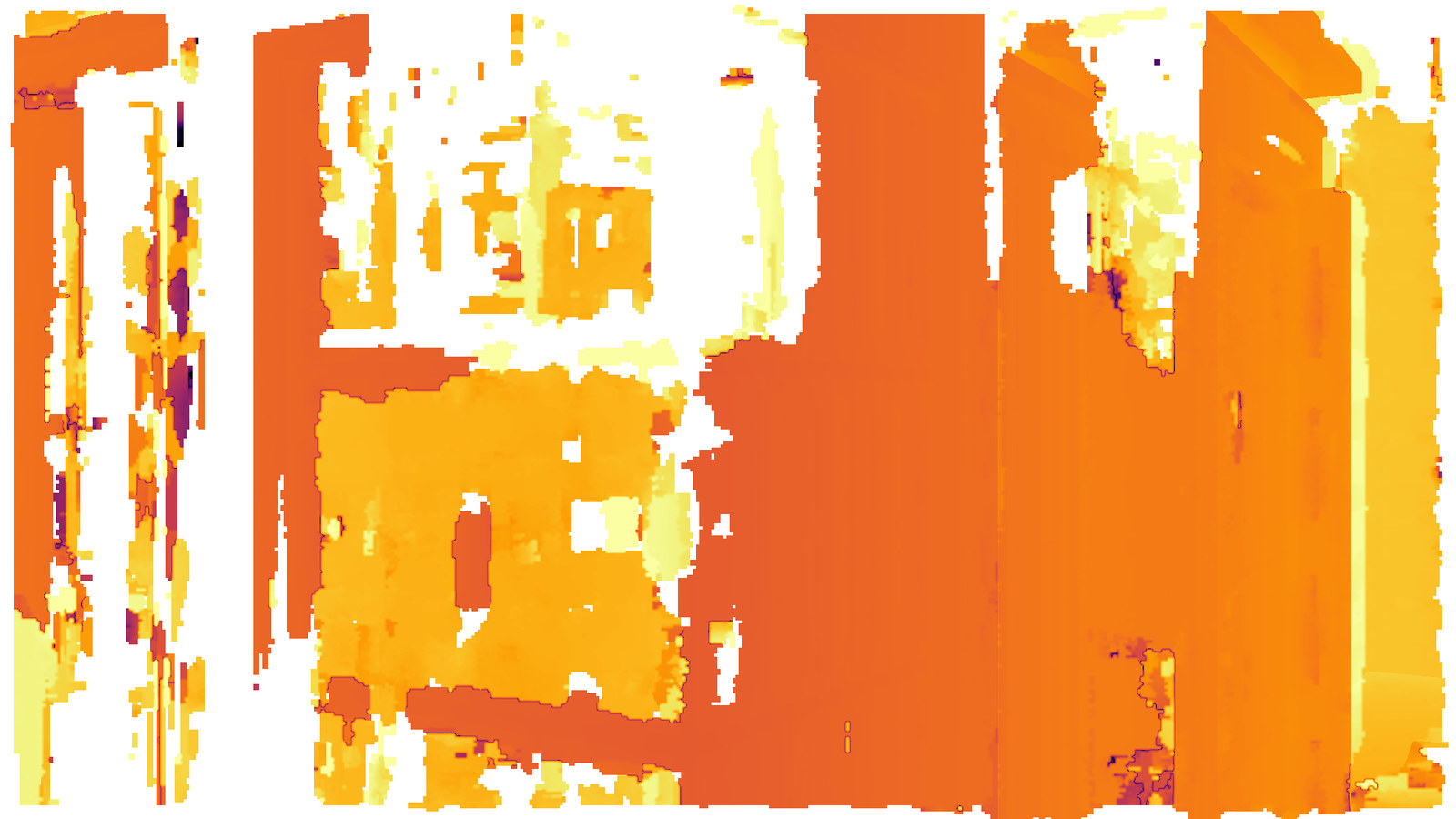}
      & \cell{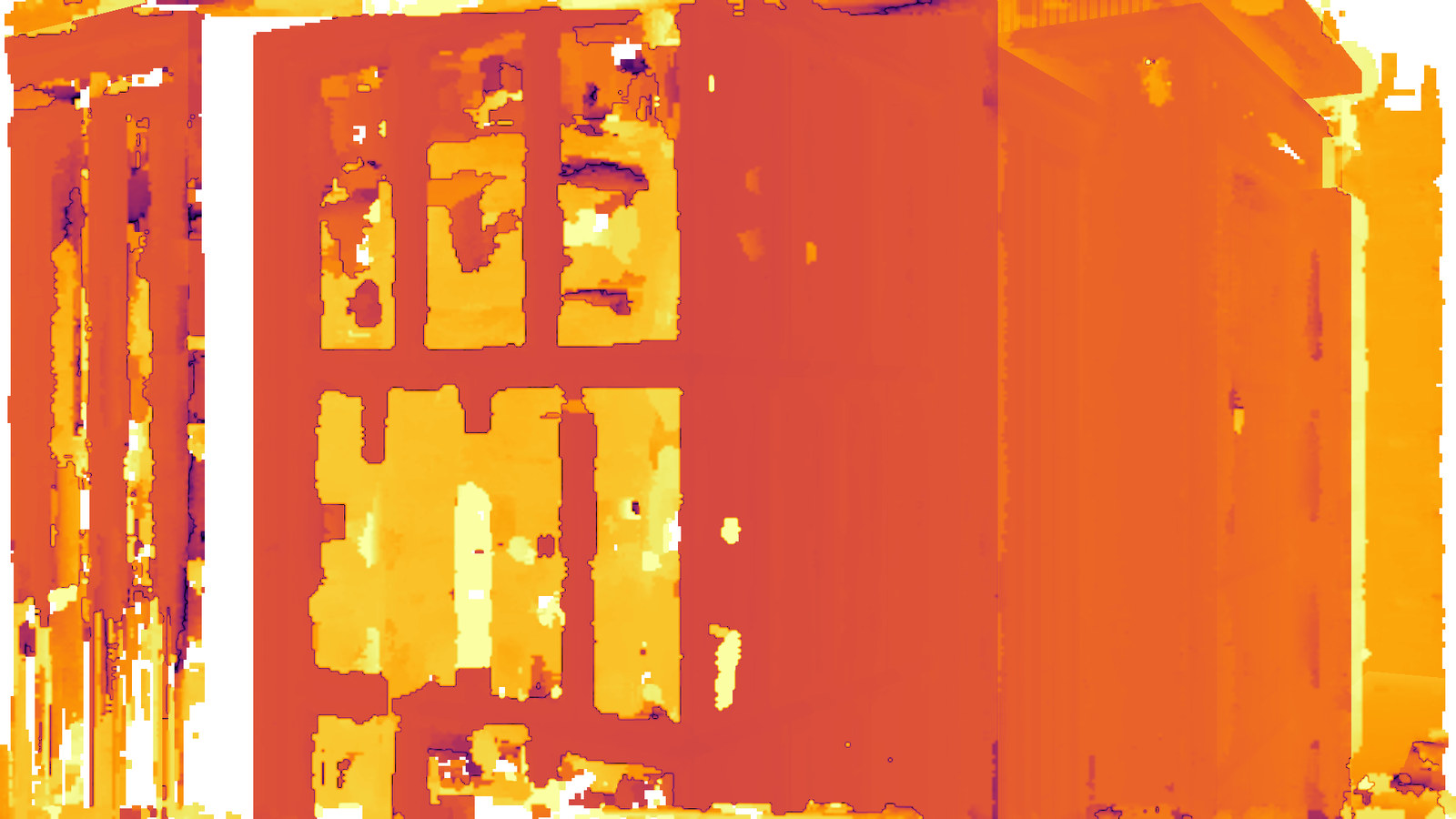}
      & \cell{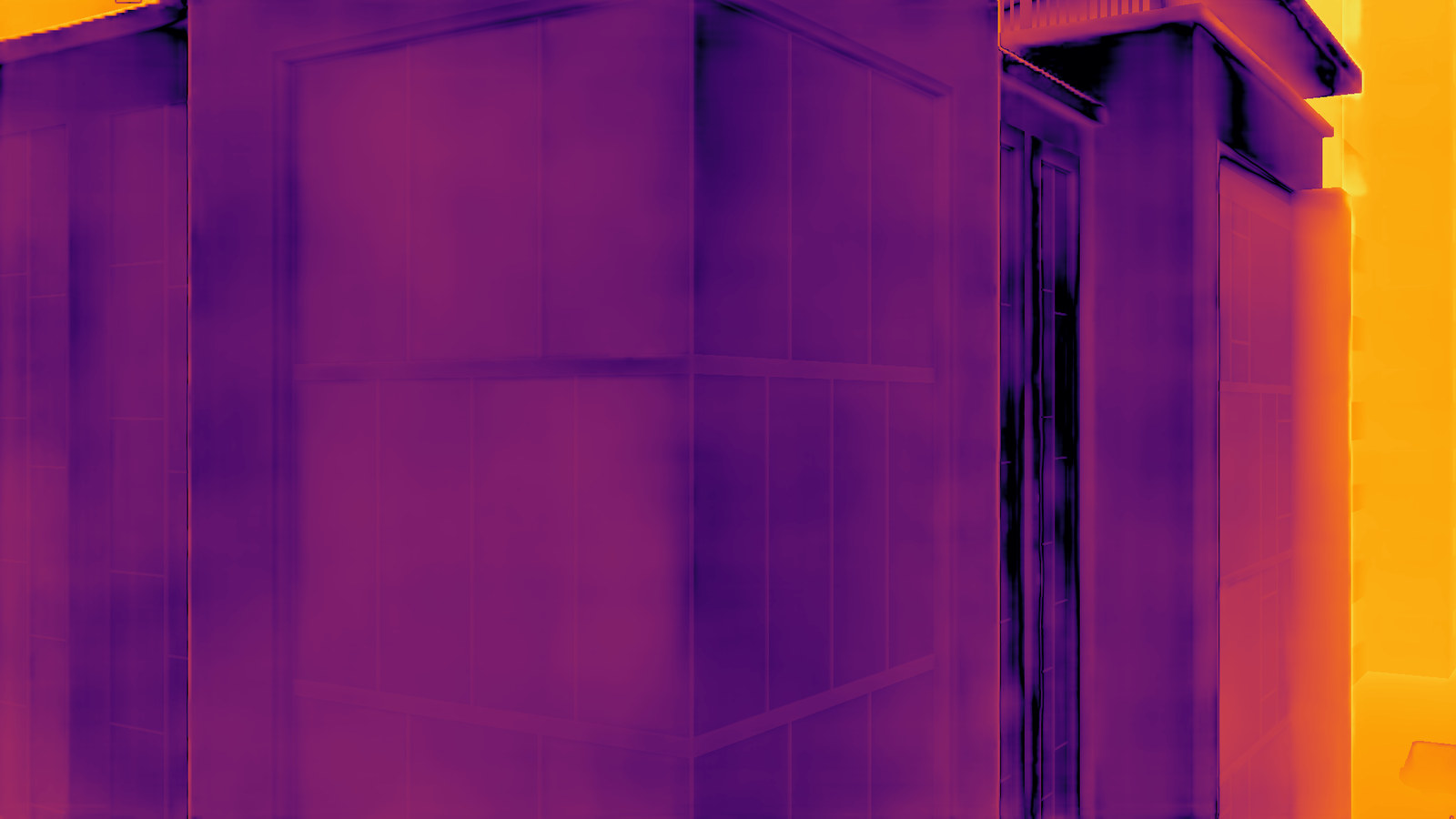}  \\
    & \rlabel{1.0}
      & \cell{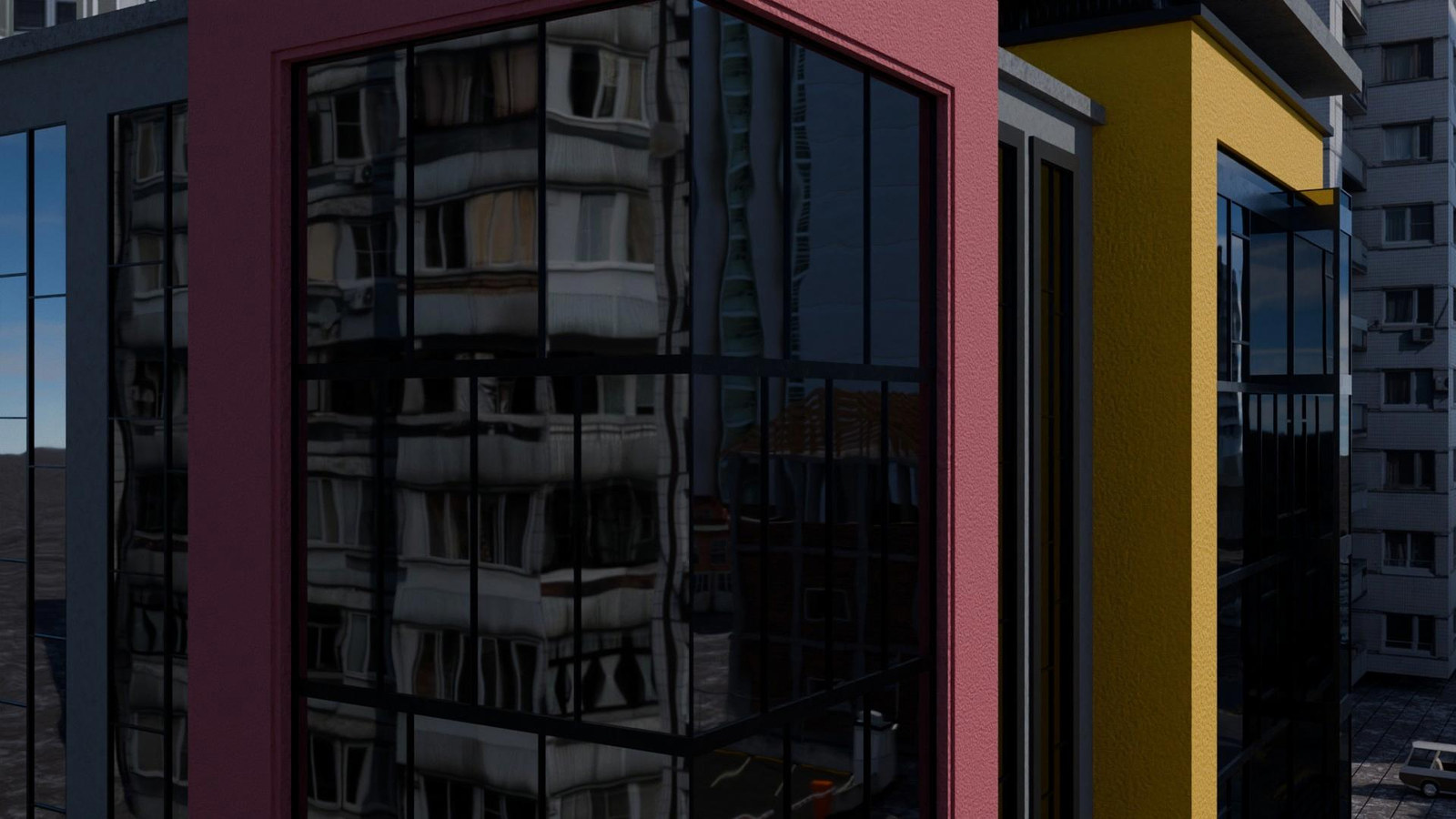}
      & \cell{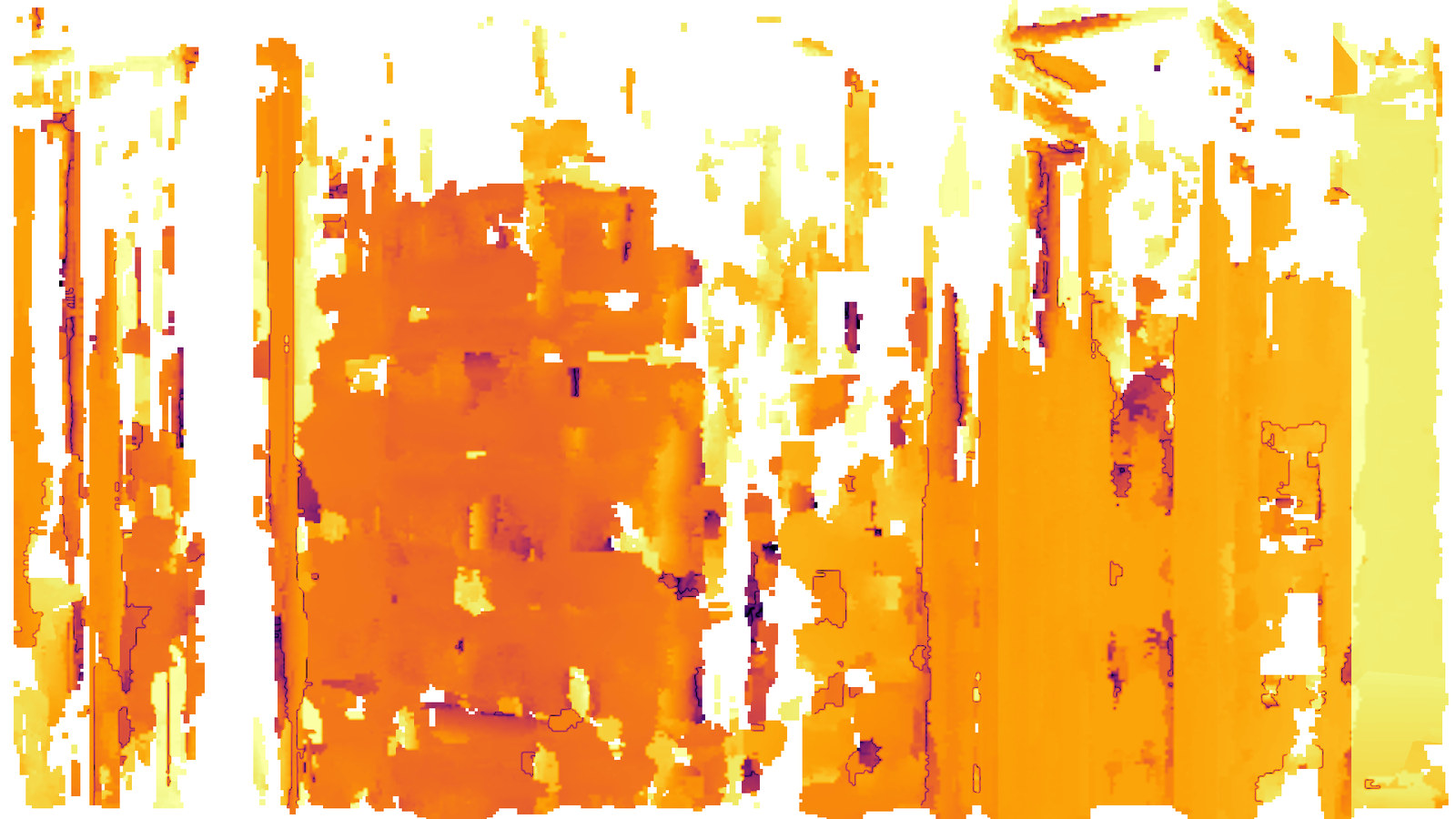}
      & \cell{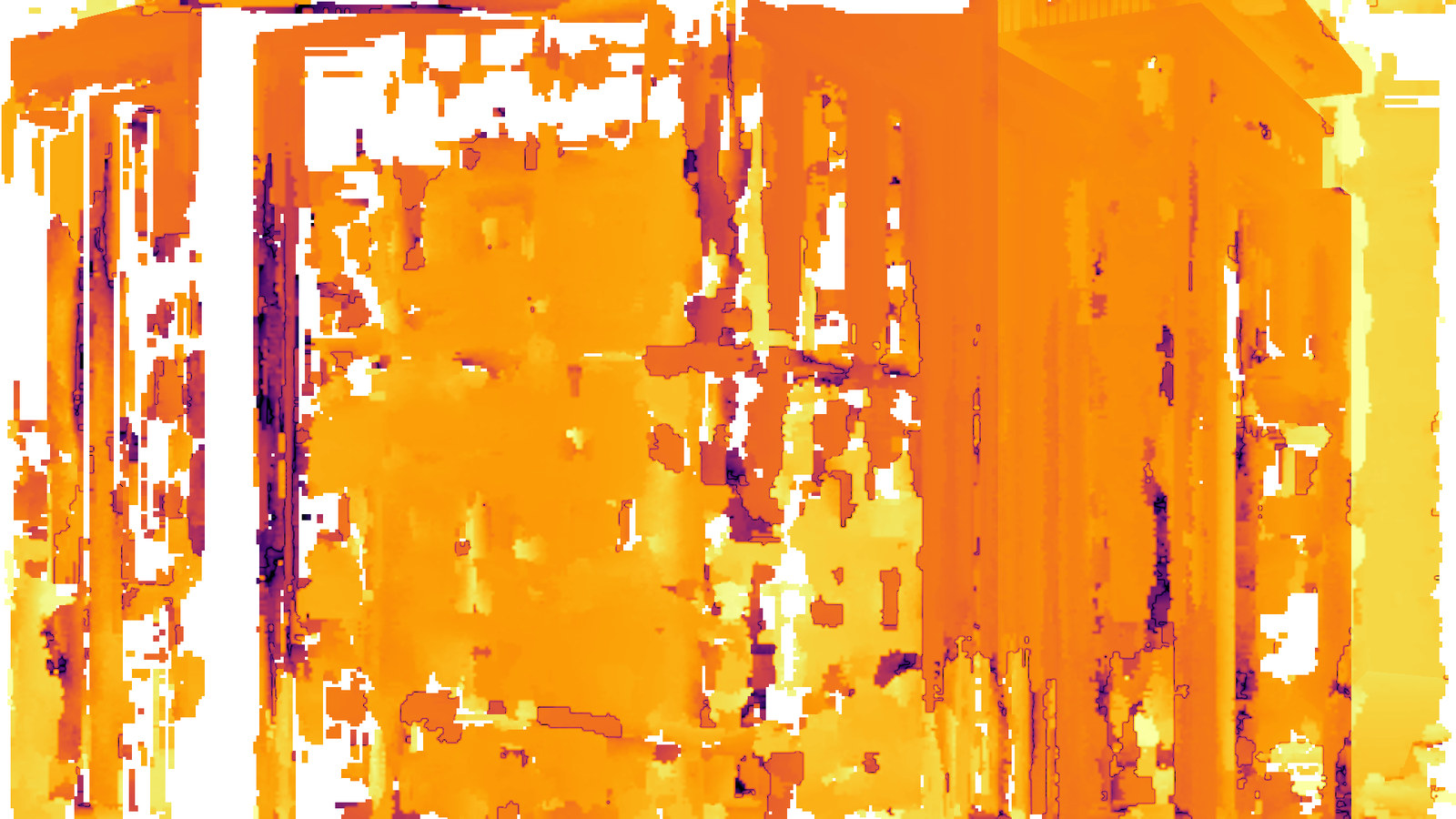}
      & \cell{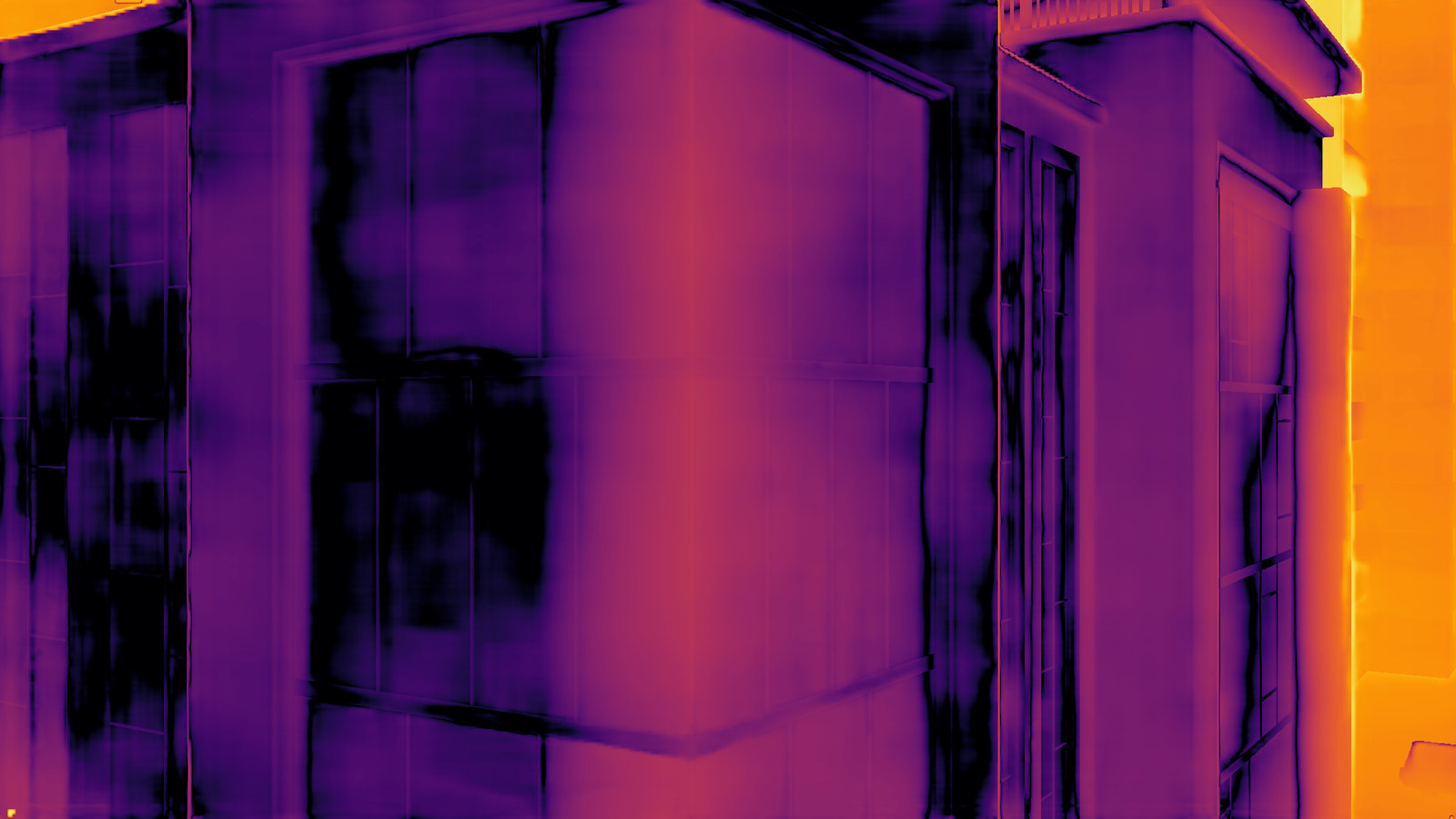}  \\[6pt]
            & & \multicolumn{4}{c}{\includegraphics[width=0.7\linewidth]{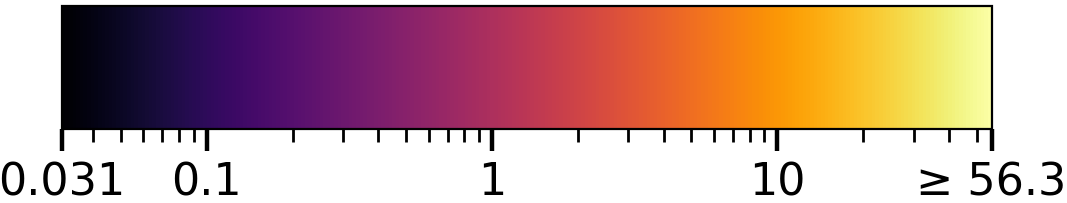}} \\[2pt]
      & & \multicolumn{4}{c}{\small Scale-adjusted depth error (m)} \\
  \end{tabular}
  \caption{\small
    Depth error for each SfM method after scales are adjusted to match the ground truth reference frame. White indicates a region with invalid estimated depth. 
    This frame is 802 from the \textit{Office} scene with \textit{far} trajectory under the \textit{Midday} lighting condition. 
    COLMAP and GLOMAP show substantially reduced accuracy with increased reflectivity. 
    VidMap uses \ac{DA3} as part of its framework and consequently, it is comparatively robust to changes in reflectivity.
    }
    \label{fig:reflectivty_vs_depth_qualitative_sfm}
\end{figure}

\begin{figure}[t]
  \centering
  \setlength{\tabcolsep}{1pt}
  \renewcommand{\arraystretch}{0}

  \newlength{\wA}\setlength{\wA}{0.37\linewidth}
  \newlength{\wB}\setlength{\wB}{0.32\wA}
  \newlength{\tw}\setlength{\tw}{\dimexpr\wA+4\wB+10pt\relax}  

  \newcommand{\cropL}{0.68}  
  \newcommand{\fcell}[1]{\includegraphics[width=\wA]{#1}}
  \newlength{\natw}
  \newcommand{\ccell}[1]{%
  \settowidth{\natw}{\includegraphics{#1}}%
  \setlength{\natw}{\cropL\natw}%
  \includegraphics[trim={\natw} 0 0 0, clip, width=\wB]{#1}}

  \settoheight{\imgh}{\fcell{office_image_ref0.0}}

  \newcommand{\rlabel}[1]{%
    \raisebox{0.35\imgh}[0pt][0pt]{\rotatebox[origin=c]{90}{\small\shortstack{#1}}}}
  \newcommand{\hdrA}[1]{\makebox[\wA][c]{\small #1\rule{0pt}{2.5ex}}}
  \newcommand{\hdrB}[1]{\makebox[\wB][c]{\small #1\rule{0pt}{2.5ex}}}

  \begin{tabular}{@{}c c c c c c@{}}
    & \hdrA{Reflectivity: 0.0} & \hdrB{0.25} & \hdrB{0.5} & \hdrB{0.75} & \hdrB{1.0} \\[2pt]

    \rlabel{Image}
      & \fcell{office_image_ref0.0}
      & \ccell{office_image_ref0.25}
      & \ccell{office_image_ref0.5}
      & \ccell{office_image_ref0.75}
      & \ccell{office_image_ref1.0} \\[2pt]

    \rlabel{DA3-\\Streaming}
      & \fcell{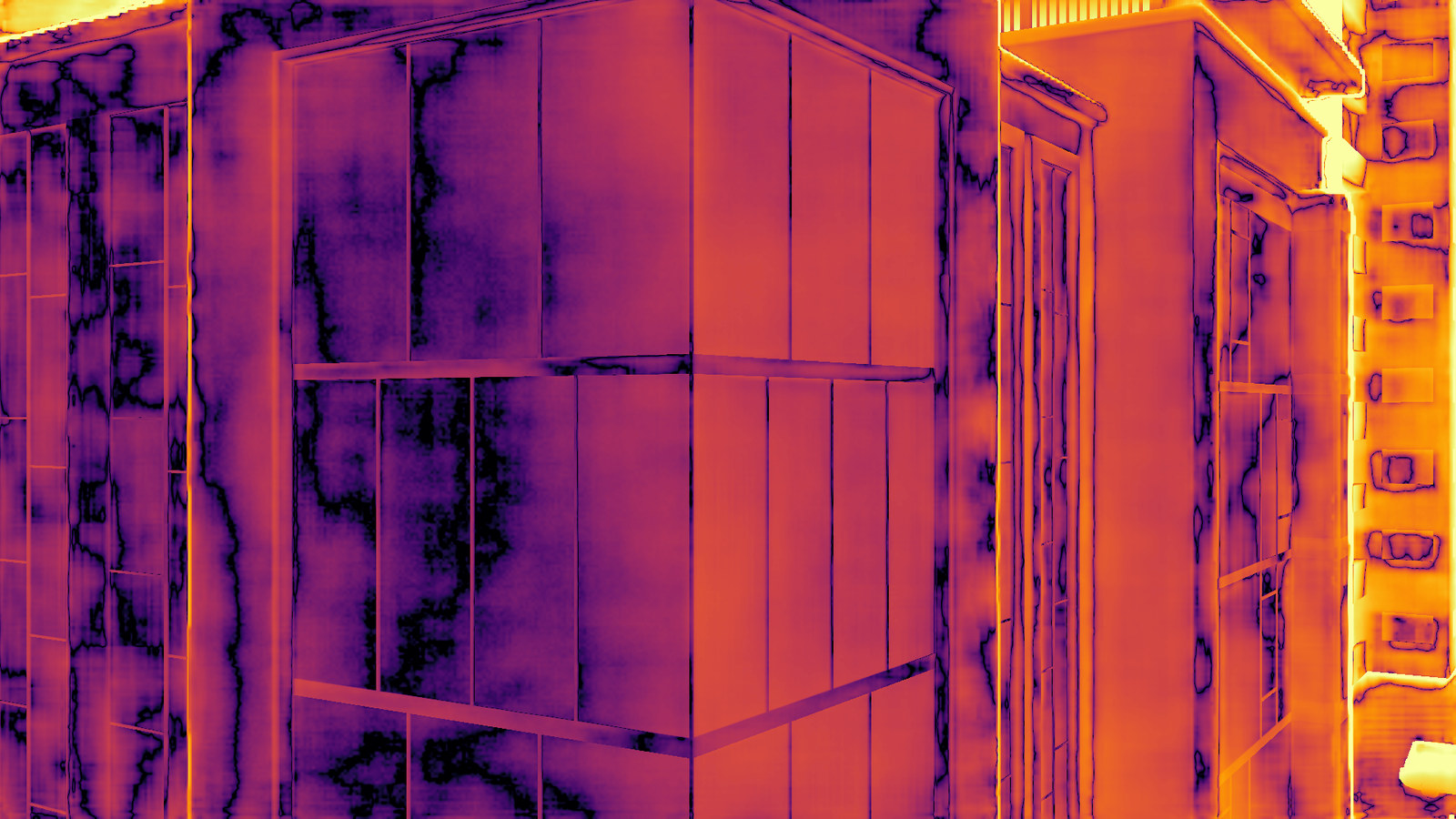}
      & \ccell{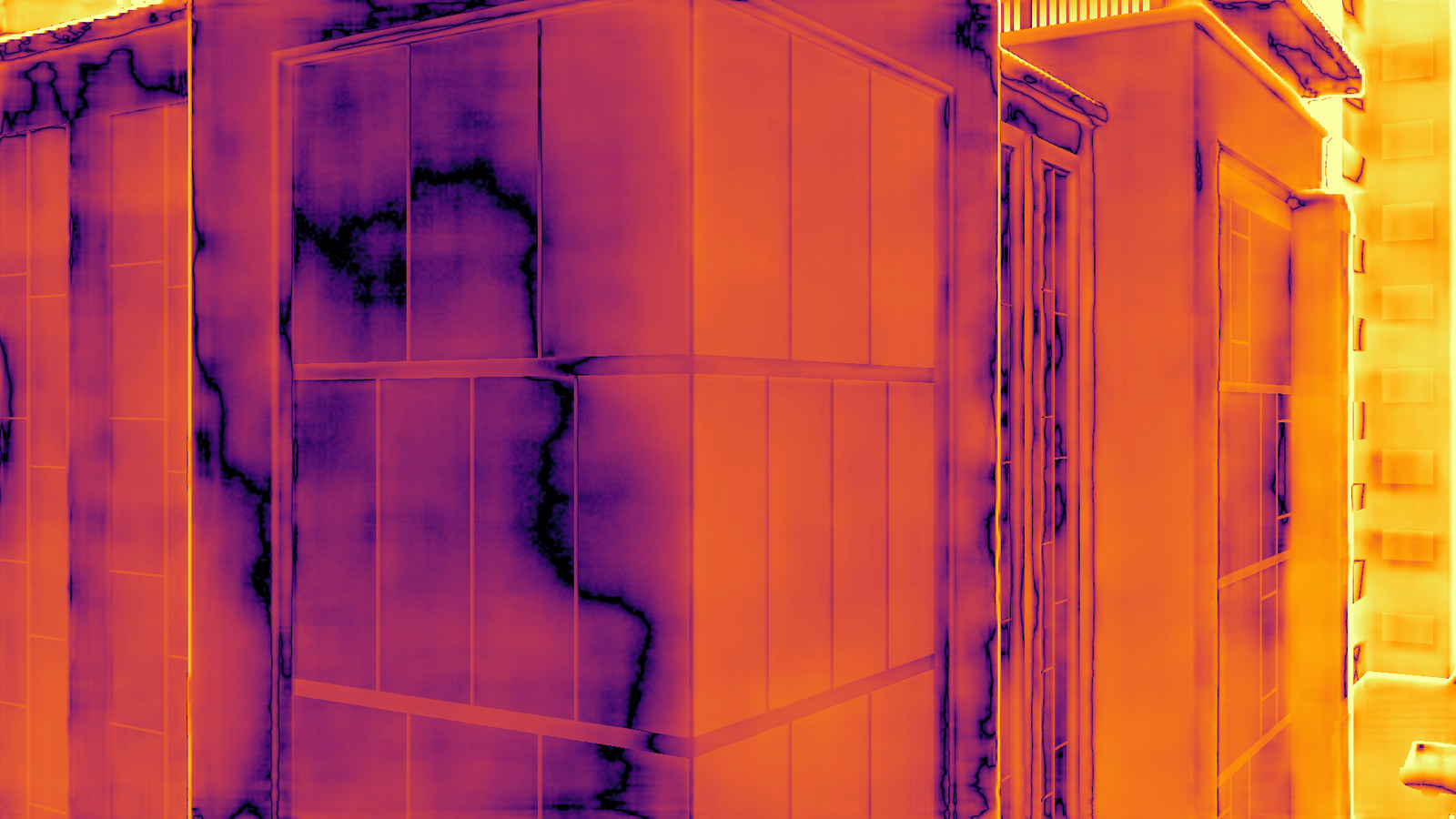}
      & \ccell{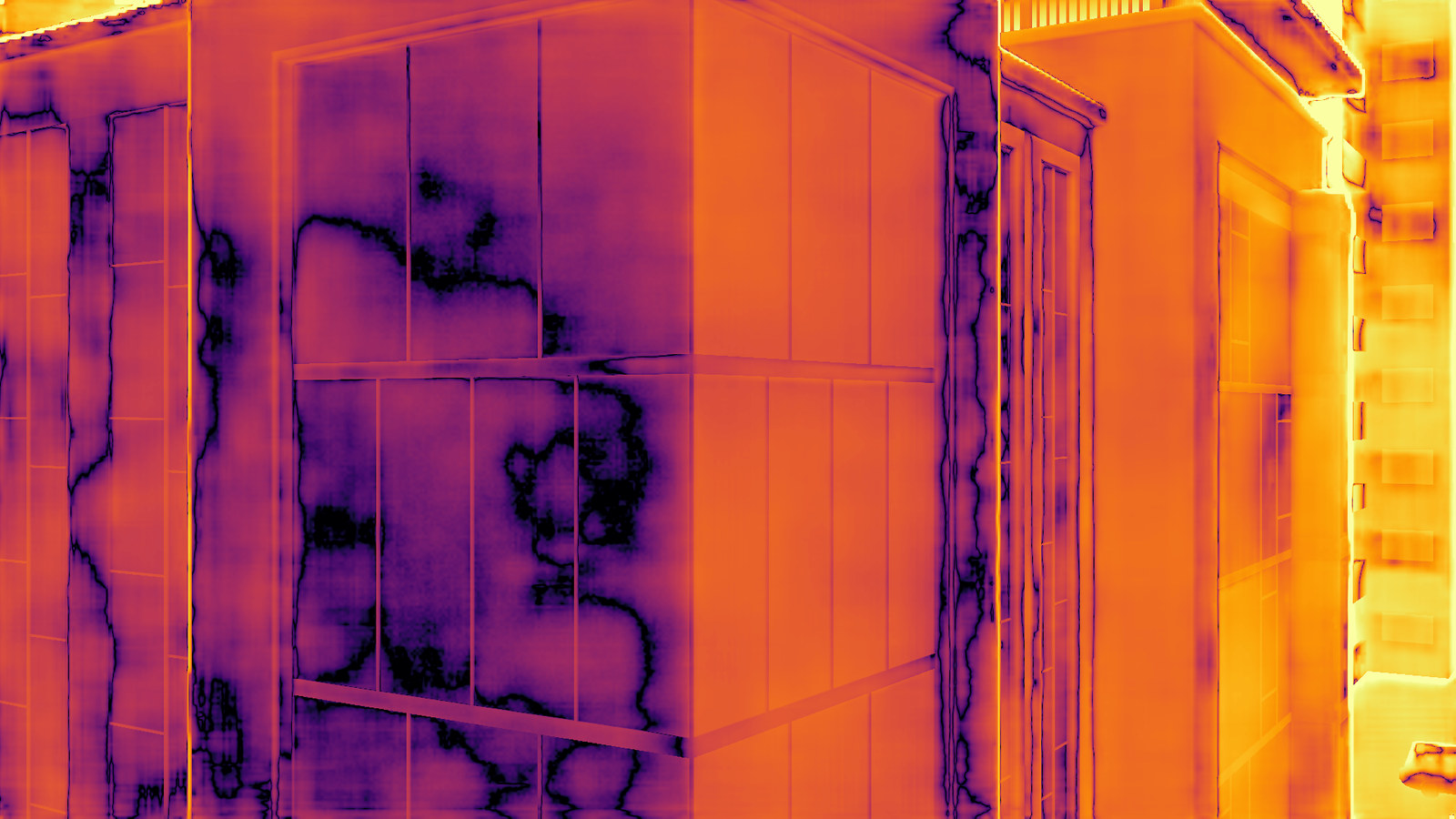}
      & \ccell{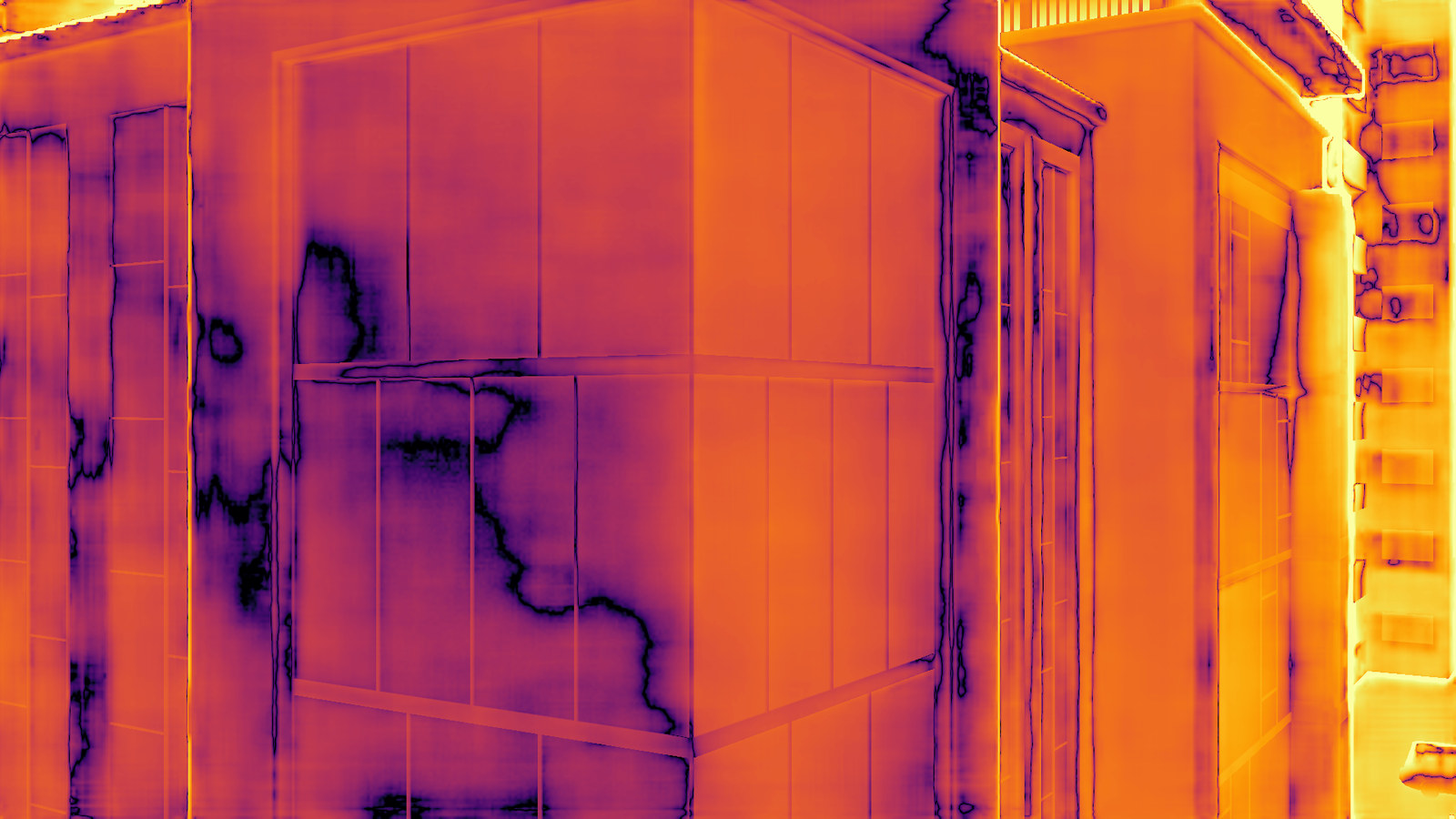}
      & \ccell{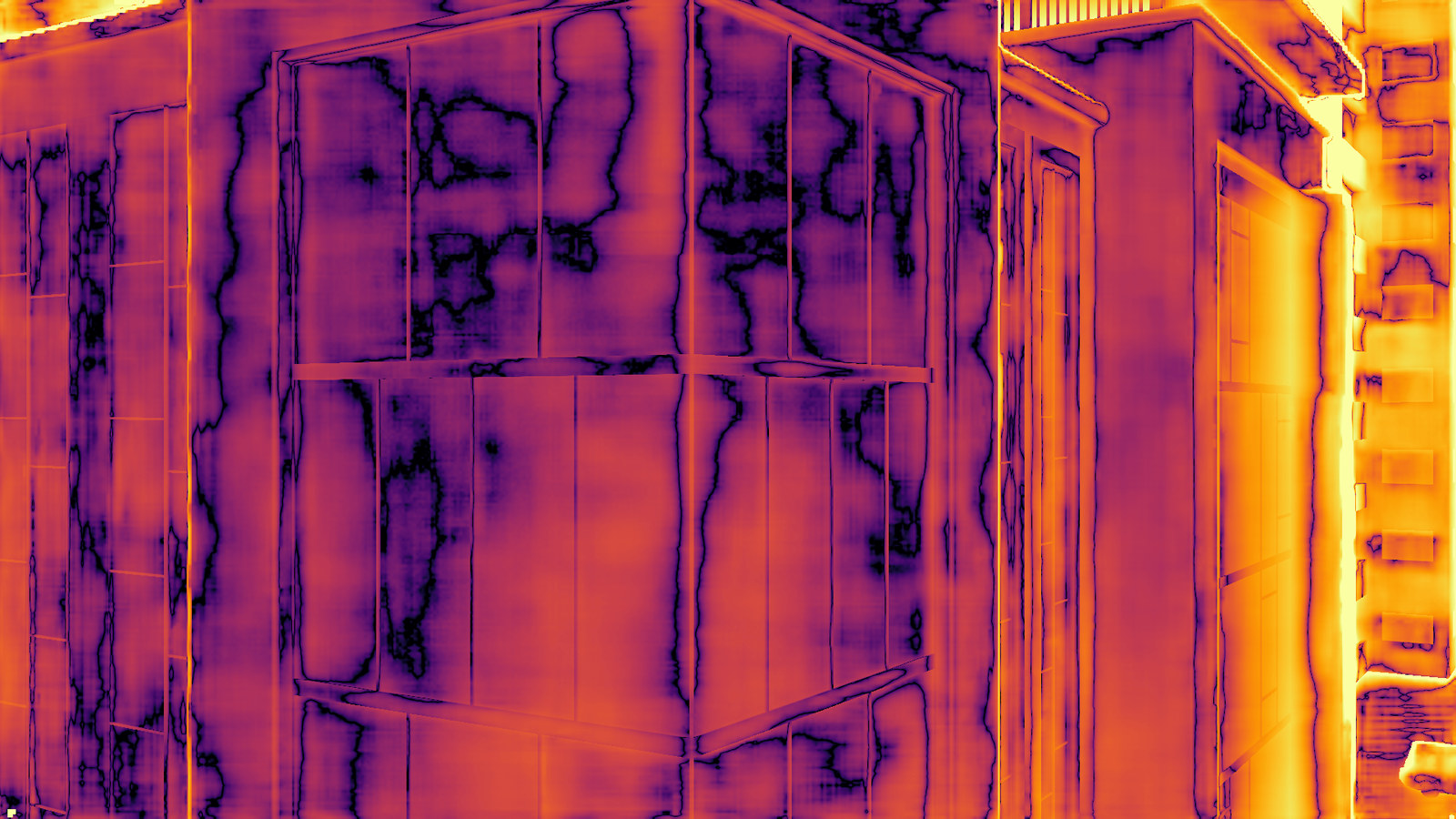} \\[2pt]

    \rlabel{Pi-Long}
      & \fcell{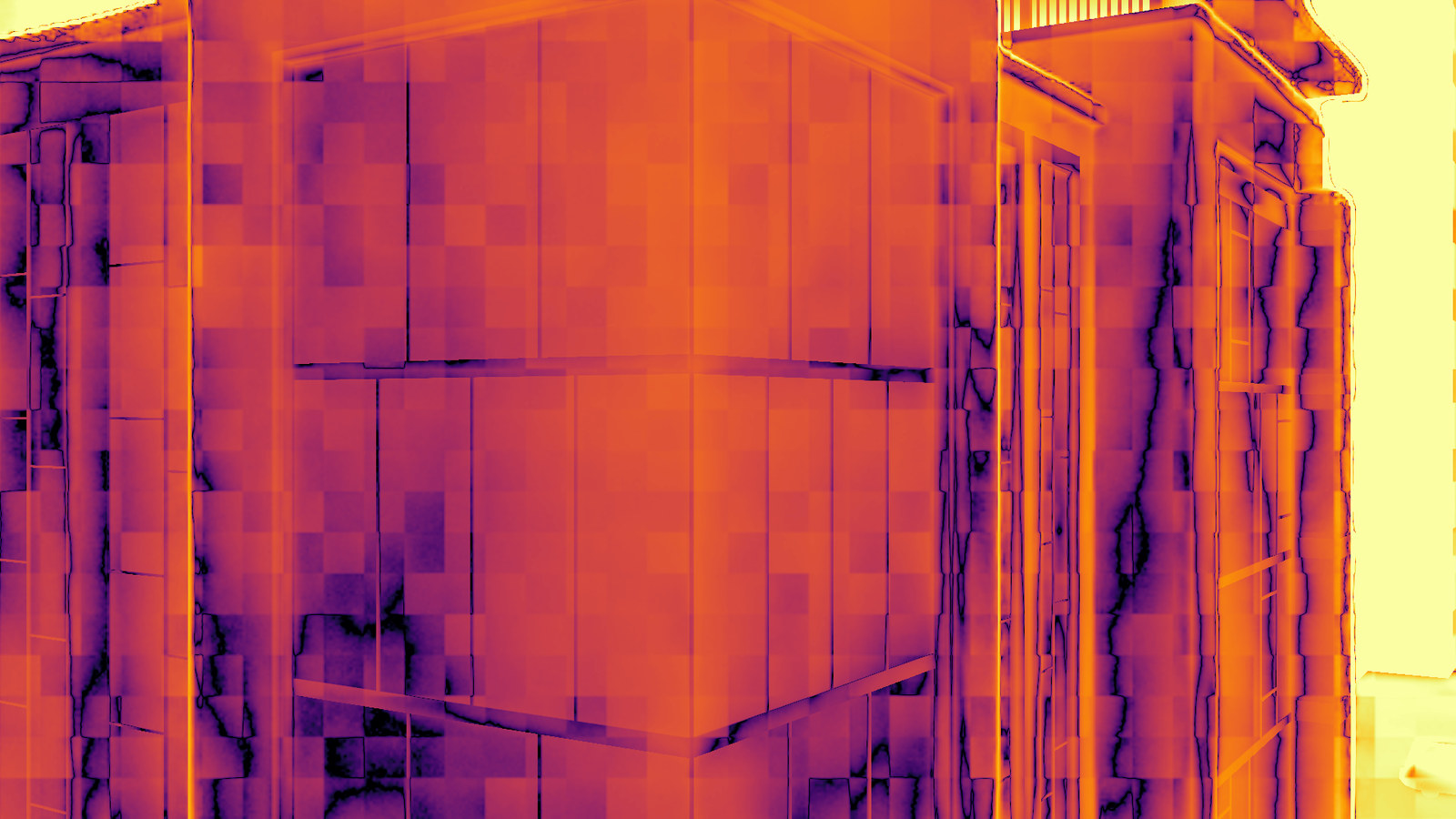}
      & \ccell{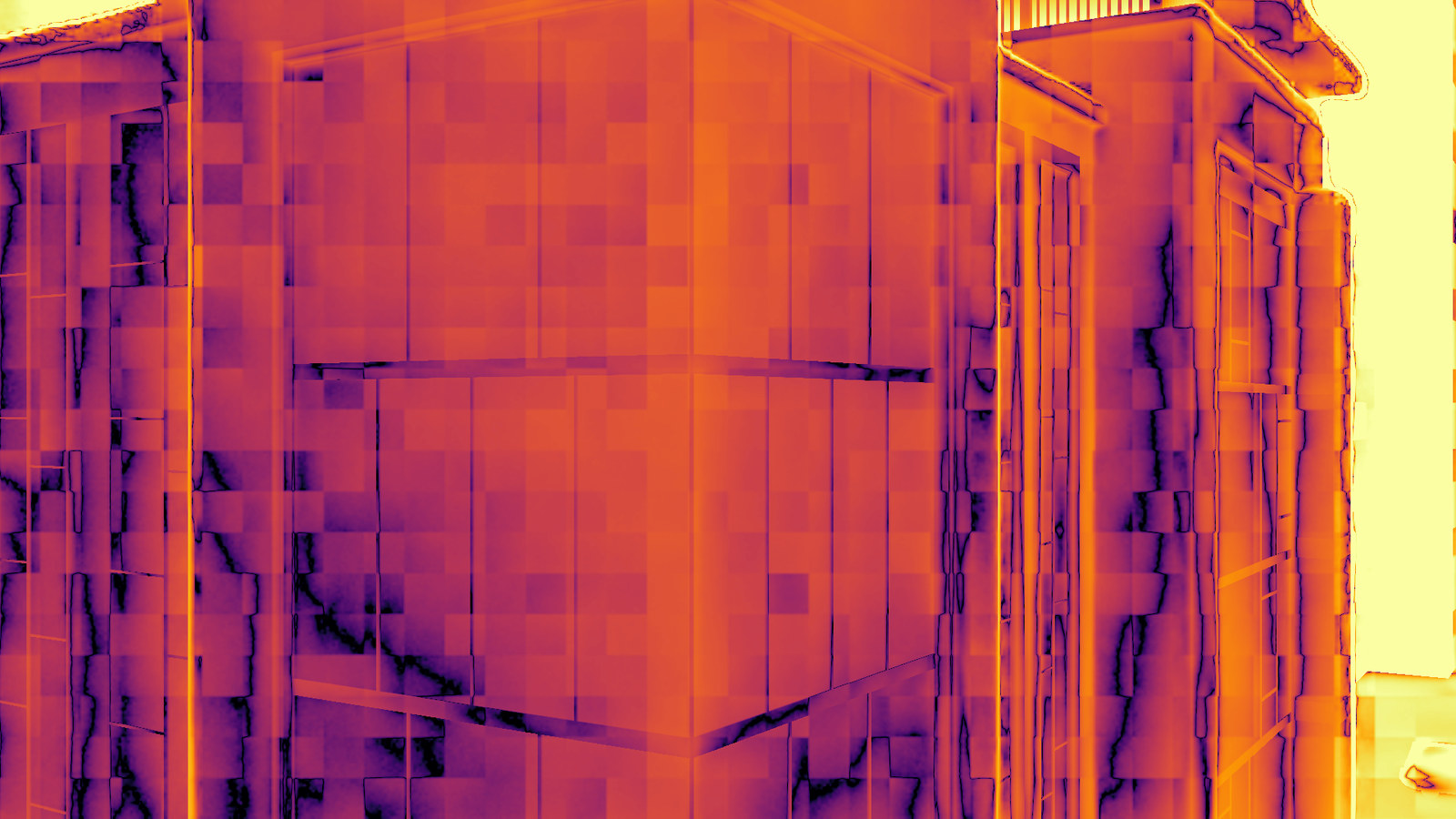}
      & \ccell{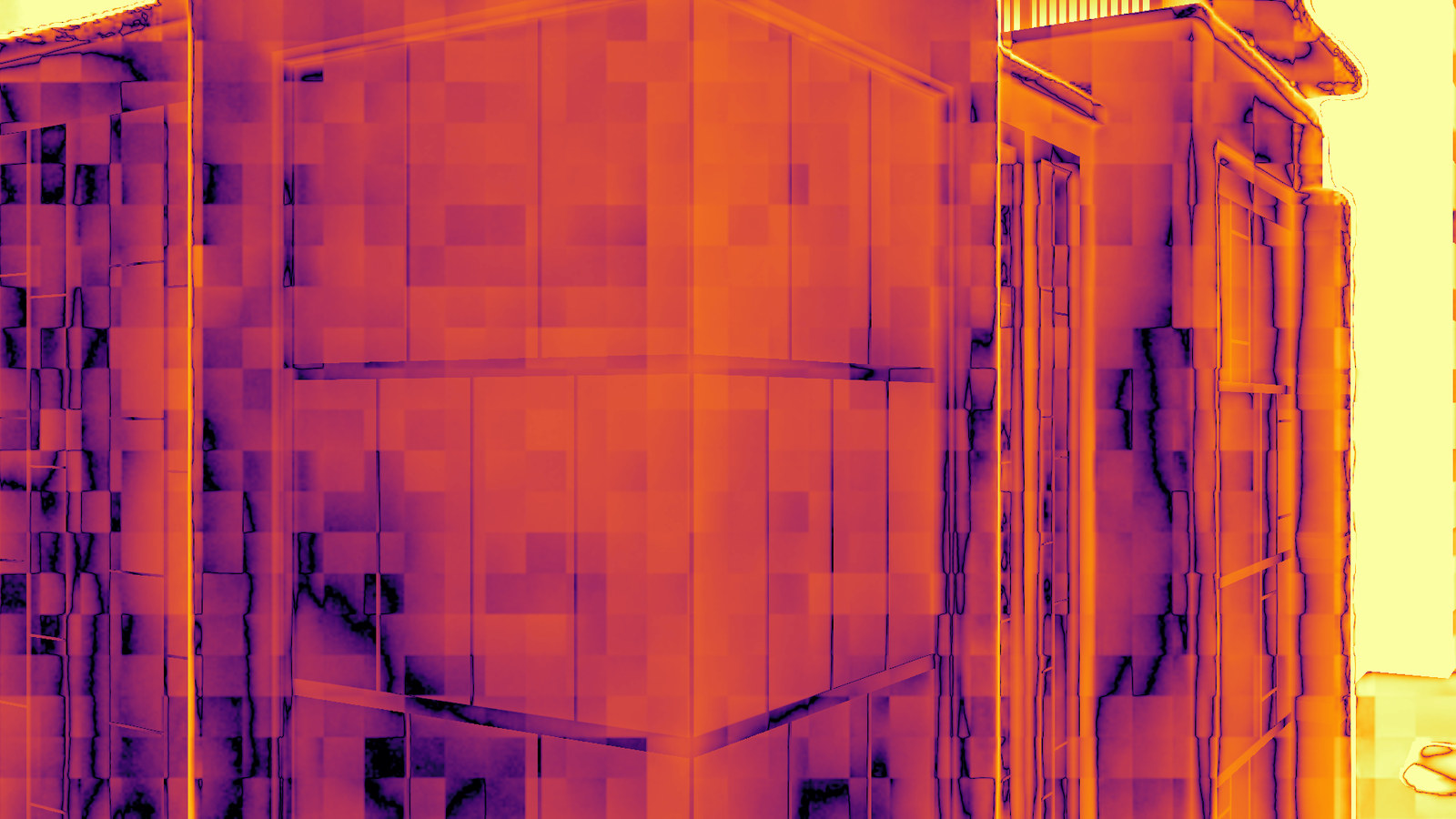}
      & \ccell{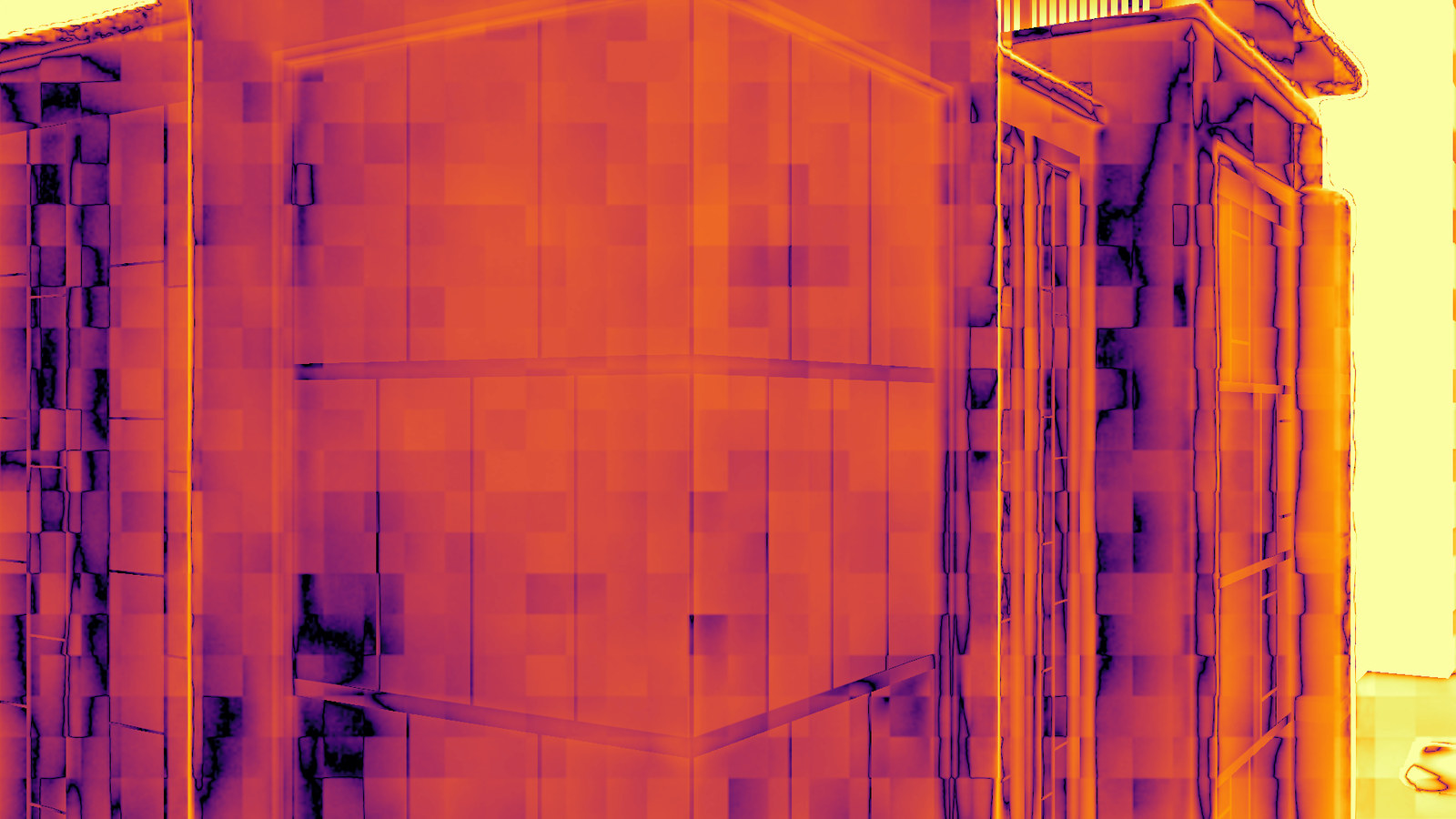}
      & \ccell{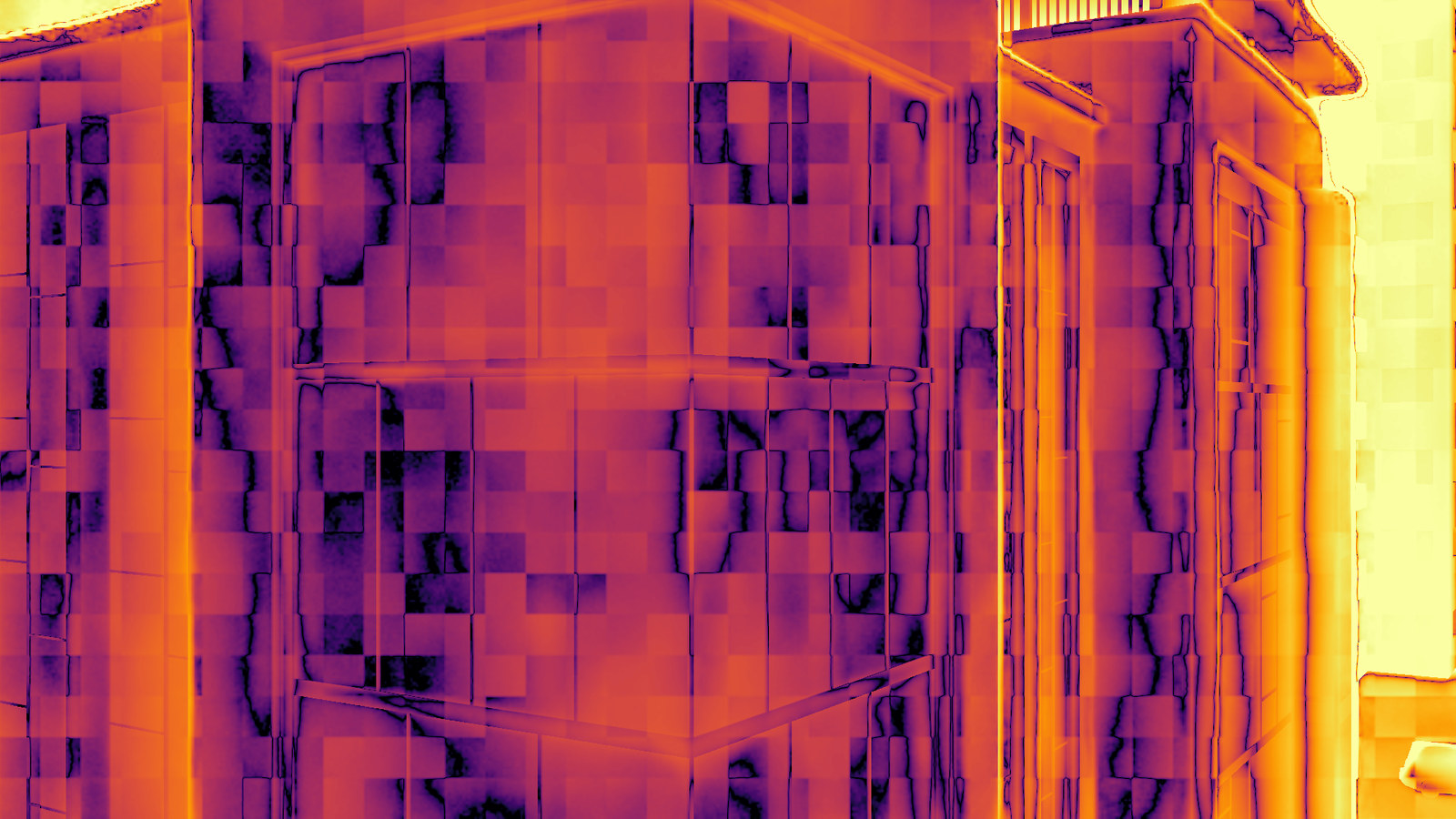} \\[2pt]

    \rlabel{VGGT-\\Long}
      & \fcell{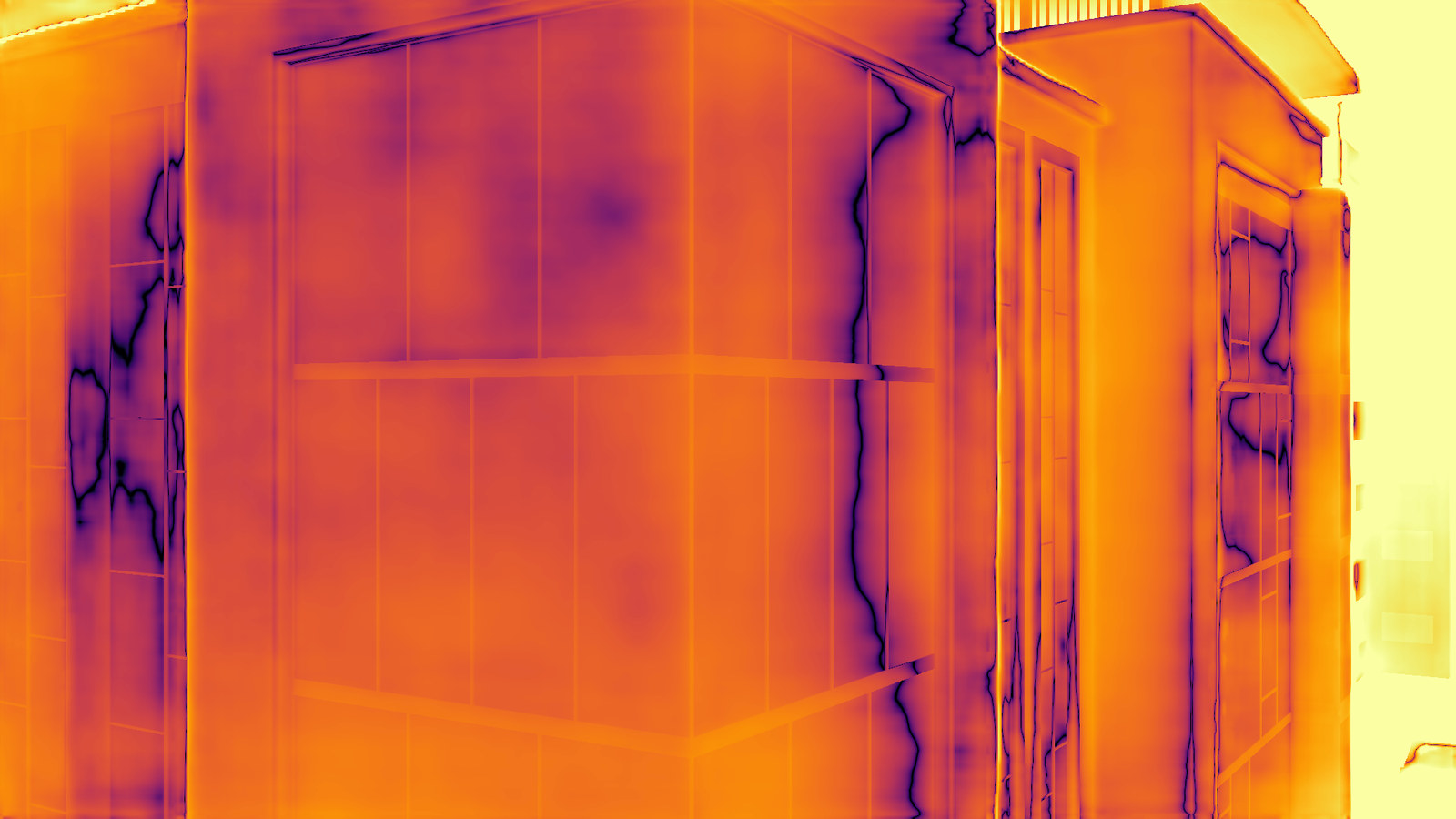}
      & \ccell{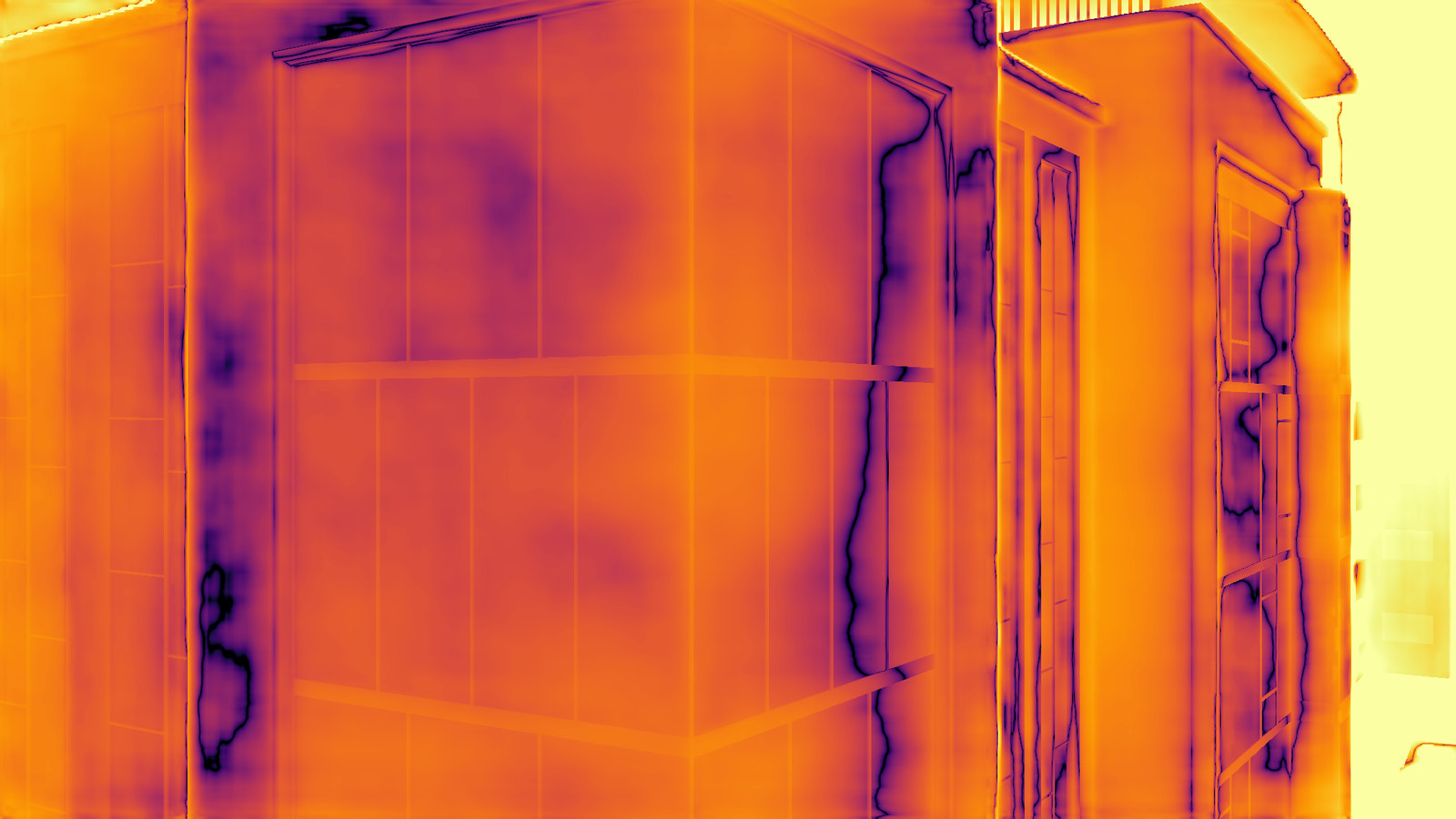}
      & \ccell{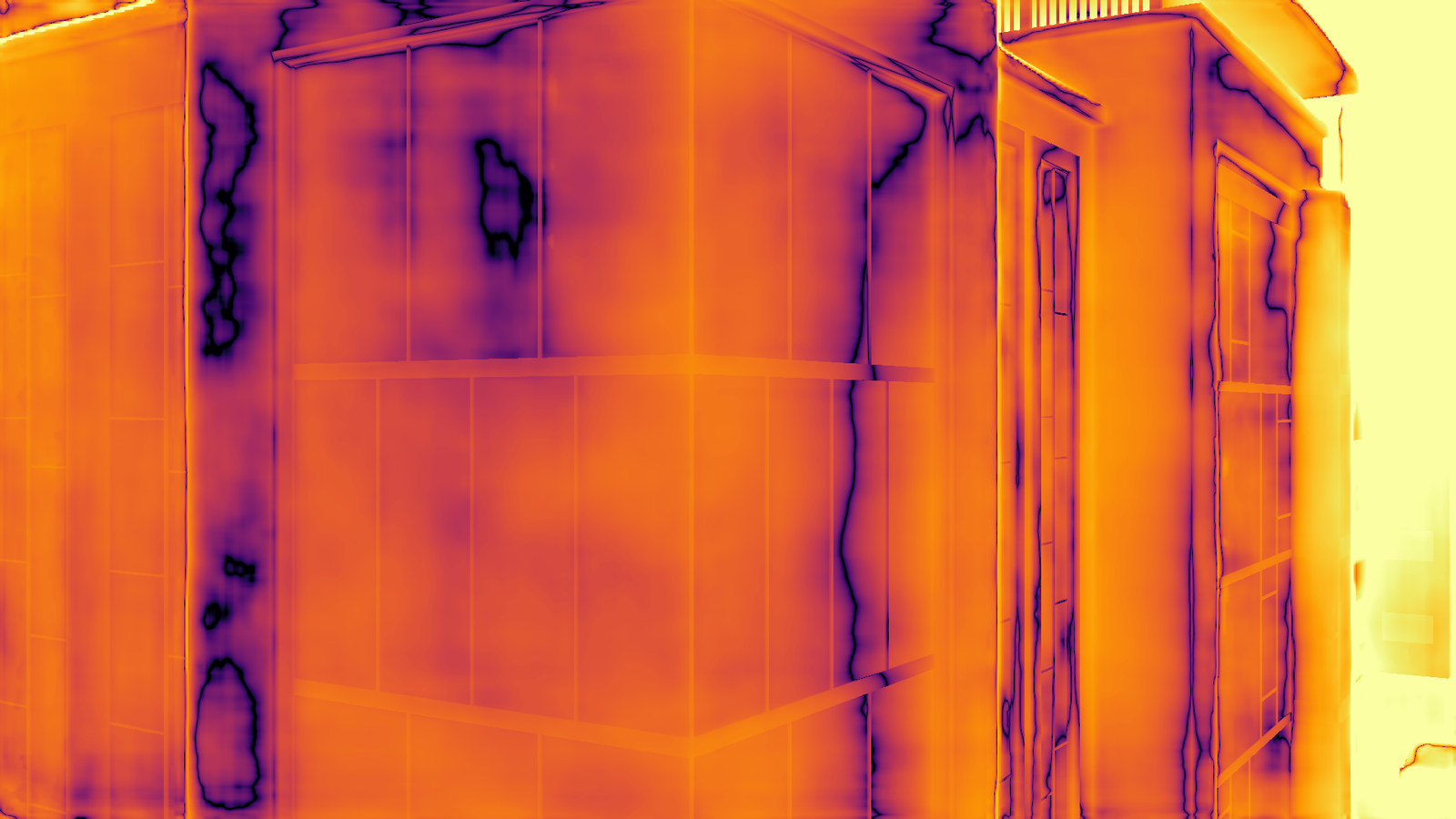}
      & \ccell{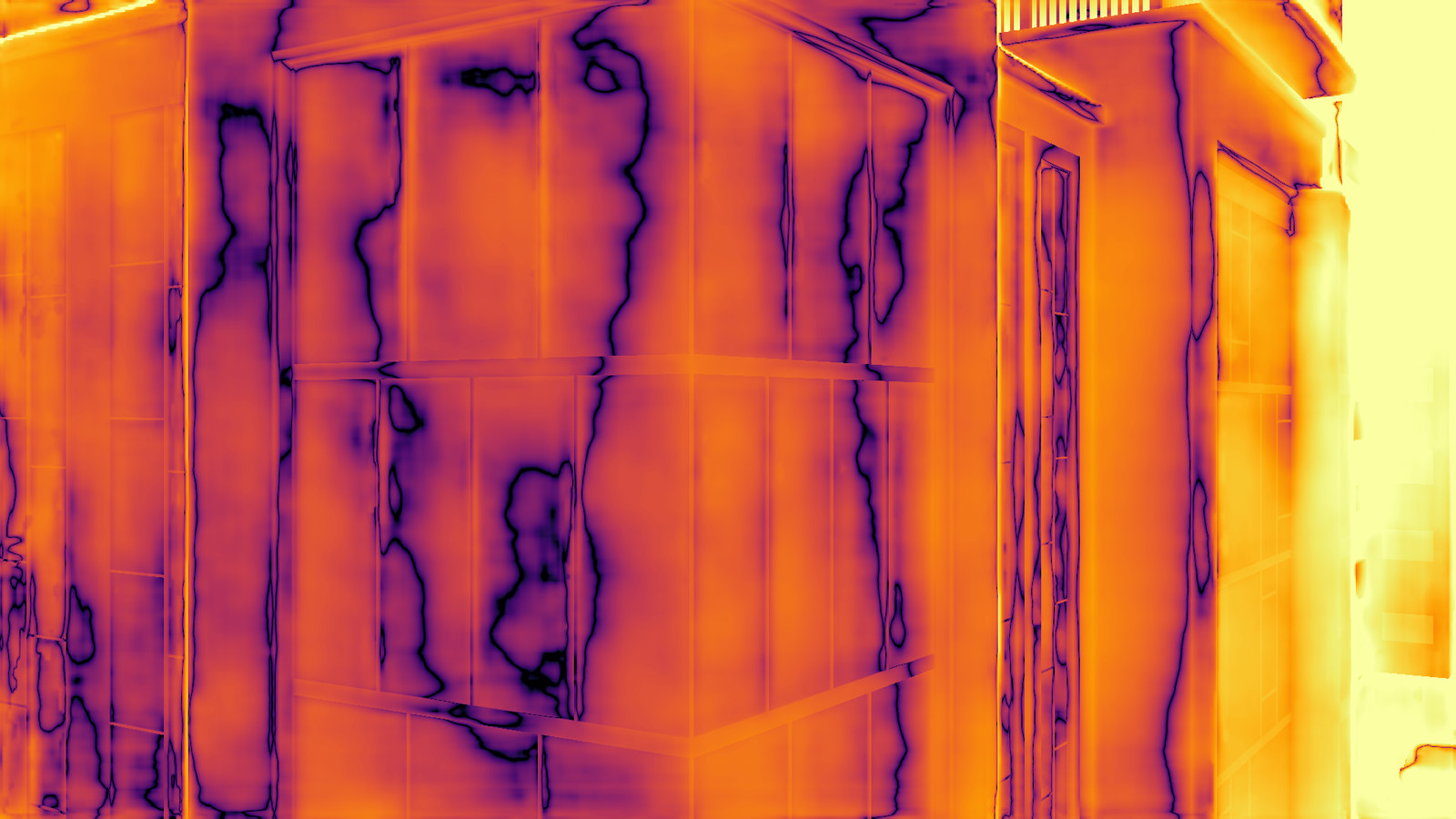}
      & \ccell{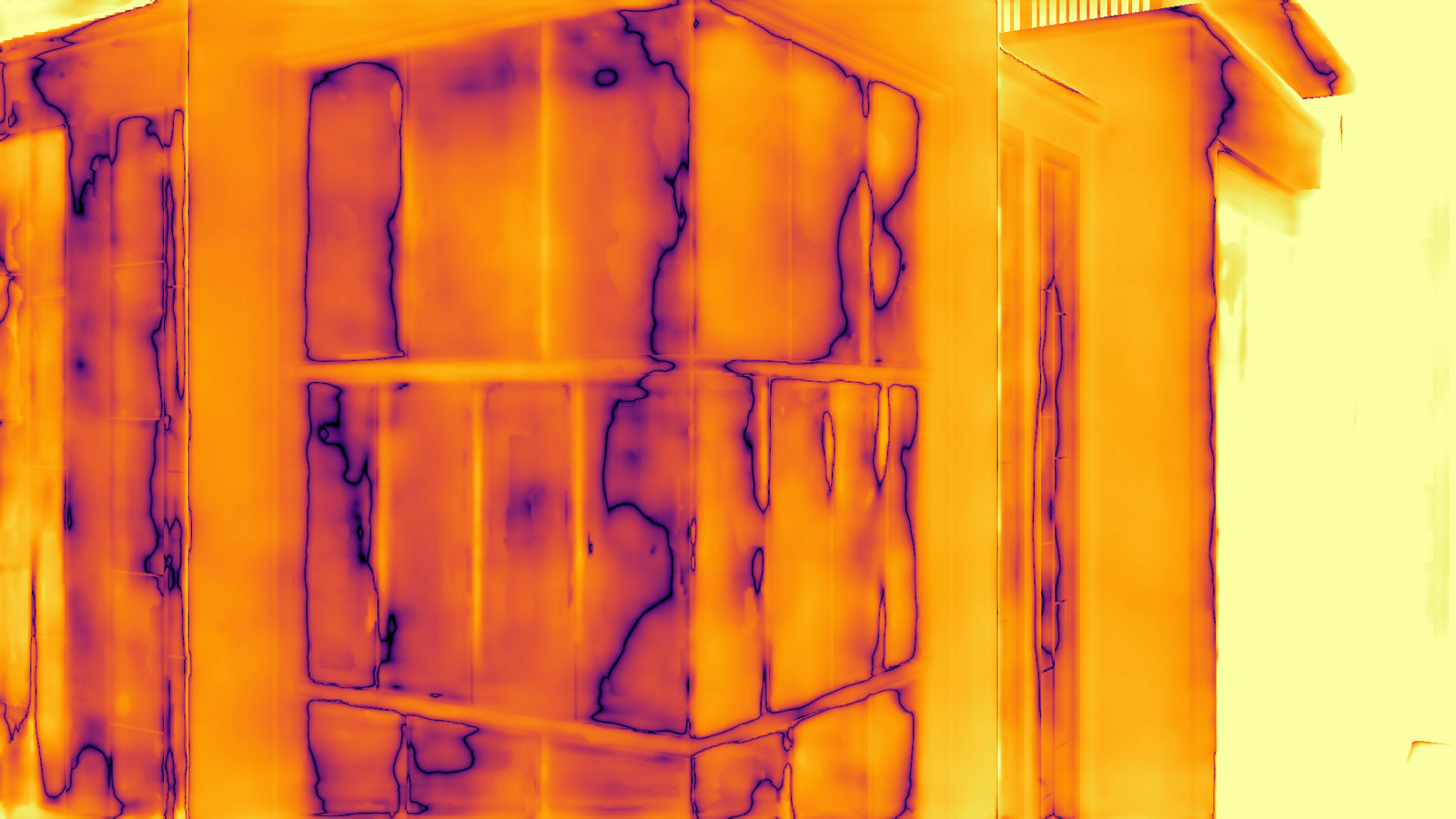} \\[2pt]

    \rlabel{VGGT-\\SLAM 2.0}
      & \fcell{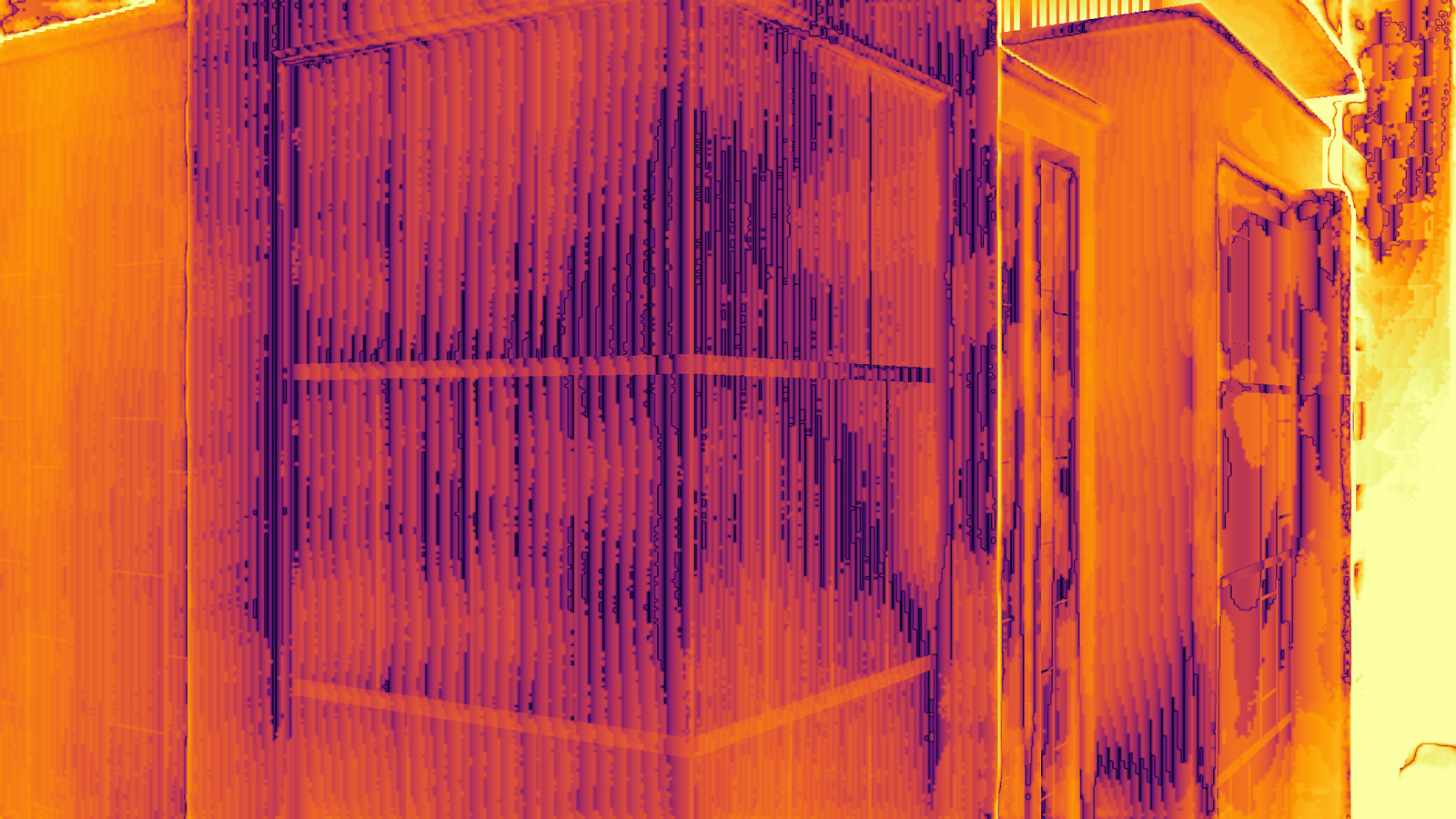}
      & \ccell{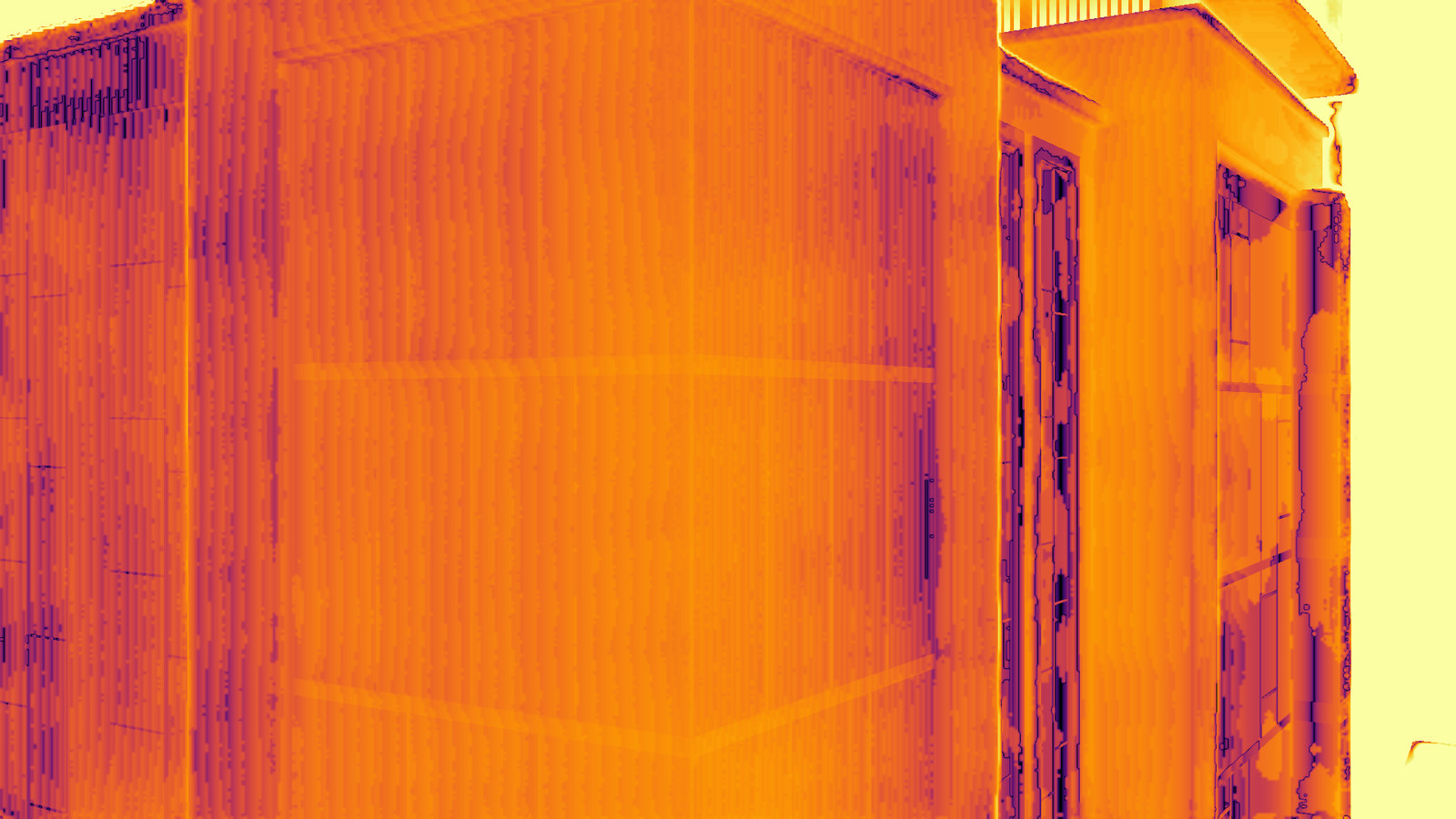}
      & \ccell{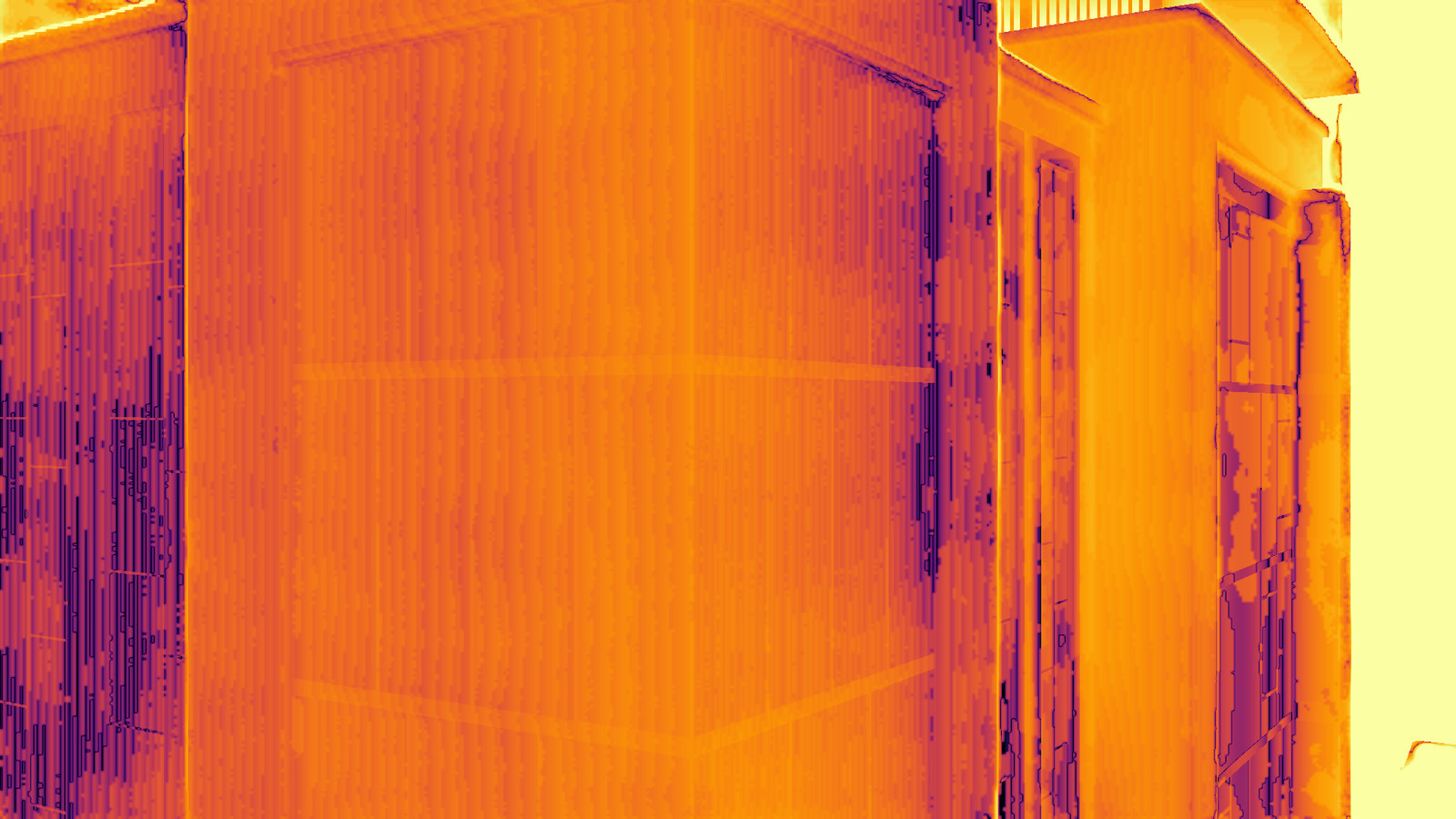}
      & \ccell{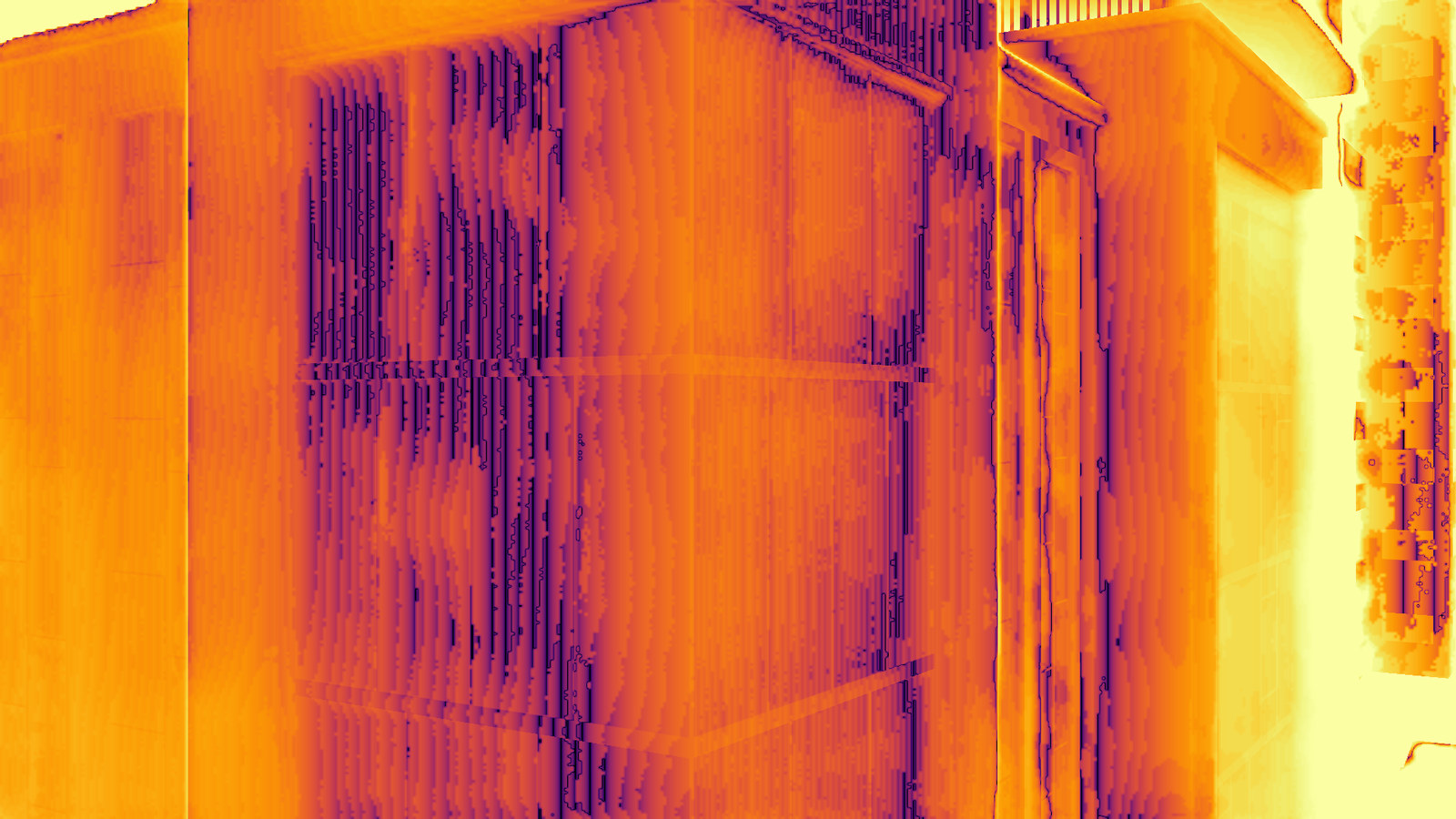}
      & \ccell{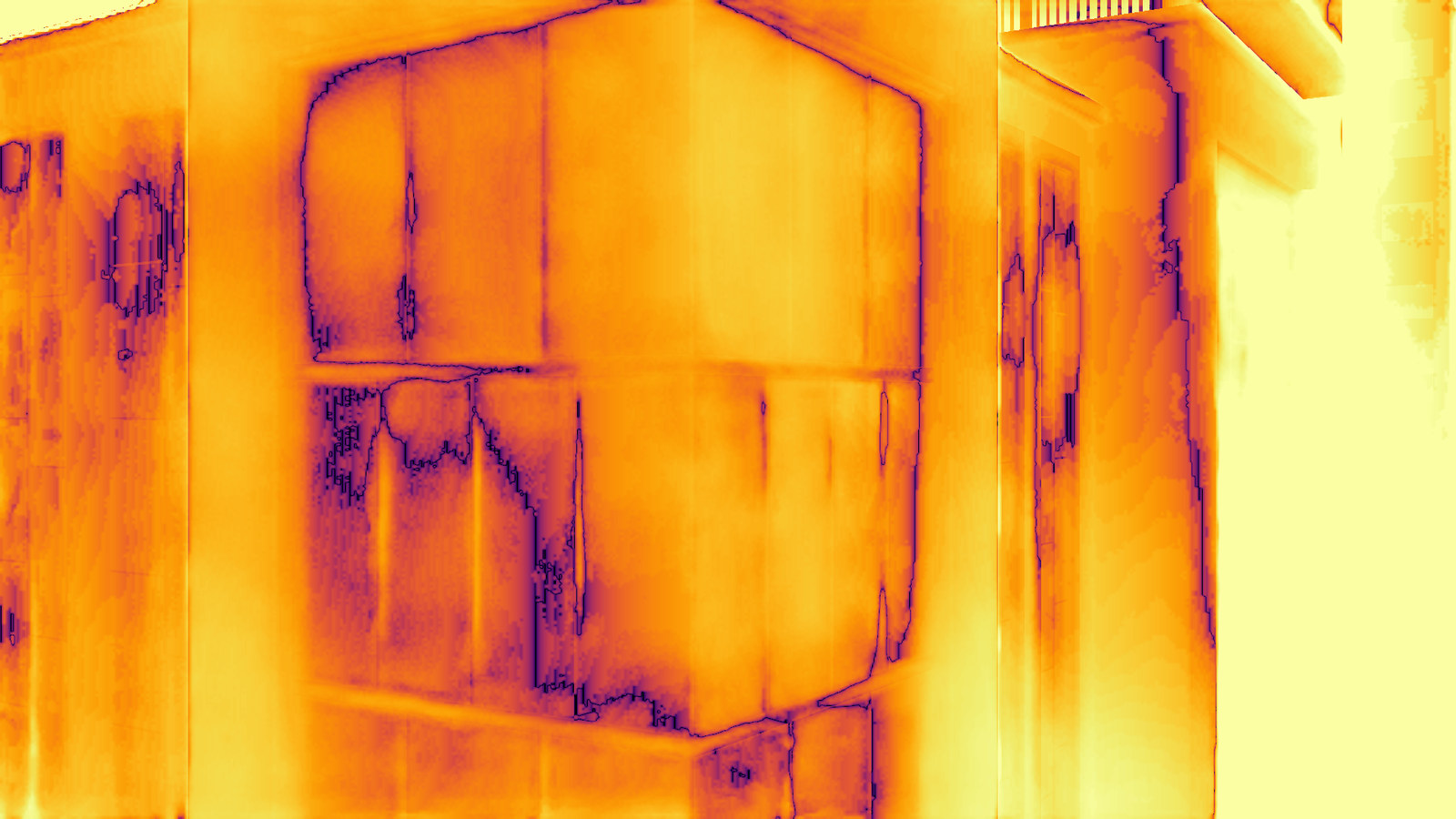} \\[6pt]

    & \multicolumn{5}{c}{\includegraphics[width=0.7\linewidth]{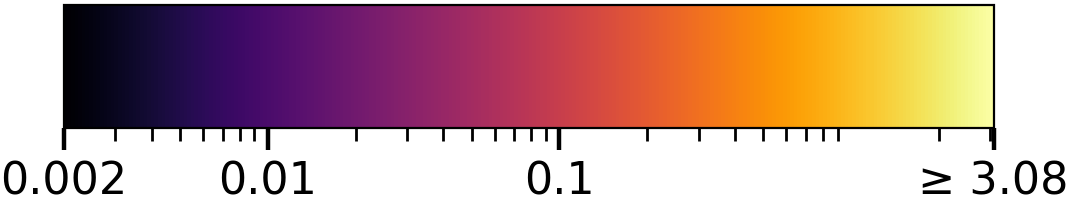}} \\[2pt]
    & \multicolumn{5}{c}{\small Scale-adjusted depth error (m)\rule{0pt}{2.5ex}} \\
  \end{tabular}

  \caption{\small
  Depth error for each SfM method after scales are adjusted to match the ground truth reference frame.
  This frame is 802 from the \textit{Office} scene with \textit{far} trajectory under the \textit{Midday} lighting condition. 
  As reflectivity increases, depth accuracy decreases on the right side of the building for all models, but most noticeably for VGGT-Long and VGGT-SLAM 2.0. 
  However, this decrease is substantially smaller than for COLMAP and GLOMAP.
}
    \label{fig:reflectivty_vs_depth_qualitative_learning}
\end{figure}

These broad trends are apparent in \Cref{fig:reflectivty_vs_depth_qualitative_sfm}, where COLMAP's and GLOMAP's depth maps become significantly less accurate and complete as reflectivity increases. 
Comparing \Cref{fig:reflectivty_vs_depth_qualitative_sfm} with \Cref{fig:reflectivty_vs_depth_qualitative_learning}, we see that the monocular depth-enabled methods are much more robust to reflectivity than the classical methods, as visible from the differences in the colour scales.
However, we can observe that some of these methods are still negatively affected by reflectivity.
In the case of VGGT-Long and VGGT-SLAM 2.0, this manifests on the glass along the right side of the building being less accurate at higher levels of reflectivity. 

\begin{figure}
    \centering
        \centering
        \includegraphics[width=\linewidth]{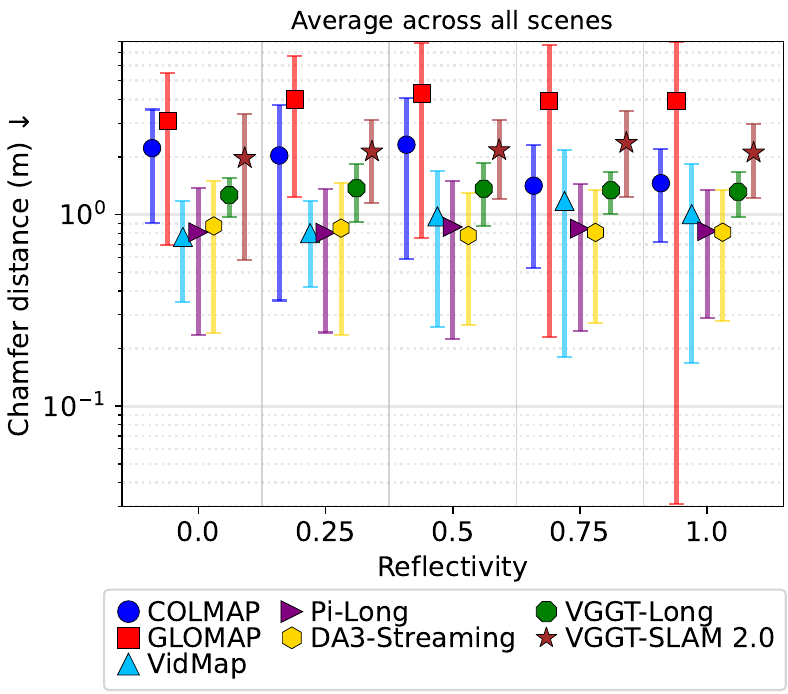}
        \caption{\small
        Average chamfer distance vs reflectivity across all sequences of \textit{Glass House} and \textit{Office}. 
        }
        \label{fig:ref_cd_ave}
\end{figure}

Despite these clear trends in the depth estimation results, the relationship between reflectivity and point cloud quality is more complex and unpredictable.
We note the ambiguous trends in \Cref{fig:ref_cd_ave} for several methods, such as GLOMAP, VidMap and VGGT-SLAM 2.0.
In \Cref{supp:reflectivity} of the supplementary information, we break down these results by scene.
For these methods, there are contradictory trends between \textit{Office} and \textit{Glass House}, which may explain some of the large variances.
On the other hand, for DA3-Streaming, Pi-Long and VGGT-Long, we observe that point cloud quality is not strongly affected by changes in reflectivity.
This is likely due to their training on reflective content.

\begin{figure}
    \centering
        \centering
        \includegraphics[width=\linewidth]{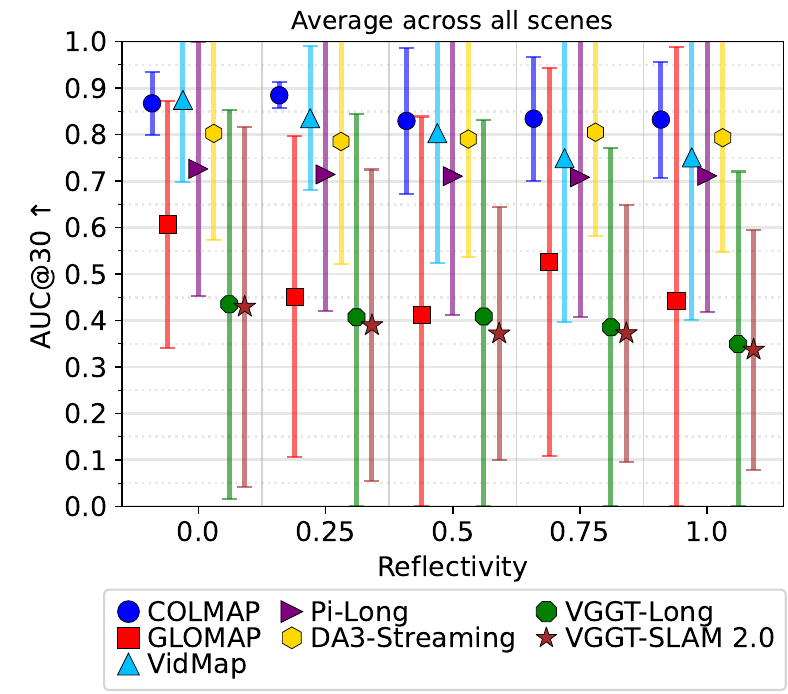}
        \caption{\small
        Average AUC@30 vs reflectivity across all sequences of \textit{Glass House} and \textit{Office}.}
        \label{fig:ref_AUC@30_ave}
\end{figure}

Interestingly, the relationship between camera pose accuracy and reflectivity is more straightforward, as shown in \Cref{fig:ref_AUC@30_ave}.
For Pi-Long and DA3-Streaming, pose accuracy remains robust to increases in reflectivity.
However, for VGGT-Long, VGGT-SLAM 2.0 and VidMap, there is an overall trend of pose accuracy decreasing as reflectivity increases.
On the other hand, the purely classical approaches of COLMAP and especially GLOMAP show less predictable behaviour with no strong trend but increasing standard deviations as reflectivity increases.
In \Cref{supp:reflectivity} of the supplementary information, we break down these results by scene, and we note that \textit{Glass House} and \textit{Office} COLMAP and GLOMAP show different trends on each scene.

\bmhead{Effect of soiling}

\begin{figure}
    \centering
        \centering
        \includegraphics[width=\linewidth]{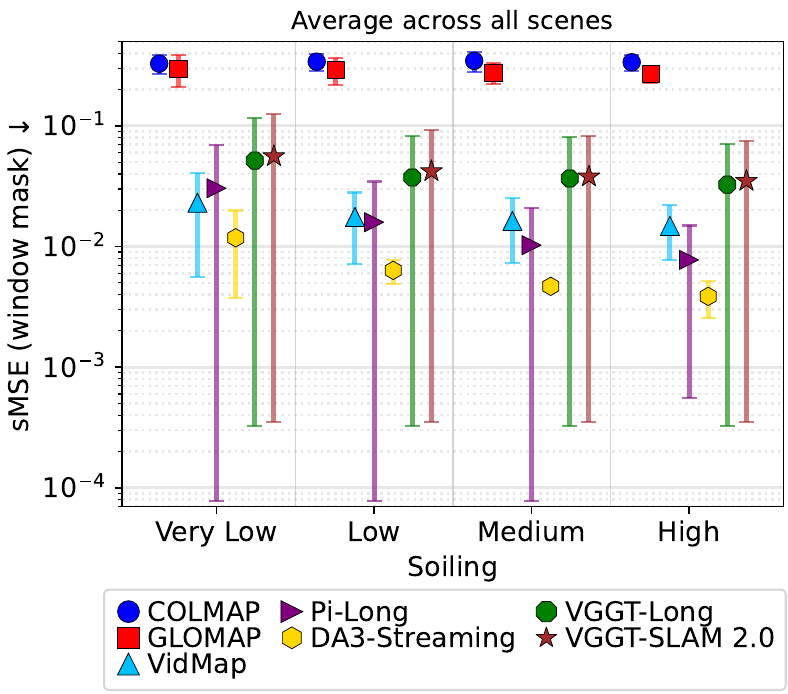}
        \caption{\small
        Average sMSE vs. soiling across all sequences of \textit{Glass House} and \textit{Office}.
        }
        \label{fig:soiling_smse_ave}
\end{figure}

Unlike the effect of reflectivity on the depth estimation results, which primarily affects classical approaches, soiling seems to affect monocular depth-enabled methods to a greater extent, as shown in \Cref{fig:soiling_smse_ave}.
Increasing soiling improves sMSE for all learning-based methods and for VidMap, which we set to use \ac{DA3} in its monocular depth estimation stage.
The greater effect on monocular depth-enabled methods than classical methods seems counter-intuitive, considering that soiling reduces the reflectivity of the windows by simulating randomly distributed smudges, and reflectivity has a stronger effect on the classical approaches.
However, these smudges are highly similar and have little texture. 
Therefore, the individual smudges may not be sufficiently distinct for the patch matching-based \ac{MVS} stages of COLMAP and GLOMAP to properly take advantage of.

\begin{figure}
    \centering
        \centering
        \includegraphics[width=\linewidth]{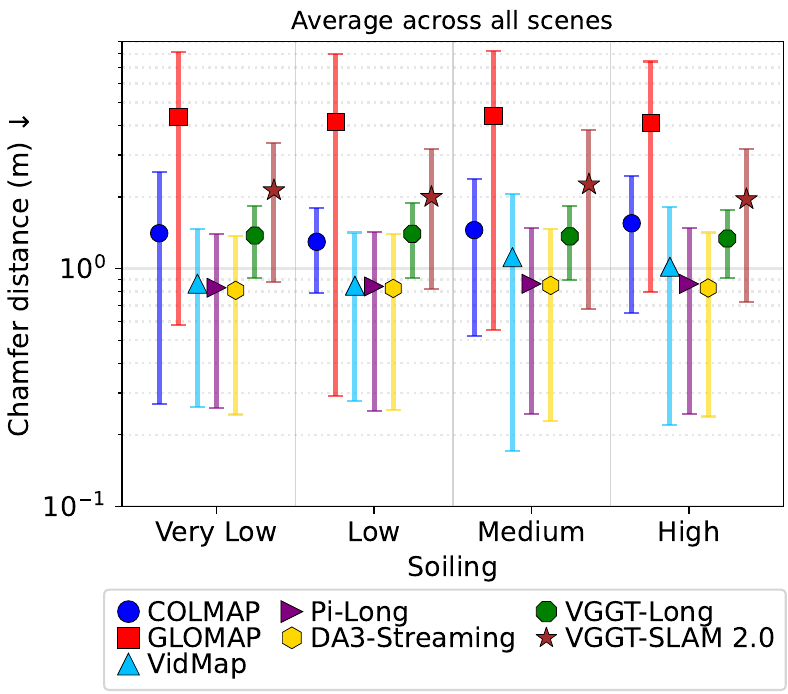}
        \caption{\small
        Average Chamfer distance vs. soiling across all sequences of \textit{Glass House} and \textit{Office}.
        }
        \label{fig:soiling_chamfer_ave}
\end{figure}

Although increasing soiling seems to improve the depth estimation of learning-based methods, it has a limited effect on their performance at the scene level.
For several of these methods, such as DA3-Streaming, VGGT-Long, and Pi-Long, their average Chamfer distance remains largely unaffected by soiling, as shown in \Cref{fig:soiling_chamfer_ave}.
VGGT-SLAM 2.0 shows ambiguous results in terms of the effect of soiling on point cloud quality; medium and very low soiling levels have lower-quality point clouds than low and high levels of soiling. 
Similarly, GLOMAP and VidMap also have ambiguous results, with medium yielding the worst point clouds.
In \Cref{supp:soiling} of the supplementary information, we break down these results by scene.
We note that some methods, such as COLMAP, show different trends on each scene.

\begin{figure}
    \centering
        \centering
        \includegraphics[width=\linewidth]{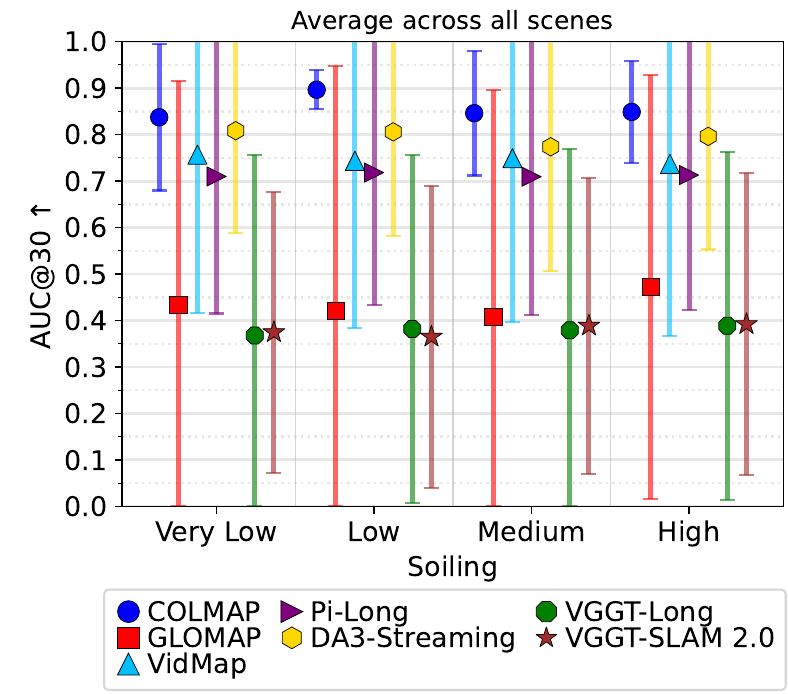}
        \caption{\small
        Average AUC@30 vs. soiling across all sequences of \textit{Glass House} and \textit{Office}.
        }
        \label{fig:soiling_poses_ave}
\end{figure}

As shown in \Cref{fig:soiling_poses_ave}, the impact of soiling on pose accuracy is relatively small for most methods, except for COLMAP and GLOMAP, which behave unpredictably.
In \Cref{supp:soiling} of the supplementary information, we break down these results by scene, and we find that COLMAP and GLOMAP show different relationships between soiling and pose accuracy between scenes. 
These results suggest that further benchmarking is required to properly understand the effect of surface soiling on pose accuracy and point cloud quality.

\bmhead{Effect of lighting conditions}

\begin{figure}
    \centering
        \centering
        \includegraphics[width=\linewidth]{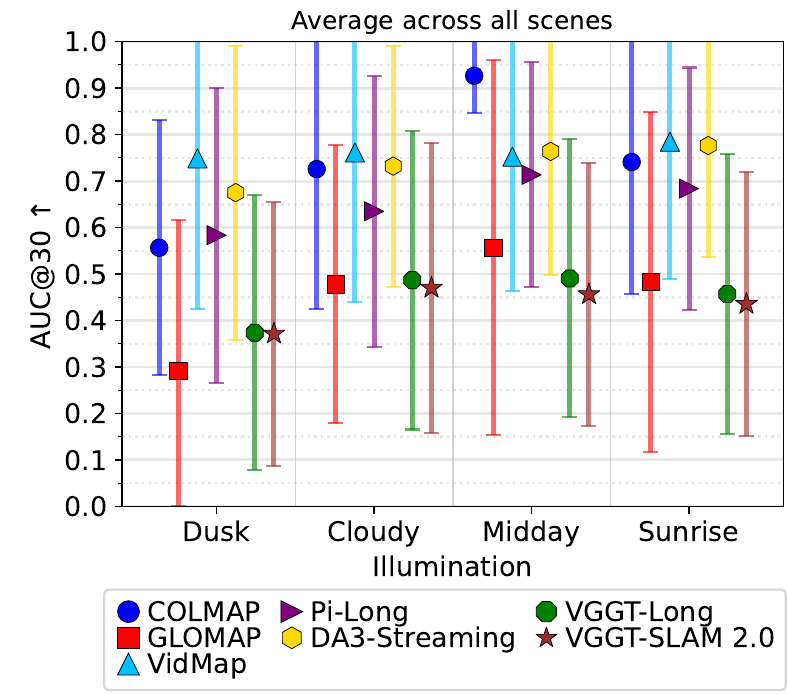}
        \caption{\small
        AUC@30 vs. lighting across all sequences in the dataset.
        }
        \label{fig:lighting_auc_ave}
\end{figure}

\begin{figure}
    \centering
        \centering
        \includegraphics[width=\linewidth]{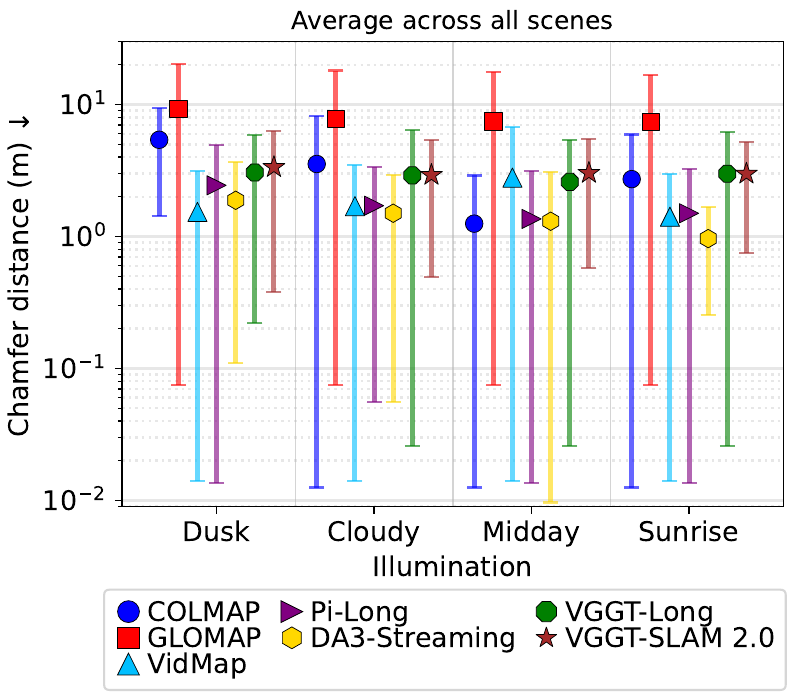}
        \caption{\small Chamfer distance vs. lighting across all sequences in the dataset.
        }
        \label{fig:lighting_chamfer_ave}
\end{figure}

\begin{figure}
    \centering
        \centering
        \includegraphics[width=\linewidth]{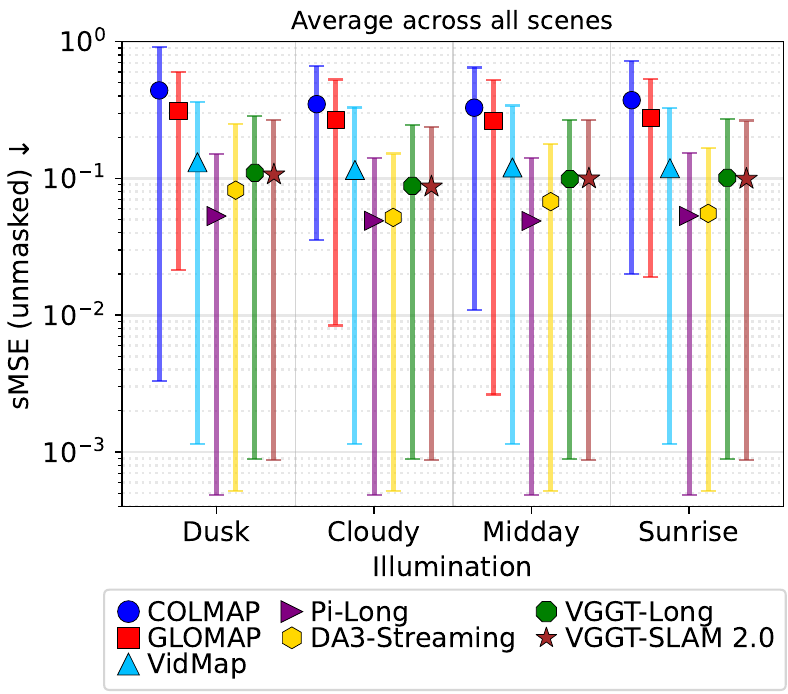}
        \caption{\small sMSE vs. lighting across all sequences in the dataset.
        }
        \label{fig:lighting_smse_ave}
\end{figure}

From the results shown in \Cref{fig:lighting_auc_ave,fig:lighting_chamfer_ave,fig:lighting_smse_ave}, we can clearly see that lighting strongly affects 3D reconstruction results. 
Averaging across all sequences, every method produced the lowest pose accuracy with the \textit{Dusk} environment map, as shown in \Cref{fig:lighting_auc_ave}.
Additionally, for most methods, \textit{Dusk} produces the highest chamfer distances and highest sMSE metrics, shown in \Cref{fig:lighting_chamfer_ave,fig:lighting_smse_ave}. 
This is likely because \textit{Dusk} simulates low-light conditions, which usually pose a greater challenge to 3D reconstruction approaches.
VidMap is most robust to these low-light conditions, while COLMAP and GLOMAP are most affected by them.

Intuitively, one might expect that the even lighting under the \textit{Cloudy} environment map would yield the best results. 
This is true for depth estimation for every method except COLMAP and GLOMAP, which produce their second-best results with \textit{Cloudy} and perform best with \textit{Midday}. 
However, this is not the case for the scene-level metrics.
All methods except for VidMap and VGGT-SLAM 2.0 yield more accurate camera pose estimates and higher-quality point clouds under the harsh lighting conditions of \textit{Midday}, and some, such as DA3-Streaming and VidMap, do best under \textit{Sunrise}.
Interestingly, unlike the other methods, VGGT-SLAM 2.0 performs best under \textit{Cloudy} conditions.

In terms of scene-level metrics, VidMap and DA3-Streaming do best under \textit{Sunrise} lighting conditions, with VidMap producing the most accurate poses of all methods and DA3-Streaming producing the best point clouds.
On the other hand, Pi-Long, VGGT-Long, COLMAP and GLOMAP do best under \textit{Midday} conditions. 
Surprisingly, under \textit{Midday} conditions, COLMAP outperforms all other methods in terms of chamfer distance and pose accuracy.

In \Cref{supp:lighting} of the supplementary information, we break down these results by scene.

\bmhead{Open Challenges and Failure Cases}

Thin structures remain an open challenge for 3D reconstruction methods, and the \textit{Crane} scene exemplifies this, since the crane's boom is a lattice structure of thin members. 
COLMAP, VidMap, VGGT, DA3-Streaming, $\pi^3$-X and Pi-Long recorded their highest Chamfer distance on this scene, and GLOMAP, VGGT-Long and VGGT-SLAM 2.0 recorded their second-highest Chamfer distance, as shown in \Cref{tab:Scene_Level_Results_Chamfer}.
All methods that successfully ran on \textit{Crane} performed worst on this scene for both depth maps and camera pose accuracy, as shown in \Cref{tab:Scene_Level_Results_AUC,tab:DepthResults_sMSE}.
Qualitative plots of the \textit{Crane} scene are included in \Cref{supp:results} of the supplementary information.

\begin{figure}
    \centering
    \includegraphics[width=\linewidth]{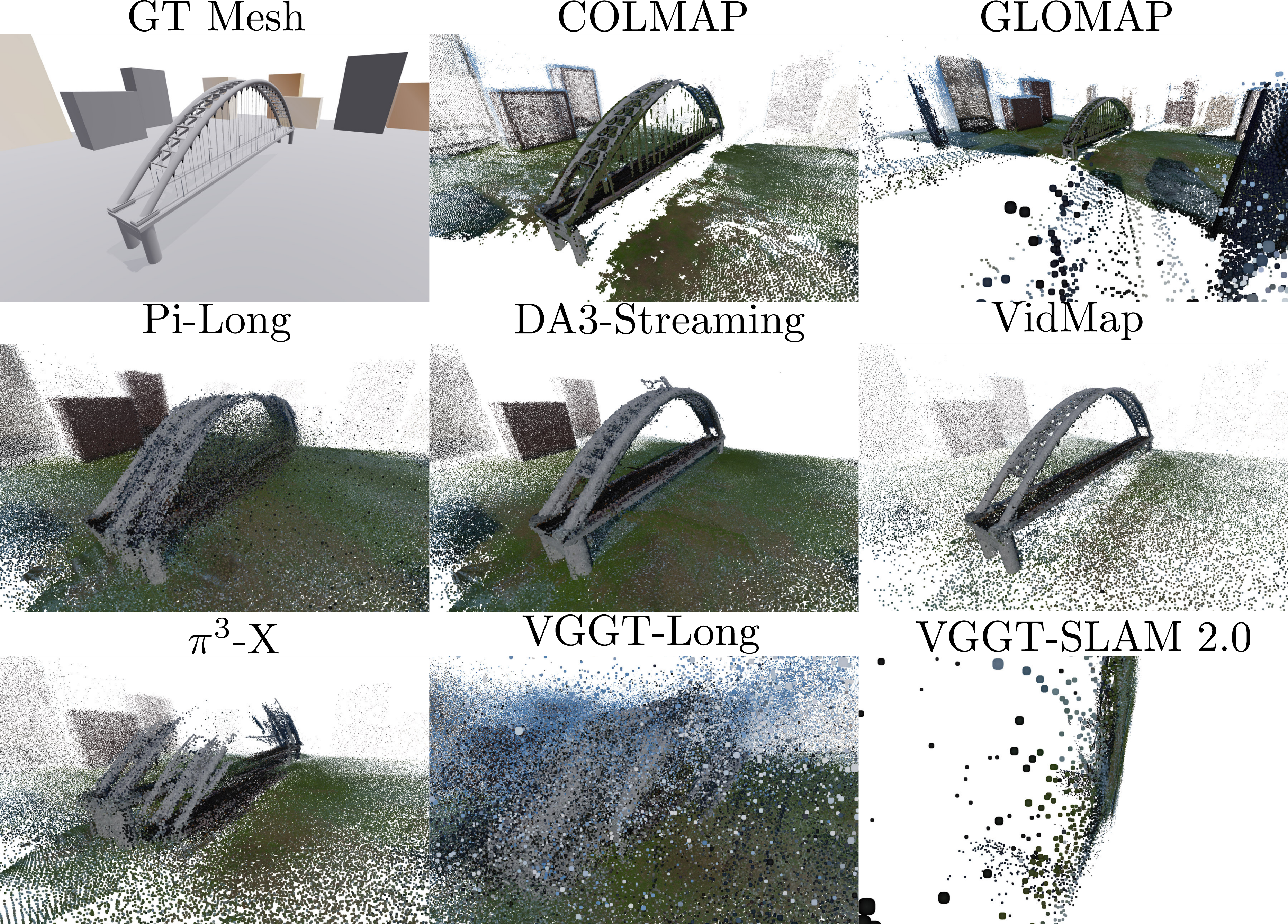}
    \caption{ \small
    Point clouds of the \textit{Bridge} scene, produced by methods with the \textit{close} capture pattern and the \textit{Midday} lighting condition.
    The ground truth mesh is included for comparison.
    Note that the colours of the ground truth mesh are inaccurate.
    No monocular depth-enabled method properly reconstructs the cables, and the classical \ac{SfM} methods fail to colour these points correctly.
    }
    \label{fig:bridge_point_clouds}
\end{figure}

\begin{figure}
    \centering
    \includegraphics[width=\linewidth]{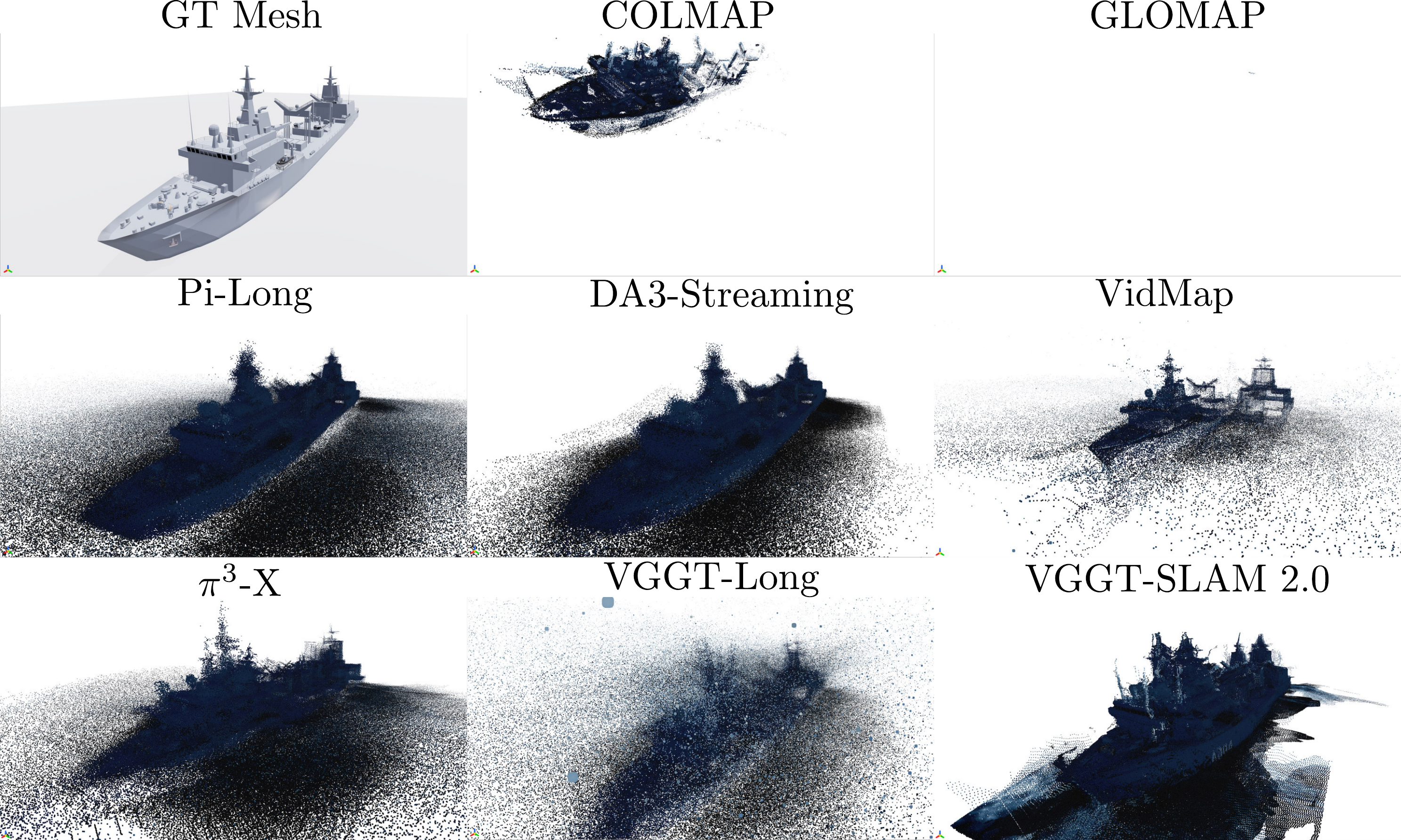}
    \caption{ \small
    Point clouds of the \textit{Ship} scene, produced by methods with the \textit{medium} capture pattern and the \textit{Dusk} lighting condition.
    The ground truth mesh is included for comparison.
    Note that the colours of the ground truth mesh are inaccurate.
    No method except for VGGT-SLAM 2.0 properly reconstructs the fine details of this scene, such as the towers and antennae.
    However, VGGT-SLAM 2.0 also duplicates key parts of the ship.
    VidMap gets the towers correct, but the two halves of the ship are misaligned.
    GLOMAP's point cloud is entirely incomplete, except for a speck which is barely visible from this viewpoint.
    }
    \label{fig:ship_point_clouds}
\end{figure}

We see similar results on the cables of \textit{Bridge} (see \Cref{fig:bridge_point_clouds}), another example of thin structures.
No monocular depth-enabled method properly reconstructed the cables; only COLMAP and GLOMAP were able to do so.
Note that GLOMAP and COLMAP colour the cable points incorrectly, making them green, when they are actually grey.
These results are also corroborated by the point clouds in \Cref{fig:construction_point_clouds_pics}, where the roof frame is often poorly reconstructed by most methods.
Another example of fine details is the \textit{Ship} scene such as the antennae and towers, shown in \Cref{fig:ship_point_clouds}.
Most of the learning-based approaches reconstruct these details coarsely, getting their overall shape right, but not the finer details.
VidMap reconstructs the towers correctly, but the back of the ship is misaligned compared to the front.
COLMAP's reconstruction is misaligned to the ground truth reference frame and roughly half of the ship is missing.
GLOMAP's reconstruction is almost entirely incomplete.
Interestingly, in both scenes, VGGT-Long produces very noisy point clouds around these fine details.

\begin{figure}[t]
  \centering
  \setlength{\tabcolsep}{0pt}
  \renewcommand{\arraystretch}{0}
  \newlength{\imgw}
  \setlength{\imgw}{0.325\linewidth}
  \newcommand{\cell}[1]{\includegraphics[width=\imgw]{#1}}
  \newcommand{\hdr}[1]{\makebox[\imgw][c]{\small #1\rule{0pt}{3ex}}}
  \newsavebox{\cbbox}
  \newlength{\cboff}
  \newlength{\cbnudge}
  \setlength{\cbnudge}{40pt}   
  \sbox{\cbbox}{%
    \shortstack{%
      \includegraphics[width=1.9\imgw]{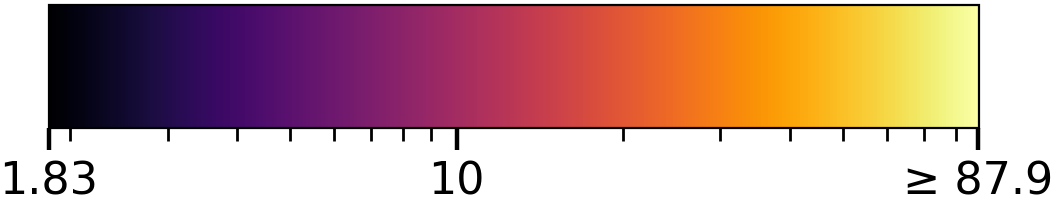}\\[2pt]
      \small Scale-adjusted depth error (m)}}
  \setlength{\cboff}{\dimexpr\imgh-\ht\cbbox+\cbnudge\relax}
  \settoheight{\imgh}{\cell{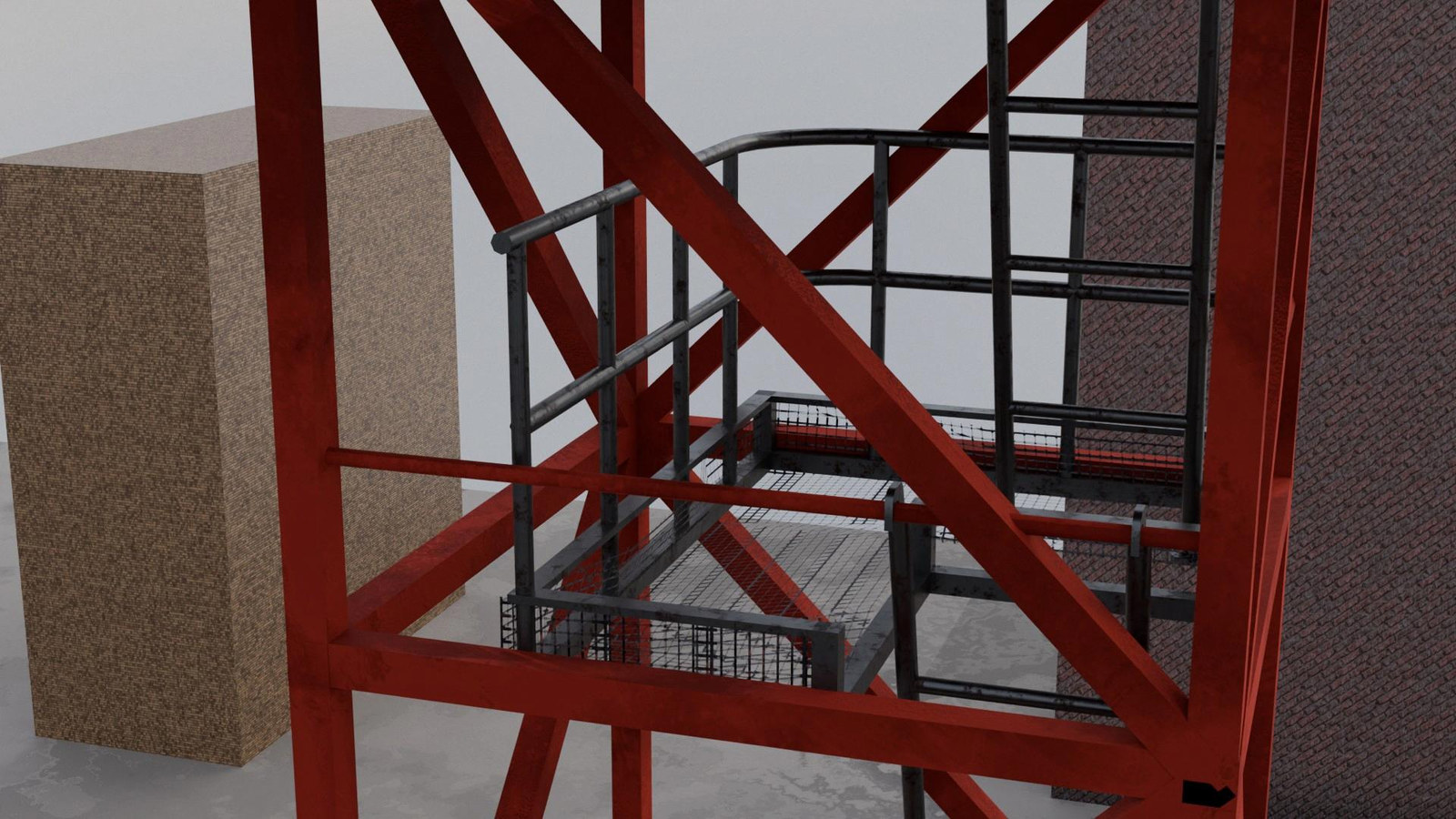}}

  \begin{tabular}{@{}c c c@{}}
    \hdr{Image} & \hdr{COLMAP} & \hdr{GLOMAP} \\
    \cell{crane_image_37} & \cell{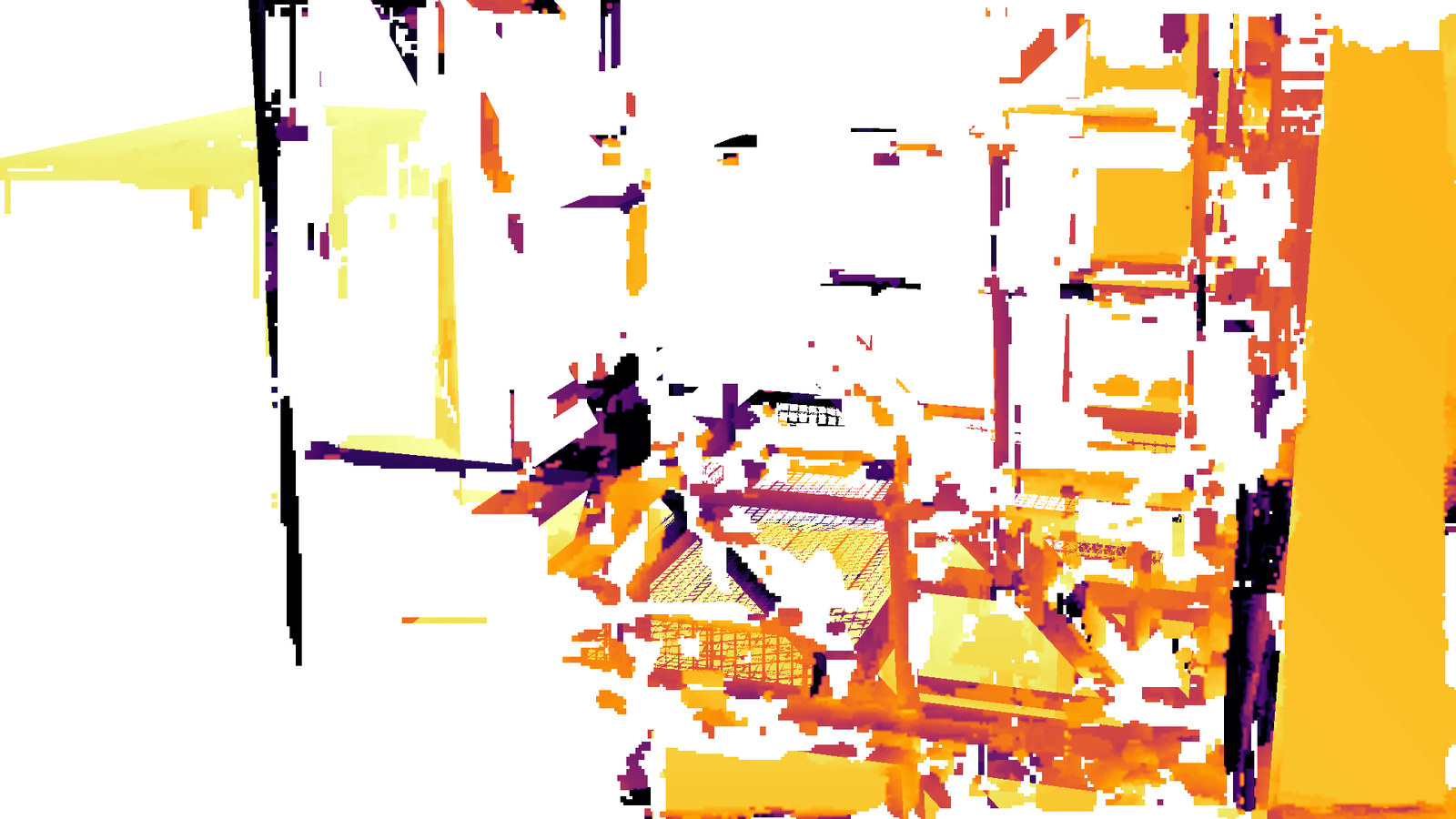} & \cell{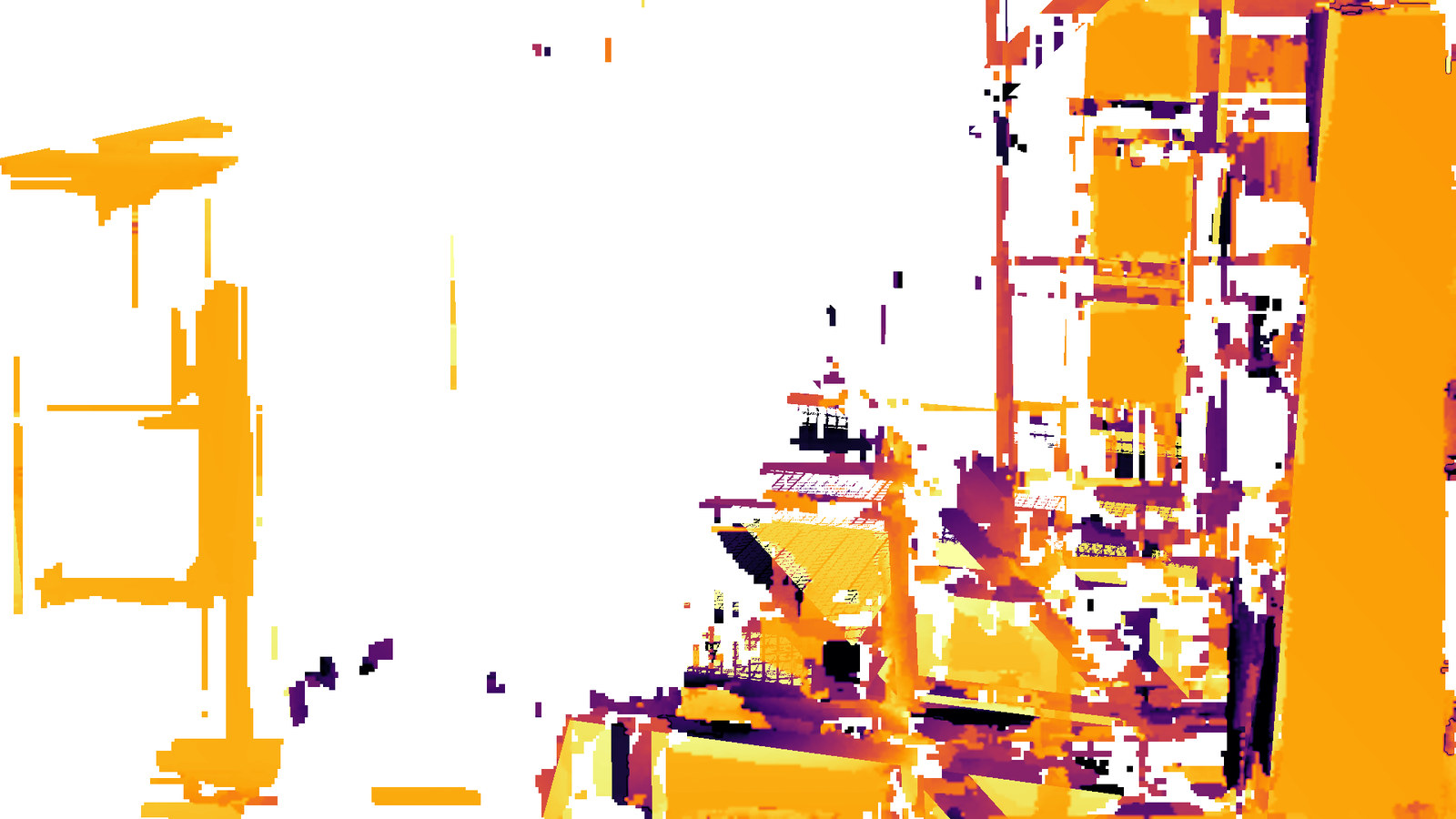} \\

    \hdr{Pi-Long} & \hdr{$\pi^3$-X} & \hdr{VidMap} \\
    \cell{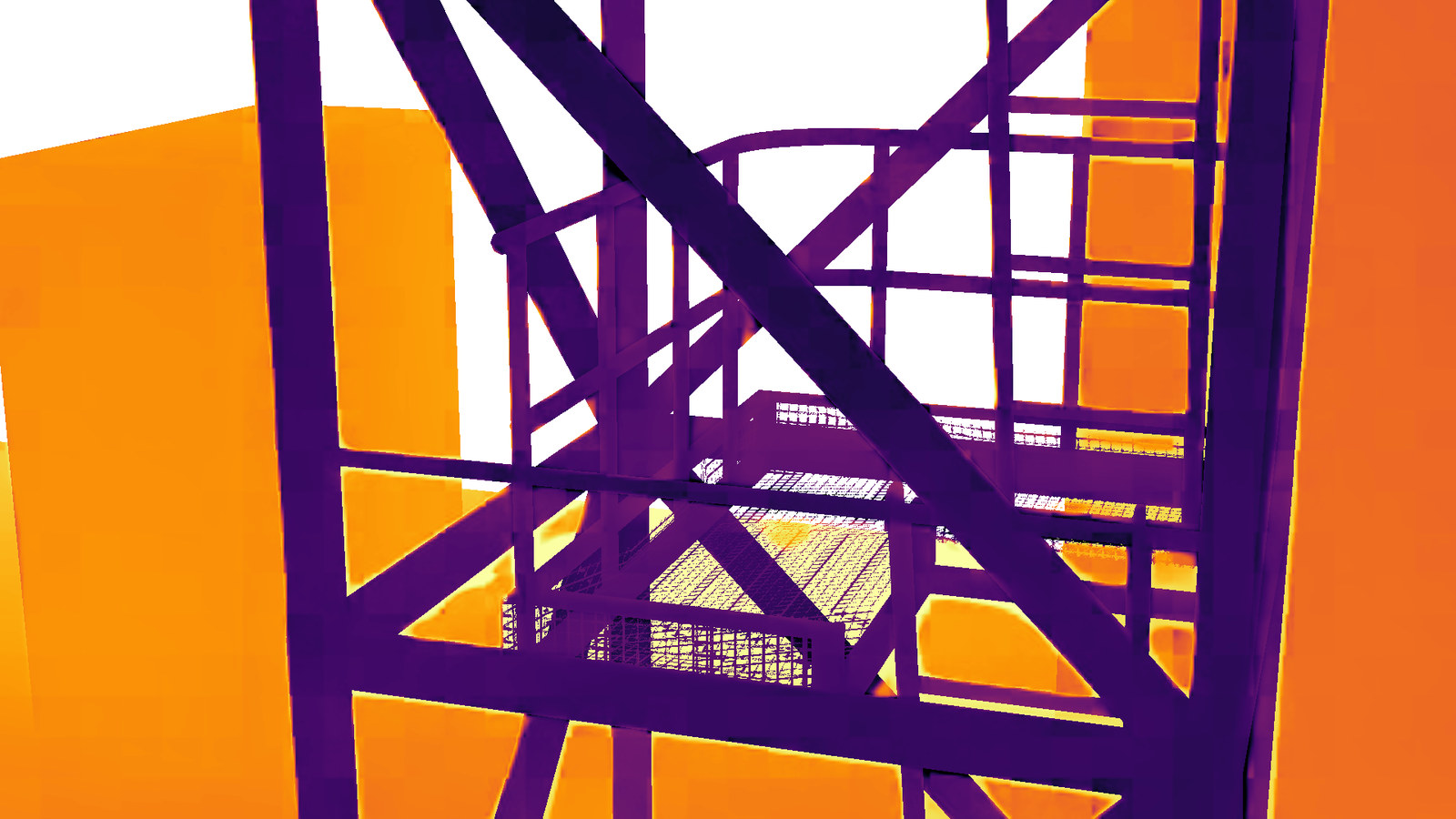} & \cell{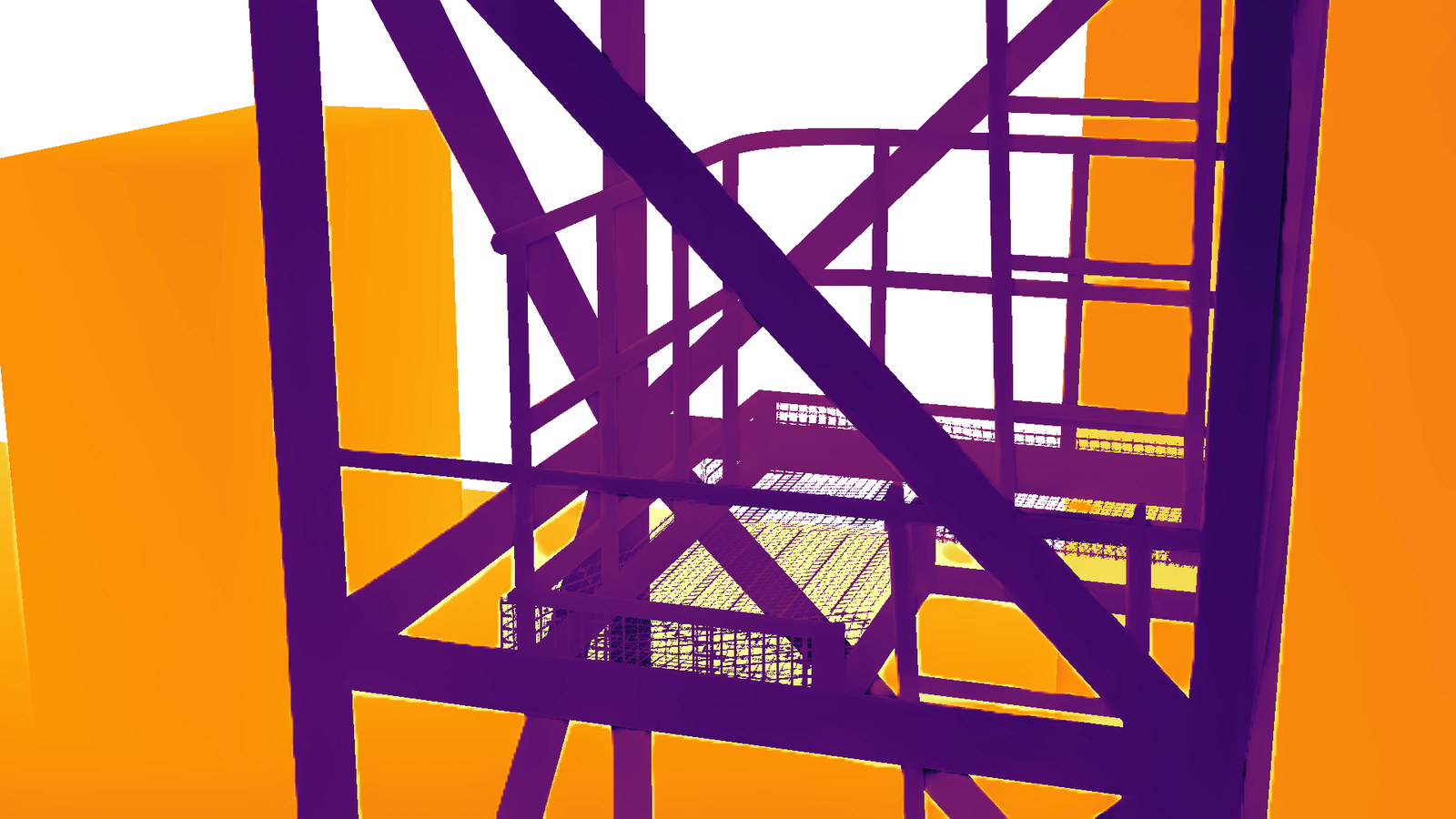} & \cell{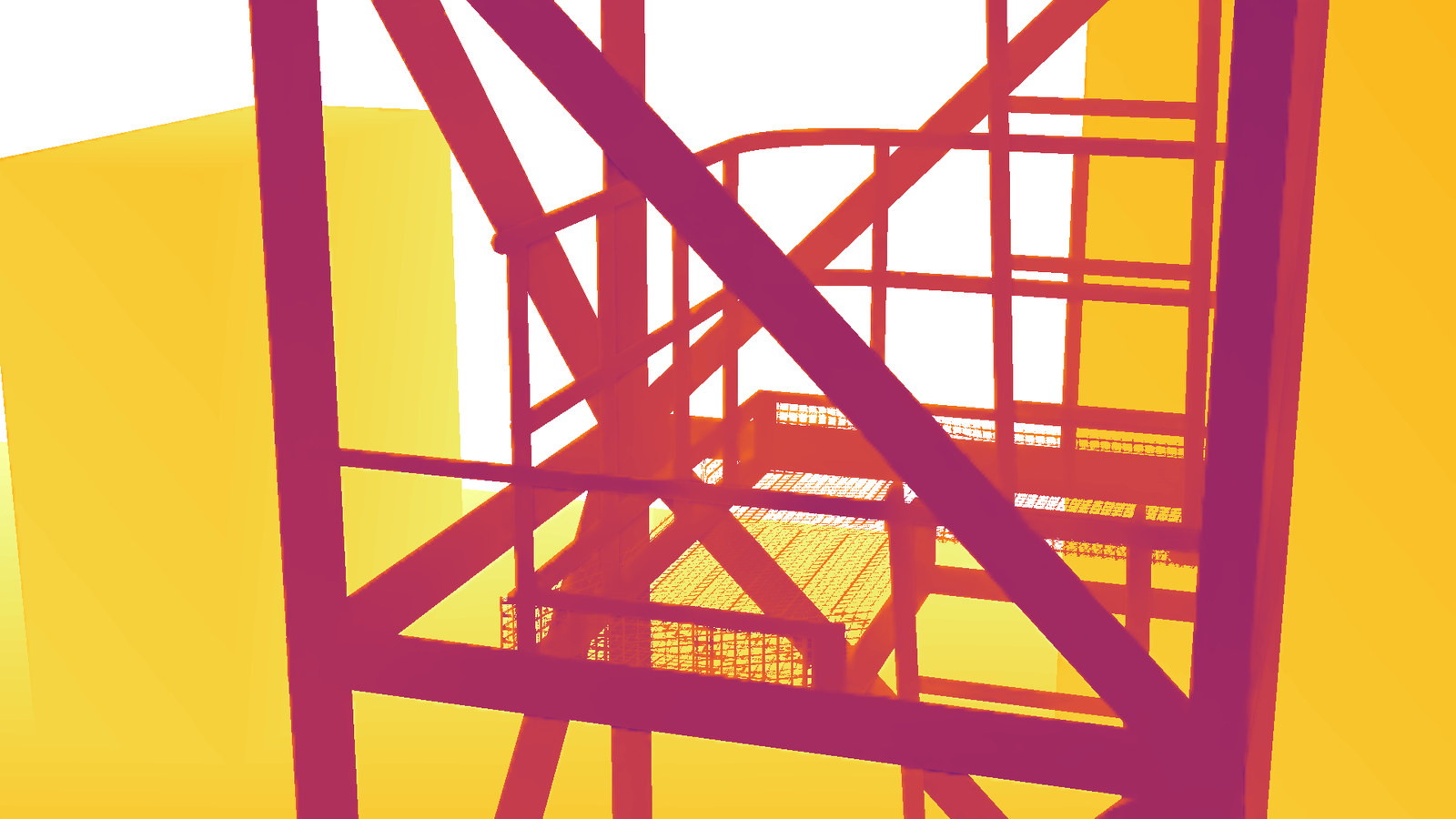} \\

    \hdr{VGGT} & \hdr{VGGT-SLAM 2.0} & \hdr{VGGT-Long} \\
    \cell{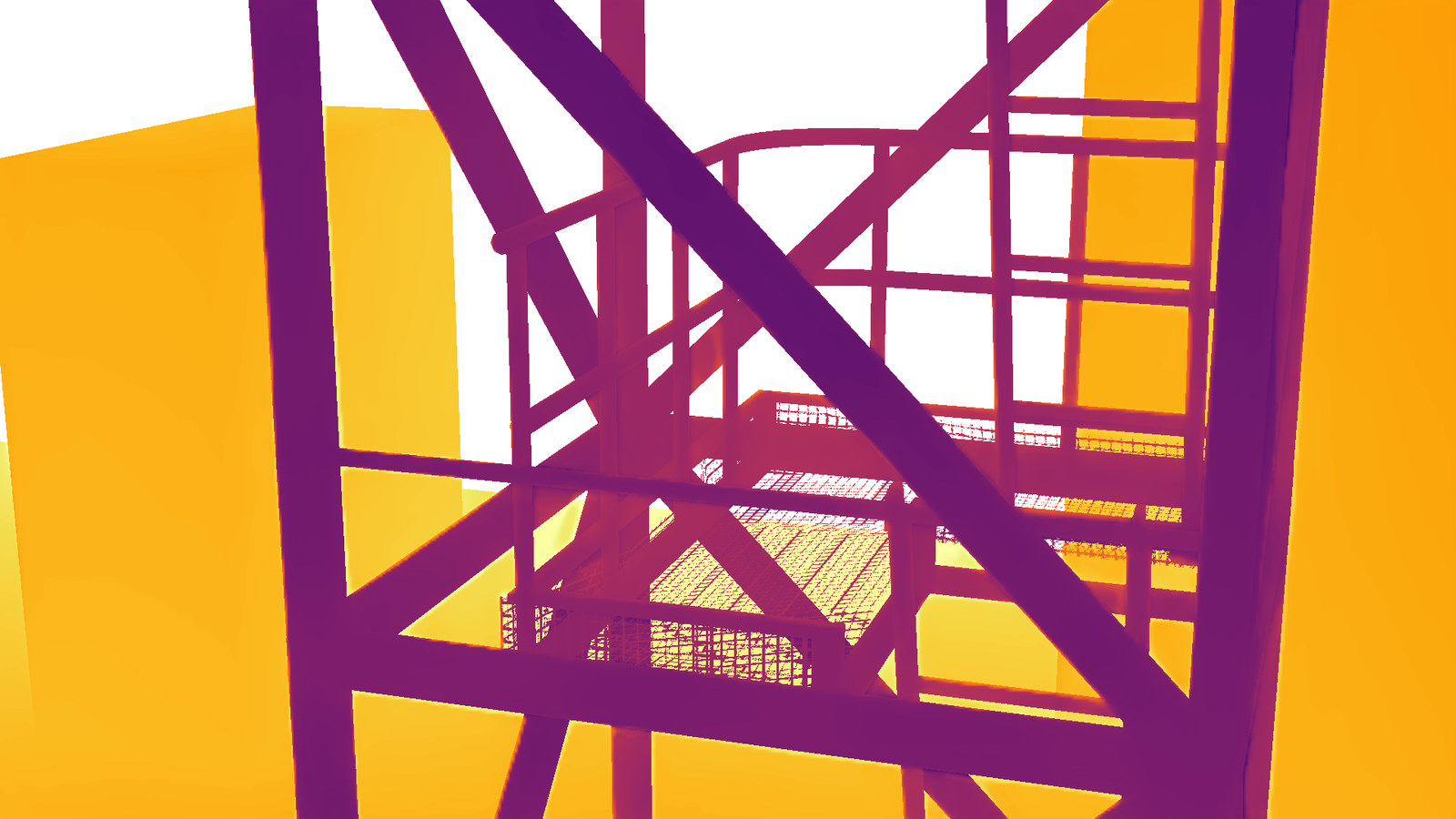} & \cell{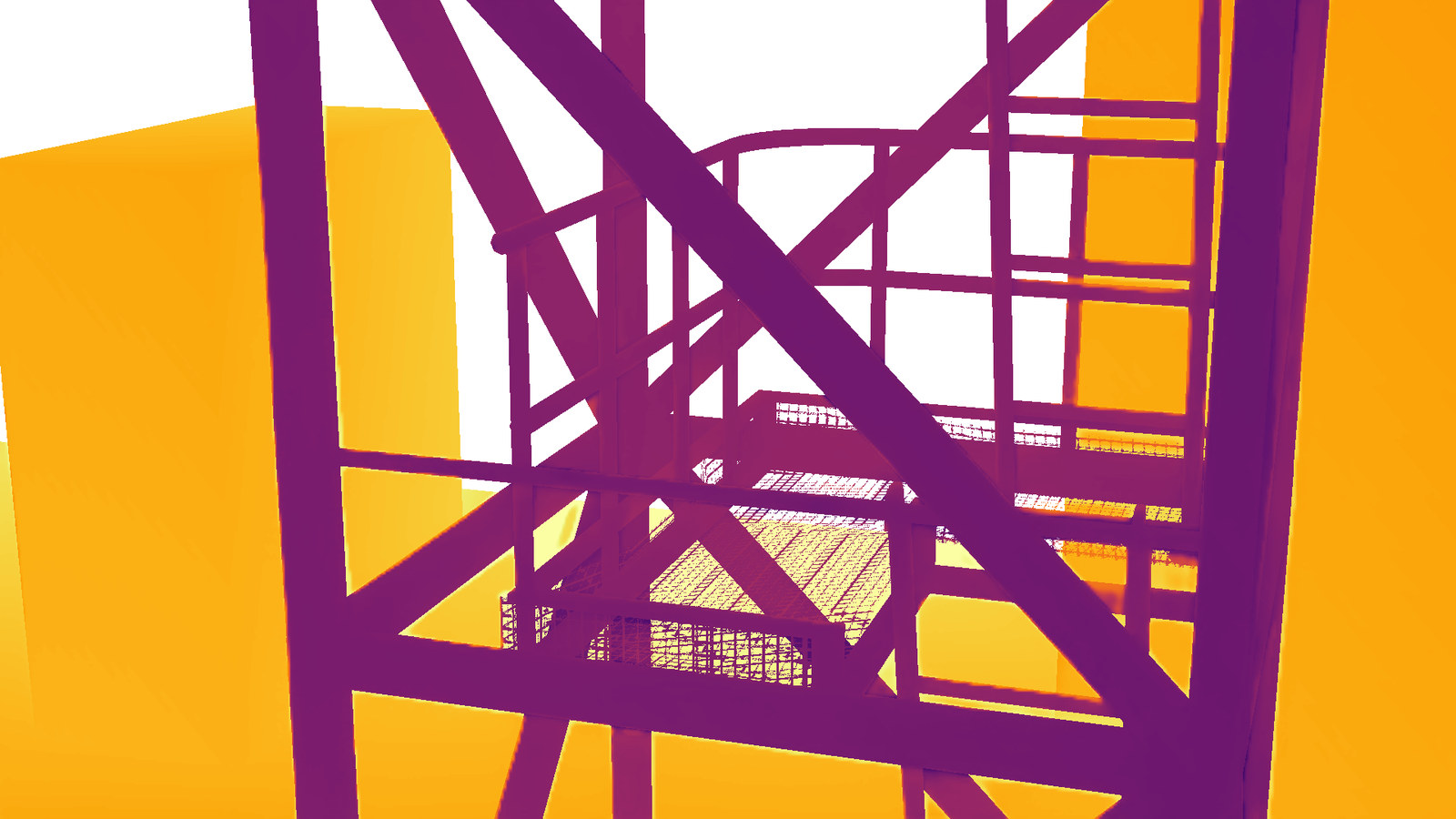} & \cell{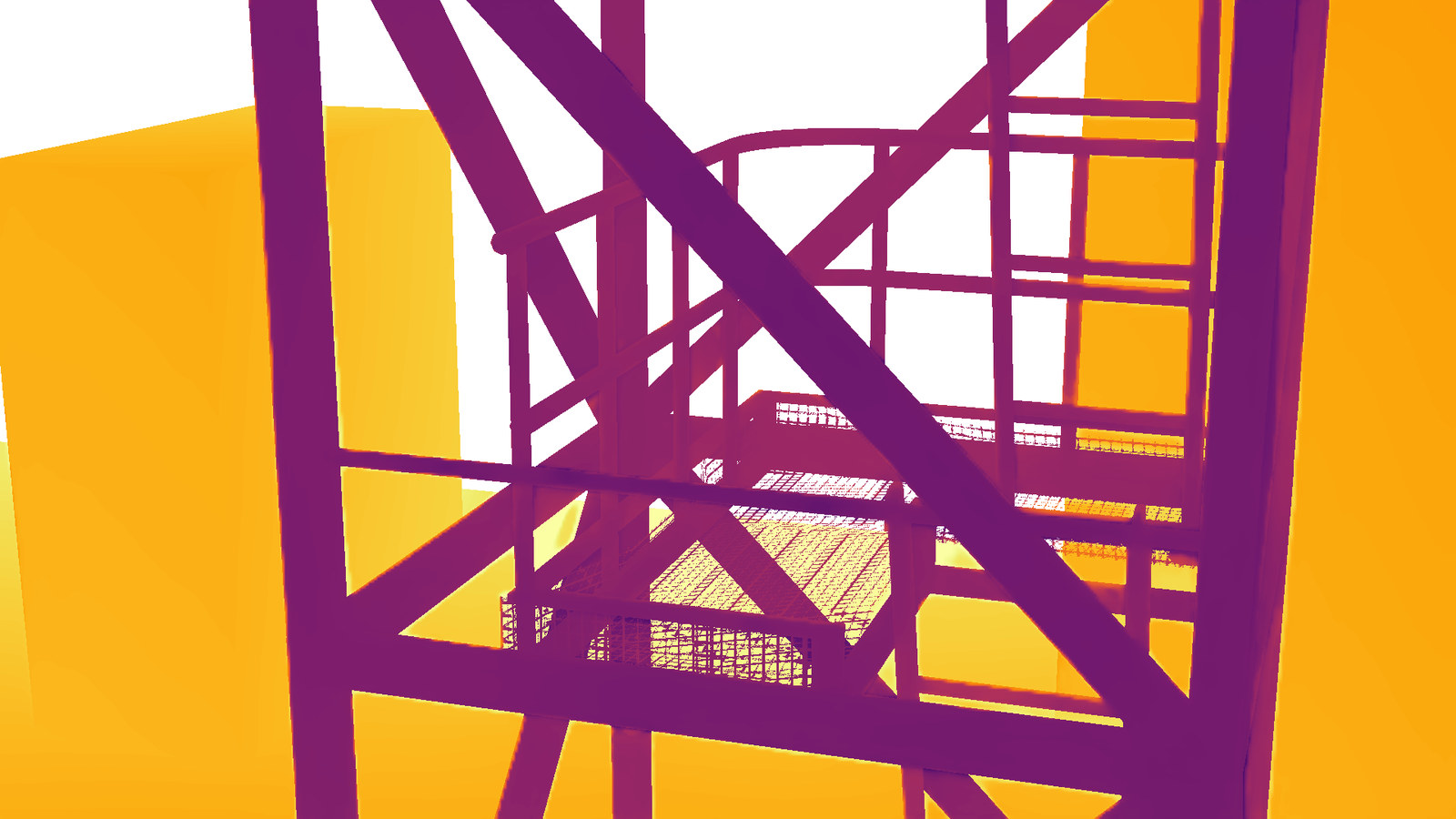} \\

    \hdr{DA3-Streaming} & \multicolumn{2}{c}{} \\
    \cell{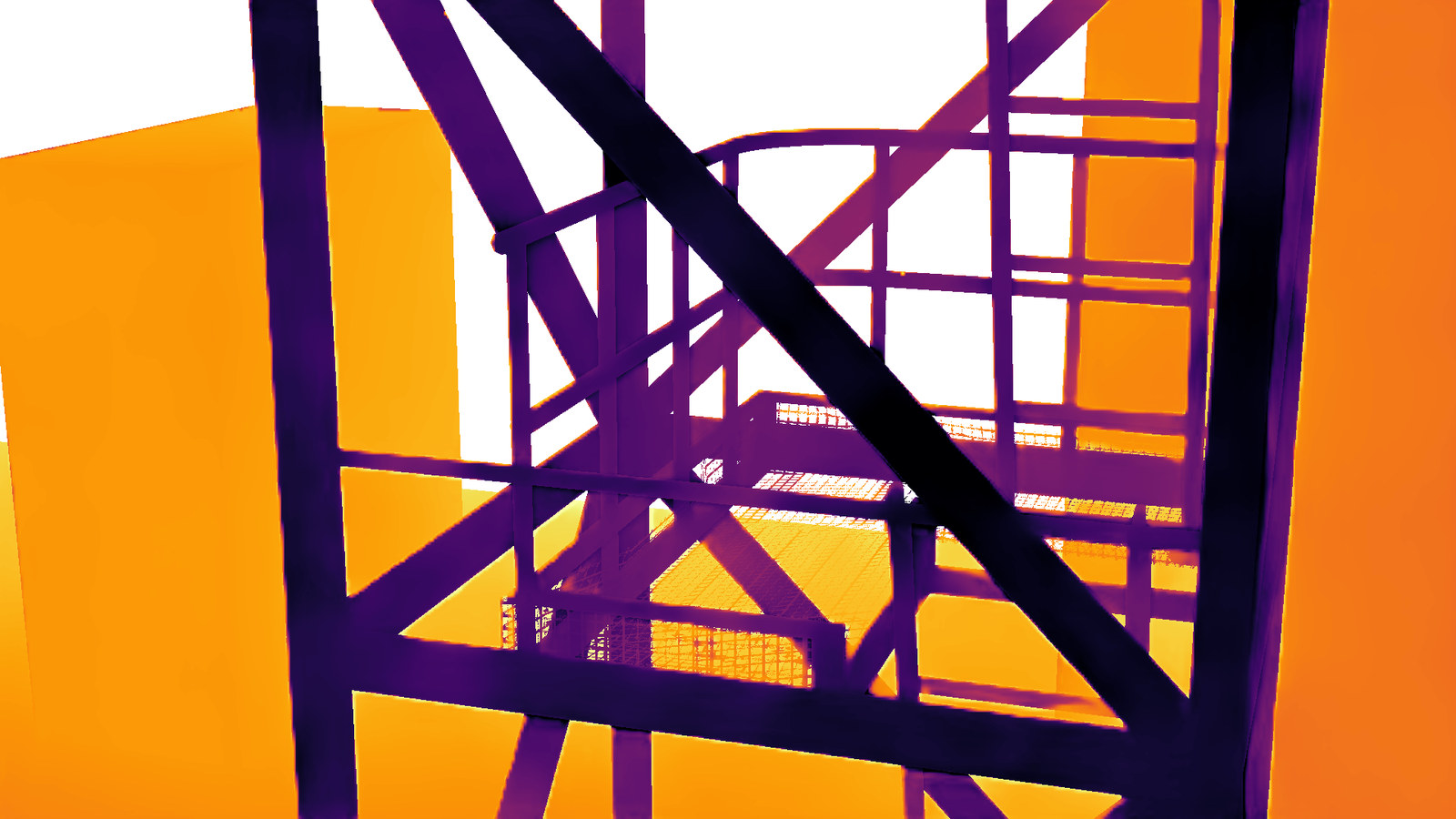}
      & \multicolumn{2}{c}{\raisebox{\cboff}{\usebox{\cbbox}}} \\
  \end{tabular}

    \caption{
    \small
    Depth map error after the scales have been aligned to the ground truth on frame 37 of the \textit{Crane} scene with the \textit{manual} capture pattern and the \textit{Cloudy} lighting condition. 
    White indicates invalid estimated or ground truth depth.
    COLMAP and GLOMAP produce mostly invalid depth maps.
    Monocular depth enabled methods estimate the depth of the foreground fairly accurately, but produce poor depth estimates of background areas such as the ground plane and the metal grating on the crane.
    }
  \label{fig:crane_depth_maps}
\end{figure}

In \Cref{fig:crane_depth_maps}, we can see how fine structures impact results at the frame level.
COLMAP and GLOMAP produced mostly invalid depth maps.
The VGGT-based methods and VidMap produce highly inaccurate depth maps in the background regions, which is consistent with their point cloud results, where the background is reconstructed very inaccurately.
The other learning-based methods produce better depth estimates of the background, but still get areas incorrect, such as the metal grating in the case of the $\pi^3$-based methods and the ground in the case of DA3-Streaming.

To summarise, by observing the reported results across the diverse scenes presented in this dataset we see that \ac{SOTA} 3D reconstruction methods continue to struggle with fine structures such as those present in the \textit{Bridge}, \textit{Crane}, \textit{Construction} and \textit{Ship} environments.  Our dataset aims to help researchers address this challenge through providing multiple assets with fine structures under a wide variety of capture patterns and illuminations, thereby contributing an essential tool for researchers to explore how to develop 3D reconstruction methods which are robust against these adversarial conditions.

\section{Conclusion}
\label{sec:Conclusion}

In this work, we present a synthetic asset inspection benchmark for 3D reconstruction methods containing 364k images, ground truth depth maps, camera poses and meshes.
This benchmark allows for the isolation of variables including camera trajectory, ground sampling distance, lighting, reflectivity and surface condition.
It thereby provides a valuable tool to researchers for exploring how both existing and emerging SOTA 3D reconstruction methods perform as these conditions are varied.
To demonstrate the value of our proposed benchmark, we provide a comprehensive evaluation of existing 3D reconstruction methods covering classical \ac{SfM} methods, recent \ac{E2E} approaches and their chunking- and SLAM-based counterparts, and report how their performance varies across the assets and permutations present in the dataset.

From our experiments, we observe several key takeaways.  First, existing \ac{E2E} methods experience OOM issues on long sequences, making them impractical for asset inspection applications.  Second, thin structures remain an open challenge for 3D reconstruction methods, often leading to high errors if not outright failure.  Third, higher reflectivity reduces depth map accuracy for both classical and (to a lesser extent) learning-based methods and has unpredictable effects on point cloud quality.  Finally, low-light conditions remain a challenge to 3D reconstruction methods across both classical and learning-based approaches. 
In order to allow others to readily test their methods on this dataset, we release our evaluation pipeline as an open-source repository alongside the dataset presented in this paper.

\bmhead{Limitations and Future Work}
While the use of synthetic data allows for the proposed dataset to be used to isolate the impact that scene permutations such as illumination, capture pattern and reflectivity has on performance, it could be further strengthened by the inclusion of real scenes which also contain permutations along these axes.  Another direction for future work for this dataset is to extend the data generation pipeline to include simulation of pose priors such as GPS or INS, for exploration of how they can be incorporated into 3D reconstruction in an asset inspection context.

\backmatter

\bmhead{Supplementary information}

Supplementary information is included after the references in this document.

\bmhead{Acknowledgements}

James L. Gray and Alexandre Cardaillac would like to acknowledge funding support of the Australian Research Council Industry Transformation Research Hub IH210100030.
Furthermore, this research was supported in part by funding from Ford Motor Company and through the NVIDIA Academic Grant Program.

\bibliography{sn-bibliography}

\clearpage
\section*{Supplementary Information}
\phantomsection
\addcontentsline{toc}{section}{Supplementary Information}
\setcounter{section}{0}
\setcounter{subsection}{0}
\setcounter{figure}{0}
\setcounter{table}{0}
\setcounter{equation}{0}
\renewcommand{\thesection}{S\arabic{section}}
\renewcommand{\thefigure}{S\arabic{figure}}
\renewcommand{\thetable}{S\arabic{table}}
\renewcommand{\theequation}{S\arabic{equation}}
\renewcommand{\theHsection}{supp.\arabic{section}}
\renewcommand{\theHfigure}{supp.\arabic{figure}}
\renewcommand{\theHtable}{supp.\arabic{table}}
\renewcommand{\theHequation}{supp.\arabic{equation}}

\section{Scene Details}
\label{supp:scenes}

This dataset consists of seven scenes: \textit{Bridge}, \textit{Crane}, \textit{Construction}, \textit{Glass House}, \textit{Office}, \textit{Old House} and \textit{Ship}.
Each scene is rendered with four capture patterns under each of four lighting conditions, giving 16 sequences per scene.
The two scenes with reflective assets are additionally rendered with nine surface-condition settings, corresponding to varying degrees of soiling or reflectivity.

\paragraph{Capture patterns}
The \textit{manual} pattern was animated by hand for each scene and follows industry guidelines for inspection capture~\cite{Trendspek_capture_guide, t2_capture_guide}. 
The \textit{close}, \textit{medium}, and \textit{far} patterns benchmark the effect of imaging resolution: camera positions are sampled on proxy surfaces offset from the asset at three increasing distances, giving progressively coarser ground sampling distances on the asset surface (see \cref{tab:gsd} of the main paper).
The camera poses are determined by Poisson-disc sampling, which produces an even but pseudorandom distribution of viewpoints, and are ordered into a continuous, drone-like trajectory by solving a traveling salesperson problem. 
The \textit{close} offset was chosen so that each view covers a similar area of the asset to the \textit{manual} pattern while still covering the entire asset, and the \textit{medium} and \textit{far} offsets increase this distance uniformly.
For all scenes except for \textit{Construction}, the proxy-surface patterns use a wider-angle lens than the \textit{manual} pattern to maintain coverage without cameras intersecting background objects.
For \textit{Bridge}, the ground beneath the roadway leaves insufficient clearance for close-range views of the deck's underside, so its \textit{close} cameras are placed further from the asset, and its \textit{close} pattern has a coarser resolution than in other scenes.

\paragraph{Lighting}
To simulate different illumination conditions, we render each scene under four HDR environment maps corresponding to \textit{Dusk}, \textit{Cloudy}, \textit{Midday} and \textit{Sunrise} lighting. The sources for all environment maps and asset models are listed in~\cref{sec:BlenderAssets}. 

\paragraph{Surface condition}
We simulate realistic windows that are partially transparent, partially reflective, and partially absorptive.
To simulate the dirt and smudging found on real windows, we also render four soiling settings (very low, low, medium and high) by adjusting surface roughness on randomised patches across the window glass.
To benchmark reconstruction on surfaces of varying reflectivity, we render \textit{Glass House} and \textit{Office} with five window reflectivity settings in steps of 0.25, from 0, an opaque and diffuse wooden panel, to 1, purely reflective glass, with intermediate settings blending the two materials.
The two are varied independently, giving nine surface conditions per scene.

\begin{figure}[htbp]
    \centering
    \includegraphics[width=\linewidth]{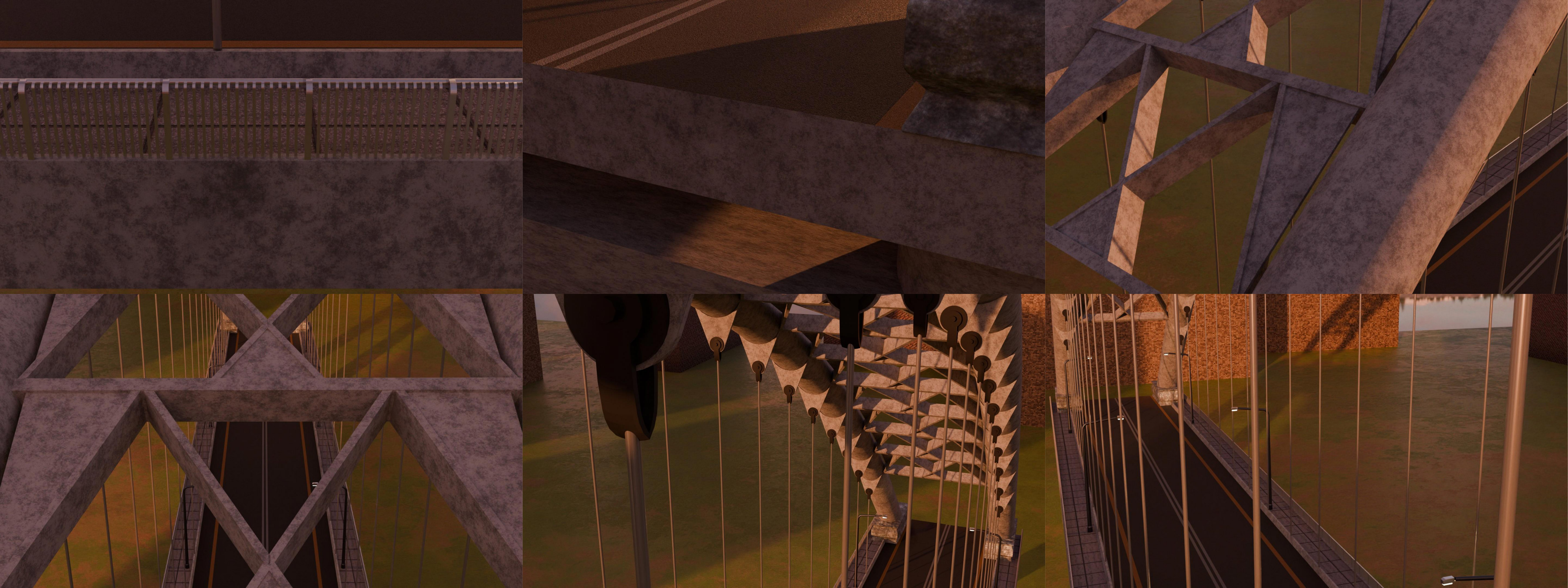}
    \caption{Several frames from the \textit{Bridge} scene with the \textit{Sunrise} lighting condition and \textit{manual}-style capture pattern.}
    \label{fig:Bridge_scene}
\end{figure}

\paragraph{Bridge}
Bridges are a core asset class for inspection, with cables, bearings and the underside of the span all requiring close-range assessment.
This arch bridge combines thin, repetitive structures, namely suspension cables, railings and truss members, which challenge feature matching and thin-structure reconstruction, with large, weakly textured planar regions such as the roadway and footpaths.
Frames from the \textit{manual} pattern are shown in~\cref{fig:Bridge_scene}.

\begin{figure}[htbp]
    \centering
    \includegraphics[width=\linewidth]{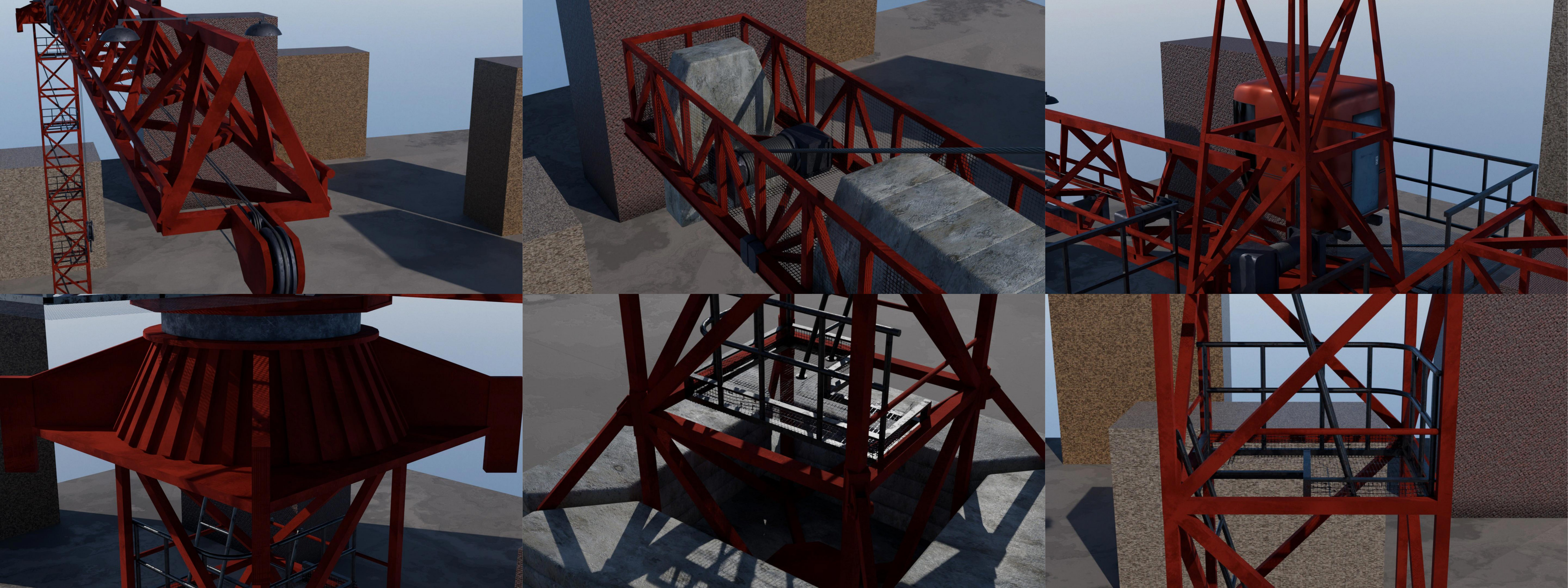}
    \caption{Representative frames from the \textit{Crane} scene with the \textit{Midday} lighting condition and \textit{manual}-style capture pattern.}
    \label{fig:Crane_scene}
\end{figure}

\paragraph{Crane}
Structures such as cranes are routinely inspected for corrosion and fatigue cracking in individual members. 
The crane's boom consists almost entirely of thin, highly repetitive members with few planar regions, so near-identical structure appears across many views, producing ambiguous matches and camera localisation errors, while the background visible through the lattice further complicates depth estimation.
Frames from the \textit{manual} pattern are shown in~\cref{fig:Crane_scene}.

\begin{figure}[htbp]
    \centering
    \includegraphics[width=\linewidth]{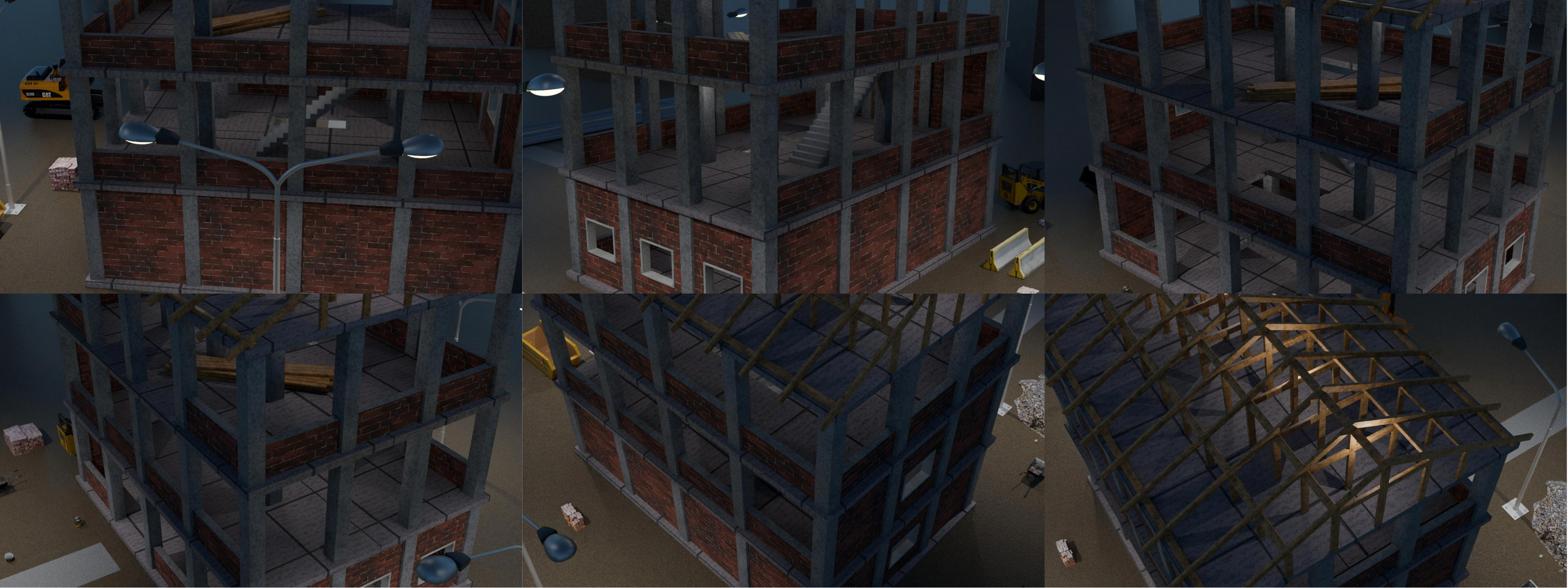}
    \caption{Representative frames from the \textit{Construction} scene with the \textit{Dusk} lighting condition and  \textit{medium} capture pattern.}
    \label{fig:construction_scene}
\end{figure}

\paragraph{Construction}
Drone-based reconstruction is widely used to monitor construction progress, where partially completed structures expose columns, beams and open floors rather than closed facades.
This scene presents such a structure: its open frame exposes the interior, creating depth discontinuities and self-occlusion, and animated construction machinery introduces scene dynamics.
Interior ceiling lights and exterior street lamps keep the structure visible under the dimmer dusk and sunrise conditions.
Frames from the \textit{medium} pattern under dusk lighting are shown in~\cref{fig:construction_scene}.

\begin{figure}[htbp]
    \centering
    \includegraphics[width=\linewidth]{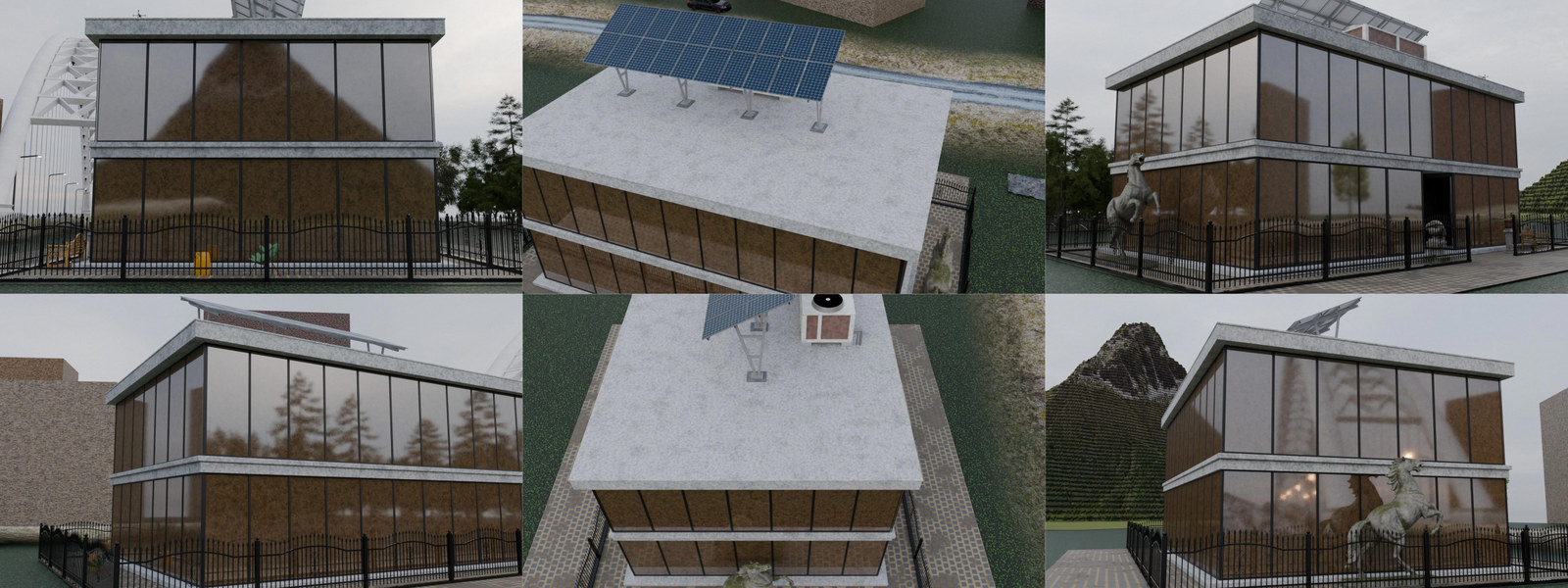}
    \caption{Representative frames from the \textit{Glass House} scene with the \textit{Cloudy} lighting condition and \textit{close} camera capture pattern. The reflectivity is set to 0.75.}
    \label{fig:Glass_house}
\end{figure}

\paragraph{Glass House}
This scene represents an extreme example of a glass-dominant building: a small structure whose walls are almost entirely glass, meaning that most of its visual appearance comes from reflections of its surroundings.
Dense background clutter surrounds the building so that, as with real glass façades, the glass has substantial content to reflect.
The scene is rendered with all nine surface-condition settings, and an animated car provides scene dynamics.
Frames from the \textit{close} pattern at reflectivity 0.75 are shown in~\cref{fig:Glass_house}.

\begin{figure}[htbp]
    \centering
    \includegraphics[width=\linewidth]{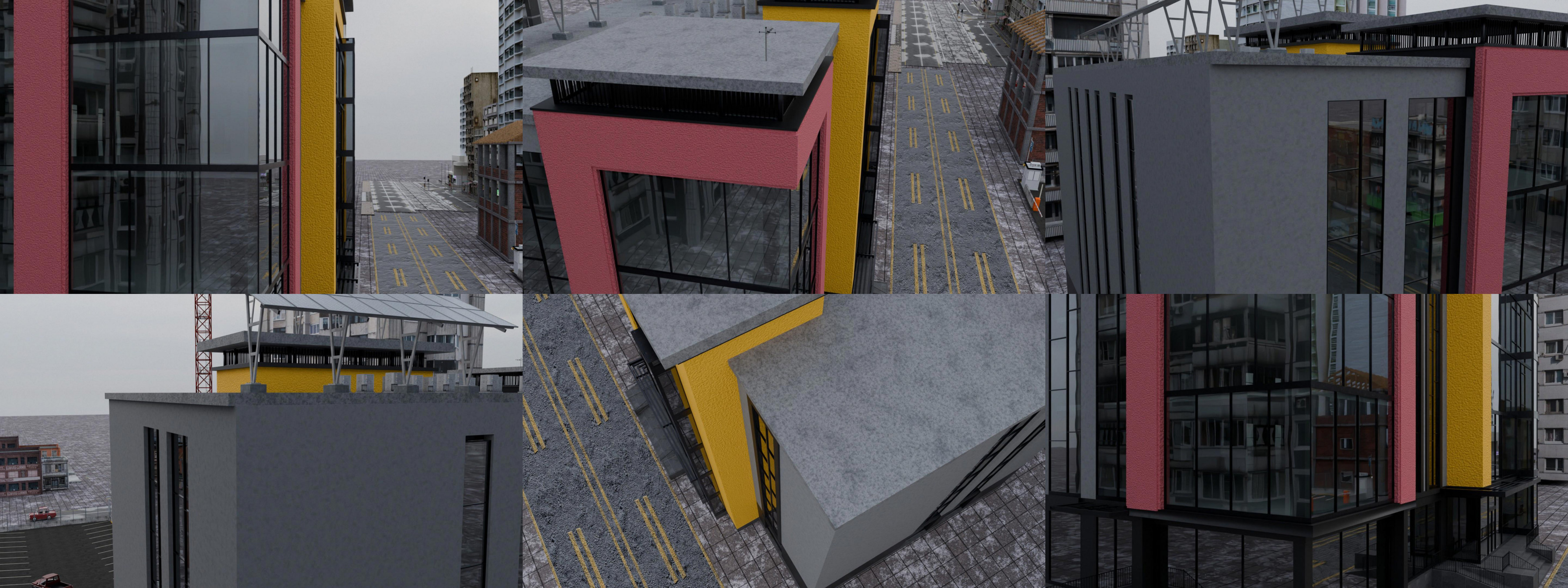}
    \caption{Representative frames from the \textit{Office} scene with the \textit{Cloudy} lighting condition and \textit{far} capture pattern. The surface condition soiling setting is very low.}
    \label{fig:Office_scene}
\end{figure}

\paragraph{Office}
Commercial buildings with curtain-wall facades are a common inspection target, combining large areas of reflective glass surfaces with flat, weakly textured wall panels.
This scene therefore challenges both the photometric-consistency and feature-distinctiveness assumptions of 3D reconstruction methods, and tests their scalability, as the dataset \textit{manual} sequence is the longest at 4{,}763 frames.
It is rendered with all nine surface-condition settings, and two cars and a pick-up truck provide scene dynamics.
Frames from the \textit{far} pattern with very low soiling are shown in~\cref{fig:Office_scene}.

\begin{figure}[htbp]
    \centering
    \includegraphics[width=\linewidth]{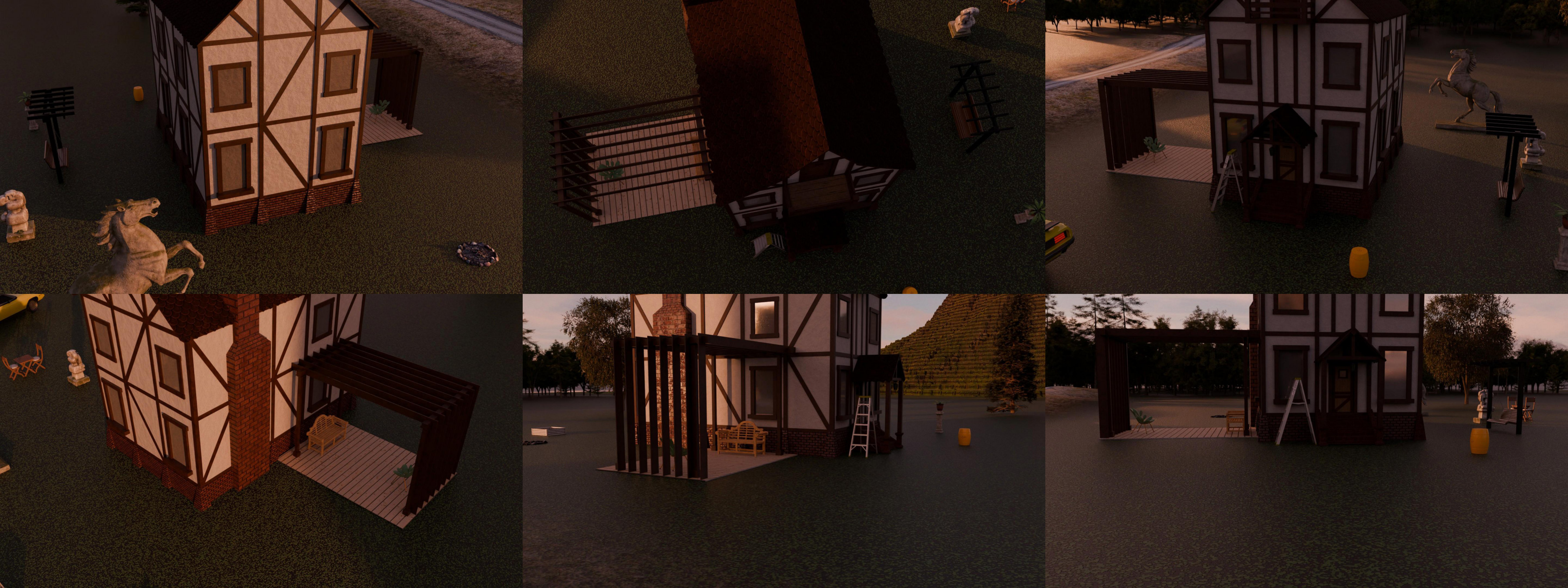}
    \caption{Representative frames from the \textit{Old House} scene with the \textit{Sunrise} lighting condition and \textit{medium} capture pattern.}
    \label{fig:Old_house_scene}
\end{figure}

\paragraph{Old House}
Residential and heritage buildings are inspected for facade, roof and timber condition.
This richly textured house is small, and a whole facade is often visible in a single view even at close range.
It nevertheless contains small reflective windows and a pergola of thin, truss-like members, and an animated car provides scene dynamics.
Frames from the \textit{medium} pattern are shown in~\cref{fig:Old_house_scene}.

\begin{figure}[htbp]
    \centering
    \includegraphics[width=\linewidth]{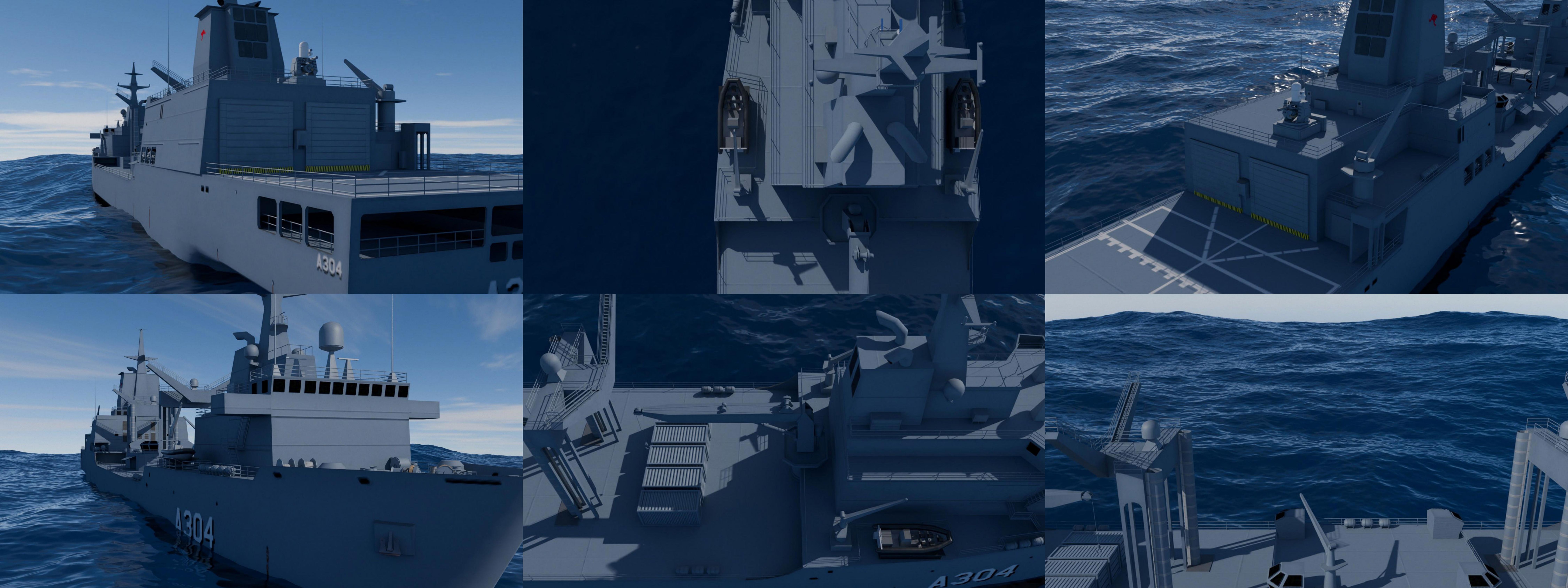}
    \caption{
    Representative frames from the \textit{Ship} scene with the \textit{Midday} lighting condition and \textit{close} capture pattern.
    }
    \label{fig:ship_scene}
\end{figure}

\paragraph{Ship}
Maritime vessels are inspected for corrosion and damage to the hull and superstructure.
The naval vessel combines large areas of uniform grey paint, which provide little texture for matching, with dense fine structures such as railings and masts.
The surrounding sea, animated with Blender's ocean simulation, is a dynamic, partly specular background that violates the static-scene assumption of most 3D reconstruction methods.
Frames from the \textit{close} pattern are shown in~\cref{fig:ship_scene}.

\section{Metrics}
\label{supp:metrics}

The goal of 3D reconstruction for asset inspection and management is to construct a digital twin of the asset for remote inspection.
For the digital twin to be useful, it must be geometrically accurate at the scene level and sufficiently detailed to identify issues requiring maintenance or remedial work.
Therefore, we evaluate both global and local accuracy metrics.

\subsection{Global reconstruction metrics}
Given the challenges of asset inspection, 3D reconstruction methods often fail by producing inaccurate geometry, \ie point clouds and camera poses, as well as failing to register images. 
Accordingly, we consider the performance on these tasks as indicative of global performance. 

\paragraph{Point cloud accuracy} 
We evaluate the fidelity of the reconstructed 3D point clouds, using the Chamfer distance, since it is an overall measure of accuracy and completeness for an estimated point cloud.
However, we also report the accuracy and completeness.

Each scene has two ground truth point clouds, one for evaluating the accuracy and one for evaluating the completeness of an estimated point cloud.
The accuracy point cloud is sampled from a mesh containing all static objects in the scene, while the completeness point cloud contains only the asset.
The accuracy is defined by,
\begin{equation}
    \mathrm{Acc}(\mathcal{R}, \mathcal{G}_c) = 
    \frac{1}{\left|\mathcal{R}\right|} \sum_{\mathbf{r} \in \mathcal{R}} e_{\mathbf{r} \rightarrow \mathcal{G}_a},
\end{equation}
where $e_{\mathbf{r} \rightarrow \mathcal{G}_a}$ refers to the Euclidean distance between an reconstructed point $\mathbf{r}$ in the set of reconstructed points $\mathcal{R}$ and the accuracy ground truth point cloud $\mathcal{G}_a$.
Specifically,
\begin{equation}
    e_{\mathbf{r} \rightarrow \mathcal{G}_a} = \min _{\mathbf{g} \in \mathcal{G}_a}\|\mathbf{r}-\mathbf{g}\|_2 .
\end{equation}
Therefore, accuracy measures the average distance between each reconstructed point and its nearest scene point.
On the other hand, completeness measures the average distance between each asset point and its nearest reconstructed point.
Specifically,
\begin{equation}
    \mathrm{Comp}(\mathcal{R}, \mathcal{G}_c) = 
    \frac{1}{\left|\mathcal{G}_c\right|} \sum_{\mathbf{g} \in \mathcal{G}_c} e_{\mathbf{g} \rightarrow \mathcal{R}}.
\end{equation}
This strategy of distinct point clouds for the accuracy and completeness metrics rewards methods that completely reconstruct the asset while maintaining accuracy throughout the scene overall. 
It would be possible to use a bounding box and ignore points beyond it; however, the choice of the bounding box is arbitrary and can strongly influence results. 
Consider a method that accurately reproduces the asset, but places lots of erroneous points in the sky. 
With a tight bounding box around the asset, the accuracy and completeness metrics would be unaffected by these erroneous points. 

Our primary measure of point cloud quality is the chamfer distance, which is simply the average of the accuracy and the completeness, \ie
\begin{equation}
    \mathrm{CD}(\mathcal{R}, \mathcal{G}_a, \mathcal{G}_c) = \frac{1}{2}\left(\mathrm{Acc}(\mathcal{R}, \mathcal{G}_c)
    + \mathrm{Comp}(\mathcal{R}, \mathcal{G}_c)\right) = 
\end{equation}

\paragraph{Camera pose accuracy} 
We use the Area under Curve (AUC@30) and RTA and RRA metrics to evaluate the camera pose accuracy. 
The AUC is defined by,
\begin{equation}
    \mathrm{AUC@30} = \sum_{\tau \in [1, 30]}  \min (\mathrm{RRA}@\tau, \mathrm{RTA}@\tau),
\end{equation}
where RTA and RRA refer to the relative translation accuracy and relative rotation accuracy, respectively.
For a given error threshold $\tau$, the RRA@$\tau$ and RTA@$\tau$ represent the percentage of all camera pairs $(i, j)$ whose errors fall strictly below $\tau$.
The relative translation error $\mathrm{RTA}_{i,j}$ between camera pose $i$ and $j$ is defined by,
\begin{equation} \mathrm{RTA}_{i,j} = 
                \arccos \left(
\frac{\mathbf{t}_{ij}^{\top}\mathbf{t}^{\star}_{ij}}
{\lVert \mathbf{t}_{ij} \rVert \, \lVert \mathbf{t}^{\star}_{ij} \rVert}
\right).
\end{equation}
The $\mathrm{RRA}_{i,j}$ is defined as the angle between the estimated relative rotation $R_iR_j^T$ and the ground truth relative rotation $R_i^\star (R_j^\star)^T$.
A higher AUC@30 indicates that a method reliably produces poses with minimal error bounds, making it a robust single-value summary of trajectory quality.

The ground truth camera poses are exported from Blender and are perfectly accurate.

\paragraph{Registration Rate} 
Some methods discard frames; this may be because of perceived redundancy with previous frames or low feature matches and high uncertainty.
In any case, discarding frames results in an incomplete set of poses. 
Evaluating the accuracy strictly on a successful subset can significantly increase the score of a method by ignoring the challenging frames on which it failed. 
To account for this, we evaluate only valid estimated poses and report the registration rate, defined as the proportion of successfully estimated camera poses relative to the total number of images in the sequence.

\subsection{Local meconstruction metrics} 
To evaluate the local reconstruction accuracy, we evaluate the estimated depth maps using \ac{sMSE}, \textit{AbsRel} and \textit{SqRel}.

\paragraph{\ac{sMSE}.}
In 3D reconstruction, depth predictions often exhibit an arbitrary global scale. 
We employ \ac{sMSE} to ensure that this scale ambiguity does not penalise the underlying geometric accuracy of the depth maps. The \ac{sMSE} is defined as,
\begin{equation}
    \mathrm{s M S E}= \sum_{p \in \mathcal{V}} 
    \frac{(\log  d(p) - \log \hat d(p) + a(d, \hat d))^2}
    {|\mathcal{V}|},
\end{equation}
where $a(d, \hat d ) =\frac{1}{\mathcal{V}} \sum_{p \in \mathcal{V}} (\log \hat d(p) - \log d(p) )$ 
 is the scale alignment value
 $d$ and $\hat d$ are the ground truth and estimated depth maps, respectively, 
 and $\mathcal{V}$ is the set of valid pixels.
 A valid pixel is a pixel location where both the ground truth depth and estimated depth maps have depths that are positive and finite.
 Masks, including window and asset masks, when considered, limit valid pixels to within the window or asset regions in the mask.

 \paragraph{AbsRel and SqRel}
 Some \ac{E2E} methods \eg \ac{DA3} and \ac{DA-Next} estimate both relative and metric (absolute) depth.
 Accordingly, we evaluate each method's performance at producing accurate metric depth.
 To do this, we use \textit{AbsRel} 
 which is defined as,
\begin{equation}
    \mathrm{AbsRel} = \frac{1}{|\mathcal{V}|} \sum_{p \in \mathcal{V}} \frac{| d(p) - \hat{d}(p)|}{d(p)}.
\end{equation}

\section{Hyperparameters and settings}
\label{supp:hyperparameters}

All learning-based methods were evaluated on NVIDIA RTX 600 Ada Generation GPUs with 48GB of VRAM.
COLMAP, GLOMAP and VidMap were evaluated on several computers, with different GPUs including: NVIDIA GeForce RTX 4070, NVIDIA RTX 600 Ada Generation, NVIDIA RTX PRO 6000 Blackwell Max Q-Workstation Edition. 
Since COLMAP, GLOMAP and VidMap handles inputs sequentially, even on limited hardware and vast sequences, they does not produce OOM errors.
However, owing to differences in hardware, we do not include timing comparisons. 

All methods use default parameters and settings unless indicated otherwise. 

\subsection{COLMAP and GLOMAP}

Asset inspection data is typically taken by flying a drone around an asset. 
Photos captured along the flight trajectory sequentially are therefore likely to have camera poses close to each other. 
Additionally, the drone's camera will remain the same over the course of the flight.
Therefore, to reduce computational expense and to allow COLMAP to make use of this prior information, we use COLMAP's automatic reconstructor, with default settings except for:
\vspace{4mm}
\begin{itemize}
    \item \texttt{automatic\_reconstructor}
        \begin{itemize}
            \item \texttt{camera\_model: SIMPLE\_PINHOLE}
            \item \texttt{single\_camera: 1}
            \item \texttt{dense: 0}
            \item \texttt{data\_type: video}
            \item \texttt{quality: medium}
        \end{itemize}
\end{itemize}
By selecting the video data type, we specify that COLMAP will use its sequential matcher rather than its exhaustive matcher.

For GLOMAP we use the lower-level modules instead of the automatic reconstructor. 
Specifically,
\begin{enumerate}
    \item \texttt{feature\_extractor}:
        \texttt{ImageReader.camera\_model = SIMPLE\_PINHOLE} (default \texttt{SIMPLE\_RADIAL});
        \texttt{ImageReader.single\_camera = true} (default \texttt{false}).
  \item \texttt{exhaustive\_matcher}: all defaults.
  \item \texttt{view\_graph\_calibrator}: run on a copy of the
        database; this optional step is enabled. All its parameters are at their defaults.
  \item  \texttt{global\_mapper}: all defaults.
\end{enumerate}
Since GLOMAP offers a substantial computational saving when compared to COLMAP, it became practical to use the more comprehensive exhaustive matcher rather than the sequential one.

Often COLMAP and GLOMAP produce fragmented reconstructions. 
If this occurs, we attempt to merge them using COLMAP's \texttt{model\_merger}. 
If merging these models proves impossible, we proceed with the largest reconstruction. 

To reduce computational expense during the dense reconstruction stage, we turned off the geometric consistency pass and reduced the image resolution. 
We experimentally found that doing so did not substantially impact the results.
We also note that most learning-based methods by default substantially reduce image resolution as well.
Accordingly, for the subsequent dense reconstruction stage, we used the following stages with default parameters unless indicated otherwise:
\begin{itemize}
    \item \texttt{image\_undistorter}
    \item \texttt{patch\_match\_stereo}:
        \begin{itemize}
            \item \texttt{geom\_consistency: false}
            \item \texttt{filter: true}
            \item \texttt{max\_image\_size: 480}
        \end{itemize}
    \item \texttt{stereo\_fusion}:
        \begin{itemize}
            \item \texttt{max\_image\_size: 480}
            \item \texttt{input\_type: photometric}
        \end{itemize}
\end{itemize}

\subsection{VidMAP}
VidMap was run with its uncalib/base frontend and mapping presets, unmodified, with depth-map caching enabled.

\subsection{End-to-End Methods}
DA3 was run with the default parameters, using the DA3NESTED-GIANT-LARGE model.
All other \ac{E2E} methods use their default parameters unmodified.

\subsection{Chunking Methods}
DA3-Streaming was run with its default configuration. However, some sequences (\ie Glass House, Old House and Construction) are shorter than the default chunk size of 120, which would provoke an error during pose export when a sequence fits inside a single chunk.
To avoid this, in these sequences we reduce the chunk size to one frame fewer than the sequence length, with the overlap capped below it, so that each sequence is processed as two chunks.

\subsection{VGGT-SLAM 2.0}
We use default parameters, aside from disabling loop closure and setting the confidence threshold to 50.
This is because camera paths in our sequences may result in multiple loop closures, and VGGT-SLAM 2.0 only supports a single loop closure.

\section{Additional Results}
\label{supp:results}

\begin{figure*}
    \centering
    \includegraphics[width=\linewidth]{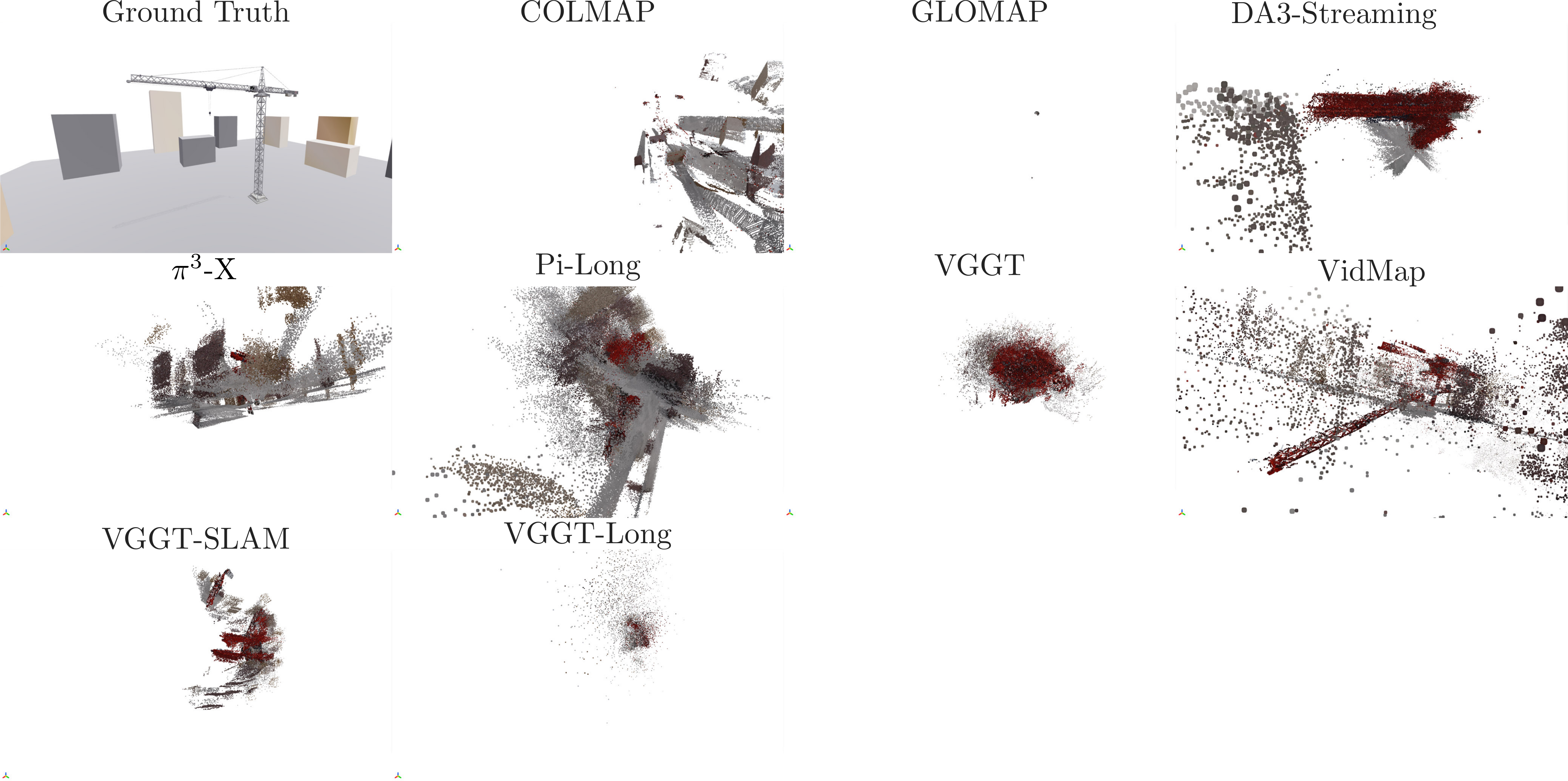}
    \caption{
    Point clouds of the \textit{Crane} scene, produced by these methods with the \textit{manual} capture pattern and the \textit{Cloudy} lighting condition.
    Note that the colours of the ground truth mesh are inaccurate. 
    All methods struggle on this sequence.
    GLOMAP's reconstruction was so small that the speck that is visible is in fact the frustums (turned off in every other plot), rather than the points.
    }
    \label{fig:crane_point_clouds}
\end{figure*}

\begin{figure*}[t]
  \centering
  \setlength{\tabcolsep}{0pt}
  \renewcommand{\arraystretch}{0}

  \setlength{\imgw}{0.245\textwidth}   
  
  \newlength{\nath}

  \newcommand{\cropL}{0.095}
  \newcommand{\cropB}{0}
  \newcommand{\cropR}{0.20}
  \newcommand{\cropT}{0.09}

  \newcommand{\cell}[1]{%
    \settowidth{\natw}{\includegraphics{#1}}%
    \settoheight{\nath}{\includegraphics{#1}}%
    \includegraphics[trim={\cropL\natw} {\cropB\nath} {\cropR\natw} {\cropT\nath},
                     clip, width=\imgw]{#1}}

  \newcommand{\hdr}[1]{\makebox[\imgw][c]{\small #1\rule{0pt}{3ex}}}
  \newcommand{\blank}{\makebox[\imgw]{}}
    
  \begin{tabular}{@{}c c c c@{}}
    \hdr{Ground Truth Mesh} & \hdr{COLMAP} & \hdr{GLOMAP} & \hdr{VidMap} \\
    \cell{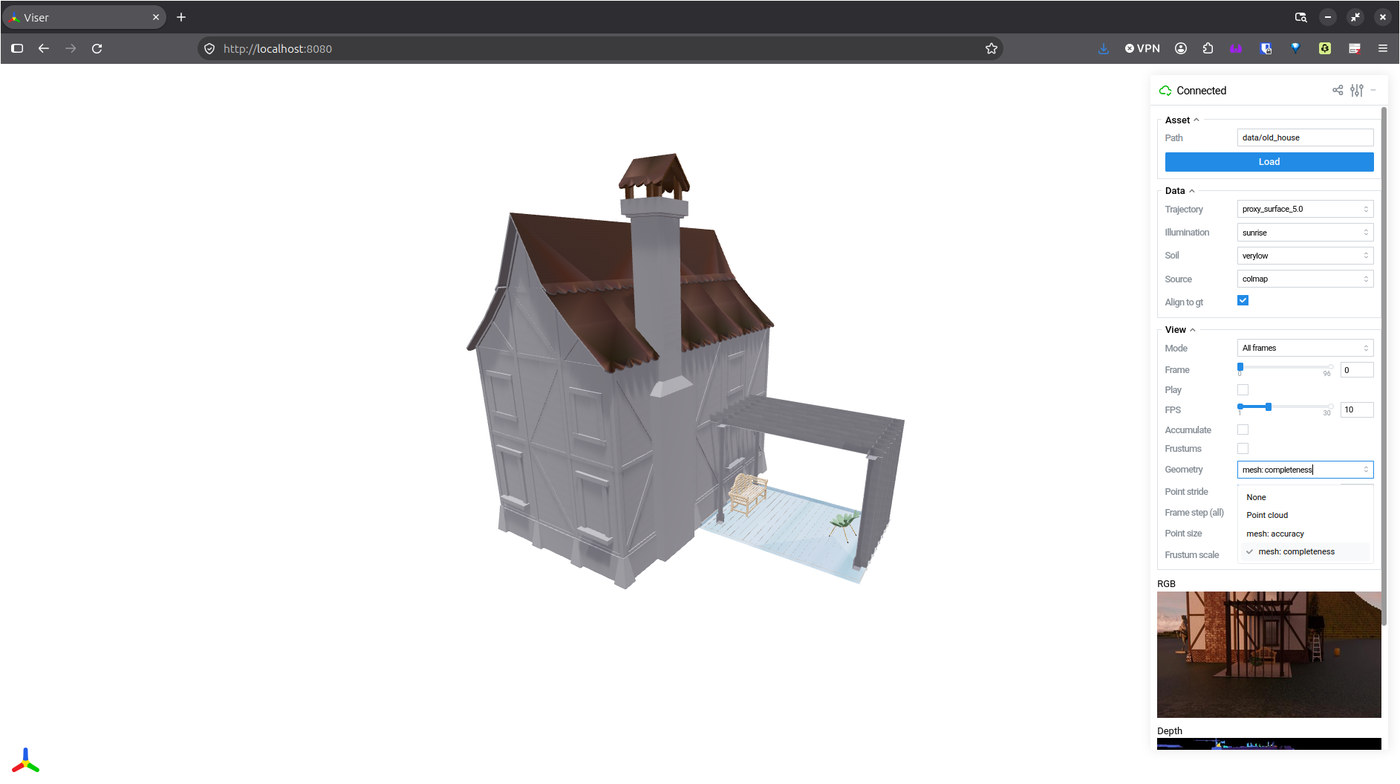} &
    \cell{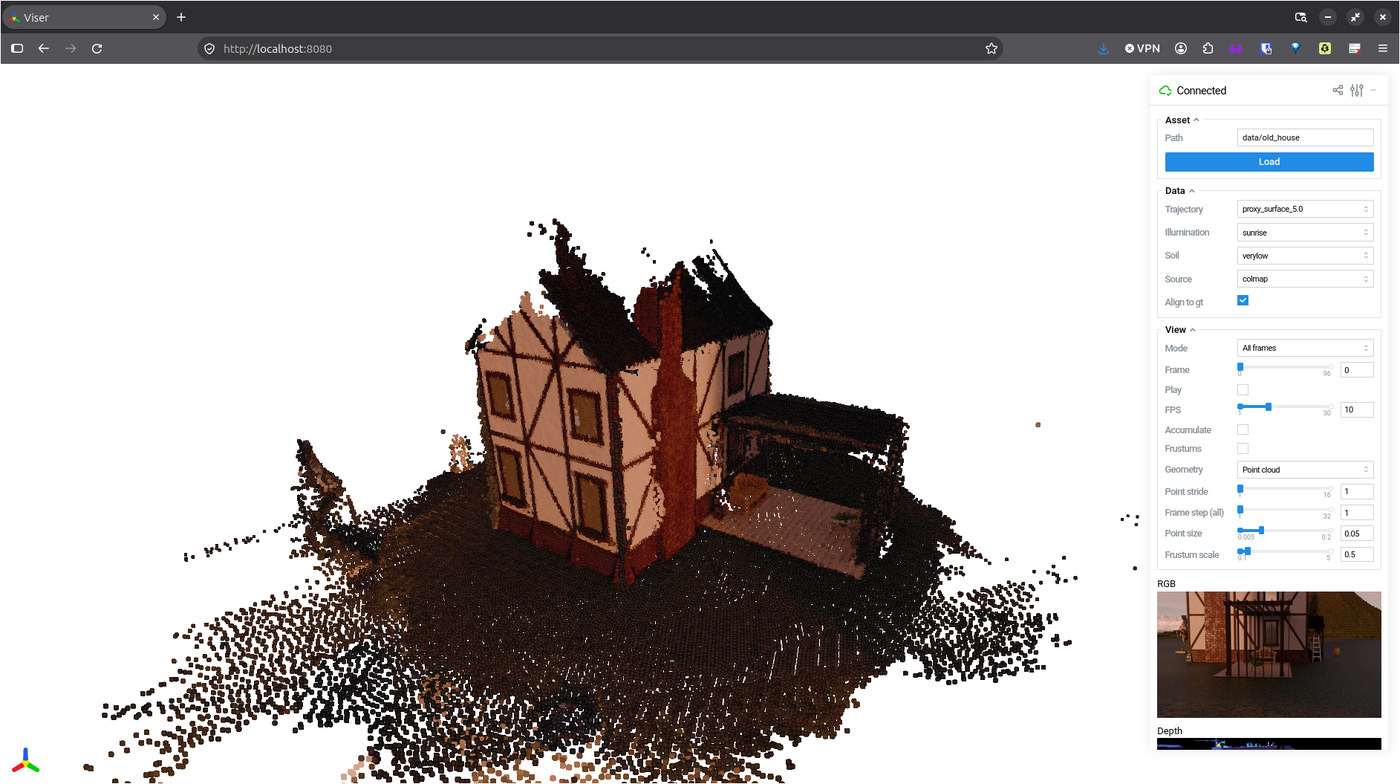} & 
    \cell{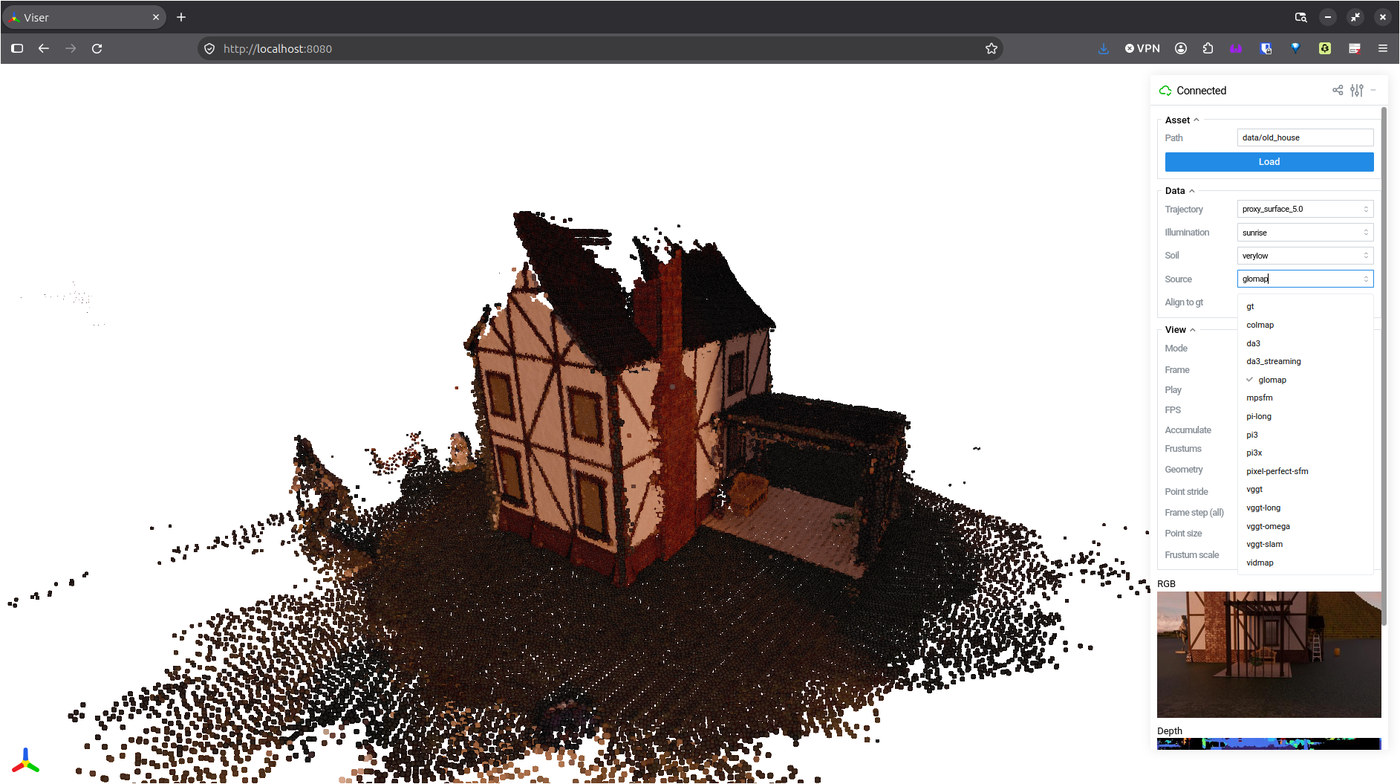} &
    \cell{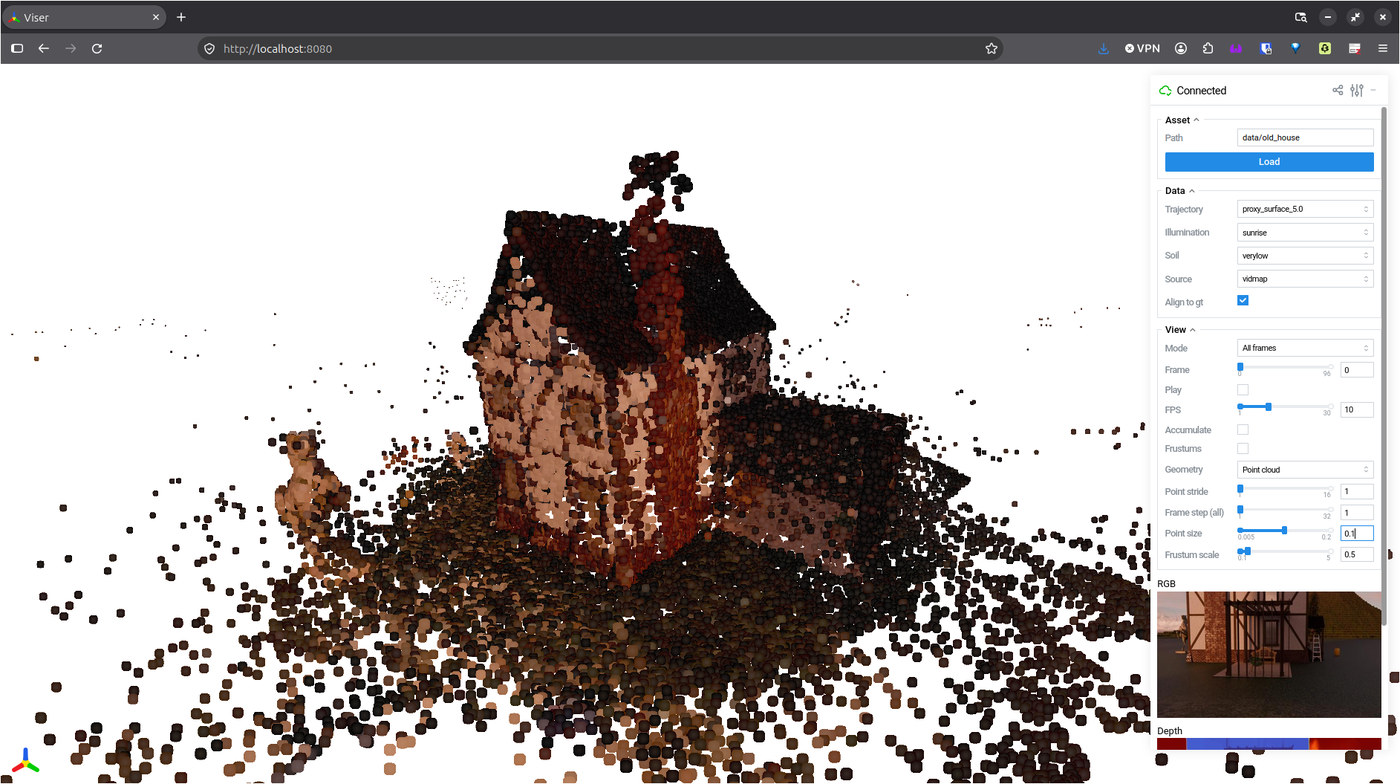} \\

    \hdr{Pi-Long} & \hdr{$\pi^3$}  & \hdr{$\pi^3$-X}  & \hdr{DA3-Streaming} \\
    \cell{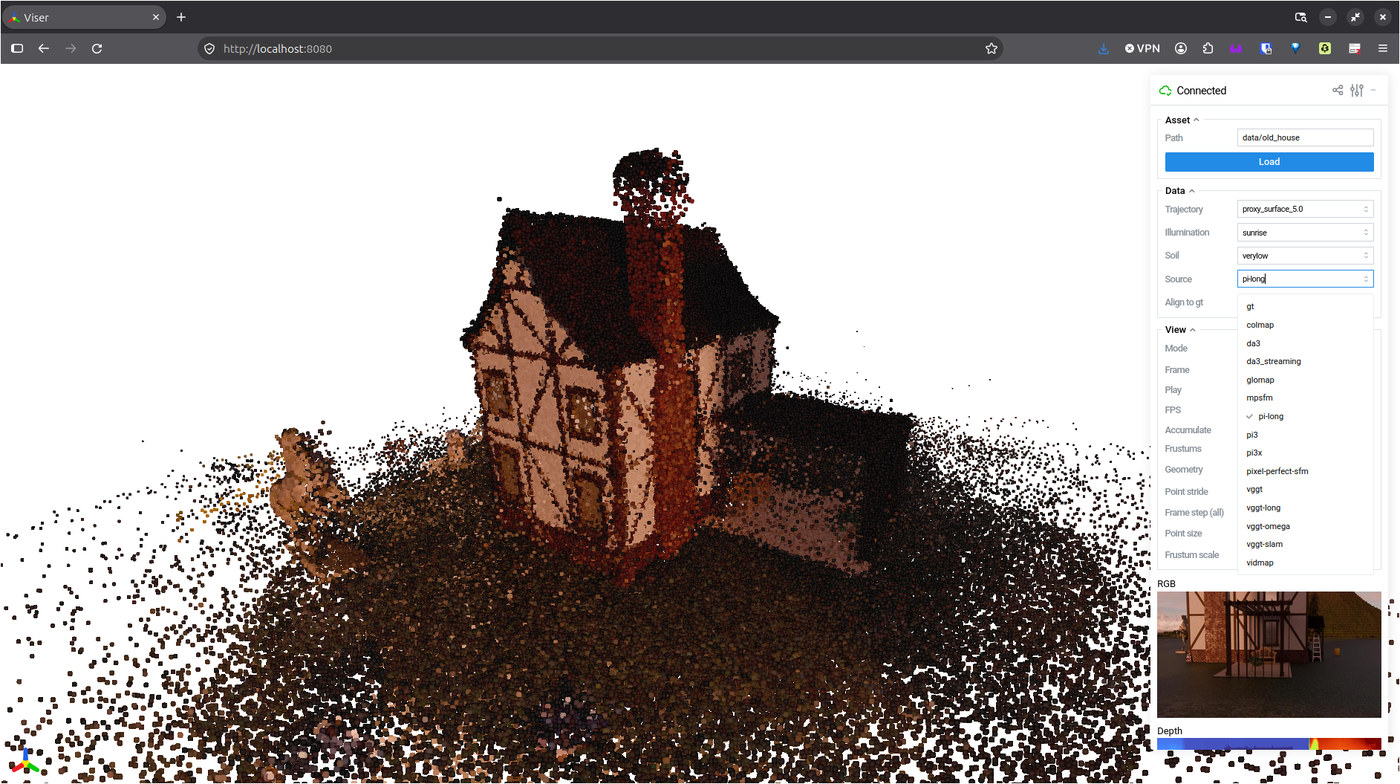} &
    \cell{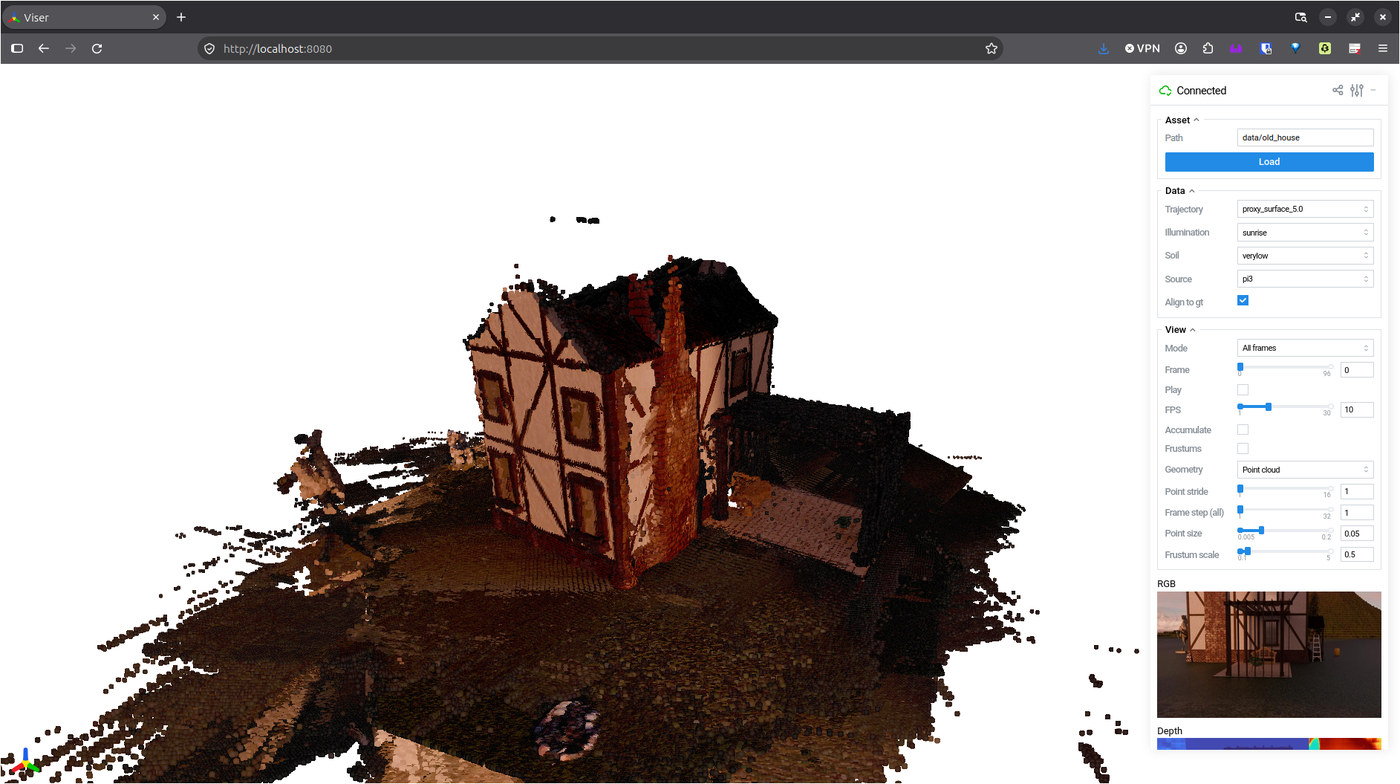} & 
    \cell{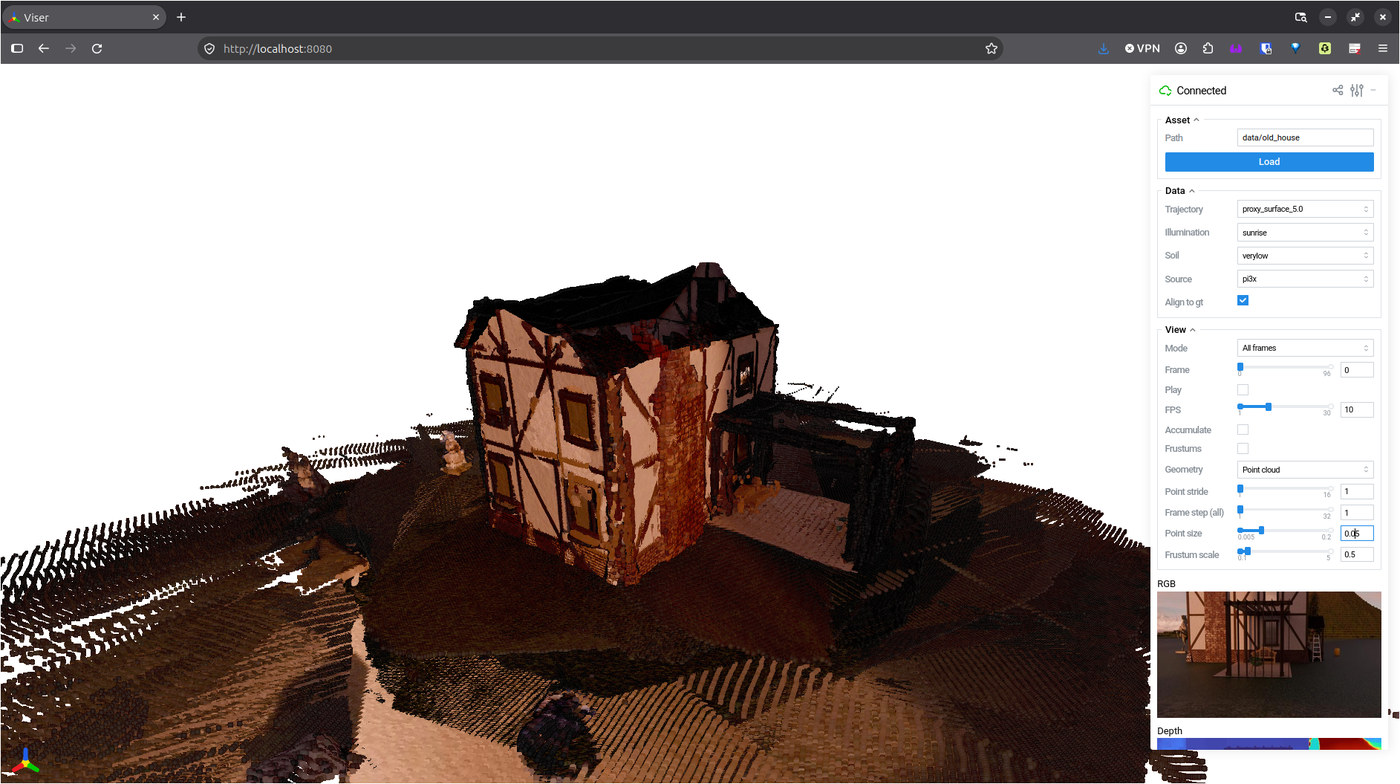} &
    \cell{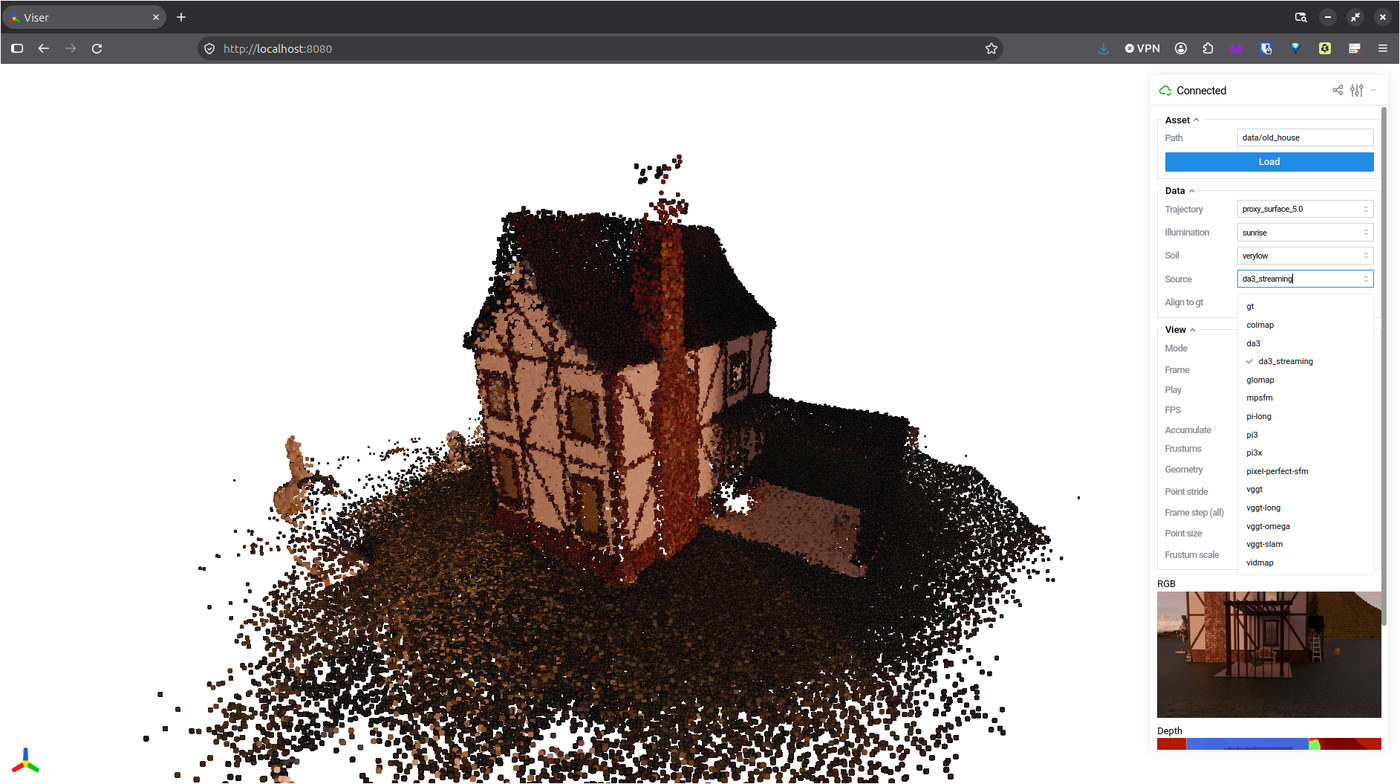} \\

    \hdr{VGGT-SLAM 2.0} & \hdr{VGGT-Long} & \hdr{VGGT} & \hdr{DA3} \\
    \cell{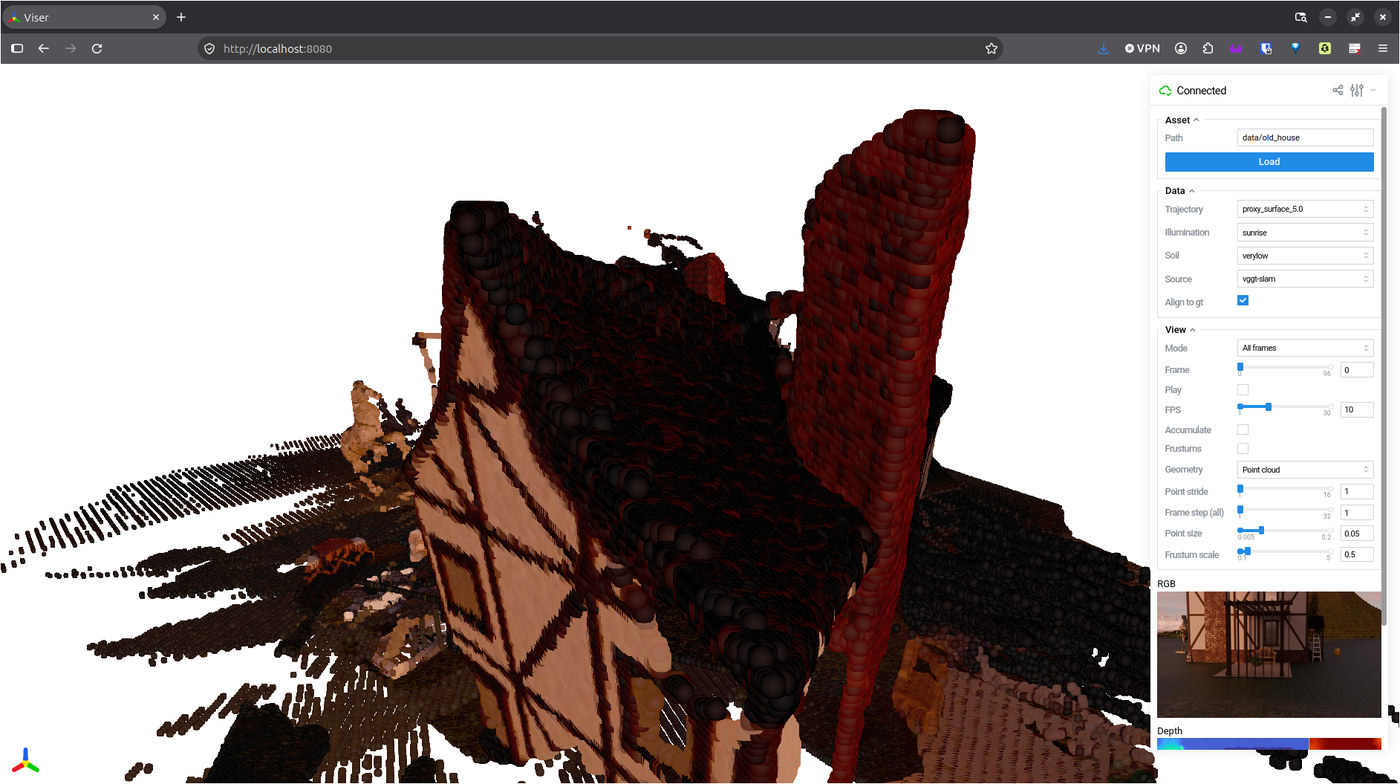} &
    \cell{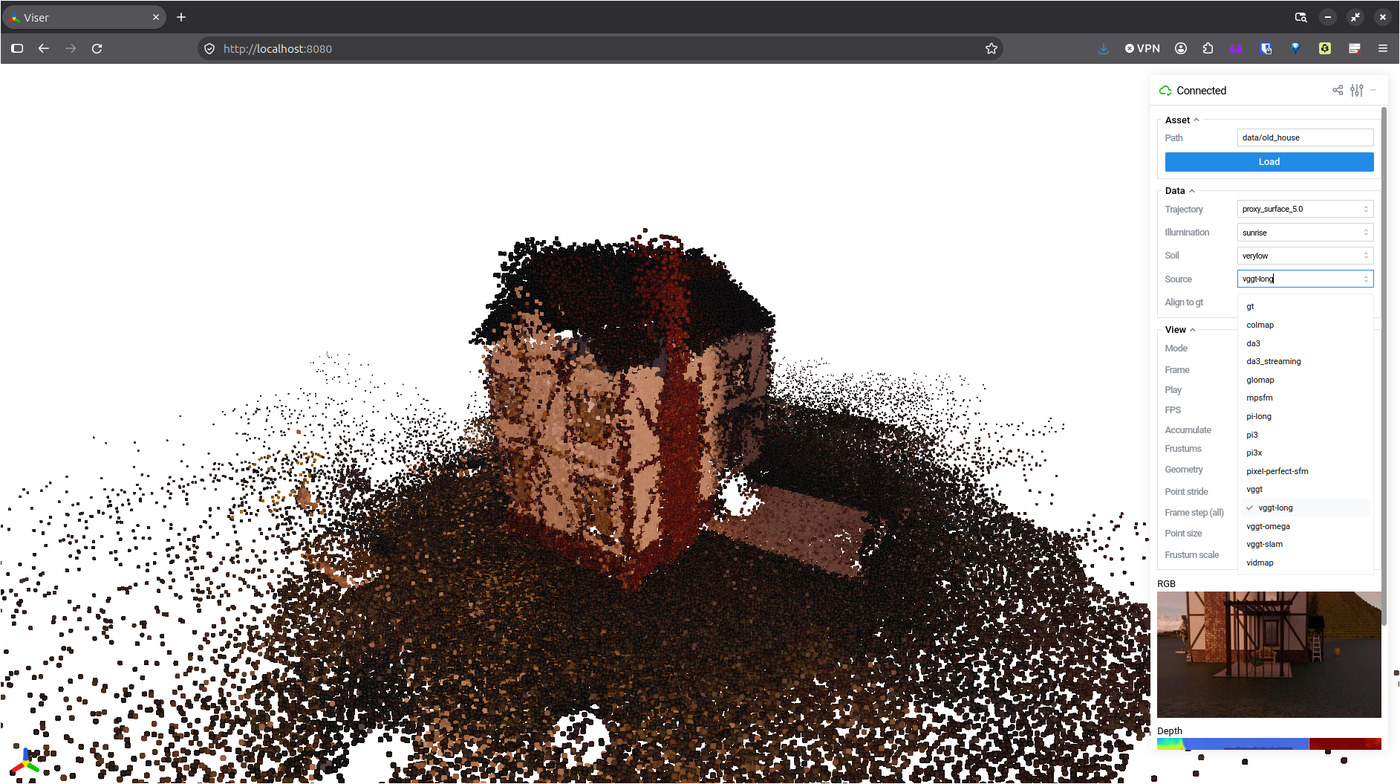} & 
    \cell{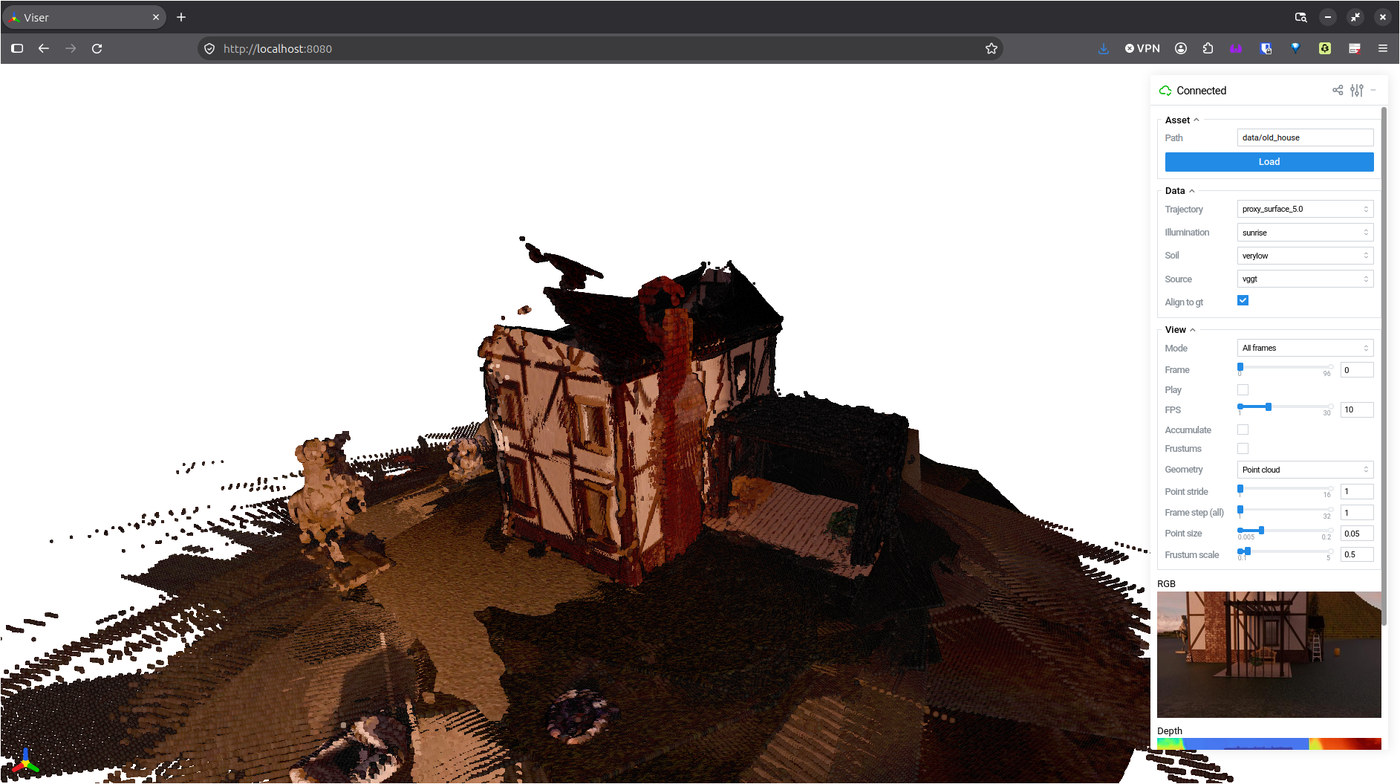} &
    \cell{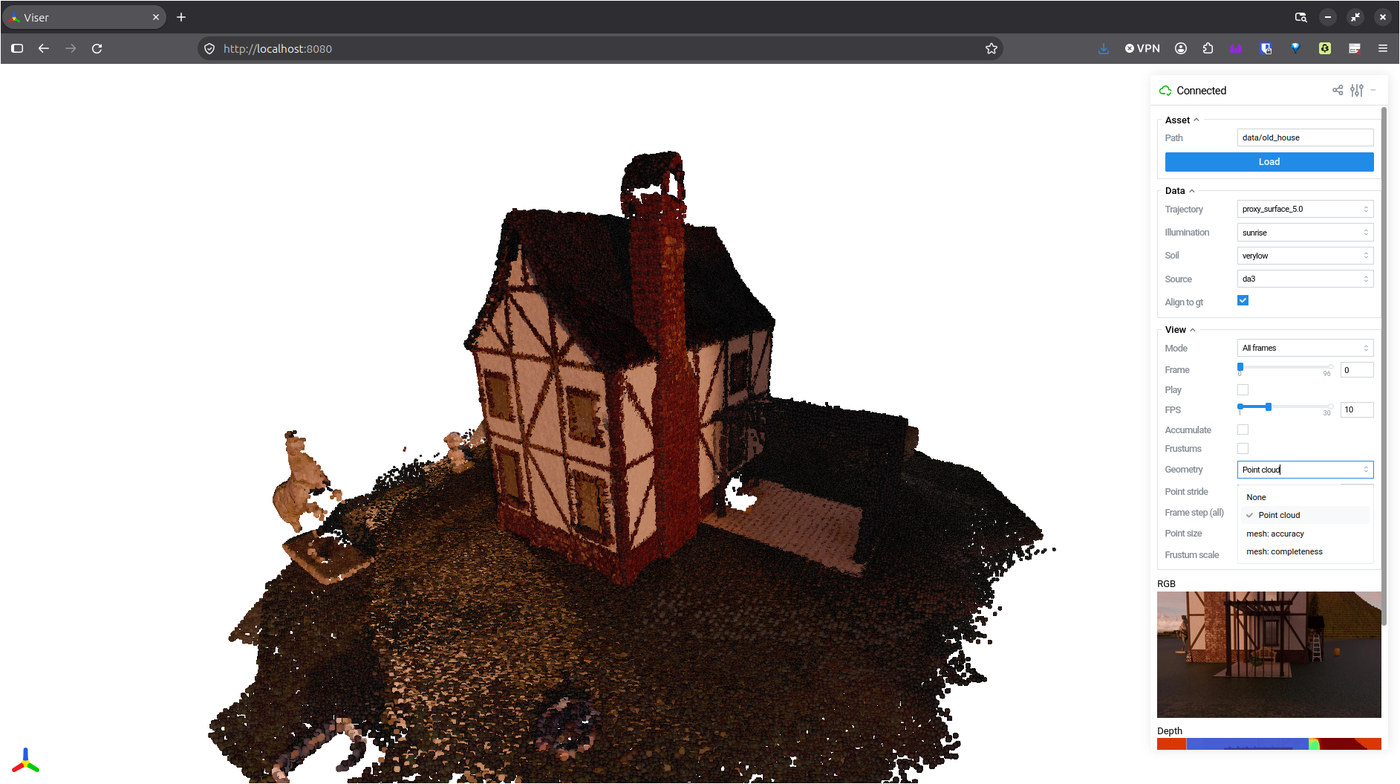} \\
    
    \hdr{VGGT-$\Omega$} &  & \blank & \blank  \\
    \cell{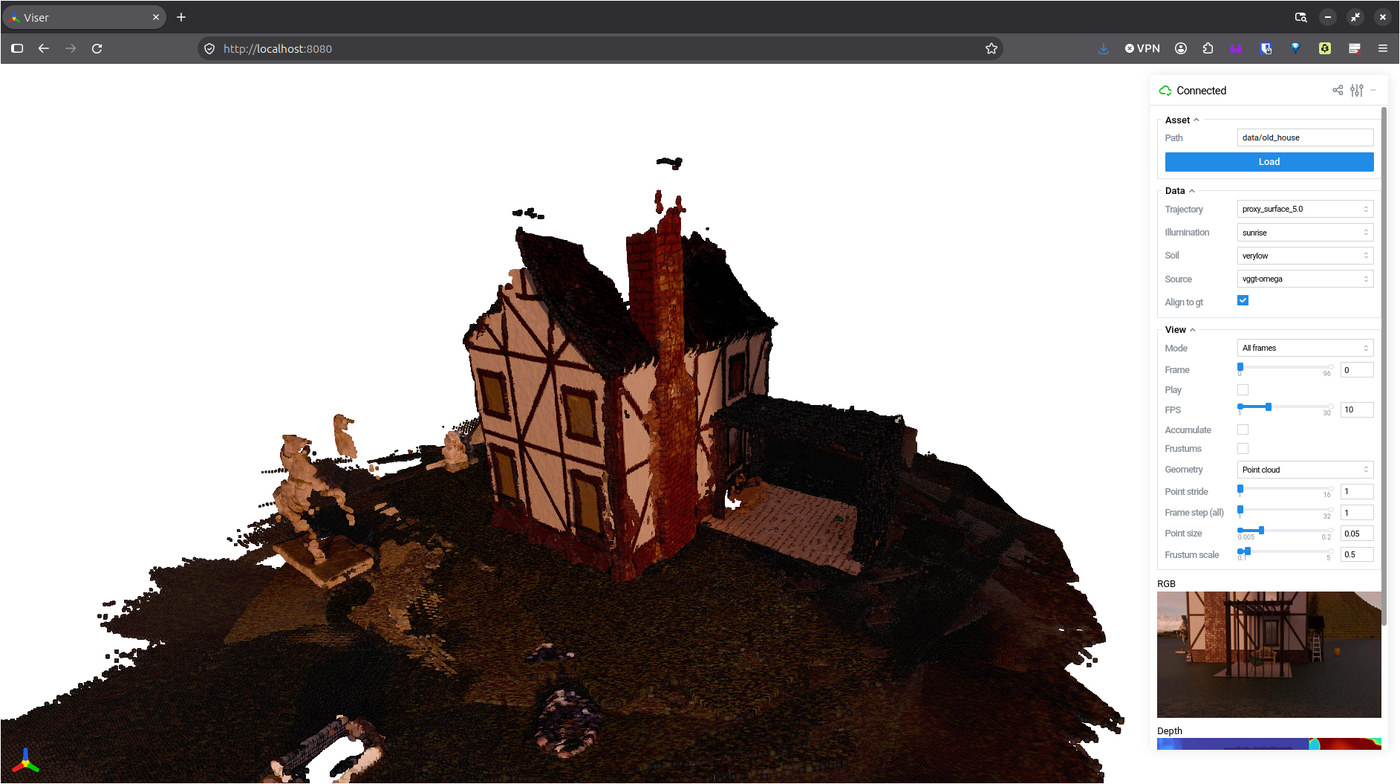} &
    \blank & \blank & \blank \\
  \end{tabular}

  \caption{
  Point clouds of the \textit{Old House} scene, produced by these methods with the \textit{medium} capture pattern and the \textit{Sunrise} lighting condition.
    Note that the colours of the ground truth mesh are inaccurate. 
    Most methods reconstruct the house accurately, but some, such as COLMAP and GLOMAP leave the roof incomplete.
    The background points visible behind the house roof in the COLMAP and GLOMAP reconstructions correspond to actual scene-objects. 
    The colour of the ground truth mesh is not accurate.}
  \label{fig:old_house_point_clouds}
\end{figure*}

\begin{figure*}[t]
  \centering
  \setlength{\tabcolsep}{0pt}
  \renewcommand{\arraystretch}{0}

  \setlength{\imgw}{0.245\textwidth}   

  \newcommand{\cropL}{0.095}
  \newcommand{\cropB}{0}
  \newcommand{\cropR}{0.20}
  \newcommand{\cropT}{0.09}

  \newcommand{\cell}[1]{%
    \settowidth{\natw}{\includegraphics{#1}}%
    \settoheight{\nath}{\includegraphics{#1}}%
    \includegraphics[trim={\cropL\natw} {\cropB\nath} {\cropR\natw} {\cropT\nath},
                     clip, width=\imgw]{#1}}

  \newcommand{\hdr}[1]{\makebox[\imgw][c]{\small #1\rule{0pt}{3ex}}}
  \newcommand{\blank}{\makebox[\imgw]{}}
    
  \begin{tabular}{@{}c c c c@{}}
    \hdr{Ground Truth Mesh} & \hdr{COLMAP} & \hdr{GLOMAP} & \hdr{VidMap} \\
    \cell{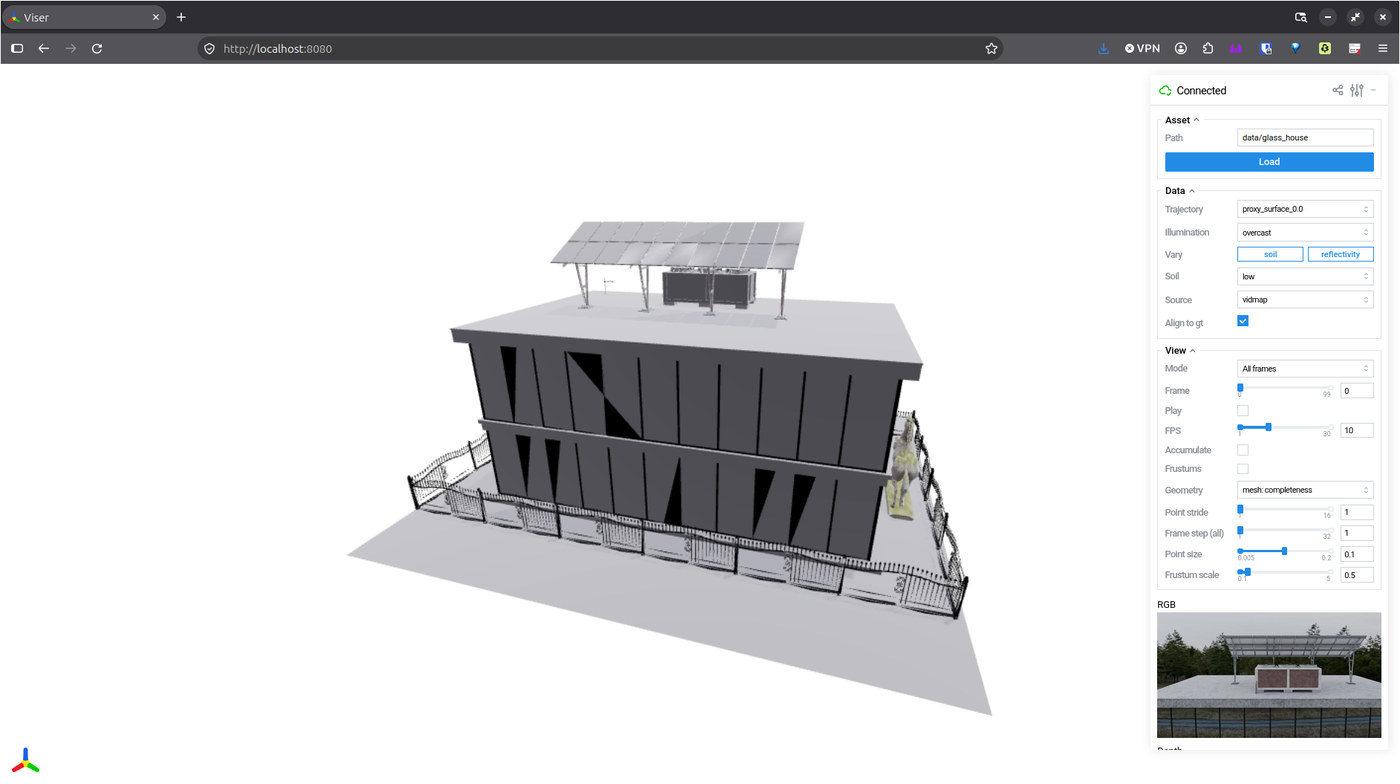} &
    \cell{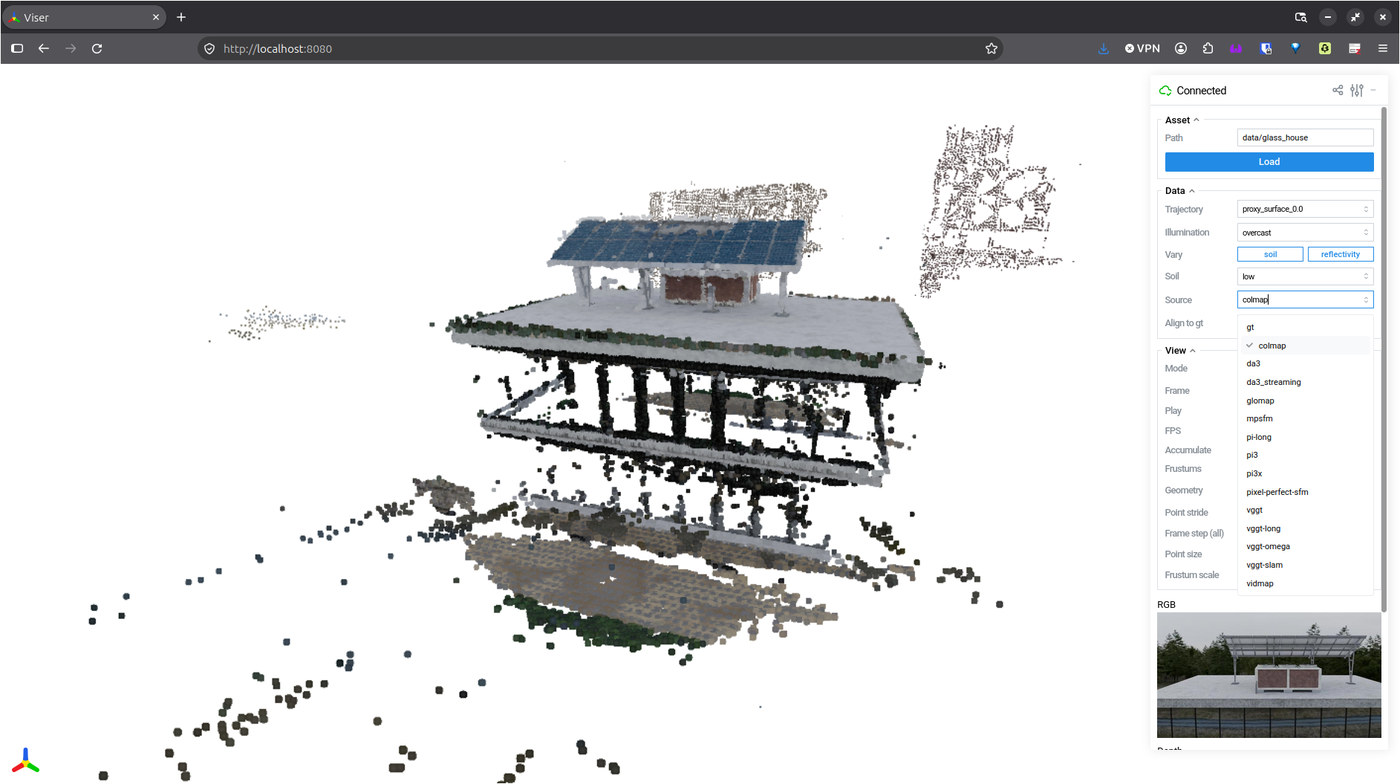} & 
    \cell{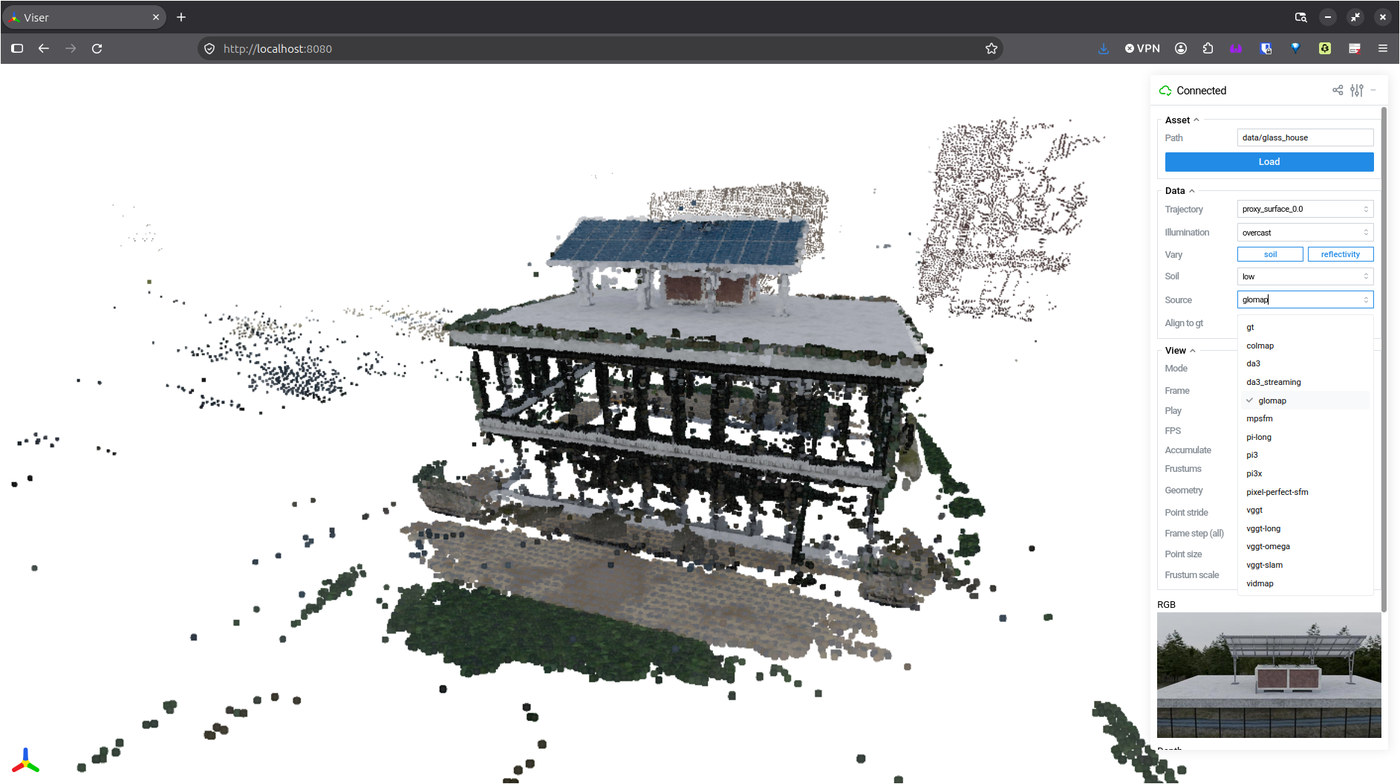} &
    \cell{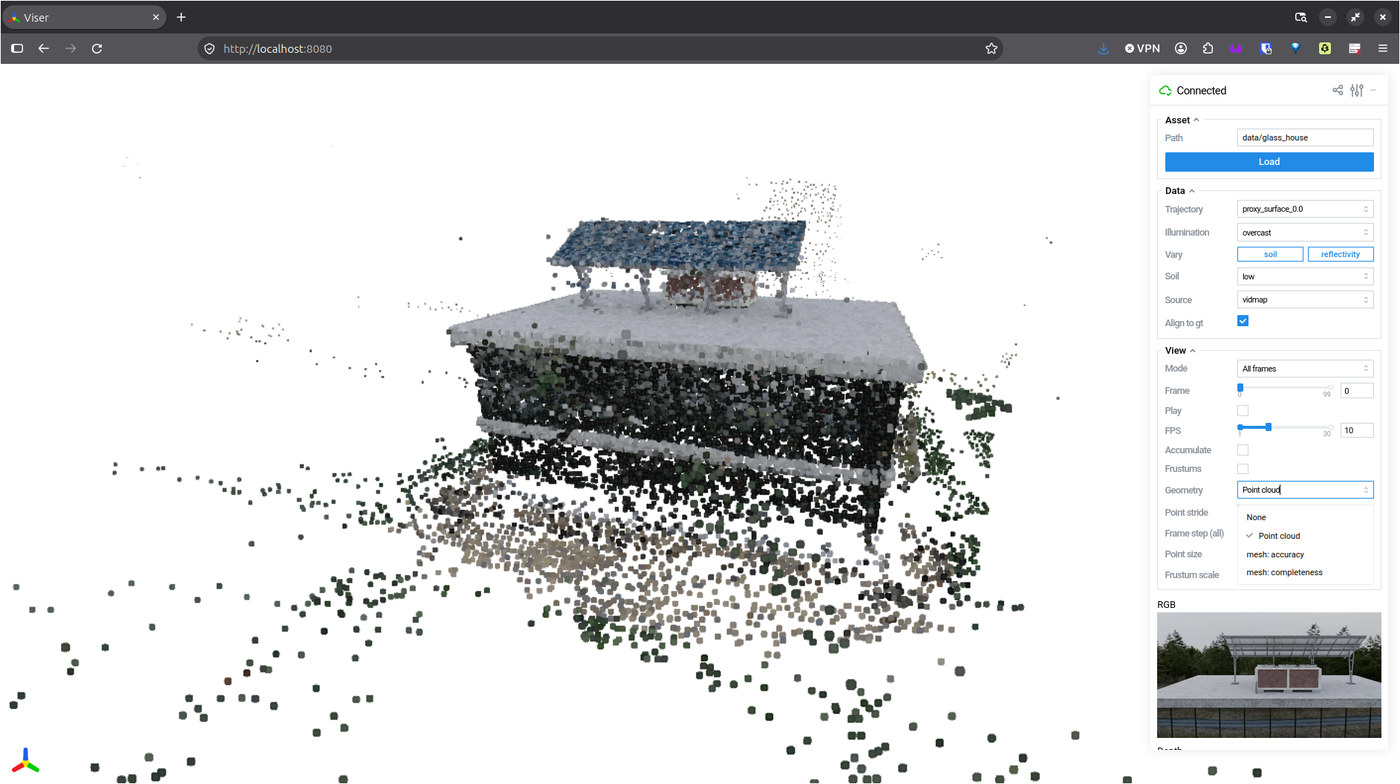} \\

    \hdr{Pi-Long} & \hdr{$\pi^3$}  & \hdr{$\pi^3$-X}  & \hdr{DA3-Streaming} \\
    \cell{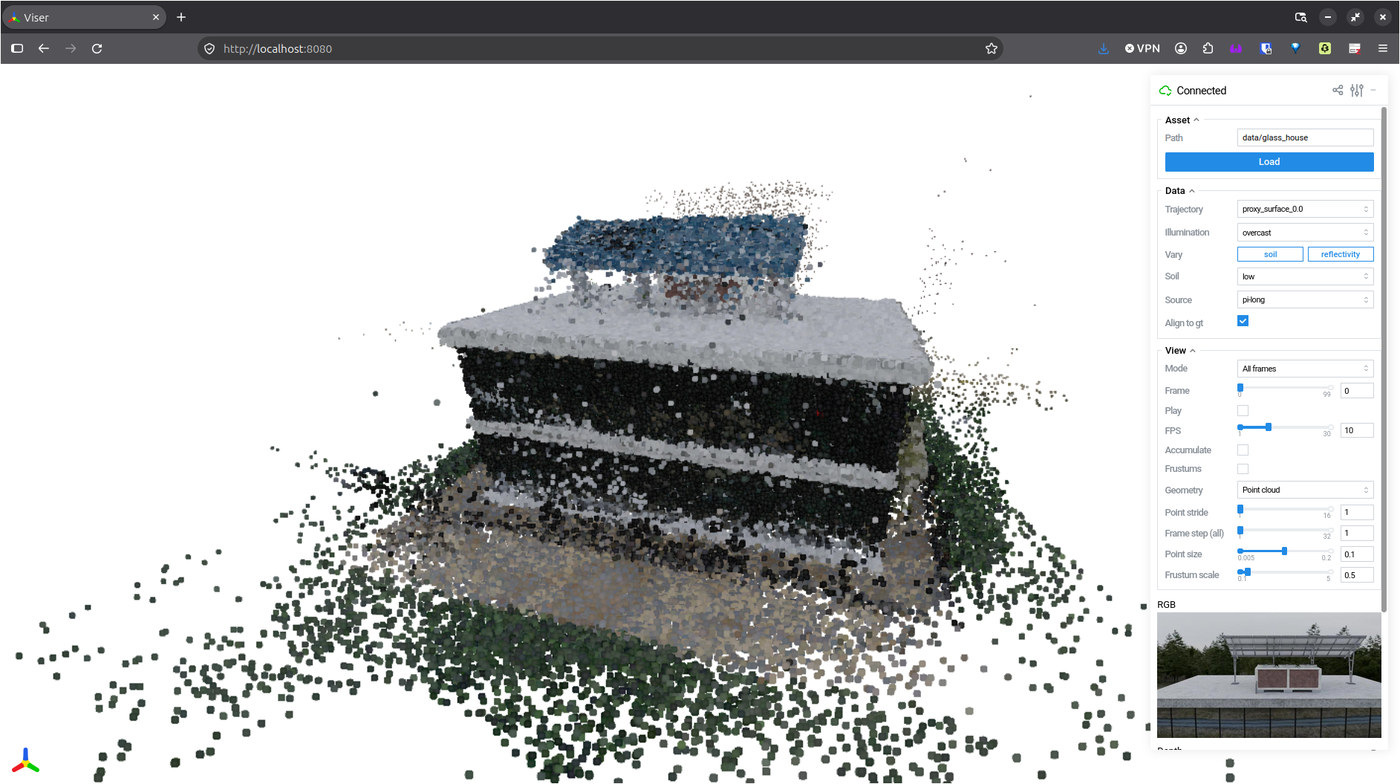} &
    \cell{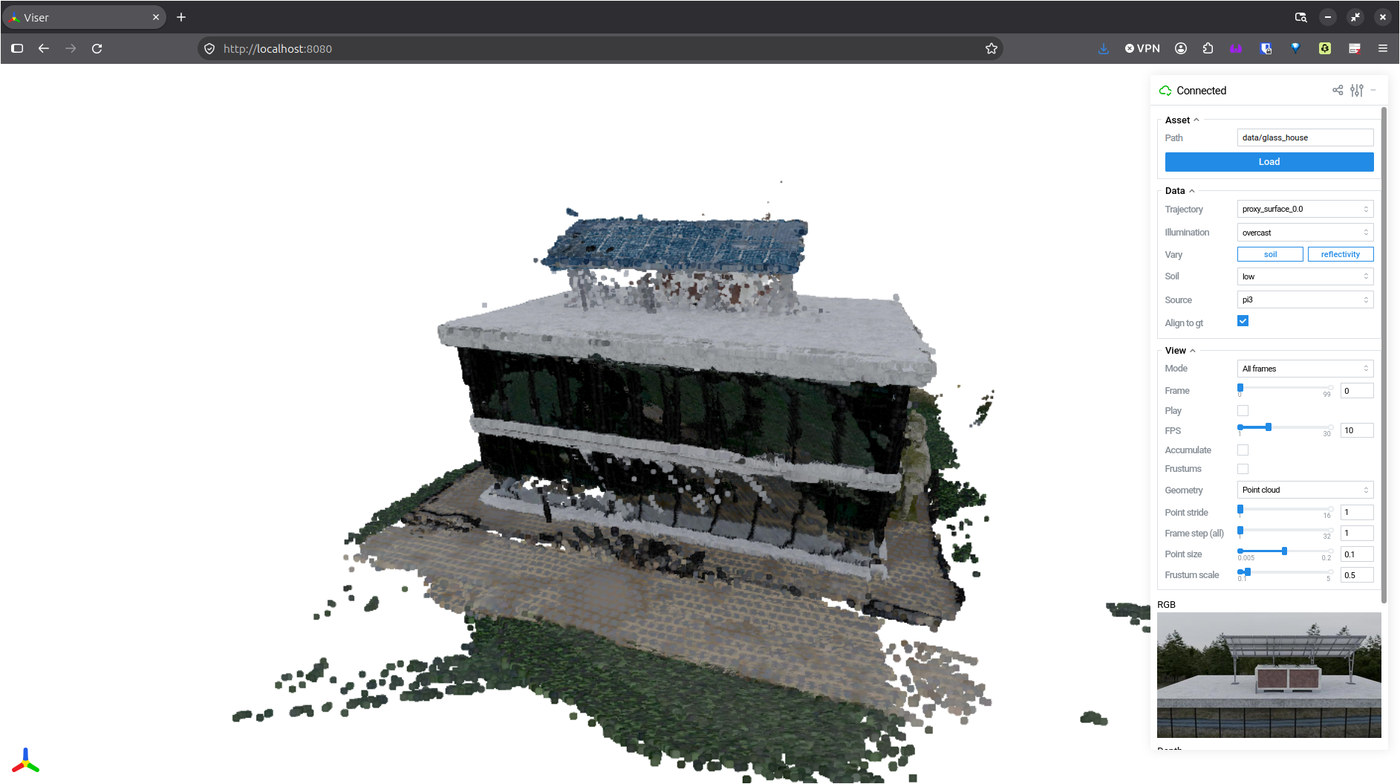} & 
    \cell{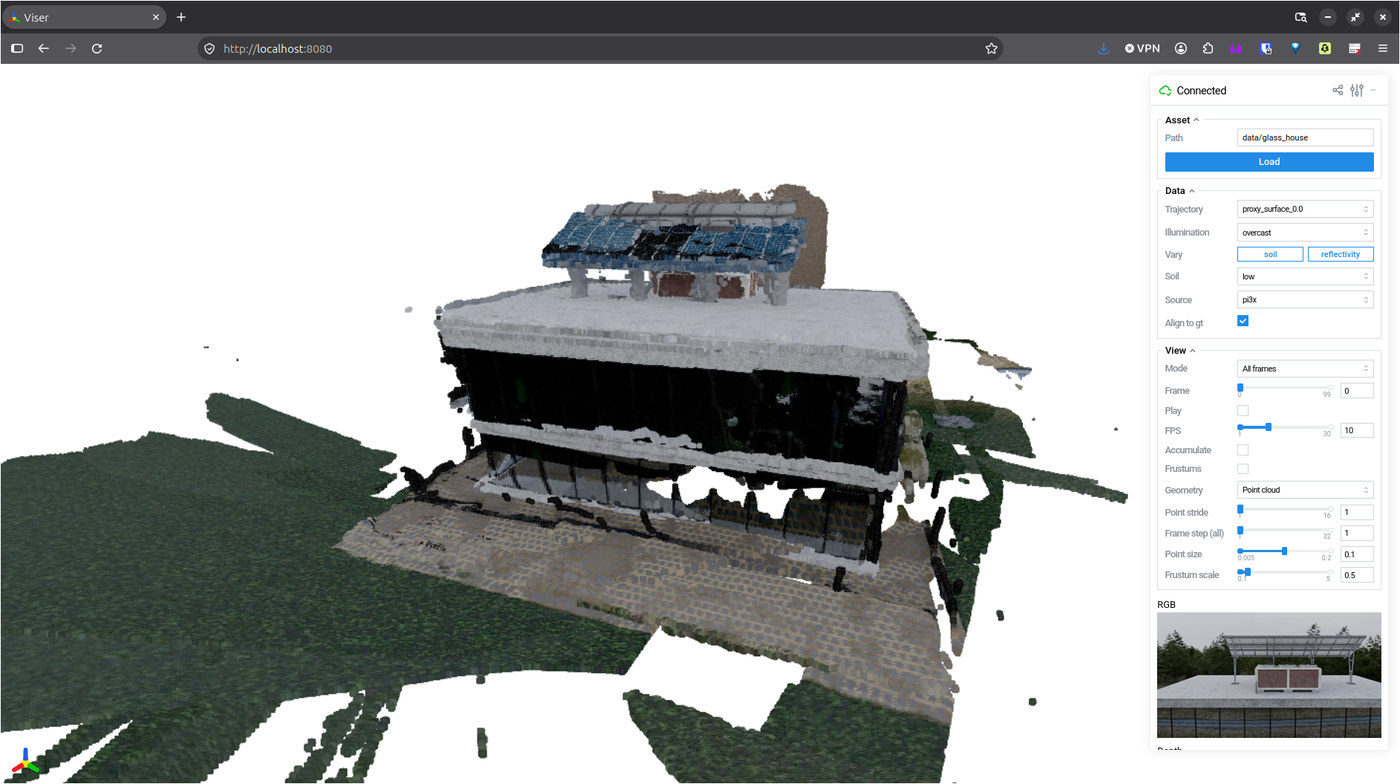} &
    \cell{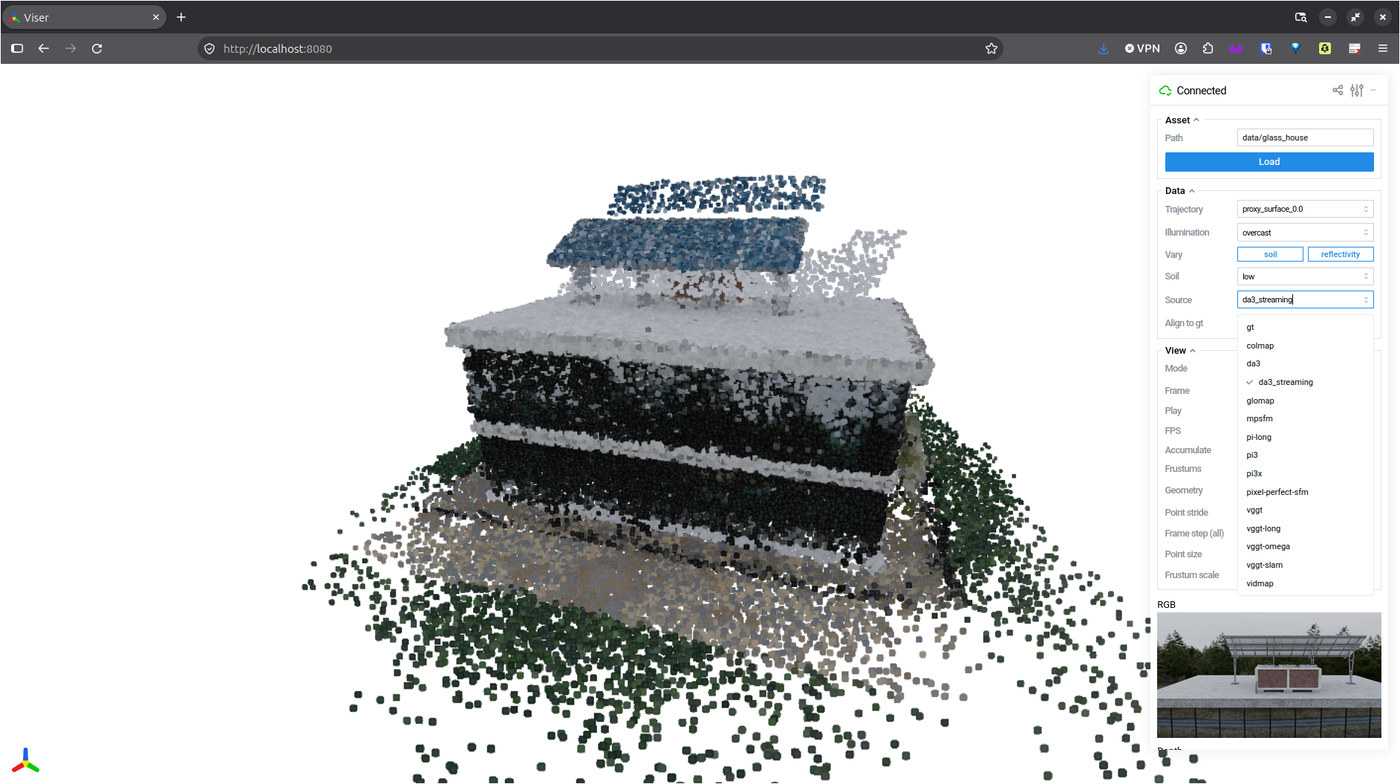} \\

    \hdr{VGGT-SLAM 2.0} & \hdr{VGGT-Long} & \hdr{VGGT} & \hdr{DA3} \\
    \cell{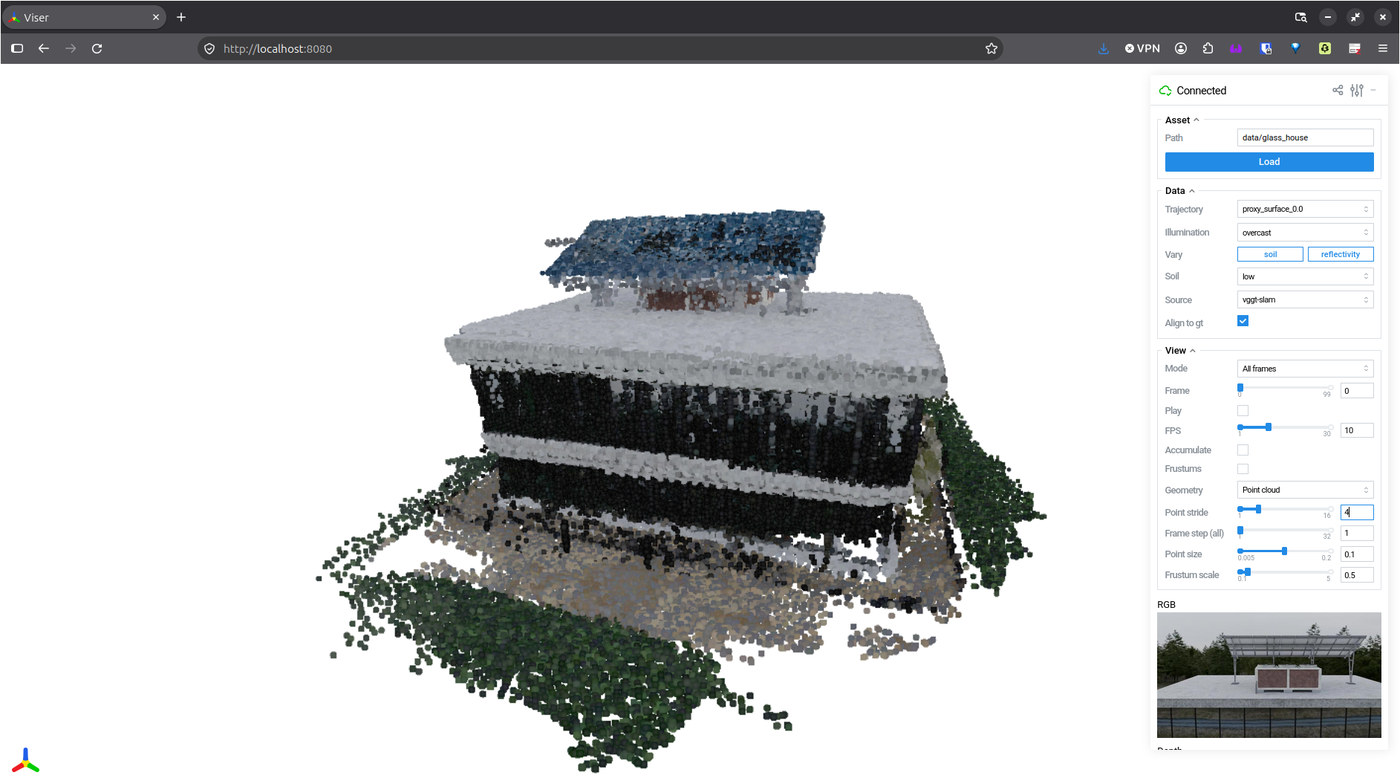} &
    \cell{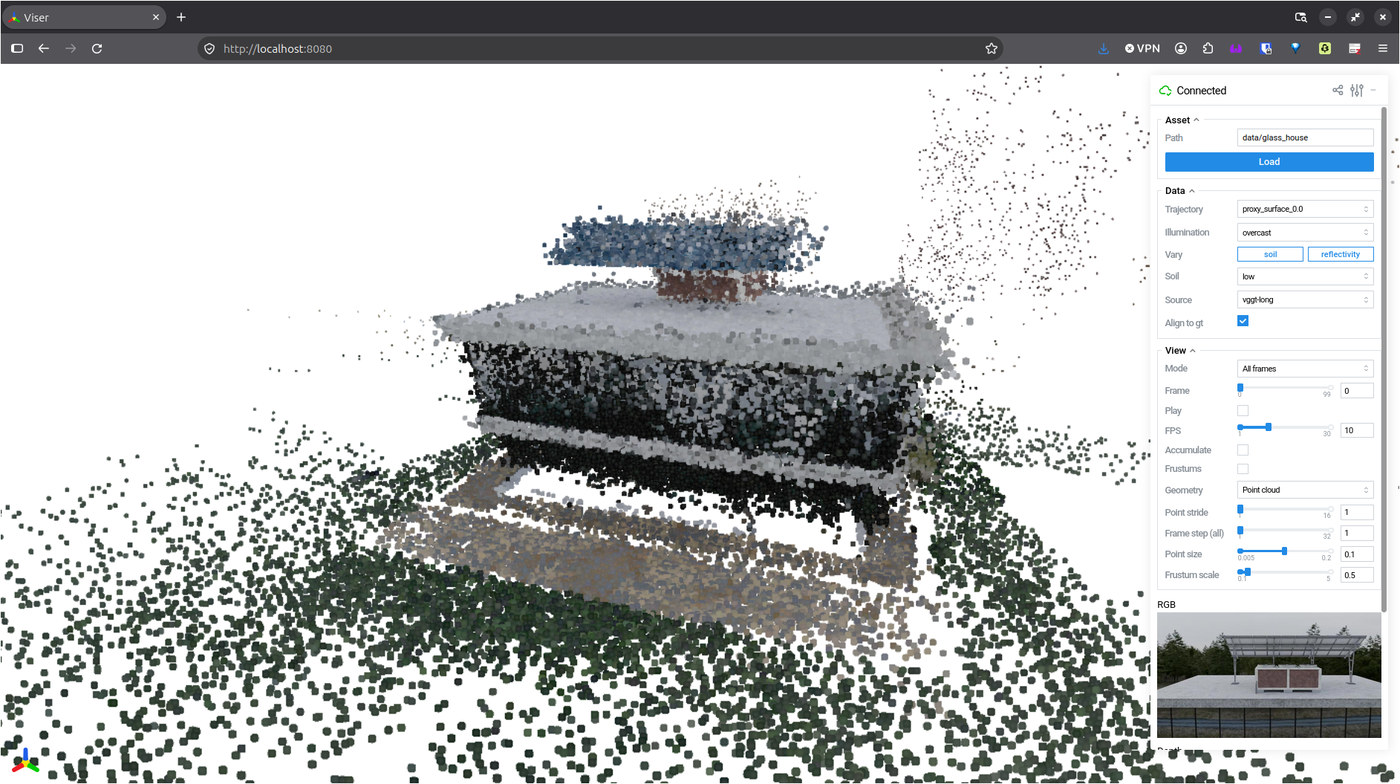} & 
    \cell{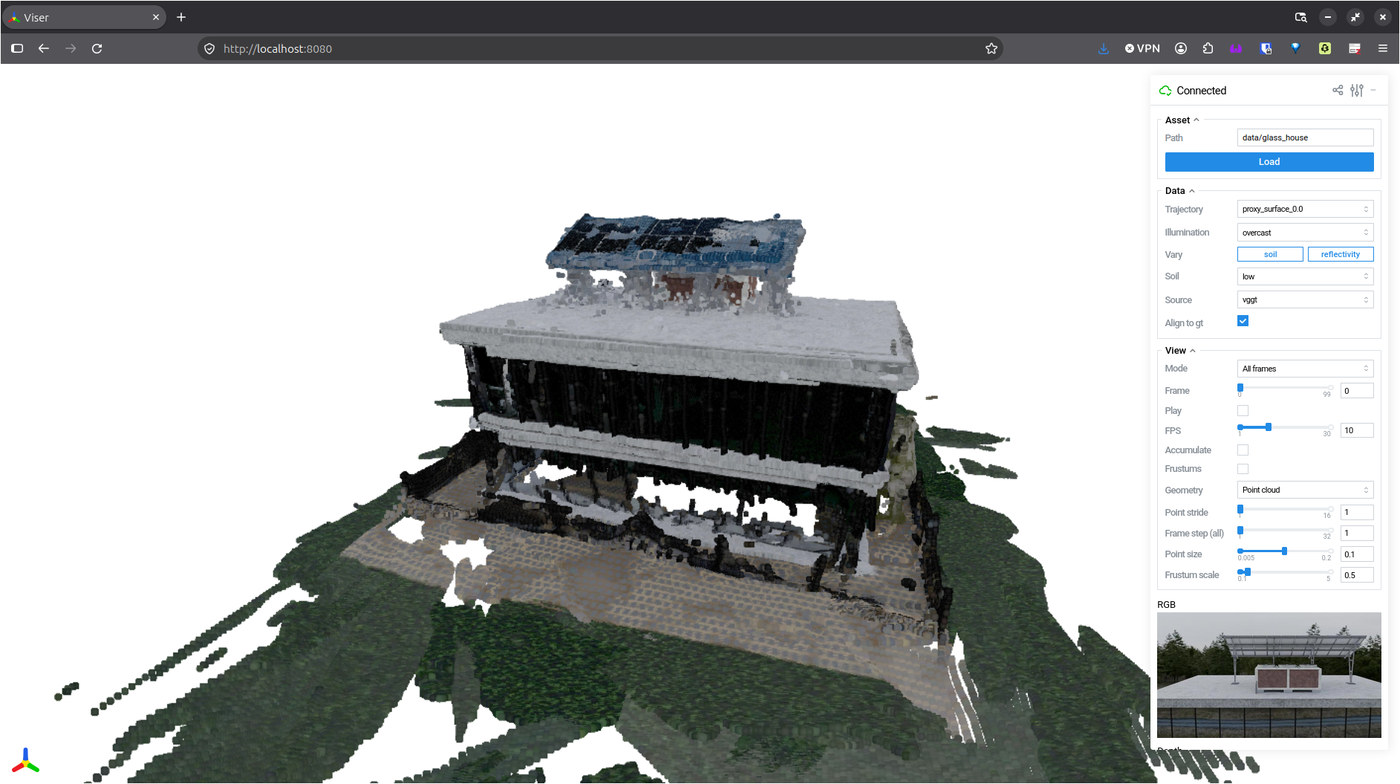} &
    \cell{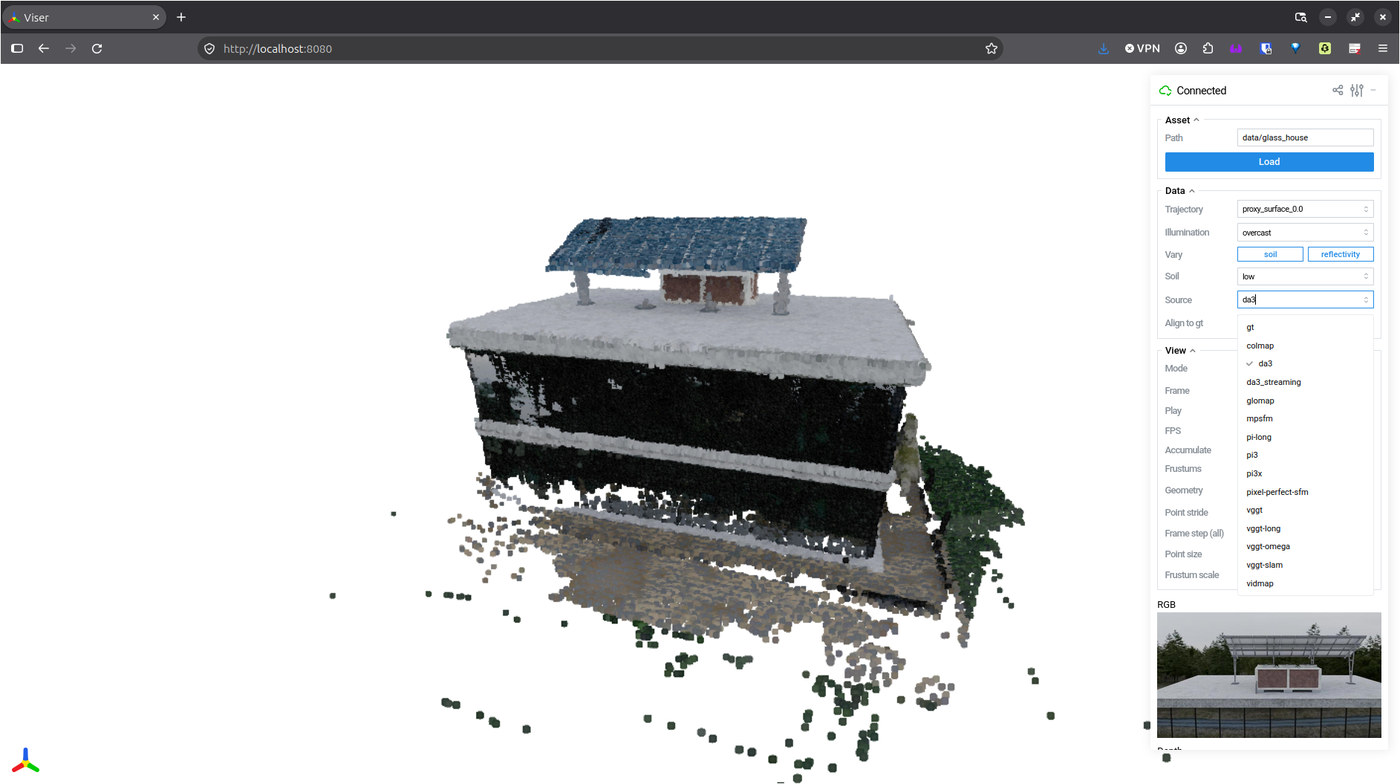} \\
    
    \hdr{VGGT-$\Omega$} &  & \blank & \blank  \\
    \cell{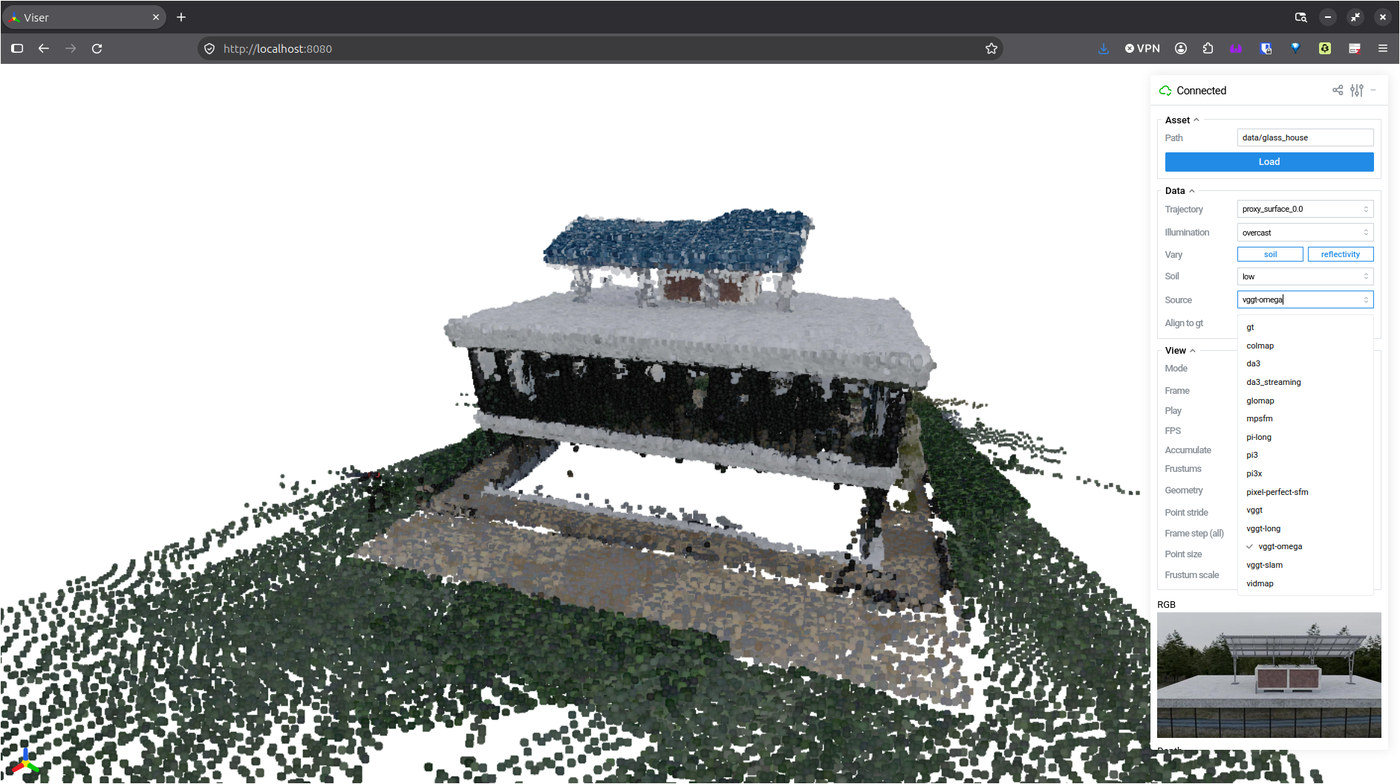} &
    \blank & \blank & \blank \\
  \end{tabular}

  \caption{
  Point clouds of the \textit{Glass House} scene, produced by these methods with the \textit{close} capture pattern and the \textit{Cloudy} lighting condition.
  The surface soiling was set to \textit{very low}.
    Note that the colours of the ground truth mesh are inaccurate. 
    Most methods reconstruct the house accurately, but COLMAP and GLOMAP leave the windows incomplete.
    The points COLMAP and GLOMAp reconstruct in the background do correspond to actual objects in the scene.
    The colour of the ground truth mesh is not accurate.}
  \label{fig:glass_house_point_clouds}
\end{figure*}

\begin{figure*}[t]
  \centering
  \setlength{\tabcolsep}{0pt}
  \renewcommand{\arraystretch}{0}

  \setlength{\imgw}{0.245\textwidth}   
  \newcommand{\cropL}{0.095}
  \newcommand{\cropB}{0}
  \newcommand{\cropR}{0.20}
  \newcommand{\cropT}{0.09}

  \newcommand{\cell}[1]{%
    \settowidth{\natw}{\includegraphics{#1}}%
    \settoheight{\nath}{\includegraphics{#1}}%
    \includegraphics[trim={\cropL\natw} {\cropB\nath} {\cropR\natw} {\cropT\nath},
                     clip, width=\imgw]{#1}}

  \newcommand{\hdr}[1]{\makebox[\imgw][c]{\small #1\rule{0pt}{3ex}}}
  \newcommand{\blank}{\makebox[\imgw]{}}
    
  \begin{tabular}{@{}c c c c@{}}
    \hdr{Ground Truth Point Cloud} & \hdr{COLMAP} & \hdr{GLOMAP} & \hdr{VidMap} \\
    \cell{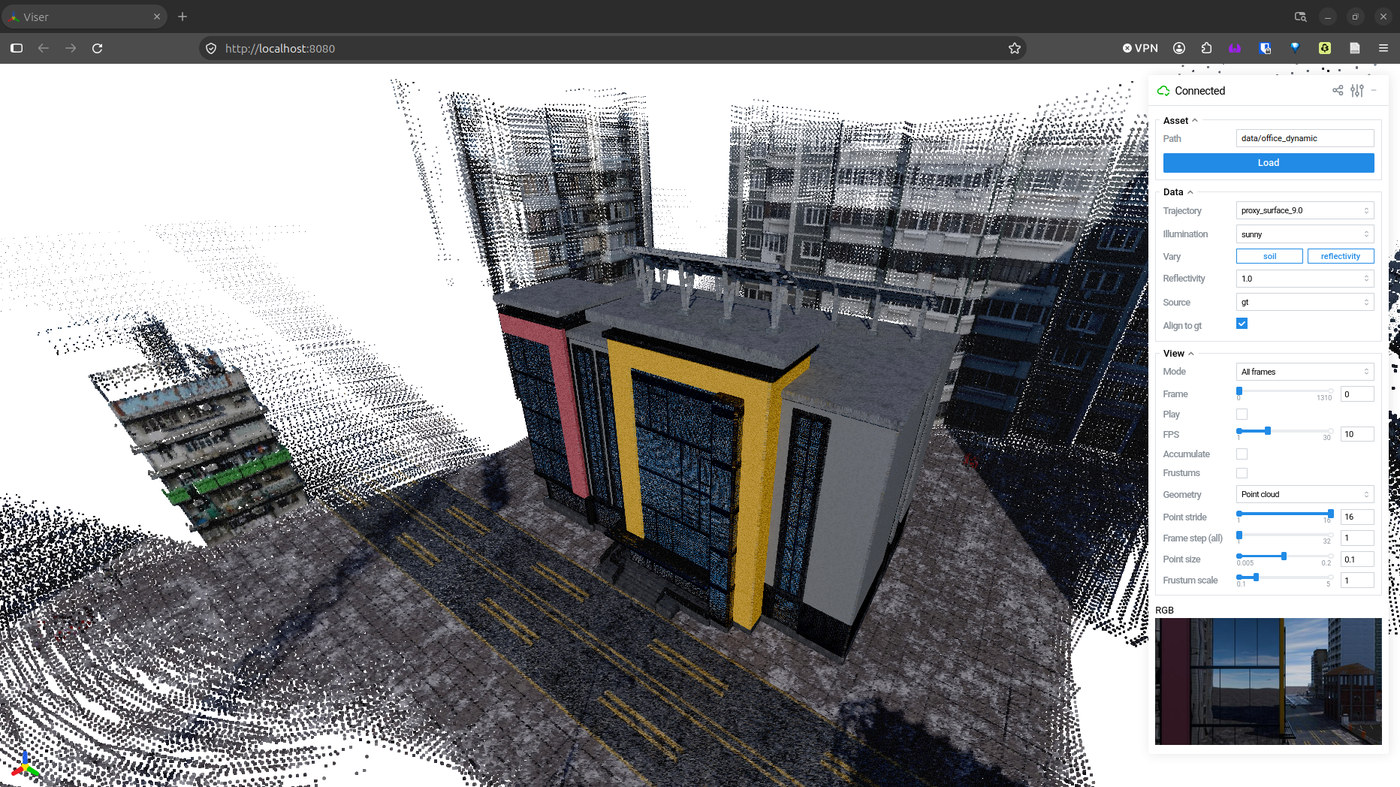} &
    \cell{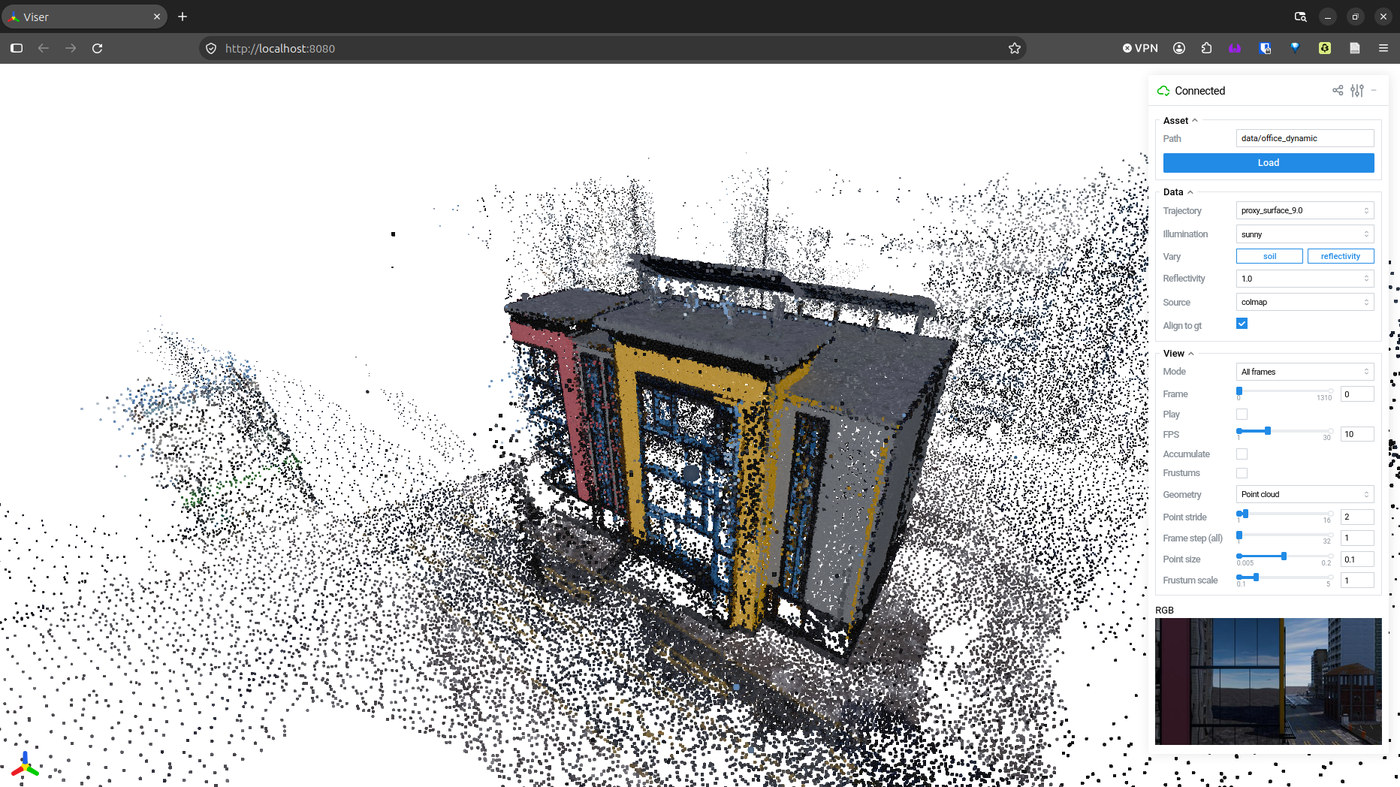} & 
    \cell{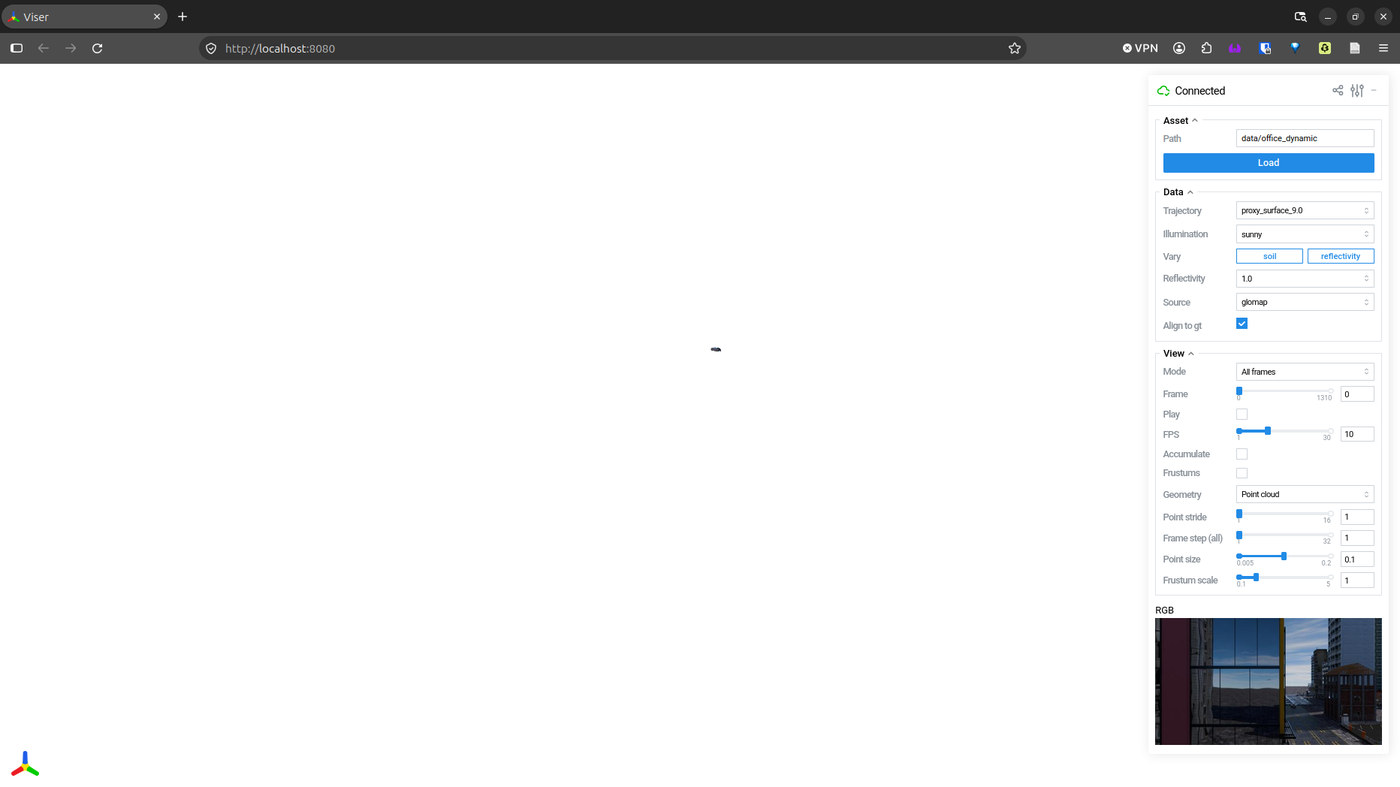} &
    \cell{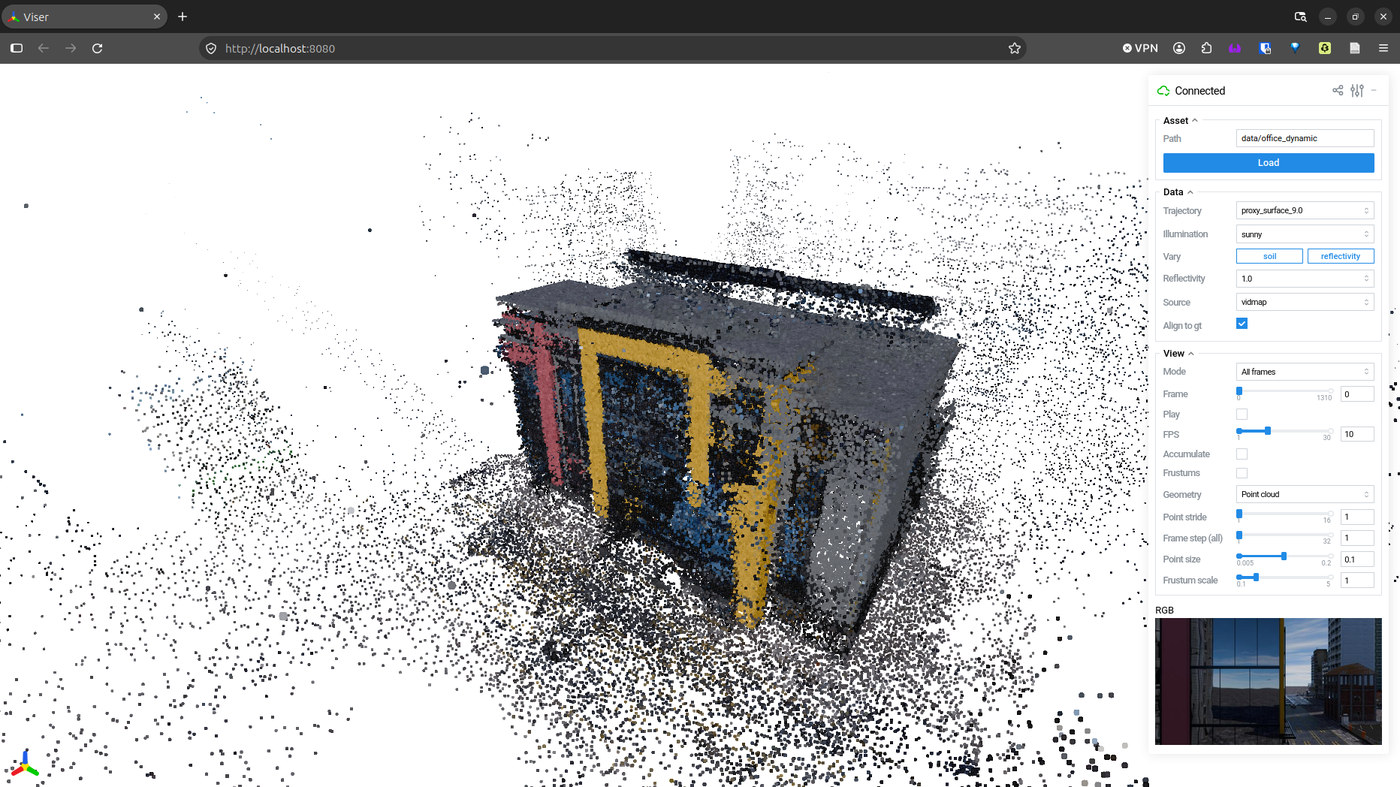} \\

    \hdr{Pi-Long} & \hdr{VGGT-Long}  & \hdr{VGGT-SLAM 2.0}  & \hdr{DA3-Streaming} \\
    \cell{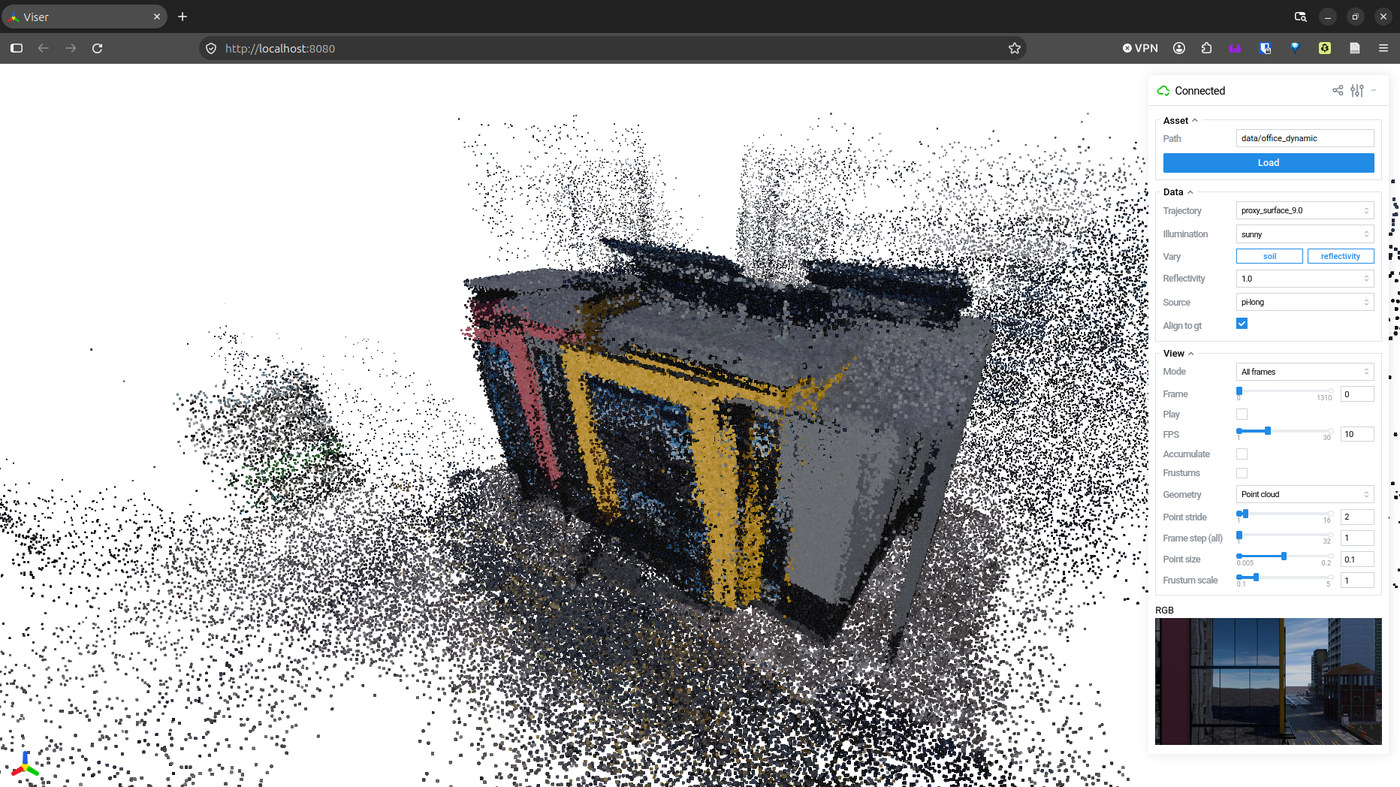} &
    \cell{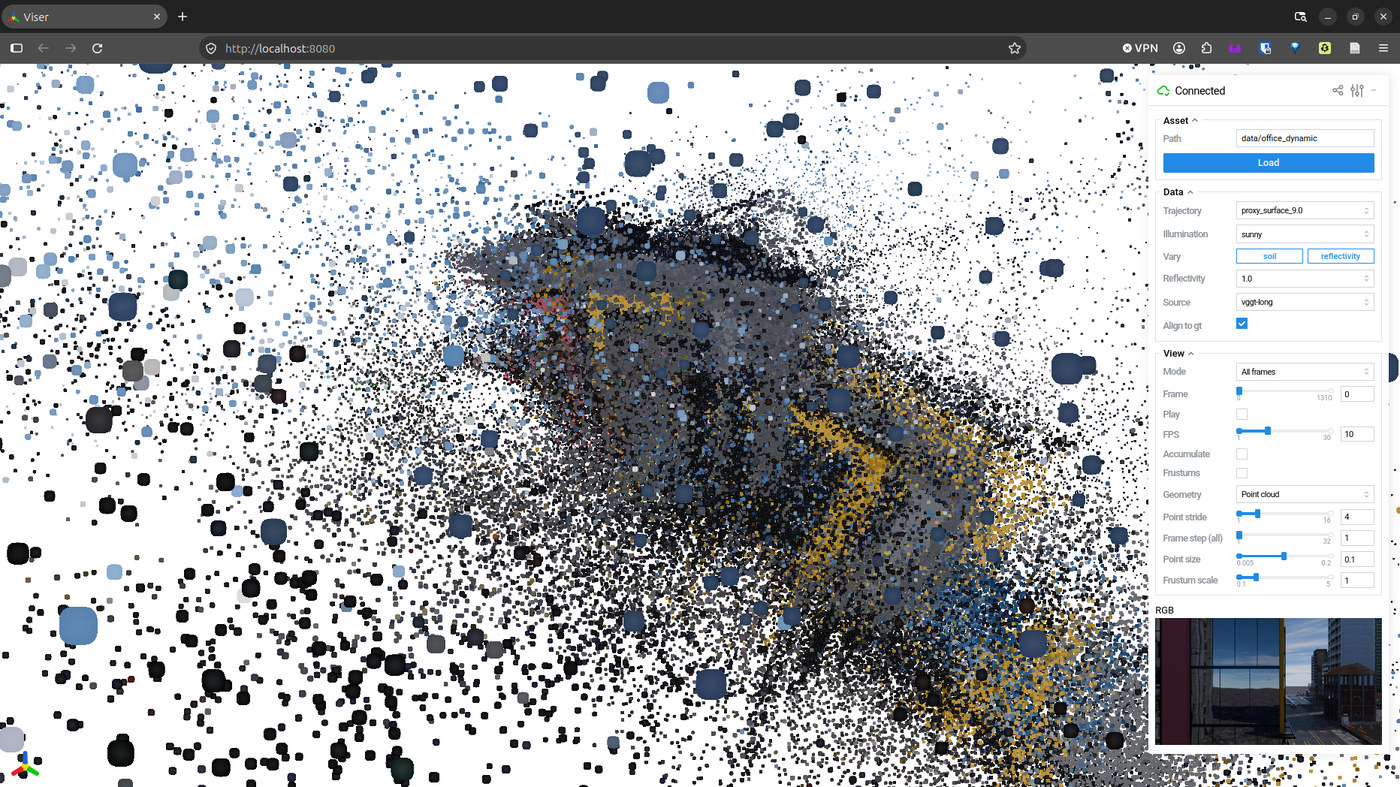} & 
    \cell{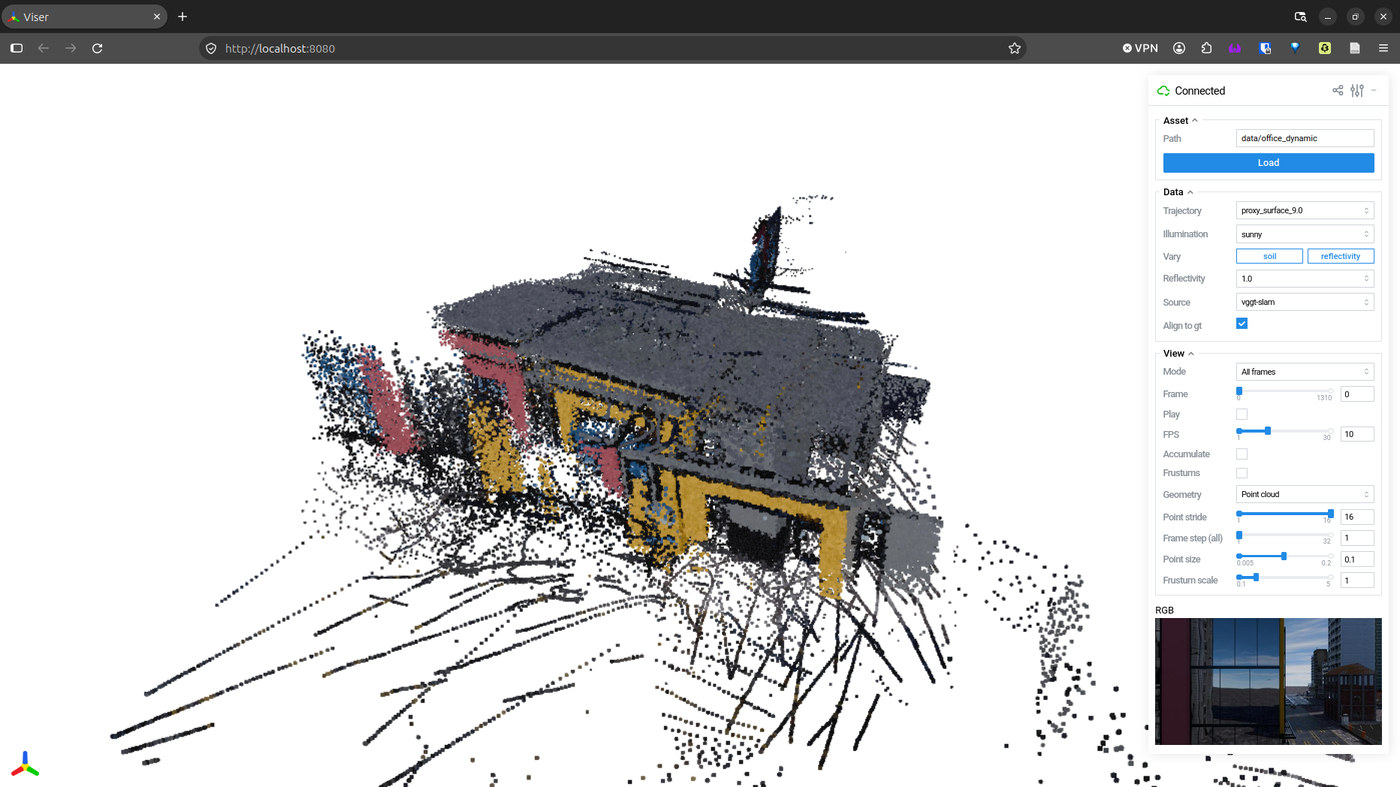} &
    \cell{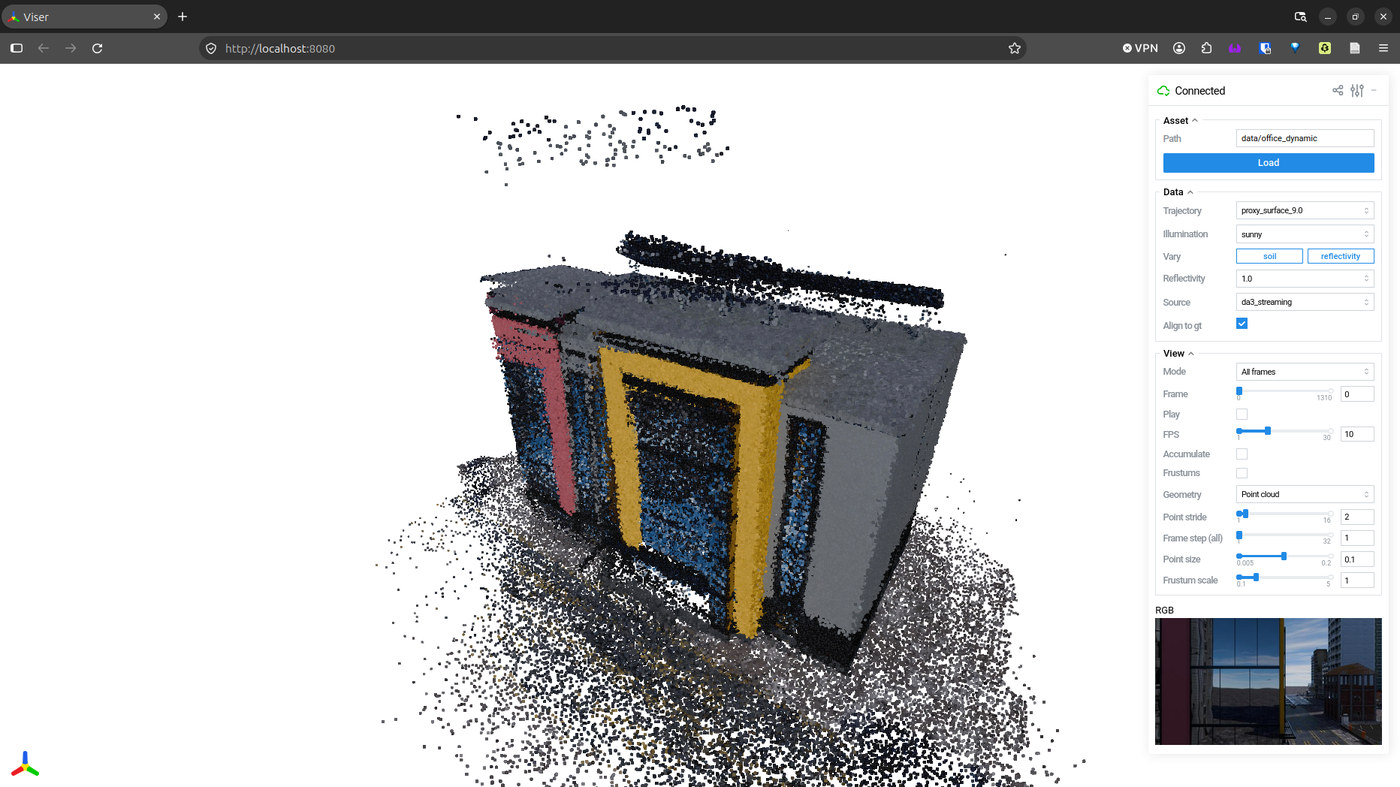} 
  \end{tabular}

  \caption{
  Point clouds of the \textit{Office} scene, produced by these methods with the \textit{far} capture pattern and the \textit{Midday} lighting condition.
  The ground truth point cloud was obtained by projecting image points back onto the scene using the ground truth depth maps.
  Reflectivity was set to 1.0.
  GLOMAP's reconstruction was mostly incomplete and is barely visible.
  The VGGT-based methods struggle with this sequence.
  DA3-Streaming, Pi-Long, VidMap and COLMAP reconstruct the overall shape of the scene at least fairly accurately.
  However, COLMAP fails to reconstruct the windows.
}
  \label{fig:office_point_clouds}
\end{figure*}

We include plots of the point clouds for three sequences.
In \Cref{fig:crane_point_clouds} we show point clouds from the \textit{Crane} under the \textit{Cloudy} lighting condition rendered from the \textit{manual} capture pattern's views.
In \Cref{fig:old_house_point_clouds} we show point clouds from the \textit{Old House} scene under the \textit{Sunrise} lighting condition rendered from the \textit{medium} capture pattern's views.
In \Cref{fig:glass_house_point_clouds} we show point clouds from the \textit{Glass House} scene under the \textit{Cloudy} lighting condition rendered from the \textit{close} capture pattern's views.
The surface soiling for this sequence was set to \textit{very low}.
In \Cref{fig:office_point_clouds} we show point clouds from the \textit{office} scene under the \textit{Sunny} lighting condition rendered from the \textit{far} capture pattern's views.
The reflectivity for this sequence was 1.0.

\subsection{Effect of Trajectory}
\label{supp:trajectory}

We present the effect of camera capture trajectory on depth map accuracy in \cref{fig:traj_depth_maps_full}.

\begin{figure*}[t]
  \centering
  \setlength{\tabcolsep}{2pt}
  \renewcommand{\arraystretch}{1.0}
  \begin{tabular}{@{}ccc@{}}
    \plotcell{trajectory_smse_unmasked_bridge.pdf} &
    \plotcell{trajectory_smse_unmasked_crane.pdf} &
    \plotcell{trajectory_smse_unmasked_unfinished_build.pdf} \\[2pt]
    \plotcell{trajectory_smse_unmasked_glass_house.pdf} &
    \plotcell{trajectory_smse_unmasked_office_dynamic.pdf} &
    \plotcell{trajectory_smse_unmasked_old_house.pdf} \\[2pt]
    \plotcell{trajectory_smse_unmasked_ship.pdf} & & \\[2pt]
  \end{tabular}
  \caption{Per scene plots of sMSE vs. capture trajectory.}
  \label{fig:traj_depth_maps_full}
\end{figure*}

We also present the effect of capture trajectory on point cloud quality in \cref{fig:traj_clouds_full}.

\begin{figure*}[t]
  \centering
  \setlength{\tabcolsep}{2pt}
  \renewcommand{\arraystretch}{1.0}
  \begin{tabular}{@{}ccc@{}}
    \plotcell{trajectory_chamfer_bridge.pdf} &
    \plotcell{trajectory_chamfer_crane.pdf} &
    \plotcell{trajectory_chamfer_unfinished_build.pdf} \\[2pt]
    \plotcell{trajectory_chamfer_glass_house.pdf} &
    \plotcell{trajectory_chamfer_office_dynamic.pdf} &
    \plotcell{trajectory_chamfer_old_house.pdf} \\[2pt]
    \plotcell{trajectory_chamfer_ship.pdf} & & \\[2pt]
  \end{tabular}
  \caption{Per scene plots of Chamfer distance vs. capture trajectory.}
  \label{fig:traj_clouds_full}
\end{figure*}

We also present the effect of capture trajectory on camera pose accuracy quality in \cref{fig:traj_poses_full}.
\begin{figure*}[t]
  \centering
  \setlength{\tabcolsep}{2pt}
  \renewcommand{\arraystretch}{1.0}
  \begin{tabular}{@{}ccc@{}}
    \plotcell{trajectory_auc30_bridge.pdf} &
    \plotcell{trajectory_auc30_crane.pdf} &
    \plotcell{trajectory_auc30_unfinished_build.pdf} \\[2pt]
    \plotcell{trajectory_auc30_glass_house.pdf} &
    \plotcell{trajectory_auc30_office_dynamic.pdf} &
    \plotcell{trajectory_auc30_old_house.pdf} \\[2pt]
    \plotcell{trajectory_auc30_ship.pdf} & & \\[2pt]
  \end{tabular}
  \caption{Per scene plots of AUC@30 vs. capture trajectory.}
  \label{fig:traj_poses_full}
\end{figure*}

\subsection{Effect of reflectivity}
\label{supp:reflectivity}

\begin{figure}[htbp]
    \centering
    \includegraphics[width=0.8\linewidth]{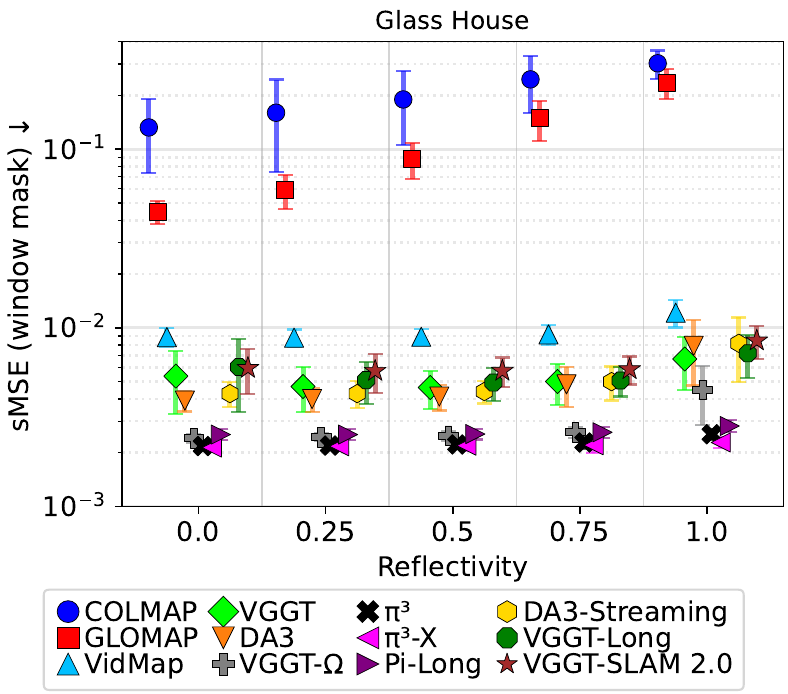}
    \caption{sMSE vs reflectivity on \textit{Glass House} on the window regions.}
    \label{fig:ref_smse_glass}
\end{figure}

\begin{figure}[htbp]
    \centering
    \includegraphics[width=0.8\linewidth]{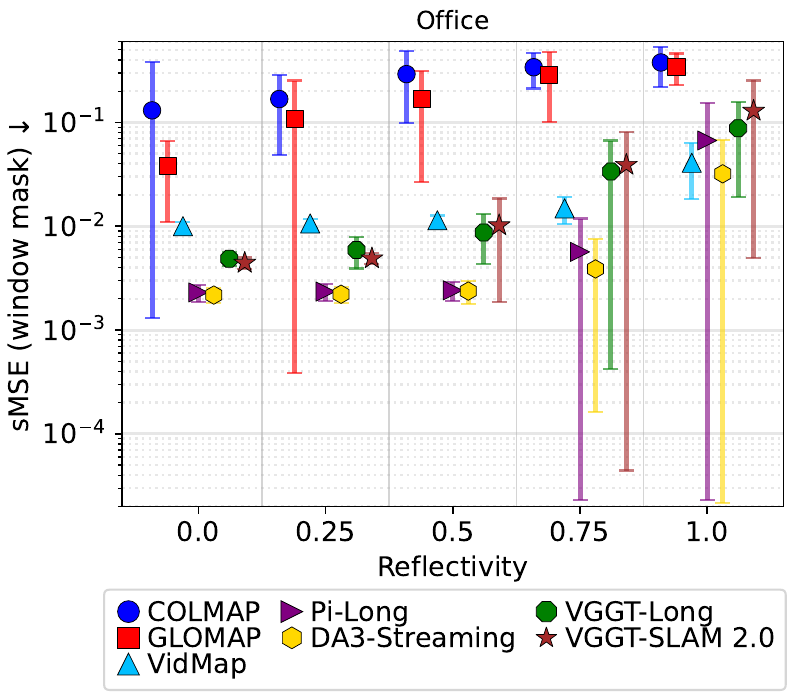}
    \caption{sMSE  vs reflectivity on \textit{Office} on the window regions.}
    \label{fig:ref_smse_office}
\end{figure}

We present the effect of reflectivity on depth map accuracy of both \textit{Glass House} and \textit{Office} in the windowed regions in \cref{fig:ref_smse_glass,fig:ref_smse_office}.

\begin{figure}[htbp]
    \centering
    \includegraphics[width=0.8\linewidth]{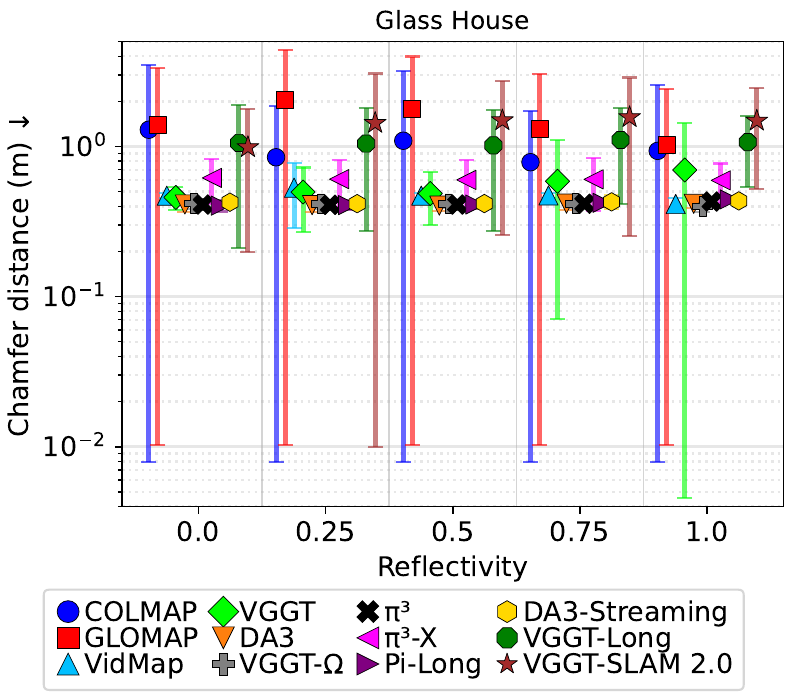}
    \caption{Chamfer distance vs reflectivity on \textit{Glass House}}
    \label{fig:ref_chamfer_glass}
\end{figure}

\begin{figure}[htbp]
    \centering
    \includegraphics[width=0.8\linewidth]{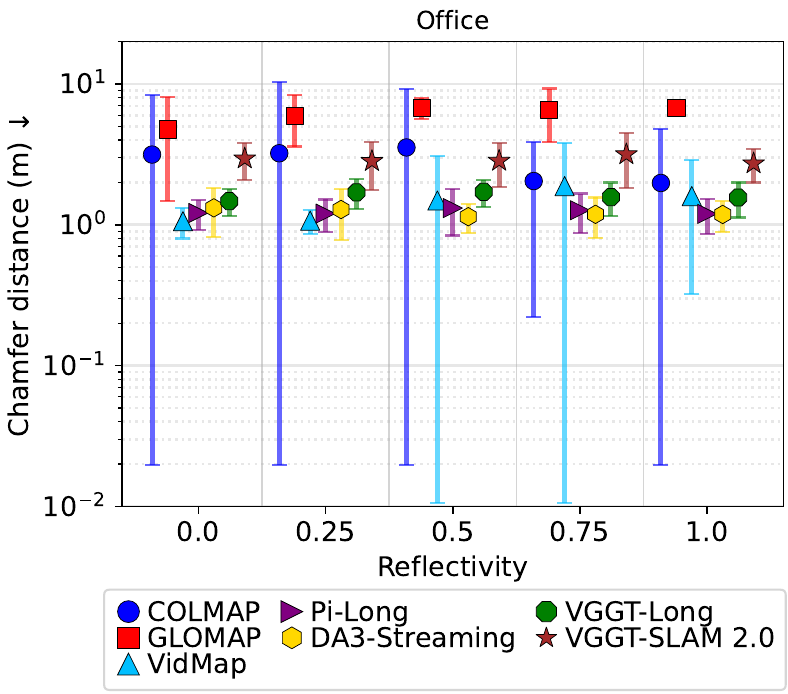}
    \caption{Chamfer distance vs reflectivity on \textit{Office}}
    \label{fig:ref_chamfer_office}
\end{figure}

\begin{figure}[htbp]
    \centering
    \includegraphics[width=0.8\linewidth]{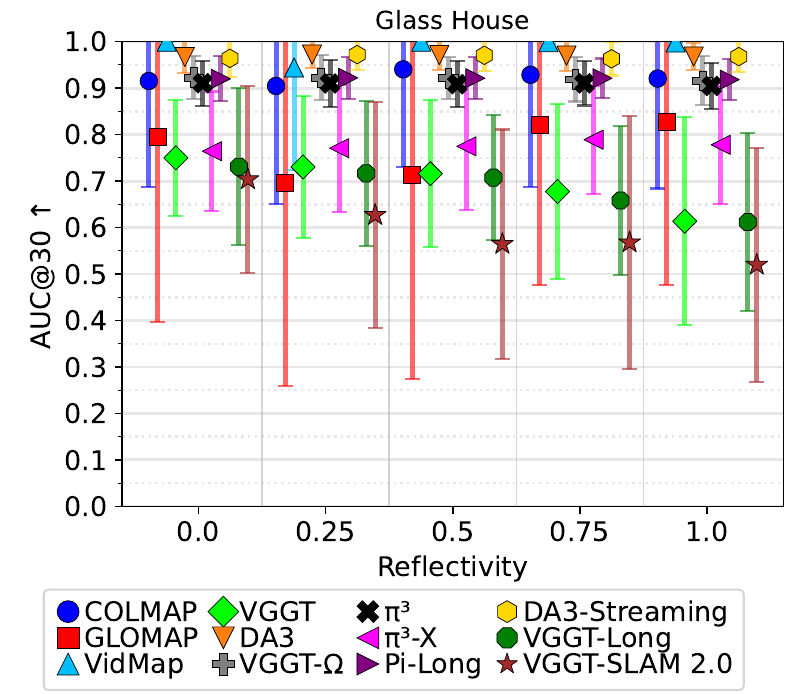}
    \caption{AUC@30 vs reflectivity on \textit{Glass House}}
    \label{fig:ref_auc30_glass}
\end{figure}

\begin{figure}[htbp]
    \centering
    \includegraphics[width=0.8\linewidth]{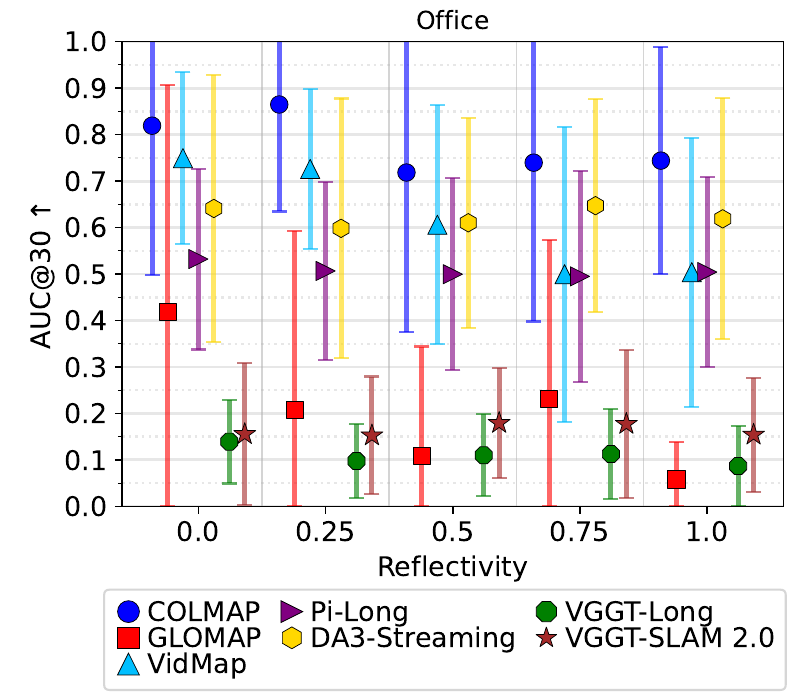}
    \caption{AUC@30 vs reflectivity on \textit{Office}}
    \label{fig:ref_auc30_office}
\end{figure}

The effect of reflectivity on scene-level reconstruction is shown in \cref{fig:ref_chamfer_glass,fig:ref_chamfer_office,fig:ref_auc30_glass,fig:ref_auc30_office}.

\subsection{Effect of Soiling}
\label{supp:soiling}

\begin{figure}[htbp]
    \centering
    \includegraphics[width=0.8\linewidth]{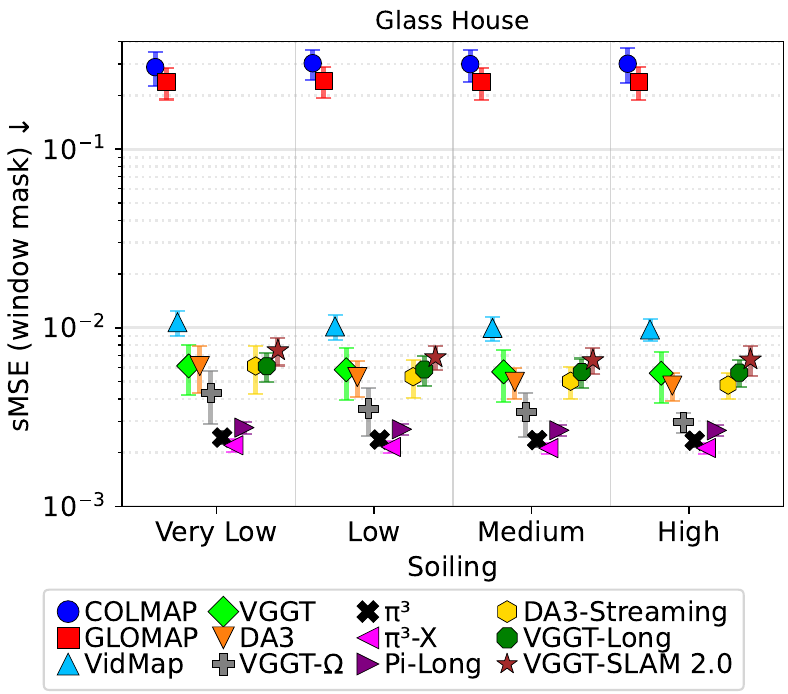}
    \caption{SMSE (window masks) vs soiling on \textit{Glass House}}
    \label{fig:soiling_smse_glass}
\end{figure}

\begin{figure}[htbp]
    \centering
    \includegraphics[width=0.8\linewidth]{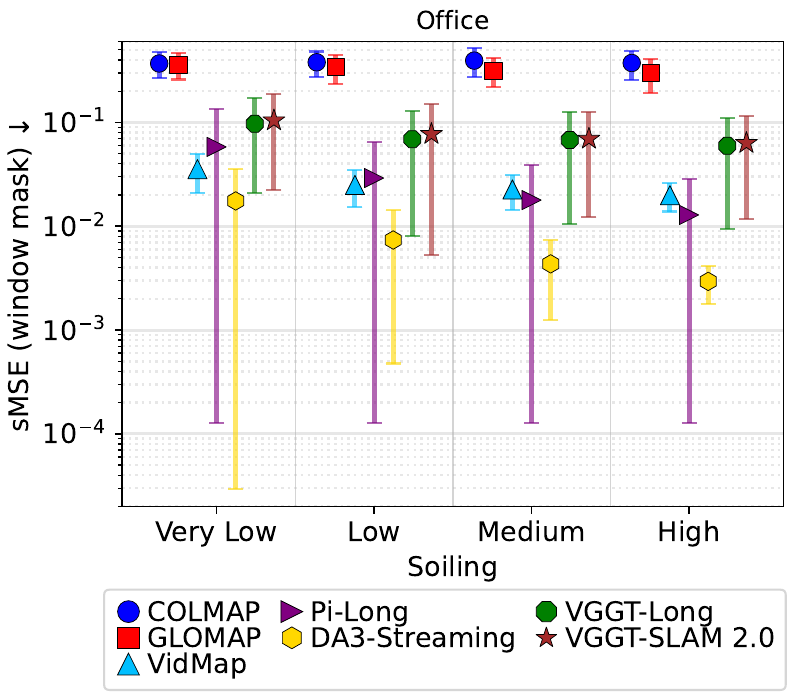}
    \caption{SMSE (window masks) vs soiling on \textit{Office}}
    \label{fig:soiling_smse_office}
\end{figure}

We present the impact of soiling on depth map accuracy in windowed regions in \cref{fig:soiling_smse_glass,fig:soiling_smse_office}. 

\begin{figure}[htbp]
    \centering
    \includegraphics[width=0.8\linewidth]{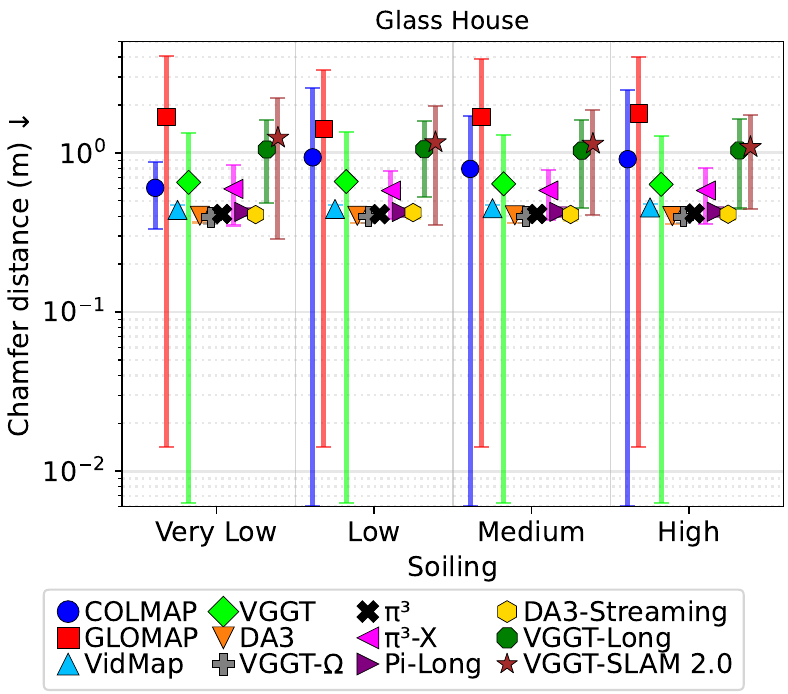}
    \caption{Chamfer distance vs soiling on \textit{Glass House}}
    \label{fig:soiling_chamfer_glass}
\end{figure}

\begin{figure}[htbp]
    \centering
    \includegraphics[width=0.8\linewidth]{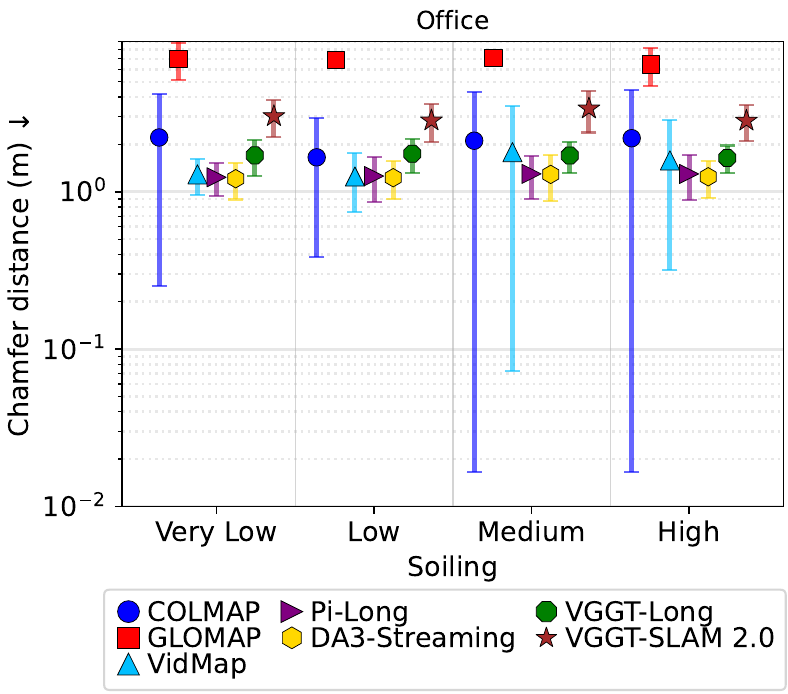}
    \caption{Chamfer distance vs soiling on \textit{Office}}
    \label{fig:soiling_chamfer_office}
\end{figure}

\begin{figure}[htbp]
    \centering
    \includegraphics[width=0.8\linewidth]{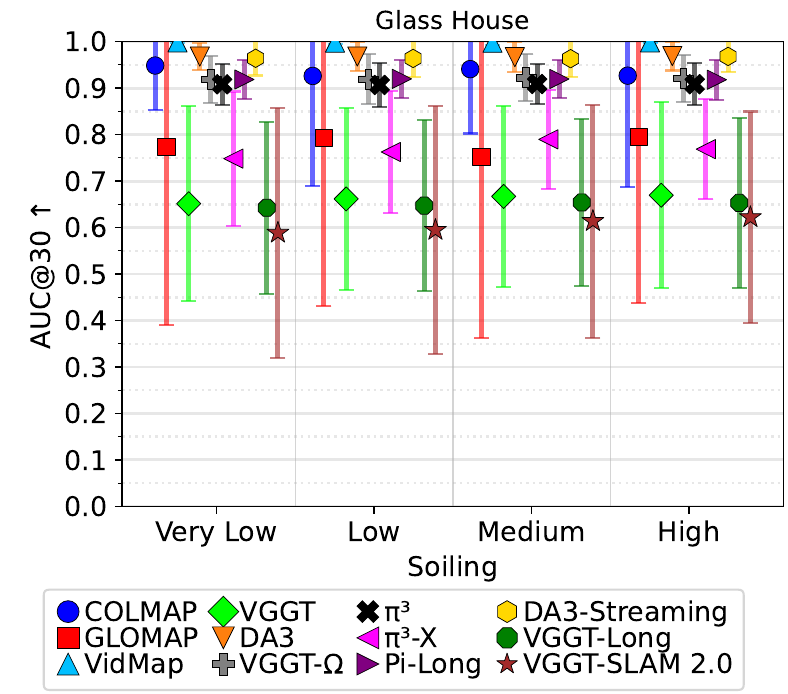}
    \caption{AUC@30 vs soiling on \textit{Glass House}}
    \label{fig:soiling_auc30_glass}
\end{figure}

\begin{figure}[htbp]
    \centering
    \includegraphics[width=0.8\linewidth]{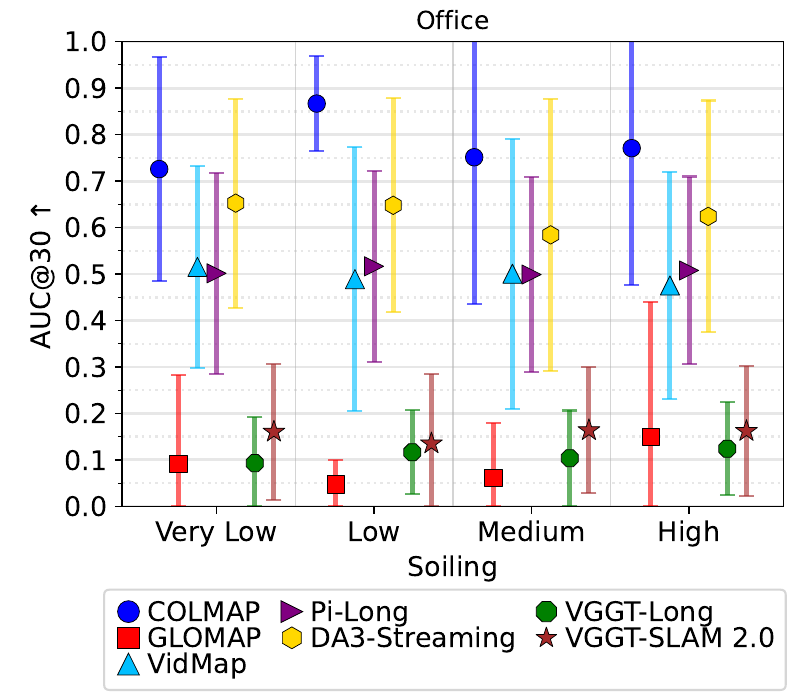}
    \caption{AUC@30 vs soiling on \textit{Office}}
    \label{fig:soiling_auc30_office}
\end{figure}

The effect of soiling on point cloud quality is shown in \cref{fig:soiling_chamfer_glass,fig:soiling_chamfer_office} and its effect on camera pose accuracy is shown in \cref{fig:soiling_auc30_glass,fig:soiling_auc30_office}.

\subsection{Effect of Lighting}
\label{supp:lighting}

We also present the effect of different lighting conditions on depth map accuracy in \cref{fig:illumination_depth_map_full}.

\begin{figure*}[t]
  \centering
  \setlength{\tabcolsep}{2pt}
  \renewcommand{\arraystretch}{1.0}
  \begin{tabular}{@{}ccc@{}}
    \plotcell{illumination_smse_unmasked_bridge.pdf} &
    \plotcell{illumination_smse_unmasked_crane.pdf} &
    \plotcell{illumination_smse_unmasked_unfinished_build.pdf} \\[2pt]
    \plotcell{illumination_smse_unmasked_glass_house.pdf} &
    \plotcell{illumination_smse_unmasked_office_dynamic.pdf} &
    \plotcell{illumination_smse_unmasked_old_house.pdf} \\[2pt]
    \plotcell{illumination_smse_unmasked_ship.pdf} & & \\[2pt]
  \end{tabular}
  \caption{Per scene plots of sMSE vs. lighting.}
  \label{fig:illumination_depth_map_full}
\end{figure*}

We also present the effect of different lighting conditions on point cloud quality and pose accuracy in \cref{fig:illumination_clouds_full,fig:illumination_poses_full}.

\begin{figure*}[t]
  \centering
  \setlength{\tabcolsep}{2pt}
  \renewcommand{\arraystretch}{1.0}
  \begin{tabular}{@{}ccc@{}}
    \plotcell{illumination_chamfer_bridge.pdf} &
    \plotcell{illumination_chamfer_crane.pdf} &
    \plotcell{illumination_chamfer_unfinished_build.pdf} \\[2pt]
    \plotcell{illumination_chamfer_glass_house.pdf} &
    \plotcell{illumination_chamfer_office_dynamic.pdf} &
    \plotcell{illumination_chamfer_old_house.pdf} \\[2pt]
    \plotcell{illumination_chamfer_ship.pdf} & & \\[2pt]
  \end{tabular}
  \caption{Per scene plots of Chamfer distance vs. lighting.}
  \label{fig:illumination_clouds_full}
\end{figure*}

\begin{figure*}[t]
  \centering
  \setlength{\tabcolsep}{2pt}
  \renewcommand{\arraystretch}{1.0}
  \begin{tabular}{@{}ccc@{}}
    \plotcell{illumination_auc30_bridge.pdf} &
    \plotcell{illumination_auc30_crane.pdf} &
    \plotcell{illumination_auc30_unfinished_build.pdf} \\[2pt]
    \plotcell{illumination_auc30_glass_house.pdf} &
    \plotcell{illumination_auc30_office_dynamic.pdf} &
    \plotcell{illumination_auc30_old_house.pdf} \\[2pt]
    \plotcell{illumination_auc30_ship.pdf} & & \\[2pt]
  \end{tabular}
  \caption{Per scene plots of AUC@30 vs. lighting.}
  \label{fig:illumination_poses_full}
\end{figure*}

\subsection{Open Challenges and Failure Cases}
\label{supp:failure}

We include plots of estimated point clouds and camera poses for \textit{Crane} scene under the \textit{manual} capture pattern and the \textit{Cloudy} lighting condition in \Cref{fig:crane_point_clouds}.
We note that no method evaluated performs well on this sequence.

\clearpage
\section{Blender Assets Used}
\label{supp:assets}
\label{sec:BlenderAssets}
To make the scenes, we primarily used open-source resources made by artists available on the blendkit platform: \url{www.blendkit.com}, \url{sketchfab.com}. All assets from blendkit have an RF or CC0 license. 
Assets from Sketchfab have a CC-Attribution license. 
These resources include materials, models, and HDRis, which are listed below:

\begin{enumerate}
    \item \label{itm:blenderBridge} Cosmo --- \href{https://www.blendkit.com/asset-gallery-detail/5358ddd3-6dbf-4a1c-8731-4c6bdcfa8338/}{Suspension/Arch Bridge}
    \item \label{itm:blenderCrane} MapacheDRelease --- \href{https://www.blendkit.com/asset-gallery-detail/eb5c137e-007e-485b-bbb9-675e76fcdac9/}{Crane Tower}
    \item \label{itm:blenderGrass} Mohammadi, S. --- \href{https://www.blendkit.com/asset-gallery-detail/a465bd1e-a14d-442f-90b4-0bfb02dffa9f/}{Realistic Procedural Grass (Cycles/Eevee)}
    \item \label{itm:blenderConcrete} Godoi, M. --- \href{https://www.blendkit.com/asset-gallery-detail/6312d40c-7afb-49e3-b167-c772147ce40d/}{Procedural Concrete}
    \item \label{itm:blendeBrick1} Russo 3D --- \href{https://www.blendkit.com/asset-gallery-detail/f5ab3780-1489-4c62-8050-b400bbd14e95/}{Procedural Brick}
    \item \label{itm:blendeBrick2} Middleton, J. --- \href{https://www.blendkit.com/asset-gallery-detail/fa2d62e6-fc17-4310-a0bf-b35eaefcfd3c/}{Bricks Procedural Wall}
    \item \label{itm:blendeBrick3} Middleton, J. --- \href{https://www.blendkit.com/asset-gallery-detail/79f3e321-4923-45dd-8508-8082163db23d/}{Brick Wall Procedural}
    \item \label{itm:blendeHdriCrane} FreePoly --- \href{https://www.blendkit.com/asset-gallery-detail/66411e15-4a1f-46cb-8e3a-0271e83dc8e4/}{Over The Construction Site Yellow Mud}
    \item \label{itm:blendeHdriBridge} FreePoly --- \href{https://www.blendkit.com/asset-gallery-detail/aa82da45-2e29-4e4d-9ec4-631977b4d29b/}{Lake Highsky Blue Sky}
    \item \label{itm:blenderModernBuilding} Mirazev, A. --- \href{https://www.blendkit.com/asset-gallery-detail/bab4567f-7f71-44f7-af6b-96f5d60cb66e/}{Modern Building}
    \item \label{itm:pegasusSimulator} Jacinto, M. et al. --- \href{https://github.com/PegasusSimulator/PegasusSimulator/blob/main/extensions/pegasus.simulator/pegasus/simulator/assets/Robots/Iris/iris.usd}{Iris Drone Model}
    \item \label{itm:blenderWagonCar} Se-Cam, J. --- \href{https://www.blendkit.com/asset-gallery-detail/5150ca16-f5c0-49e9-b2a4-bdf9d6a51e1a/}{80s Cartoon Wagon Car}
    \item \label{itm:blenderToyCar} Se-Cam, J. --- \href{https://www.blendkit.com/asset-gallery-detail/b24f0fab-da30-4a85-bdec-c3b2fced3bb7/}{Cartoon Car / Toy Car}
    \item \label{itm:blenderPickUp} Se-Cam, J. --- \href{https://www.blendkit.com/asset-gallery-detail/c587fce1-af5a-4f77-9e78-af10526cf4a5/}{80s Cartoon Pick-Up}
    \item \label{itm:blenderSolarPanels} Tirindelli, D. --- \href{https://www.blendkit.com/asset-gallery-detail/ba8093b7-416c-461c-af22-ac311df86b03/}{6kw Solar Panels Structure}
    \item \label{itm:blenderAntenna} Rexodus --- \href{https://www.blendkit.com/asset-gallery-detail/c67c9184-3ebd-4809-adf5-91b1dd0ae759/}{Antenna}
    \item \label{itm:blenderAcUnit} Mitroi, R. --- \href{https://www.blendkit.com/asset-gallery-detail/1a06752c-d093-474a-b870-a011deb2366e/}{AC Unit Big}
    \item \label{itm:blenderConcreteTiles} Wells, C. --- \href{https://www.blendkit.com/asset-gallery-detail/3b90f722-6d01-4f62-be56-f392e9312bac/}{Procedural Concrete Tiles}
    \item \label{itm:blenderSimpleConcrete} Paludo, L. --- \href{https://www.blendkit.com/asset-gallery-detail/6156591b-e7a4-4aee-a6cc-0380517e4003/}{Simple Concrete Procedural}
    \item \label{itm:blenderDarkSteel} Share Textures --- \href{https://www.blendkit.com/asset-gallery-detail/f170e0bf-f784-482e-86eb-1d7e307a22cc/}{Dark Steel}
    \item \label{itm:blenderPaintedWall} Michaud, A. --- \href{https://www.blendkit.com/asset-gallery-detail/9891df1f-b429-461a-bc5e-b0520e25a120/}{Textured Painted Wall (Procedural)}
    \item \label{itm:blenderWhiteWall} Yang, R. --- \href{https://www.blendkit.com/asset-gallery-detail/9891df1f-b429-461a-bc5e-b0520e25a120/}{White Wall 02}
    \item \label{itm:blenderBlackWall} Adhe, E. --- \href{https://www.blendkit.com/asset-gallery-detail/61c2c84d-1701-4b19-9728-f0710bab1cce/}{Black Painted Plaster Wall}
    \item \label{itm:blenderWindowGlass} Pambudi, R.R. --- \href{https://www.blendkit.com/asset-gallery-detail/4367fa4b-9633-48a2-862f-224fbf7b1730/}{Material Window Glass Cycle}
    \item \label{itm:blenderUrbanHdri} Yang, R. --- \href{https://www.blendkit.com/asset-gallery-detail/139e9edb-ba7f-4848-b6c5-3b5c227ca805/}{Urban City Highsky Bluesky}
    \item \label{itm:blenderAsphaltRoad} NK Productions --- \href{https://www.blendkit.com/asset-gallery-detail/0f2ec127-987c-49ab-9478-7f777bdde2d3/}{Asphalt Road Procedural}
    \item \label{itm:Ship} sakigakefuruzawa --- \href{https://sketchfab.com/3d-models/hmas-stalwart-iii-d1b43d62ccdd40e8871becc15265fad1}{HMAS Stalwart (III)} 
    \item ali alipour --- \href{https://www.blendkit.com/asset-gallery-detail/03ca9bf6-5c44-419b-affc-ba22b7d17f2f/}{House}
    \item Amanpreet Bajwa --- \href{https://www.blendkit.com/asset-gallery-detail/66121466-a304-40e1-b0df-948f327275f3/}{Abies Alba Silver Fir Large Wild}
    \item Andrei Petrukovich --- \href{https://www.blendkit.com/asset-gallery-detail/038d27ac-1fcb-42a2-a8b8-f1b95426a044/}{Background Mountain}
    \item Eleanie --- \href{https://www.blendkit.com/asset-gallery-detail/4706ad0d-d3c8-4f15-a5a8-759eb570ddf3/}{Rearing horse stone moss statue}
    \item Abubakr Hamidov --- \href{https://www.blendkit.com/asset-gallery-detail/c2b5df64-16c9-4b10-ae98-89a1423faca1/}{Swing}
    \item jhon maycon --- \href{https://www.blendkit.com/asset-gallery-detail/f66d9730-9e8b-4481-bae6-a808e7efc66b/}{Seat garden-02}
    \item Klo Works --- \href{https://www.blendkit.com/asset-gallery-detail/782757c4-e95c-42f1-8004-f85791515eed/}{Garden Pedestal}
    \item antoine f --- \href{https://www.blendkit.com/asset-gallery-detail/4c02c37b-b2c6-4b55-9d26-a640eb0a403f/}{Modern garden sculpture}
    \item tias too --- \href{https://www.blendkit.com/asset-gallery-detail/dd4c221f-d25e-4fbf-a985-6f6a6ce5260b/}{Raised Garden Bed}
    \item Paweł Wałasiewicz --- \href{https://www.blendkit.com/asset-gallery-detail/4383c7b7-19f9-4f3d-8628-22d47dbc1bf0/}{Outdoor stone bonfire}
    \item Cambur theGrey --- \href{https://www.blendkit.com/asset-gallery-detail/6d9ba714-e422-4073-ae9f-ef81650e2c4a/}{Garden Bench}
    \item Tomasz Czerniak --- \href{https://www.blendkit.com/asset-gallery-detail/e6fea9e8-0e65-4426-a247-e7fb7d828f1d/}{Pergola garden}
    \item Joachim Bornemann --- \href{https://www.blendkit.com/asset-gallery-detail/75469739-5a02-45f5-9879-e15acdc6f8d3/}{Cactus garden chair}
    \item Rohit Gupta --- \href{https://www.blendkit.com/asset-gallery-detail/bdc55ac4-0696-4304-b616-4f91ac40808f/}{Low poly plymouth car}
    \item Radhames Del Valle --- \href{https://www.blendkit.com/asset-gallery-detail/d921f0b5-f3e6-442f-8c5c-2d900b5524fe/}{Garden Furniture Wooden Set}
    \item FriskyTater --- \href{https://www.blendkit.com/asset-gallery-detail/8380d175-4f0f-423b-b9d0-3471e896abc9/}{Folding Ladder}
    \item EB Adventure Photoscans --- \href{https://www.blendkit.com/asset-gallery-detail/346f848e-4c9a-4c5c-947c-eac6613e2c85/}{Alpine Boulder Rocks}
    \item Esther Michael --- \href{https://www.blendkit.com/asset-gallery-detail/40ffdf82-fb21-4527-9ecd-40f9f3bcf712/}{Modern Glass House}
    \item Ibrohim Toxirov --- \href{https://www.blendkit.com/asset-gallery-detail/c3c28d91-3547-4b7b-9cc3-68d00d7a6993/}{Audi e-tron GT quattro Black}
    \item Toby Noby --- \href{https://www.blendkit.com/asset-gallery-detail/ef359c9c-3e75-42e4-a7c5-2fa935fab219/}{Tree}
    \item Andrei Petrukovich --- \href{https://www.blendkit.com/asset-gallery-detail/a82e5300-c15b-4717-9979-ba609f9e0254/}{Grand Fir (8m)}
    \item hhkkh --- \href{https://www.blendkit.com/asset-gallery-detail/2688549f-f161-4736-9949-995f2f533043/}{Nature Road}
    \item Antoine Bagattini --- \href{https://www.blendkit.com/asset-gallery-detail/597e96b6-c52e-4b78-a8fe-3681119fd7ed/}{Lowpoly Forest (with random season)}
    \item Dal Roron --- \href{https://www.blendkit.com/asset-gallery-detail/0007a31c-467f-47ce-b0c4-aa9c43738d2c/}{Fence Segment}
    \item Ibrohim Toxirov --- \href{https://www.blendkit.com/asset-gallery-detail/b2a971b0-08f1-4354-aecc-0969e51c5c20/}{Grass (lawn) procedural}
    \item Joshua Whitney --- \href{https://www.blendkit.com/asset-gallery-detail/a360825a-f3ff-470d-885a-a22d99cb77a2/}{Patch of Photoscanned Dirt}
    \item 99.Miles --- \href{https://sketchfab.com/3d-models/free-city-road-layout-mockup-3260e3fef3434939bc8df94a0a15ead3}{[FREE] City Road Layout Mockup}
    \item Veterock (@windofglass) --- \href{https://sketchfab.com/3d-models/parking-lot-80e54d8326ea4646949961e8ada35518}{Parking lot}
    \item Pasha (@Pasha.) --- \href{https://sketchfab.com/3d-models/building-pack-ecec5070c3564d429062f346355634c7}{Building pack}
    \item AIR3D --- \href{https://www.blendkit.com/asset-gallery-detail/4b541812-f332-4b6c-94f5-26a2f800c80e/}{Unfinished Building}
    \item EB Adventure Photoscans --- \href{https://www.blendkit.com/asset-gallery-detail/58e7a02b-c565-461d-a516-ea34aafbc451/}{Rural Highway Road Photoscan}
    \item Free poly --- \href{https://www.blendkit.com/asset-gallery-detail/0ca8370d-fe5d-4428-9ac3-d0a18976a4bf/}{Scan Stone sand pile}
    \item Hamid Bairouk --- \href{https://www.blendkit.com/asset-gallery-detail/909d84f9-6993-4051-93f1-0c6e378cf15b/}{Excavator caterpillar 323d}
    \item Andrew Crook --- \href{https://www.blendkit.com/asset-gallery-detail/af5173fc-4a0b-4fcb-89af-7b863af5b079/}{Industrial Skip}
    \item REAL SCANS --- \href{https://www.blendkit.com/asset-gallery-detail/2a857782-3f95-4947-9087-a23cf20ed223/}{Bricks stack}
    \item The Doctor By Design --- \href{https://www.blendkit.com/asset-gallery-detail/62672748-2e7c-4239-987b-e38a7a60046d/}{Rubbish Skip}
    \item REAL SCANS --- \href{https://www.blendkit.com/asset-gallery-detail/d2652ecf-41ce-43ad-a5f9-cdcfd88c6ff0/}{Bricks large pile}
    \item Mohammed Hossain --- \href{https://www.blendkit.com/asset-gallery-detail/0438017c-c647-4a10-87a9-27adfd0ca13a/}{Street Lamp Post}
    \item Joshua Whitney --- \href{https://www.blendkit.com/asset-gallery-detail/e0478006-774f-4120-b432-ff3cdf22611a/}{Photoscanned Pile Of Clumpy Dirt 02}
    \item REAL SCANS --- \href{https://www.blendkit.com/asset-gallery-detail/1748ffad-48f8-49d8-8fe2-603936876bfc/}{Bricks medium Pile}
    \item Jordi Se-Cam --- \href{https://www.blendkit.com/asset-gallery-detail/0ff3bb57-8fa6-42cc-8023-5502408bac6d/}{Cartoon Car / toy car}
    \item GL3D Gabriela Linhares --- \href{https://www.blendkit.com/asset-gallery-detail/254b4226-73a6-4b8b-b8ff-dbcd9abd6b07/}{Black Bucket with Construction Tools}
    \item GL3D Gabriela Linhares --- \href{https://www.blendkit.com/asset-gallery-detail/55ac2903-74fc-45d6-b44e-2ef7a4ea2460/}{Realstic Debris Bucket}
    \item SPLEEN.VISION Adam --- \href{https://www.blendkit.com/asset-gallery-detail/767958e0-4944-44a8-9f5b-e93669a321e3/}{Lowpoly concrete barriers}
    \item ALEX SAMUSENKO --- \href{https://www.blendkit.com/asset-gallery-detail/6408d7fc-85a3-43db-826c-979cf72cfa9d/}{Construction trolley}
     \item diaverx miky --- \href{https://www.blendkit.com/asset-gallery-detail/3da0e1a5-513d-4f26-84a7-78520678f8ad/}{Procedural Turf}
    \item Soran Mohammadi --- \href{https://www.blendkit.com/asset-gallery-detail/c5f7c5cf-c11f-4816-bf77-315d09d371e4/}{Realsistic Procedural Dirt}
    \item Lucas Paludo --- \href{https://www.blendkit.com/asset-gallery-detail/398f6810-4774-475d-810d-34b0231d26ad/}{Simple Concrete Procedural}
    \item Albert Bozesan --- \href{https://www.blendkit.com/asset-gallery-detail/7f2f336d-cf91-4f85-9a30-2e9a3bb4e93d/}{Procedural Brick}
    \item Waldemar Dad --- \href{https://www.blendkit.com/asset-gallery-detail/46021334-e26f-4889-9f8e-36cb6c47860b/}{Customizable Procedural Wood}
    \item Chris Wells --- \href{https://www.blendkit.com/asset-gallery-detail/3b90f722-6d01-4f62-be56-f392e9312bac/}{Procedural Concrete Tiles}
    \item Russo 3D --- \href{https://www.blendkit.com/asset-gallery-detail/f5ab3780-1489-4c62-8050-b400bbd14e95/}{Procedural White Brick}
    \item James Middleton --- \href{https://www.blendkit.com/asset-gallery-detail/fa2d62e6-fc17-4310-a0bf-b35eaefcfd3c/}{Stone Flags or Bricks 5}
    \item Ibrohim Toxirov --- \href{https://www.blendkit.com/asset-gallery-detail/857ef1e0-649a-42ca-a2ae-0b3396f0fe2d/}{Infinite Cloudy Sky}
\item Poly Haven --- \href{https://www.blendkit.com/asset-gallery-detail/fb201776-945b-4651-8451-78ec7dc79ab8/}{Kloofendal Overcast Pure Sky}
\item EB Adventure Photoscans --- \href{https://www.blendkit.com/asset-gallery-detail/824bbb4a-9ace-4f95-8d04-7e9c4e39ee15/}{Peaceful Twilight Under Stars}
\item EB Adventure Photoscans --- \href{https://www.blendkit.com/asset-gallery-detail/b8bc5791-c263-4a3e-82a6-3de5aab19a8e/}{Glacier Lake Pink Sunrise}
    \item dasy444 --- \href{https://sketchfab.com/3d-models/street-city-buildings-8-873c5f14ec464966a11a4d772d187270}{street city buildings (8)}
    \item abhayexe --- \href{https://sketchfab.com/3d-models/full-gameready-city-buildings-iii-hongkong-03a5255bb9624d4e82cb8e15e0070b70}{Full Gameready City Buildings III [HongKong]}
    \item dasy444 --- \href{https://sketchfab.com/3d-models/street-city-7-for-games-free-493a69b451284ff88346c7b3e4e1b5a7}{street city (7) for games FREE}
\end{enumerate}
The Iris Drone Model is from a GitHub project with a \href{https://github.com/PegasusSimulator/PegasusSimulator?tab=BSD-3-Clause-1-ov-file}{BSD 3-Clause License}.
\end{document}